\documentclass[runningheads]{llncs}

\usepackage[mobile,year=2026]{eccv}

\usepackage{eccvabbrv}
\usepackage{graphicx}
\usepackage{booktabs}
\usepackage{multirow}
\usepackage{array}
\usepackage{amsmath,amssymb}
\usepackage{xspace}
\usepackage{import}
\usepackage{overpic}
\usepackage{contour}
\usepackage{varioref}
\usepackage[hidelinks]{hyperref}
\hypersetup{pdftitle={Diffusion-Based Material Regularization for Physics-Based Inverse Rendering},
            pdfauthor={Jingwang Ling, Lifan Wu, Feng Xu, Shuang Zhao}}
\usepackage[section]{placeins}
\usepackage[accsupp]{axessibility}
\usepackage[misc]{ifsym}

\contourlength{0.2pt}

\providecommand{\TextOrMath}[2]{\ifmmode #2\else #1\fi}
\newenvironment{teaserfigure}{\begin{center}\captionsetup{type=figure}}{\end{center}}
\providecommand{\citet}[1]{\cite{#1}}
\providecommand{\citep}[1]{\cite{#1}}

\usepackage{collect}
\definecollection{mymaths}
\newcommand{\mymath}[2]{
    \newcommand{#1}{\TextOrMath{$#2$\xspace}{#2}}
    \begin{collect}{mymaths}{}{}{}{}
    #1
    \end{collect}
}

\DeclareRobustCommand{\casename}[1]{%
    \begingroup
    \Urlmuskip=0mu plus 1mu\relax
    \nolinkurl{#1}%
    \endgroup
}

\mymath{\nviews}{N}
\mymath{\viewimg}{I}
\mymath{\camparams}{\mathbf{p}}
\mymath{\gbuffer}{\mathbf{G}}
\mymath{\gbufferalbedo}{\mathbf{A}}
\mymath{\gbufferrough}{\mathbf{R}}
\mymath{\gbuffermetal}{\mathbf{M}}
\mymath{\gbuffernormal}{\mathbf{N}}
\mymath{\gbufferweight}{\mathbf{W}}
\mymath{\renderimg}{\hat{I}}
\mymath{\imgweight}{\alpha}
\mymath{\loss}{\mathcal{L}}
\mymath{\Limg}{\loss_{\text{img}}}
\mymath{\relmse}{\loss_{\text{rMSE}}}
\mymath{\sg}{\operatorname{sg}}
\mymath{\Lmat}{\loss_{\text{mat}}}
\mymath{\Ljbf}{\loss_{\text{jbf}}}
\mymath{\Llight}{\loss_{\text{light}}}
\mymath{\Lshape}{\loss_{\text{shape}}}
\mymath{\nlobes}{K}
\mymath{\brdfweight}{w}
\mymath{\brdfweightmin}{w_{\min}}
\mymath{\brdfalbedo}{\mathbf{a}}
\mymath{\brdfrough}{\rho}
\mymath{\brdfmetal}{m}
\mymath{\roughmin}{\rho_{\min}}
\mymath{\roughmax}{\rho_{\max}}
\mymath{\roughint}{[\rho_{\min,k},\rho_{\max,k}]}

\def\equationautorefname~#1\null{%
  Equation~#1\null
}

\begin{document}

\title{Diffusion-Based Material Regularization for Physics-Based Inverse Rendering}
\titlerunning{Diffusion-Based Material Regularization for Physics-Based Inverse Rendering}

\author{Jingwang Ling\inst{1} \and Lifan Wu\inst{2} \and Feng Xu\inst{3} \and Shuang Zhao\inst{1}\textsuperscript{(\Letter)}}
\authorrunning{J. Ling et al.}
\institute{$^{1}$\,University of Illinois Urbana-Champaign\qquad$^{2}$\,NVIDIA\\
$^{3}$\,BNRist and School of Software, Tsinghua University\\
\email{\{ling23,shzhao\}@illinois.edu, lifanw@nvidia.com, xufeng2003@gmail.com}}

\maketitle

\begin{teaserfigure}
  \centering
\begin{tabular}{@{}c@{}}
  \begin{minipage}[t]{1.0\linewidth}\centering
\begin{tabular}{@{}c@{}c@{}c@{}c@{}c@{}}
  \multicolumn{5}{l}{\normalsize\texttt{\detokenize{Stanford-ORB cup_scene006}}} \\
  \includegraphics[width=0.197\linewidth]{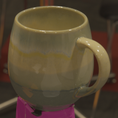} & \includegraphics[width=0.197\linewidth]{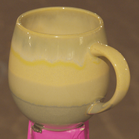} & \includegraphics[width=0.197\linewidth]{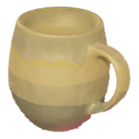} & \includegraphics[width=0.197\linewidth]{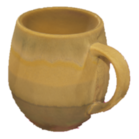} & \includegraphics[width=0.197\linewidth]{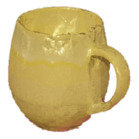}
\end{tabular}
\end{minipage}
\end{tabular}
\vspace{0.5em}
\begin{tabular}{@{}c@{}}
  \begin{minipage}[t]{1.0\linewidth}\centering
\begin{tabular}{@{}c@{}c@{}c@{}c@{}c@{}}
  \multicolumn{5}{l}{\normalsize\texttt{\detokenize{Stanford-ORB pepsi_scene004}}} \\
  \includegraphics[width=0.197\linewidth]{Input.png} & \includegraphics[width=0.197\linewidth]{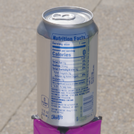} & \includegraphics[width=0.197\linewidth]{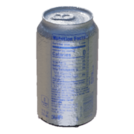} & \includegraphics[width=0.197\linewidth]{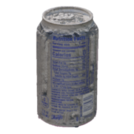} & \includegraphics[width=0.197\linewidth]{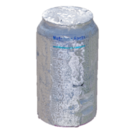} \\
  \normalsize Input & \normalsize Relight-GT & \normalsize Ours & \normalsize Neural-PBIR & \normalsize MaterialFusion
\end{tabular}
\end{minipage}
\end{tabular}
  \caption{Qualitative comparison of relighting on the Stanford-ORB dataset~\cite{DBLP:conf/nips/KuangZYAWW23} between our method, Neural-PBIR~\cite{DBLP:conf/iccv/0004CLYZM0Z023}, and MaterialFusion~\cite{DBLP:conf/3dim/LitmanPDAZTT25}. With our new material clustering regularizer, we avoid baked-in shadows while accurately modeling spatially varying materials (top). Our method is also more robust to strong highlights on glossy metallic surfaces, producing more accurate reflections (bottom).}
  \label{fig:relight-curated-comparison-teaser}
\end{teaserfigure}

\begin{abstract}
  Reconstructing physics-based 3D assets---geometry, materials, and illumination---from multi-view images is a core problem in computer graphics and vision, and a prerequisite for realistic relighting and editing. Physics-based inverse rendering offers an accurate image-formation model, but is severely underconstrained: without strong priors, illumination is baked into materials, and reconstructions generalize poorly to novel views and lighting. Data-driven diffusion models, in contrast, predict visually plausible materials, yet their predictions rarely satisfy the rendering equation and are not directly usable for physics-based rendering. We bridge these two paradigms rather than replacing either. Our key idea is to treat the predictions of a state-of-the-art diffusion model not as target material values but as a similarity kernel for optimization: we introduce a regularization loss that penalizes deviations in the optimized material over surface regions where the diffusion predictions are near-constant, while leaving the optimization free to match the input images. Built on this regularizer, our end-to-end pipeline jointly reconstructs geometry, materials, and illumination, yielding high-quality assets that drop into standard rendering pipelines and relight faithfully. On the Synthetic4Relight, Stanford-ORB, and DTC-Synthetic datasets, our method significantly outperforms state-of-the-art baselines in both reconstruction accuracy and relighting quality.

\end{abstract}

\section{Introduction}
\label{sec:intro}

Physics-based 3D assets, including object geometry, material properties, and illumination conditions, are essential to a wide array of applications in virtual/augmented reality, e-commerce, and entertainment.
Consequently, reconstructing these assets from captured images is a core problem in computer graphics and vision.

To this end, a common approach is \emph{inverse rendering}, also known as \emph{analysis by synthesis}.
By formulating reconstruction as an optimization problem, these techniques iteratively refine asset parameters, typically via some variant of stochastic gradient descent, to minimize the discrepancy between input images and their corresponding renderings.
Unfortunately, inverse-rendering optimizations are generally underconstrained, especially for material reconstruction.
Without proper priors or regularization, the reconstructed assets often generalize poorly to novel viewing or lighting conditions (e.g., suffering from severe ``baking'' artifacts), significantly limiting their usefulness.

Recently, \emph{data-driven} methods have emerged as a powerful alternative.
Trained on large quantities of real and synthetic image data, these methods directly infer physics-based parameters without the need for optimization.
However, despite being visually plausible and often free of baking artifacts, the reconstructions produced by these methods typically lack the physical accuracy required to generate renderings that faithfully match the input images.

In this paper, we introduce a technique that regularizes inverse-rendering-based material optimization using a state-of-the-art data-driven model: DiffusionRenderer~\cite{DBLP:journals/corr/abs-2501-18590}.
Instead of simply initializing inverse-rendering optimizations with diffusion predictions, we exploit the spatial consistency encoded in the model's predictions.
Specifically, we devise a novel regularization loss that penalizes deviations in optimized material parameters over surface regions where the diffusion predictions are near-constant.
Building on this regularization scheme, we develop an end-to-end inverse-rendering pipeline that reconstructs geometry, material, and illumination from multi-view images of an object.

Our method is designed to bridge the gap between data-driven material prediction and physics-based inverse rendering, rather than replacing either component alone.
Diffusion-based predictors such as DiffusionRenderer~\cite{DBLP:journals/corr/abs-2501-18590} and RGB$\leftrightarrow$X~\cite{DBLP:conf/siggraph/0005DGHHLYH24} can produce perceptually convincing intrinsic maps, but these predictions are not guaranteed to satisfy the rendering equation or reproduce held-out relighting results.
Conversely, vanilla physics-based optimization in Mitsuba~\cite{DBLP:journals/tog/JakobSRV22} provides an accurate rendering model, but remains underconstrained under sparse views and unknown illumination, making it prone to material-lighting ambiguity and baking artifacts.
The key distinction of our approach is to use diffusion-predicted G-buffers as a similarity kernel for physics-based optimization, rather than as fixed target values or a generic smoothness prior.

To demonstrate the effectiveness of our method, we show that it outperforms several state-of-the-art baselines~\cite{DBLP:conf/iccv/0004CLYZM0Z023,DBLP:conf/3dim/LitmanPDAZTT25} on the Synthetic4Relight~\cite{DBLP:conf/cvpr/ZhangSHFJZ22}, Stanford-ORB~\cite{DBLP:conf/nips/KuangZYAWW23}, and DTC-Synthetic~\cite{DBLP:conf/cvpr/DongCLYZZZTLMCF25} datasets.

\section{Related Work}
\label{sec:related_work}

\paragraph{Physics-based inverse rendering.}
Inverse rendering reconstructs scene parameters---geometry, material, and lighting---by matching renderings to observed images.
Recent advances in physics-based differentiable rendering~\cite{Zhao:2020:DRCourse} enable gradient-based optimization of these parameters via Monte Carlo light transport, an approach known as analysis by synthesis.
The problem is fundamentally ill-posed because of inherent ambiguities among these parameters~\cite{DBLP:conf/cvpr/ChungKKY23,oscanoa2023variational}, so many systems add regularization to constrain the solution space~\cite{DBLP:journals/cgf/LuanZBD21,DBLP:conf/cvpr/SchmittDRKG20,DBLP:conf/iclr/ZhengCZWWFZ0KRB25}.
Neural representations such as NeRFs~\cite{DBLP:conf/eccv/MildenhallSTBRN20} and Gaussian Splatting~\cite{DBLP:journals/tog/KerblKLD23} achieve impressive reconstruction and novel view synthesis~\cite{DBLP:conf/iccv/0004CLYZM0Z023,DBLP:conf/cvpr/JinLXZHBZX023,DBLP:journals/tog/ZhuWY26,DBLP:conf/cvpr/SunGXYW25}, but extracting 3D assets compatible with standard graphics pipelines remains challenging.
Flash Cache~\cite{DBLP:conf/eccv/AttalVMHBOS24} and IRGS~\cite{DBLP:conf/cvpr/GuWZY025} improve reconstruction quality through more accurate light transport, using a fast radiance cache and inter-reflection-aware 2D Gaussian ray tracing, respectively.
These advances are orthogonal to our focus: they target light-transport accuracy, whereas our regularization targets material-lighting ambiguity.
Our work addresses surface-based inverse rendering, producing standard triangle meshes and physically based materials (e.g., the Disney BRDF) that are directly usable in traditional rendering engines.

\paragraph{Data-driven priors for inverse rendering.}
Data-driven methods have advanced material estimation and intrinsic decomposition.
For material estimation, models such as RGB{\(\leftrightarrow\)}X~\cite{DBLP:conf/siggraph/0005DGHHLYH24} and DiffusionRenderer~\cite{DBLP:journals/corr/abs-2501-18590} predict intrinsic G-buffers (albedo, roughness, metallic, normal) from single images or videos.
For lighting, DiffusionLight~\cite{DBLP:conf/cvpr/PhongthaweeCSJR24,DBLP:journals/corr/abs-2507-01305} and LuxDiT~\cite{liang2025luxdit} use diffusion models to estimate environmental illumination.
A growing line of work integrates such priors directly into the inverse-rendering optimization loop.
MaterialFusion~\cite{DBLP:conf/3dim/LitmanPDAZTT25} and IntrinsicAnything~\cite{DBLP:conf/eccv/ChenPYLPLZ24} use diffusion priors to guide reconstruction, while VideoMat~\cite{DBLP:journals/cgf/MunkbergWLSH25} extracts PBR materials with video diffusion models.
MatMart~\cite{DBLP:journals/corr/abs-2511-18900} adopts a two-stage diffusion framework: it first predicts materials from observed views, then completes unobserved regions through prior-guided generation, using view-material cross-attention for multi-view consistency.
However, relying solely on per-view predictions can introduce inconsistencies and bias.
Rather than treating these predictions as ground truth, we use them as structural guidance to regularize our physics-based optimization, making it robust to prediction errors from the data-driven models.

A complementary line of work learns to relight images or 3D representations directly, bypassing explicit PBR material estimation.
Single-image methods~\cite{DBLP:conf/nips/JinLLXBZX0S24,DBLP:conf/siggraph/00010PKW024} and NeRF-distillation approaches~\cite{DBLP:conf/nips/ZhaoSVPMH24,DBLP:journals/corr/abs-2510-03163} produce compelling relighting but do not yield exportable PBR material maps, limiting downstream use in standard rendering pipelines.
Single-image methods such as DiLightNet~\cite{DBLP:conf/siggraph/00010PKW024} further operate on a fixed input view and do not synthesize the held-out novel views required by multi-view relighting benchmarks such as Stanford-ORB.
Feed-forward methods such as RelitLRM~\cite{DBLP:conf/iclr/ZhangKJXBTZHHFZ25} offer fast inference but rely on fixed learned priors that may not generalize to all real-world appearances.
LightSwitch~\cite{DBLP:journals/corr/abs-2508-06494} casts multi-view-consistent relighting as a denoising task, and Alzayer et al.~\cite{DBLP:conf/cvpr/AlzayerHBHSV25} handle in-the-wild captures with large illumination variation.
Like this line of work, our method leverages data-driven priors, but it combines them with physics-based optimization to improve the physical accuracy of the reconstructed material asset rather than directly synthesizing relit images.

\paragraph{Regularization of material parameters.}
Optimizing spatially varying materials from sparse views requires effective regularization to prevent overfitting and baking artifacts.
A classic strategy is reflectance sharing~\cite{DBLP:conf/rt/ZicklerERB05}, which assumes the scene contains a limited set of materials and shares observations across surface points with similar appearance.
For example, Lensch et al.~\cite{DBLP:journals/tog/LenschKGHS03} use hard cluster assignments derived from the current optimization state, which can be unreliable early on when materials still contain baked artifacts.
This idea has been formalized through basis BRDFs~\cite{DBLP:journals/tog/ZhouCDWYST16,DBLP:conf/cvpr/ChungCB25} that enforce piecewise simplicity.
Other approaches rely on explicit material segmentation~\cite{DBLP:journals/tog/SharmaPGFDD23,DBLP:journals/corr/abs-2411-19322}, which is itself difficult to obtain accurately.
Sharma et al.~\cite{DBLP:journals/tog/SharmaPGFDD23} additionally learn a perceptual similarity metric for materials, but expose it only as point-wise similarity maps or binary masks, neither of which is readily applicable to inverse rendering.
Our approach is inspired by cross-channel regularization techniques such as joint bilateral filtering (JBF)~\cite{DBLP:journals/tog/PetschniggSACHT04}, which uses guidance buffers to refine a target image and can be implemented efficiently for high-dimensional data via permutohedral lattices~\cite{DBLP:journals/cgf/AdamsBD10}.
In inverse rendering, JBF-like terms have been used to regularize specular parameters using albedo~\cite{DBLP:journals/cgf/LuanZBD21,DBLP:conf/cvpr/SchmittDRKG20}, and Wiersma et al.~\cite{DBLP:conf/siggraph/WiersmaPHMLED25} use a related similarity kernel to estimate uncertainty over BRDF parameters when geometry and illumination are known.
We instead tackle the joint reconstruction of geometry, materials, and lighting under unknown illumination by constructing a stable similarity kernel from externally predicted G-buffers.
This encourages regions with similar predicted G-buffers to converge to similar optimized parameters, constraining the solution space while still allowing physical corrections.

\section{Method}
\label{sec:method}
Taking as input $\nviews$ multi-view images $\{\viewimg_i\}_{i=1}^{\nviews}$ of a static object under unknown illumination, our goal is to reconstruct physics-based representations of the object's shape, materials, and illumination, all directly compatible with standard rendering pipelines.
Specifically, we seek asset parameters that produce renderings closely resembling the input images.
When multiple solutions fit the images equally well, we aim to obtain physically plausible and visually coherent reconstructions that minimize baked-in artifacts (when viewed under novel conditions).

\begin{figure*}[t]
    \centering
    \def\svgwidth{\linewidth}
    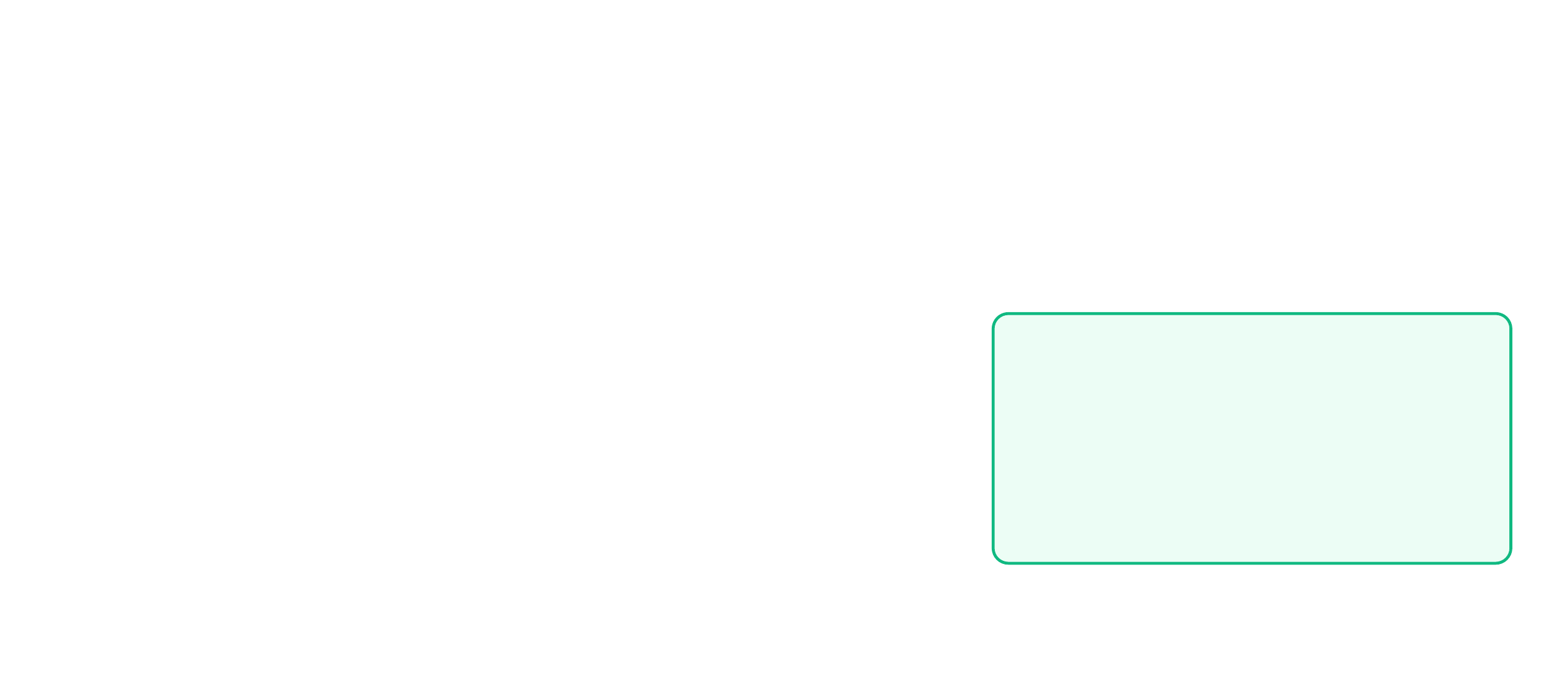
    \caption{Overview of our pipeline.
    From $\nviews$ multi-view images $\{\viewimg_i\}$ under unknown illumination, we (1) predict per-view intrinsic G-buffers $\gbuffer=[\gbufferalbedo,\gbufferrough,\gbuffermetal,\gbuffernormal]$ with a conditional diffusion model;
    (2) reconstruct a voxel-grid SDF by neural volume rendering, supervised by the predicted normals $\gbuffernormal$ ($\Lshape$);
    and (3) jointly optimize shape, spatially varying material, and an environment map by differentiable rendering, minimizing the photometric loss $\Limg$ and our material clustering regularizer $\Lmat$, with albedo regularized through the scale-agnostic transform $\psi$.
    The result is a renderer-ready PBR asset that relights faithfully under novel illumination.
    \emph{Bottom:} the diffusion-predicted G-buffer $\mathbf{g}=[\gbufferalbedo,\gbufferrough,\gbuffermetal]$ guides joint bilateral filtering of the rendered G-buffer $\hat{\mathbf{g}}$.
    The G-buffers shown are from an early training iteration for illustration purposes and do not reflect the final reconstruction quality.}
    \label{fig:pipeline}
\end{figure*}

To this end, we introduce a technique that operates in three stages: \emph{preprocessing}, \emph{neural shape reconstruction}, and \emph{physics-based inverse rendering} (PBIR).
\cref{fig:pipeline} provides an overview.
In the following, we detail each of these stages.

\paragraph{Preprocessing.}
For each view $i$, our technique calibrates the camera parameters and leverages a conditional diffusion model~\cite{DBLP:conf/siggraph/0005DGHHLYH24,DBLP:journals/corr/abs-2501-18590} to predict intrinsic G-buffers $\gbuffer_i$ comprising the base color $\gbufferalbedo_i$, roughness $\gbufferrough_i$, metallic $\gbuffermetal_i$, and surface normal $\gbuffernormal_i$.
These buffers will be used in the following stages to supervise the reconstruction of object shape and materials.

\paragraph{Neural surface reconstruction.}
With the per-view G-buffers obtained, we then reconstruct the shape of the object represented as a voxel-grid signed distance function (SDF) %
using neural volume rendering~\cite{DBLP:conf/eccv/MildenhallSTBRN20,DBLP:conf/nips/WangLLTKW21,DBLP:conf/iccv/0004CLYZM0Z023}.
To further improve reconstruction quality, we incorporate a normal-supervision loss that encourages the rendered normal buffer $\hat{\gbuffernormal}_i$ to match the diffusion-predicted normals $\gbuffernormal_i$:
\begin{equation}
    \label{eq:Lshape}
    \Lshape = \sum_{i = 1}^{\nviews} \sum_{p} H_\delta\!\left(1-\hat{\gbuffernormal}_{i,p}^{\top}\,\gbuffernormal_{i,p}\right)\,,
\end{equation}
where $\hat{\gbuffernormal}_{i,p}$ and $\gbuffernormal_{i,p}$ denote, respectively, the $p$-th pixel of $\hat{\gbuffernormal}_i$ and $\gbuffernormal_i$.
In addition, $H_\delta$ is the Huber penalty---which improves robustness to imperfect normal predictions---given by
\begin{equation}
    H_\delta(t)=
    \begin{cases}
        \tfrac{1}{2} t^2, & t\le \delta, \\
        \delta(t-\tfrac{1}{2}\delta), & \text{otherwise},
    \end{cases}\,
\end{equation}
where $t\ge 0$, and $\delta$ is set to $0.03$ to downweight angular differences larger than $15^\circ$.

In practice, the normal-supervision loss in \cref{eq:Lshape} enhances surface details and helps mitigate concave shape artifacts---especially on glossy surfaces (see \cref{fig:normal-pitcher-scene001}).
With the SDF reconstructed, we extract a mesh using Marching Cubes~\cite{DBLP:conf/siggraph/LorensenC87} as the initial shape of the object.

We note that, given the calibrated camera parameters and a reconstructed object geometry, the diffusion-predicted G-buffers $\{\gbuffer_i\}$ could be directly back-projected onto the object surface to complete the material reconstruction.
However, as we will demonstrate in \cref{sec:alternative_methods}, materials reconstructed using this strategy often fail to produce renderings that match the ground truth.
Instead, we will use these G-buffers as priors to greatly improve the robustness of our PBIR stage.

\paragraph{Physics-based inverse rendering.}
With the initial mesh obtained, the last stage of our technique jointly optimizes the object's shape, materials, and illumination to minimize:
\begin{equation}
    \loss = \Limg + \lambda_{\text{mat}} \Lmat\,,
\end{equation}
where $\Lmat$ regularizes the material using the diffusion-predicted G-buffers $\{ \gbuffer_i \}$ and $\lambda_{\text{mat}}$ is the corresponding scalar weight.
In addition, $\Limg$ is the image rendering loss measured in relative MSE~\cite{DBLP:conf/icml/LehtinenMHLKAA18,DBLP:conf/cvpr/MildenhallHMSB22}:
\begin{equation}
    \label{eq:relmse}
    \Limg = \sum_{i = 1}^{\nviews} \left\lVert \frac{\renderimg_i - \viewimg_i}{\sg(\renderimg_i) + \epsilon} \right\rVert_2^2\,,
\end{equation}
where $\renderimg_i$ denotes the image rendered under the $i$-th view, $\sg(\cdot)$ is the stop-gradient operator, and $\epsilon > 0$ is a small constant for stable normalization.
Because our datasets contain HDR images, the relative loss prevents a small number of bright pixels from dominating the optimization.

In the following, we detail our material regularization strategy in \cref{sec:material_regularization} and describe our scale-agnostic albedo transformation in \cref{sec:albedo_transf}.

\subsection{Implicit Material Clustering Regularization}
\label{sec:material_regularization}

Directly optimizing spatially varying materials from sparse multi-view observations under unknown illumination is severely underconstrained. 
Each surface location is observed from only a limited set of views and lighting conditions, often causing the optimization to bake illumination effects into the material.
Recent conditional diffusion models can infer plausible per-view intrinsic G-buffers from RGB inputs~\cite{DBLP:conf/siggraph/0005DGHHLYH24,DBLP:journals/corr/abs-2501-18590}, producing results that are often artifact-free and semantically coherent. However, these predictions are not constrained by the rendering equation and thus lack physical accuracy, leading to suboptimal visual quality when used directly for rendering.

Recent methods have integrated diffusion-based material priors into physics-based inverse rendering~\cite{DBLP:conf/3dim/LitmanPDAZTT25,DBLP:journals/cgf/MunkbergWLSH25}. 
However, directly regularizing toward the predicted G-buffers creates a tension with photometric fitting; approaches based on score distillation with annealed diffusion noise~\cite{DBLP:conf/3dim/LitmanPDAZTT25,DBLP:conf/iclr/ZhuZK24} or global scale-invariant losses~\cite{DBLP:journals/cgf/MunkbergWLSH25,DBLP:conf/eccv/ChenPYLPLZ24} do not fully address material-dependent biases in a scene with multiple materials. We instead leverage the implicit grouping of same-material regions found in diffusion predictions, and regularize our solution to remain consistent within these regions while allowing for per-region deviations. Akin to reflectance sharing~\cite{DBLP:conf/rt/ZicklerERB05}, this effectively reduces the degrees of freedom by sharing observations across similar material regions.

A practical challenge is to identify which image regions correspond to the same material without committing to an explicit segmentation. One could fit a small set of BRDF bases with spatially sparse mixing (so that each region selects a single basis)~\cite{DBLP:journals/tog/ZhouCDWYST16,DBLP:conf/cvpr/ChungCB25}. However, this introduces model selection issues (e.g., number of bases) and sensitivity to region boundaries. We instead use diffusion-predicted G-buffers to define an implicit, soft notion of material similarity across pixels as follows. For a given view, let $\mathbf{g}=[\gbufferalbedo,\gbufferrough,\gbuffermetal]$ denote the predicted material G-buffer, where each vector $\mathbf{g}_p$ is the concatenation of channels at pixel $p$. We define the similarity kernel between pixels $p$ and $q$ as
\begin{equation}
    \label{eq:kpq}
	k_{p,q}(\mathbf{g})=\exp\!\left(-\frac{\lVert \mathbf{g}_{p}-\mathbf{g}_{q}\rVert_2^2}{2\sigma_g^2}\right),
\end{equation}
where $\sigma_g$ controls the decay of similarity with respect to G-buffer distance.

Let $\hat{\mathbf{g}}=[\hat{\gbufferalbedo},\hat{\gbufferrough},\hat{\gbuffermetal}]$ be the differentiably rendered counterpart of $\mathbf{g}$ for the same view. We encourage pixels that are similar in the predicted G-buffers to share similar optimized material by computing a kernel-weighted average via a joint bilateral filter:
\begin{equation}
    \label{eq:JBF}
    \mathrm{JBF}(\hat{\mathbf{g}};\mathbf{g})_p=
    \frac{\sum_{q} k_{p,q}(\mathbf{g})\,\hat{\mathbf{g}}_{q}}{\sum_{q} k_{p,q}(\mathbf{g})}\,,
\end{equation}
where $\mathrm{JBF}(\hat{\mathbf{g}};\mathbf{g})$ denotes the filtered buffer, and subscript $p$ indicates its value at pixel $p$. This filter can be implemented efficiently and differentiably using permutohedral lattices~\cite{DBLP:journals/cgf/AdamsBD10}.
The bottom panel of \cref{fig:pipeline} illustrates these buffers during a training iteration.

In \cref{eq:kpq,eq:JBF}, we use one kernel $k_{p,q}$ obtained using the concatenated $\mathbf{g} = [\gbufferalbedo,\gbufferrough,\gbuffermetal]$ to regularize all (i.e., albedo, roughness, and metallic) channels.
This ensures the kernel considers differences in all channels and, therefore, reduces the risk of over-regularization (e.g., when some predicted G-buffer channels are overly smooth).

Finally, we define the per-view material regularization term as the L1 distance between the rendered and filtered G-buffers:
\begin{equation}
    \label{eq:mat_reg}
    \Lmat=\left\lVert \hat{\mathbf{g}}-\mathrm{JBF}(\hat{\mathbf{g}};\mathbf{g})\right\rVert_1\,.
\end{equation}

\subsection{Scale-Agnostic Albedo Transformation}
\label{sec:albedo_transf}

Applying material regularization directly to albedo under unknown illumination can bias optimization: the optimizer might scale down the albedo to reduce the regularization loss, while the lighting compensates to match the rendered appearance, resulting in overly intense light sources. 
To address this issue, we regularize albedo in a scale-agnostic space by transforming the rendered albedo as
\begin{equation}
    \psi(\hat{\gbufferalbedo})=\sg(\hat{\gbufferalbedo})\odot \log([\hat{\gbufferalbedo}]_{\epsilon})\,,
\end{equation}
where $\odot$ denotes element-wise multiplication, $\sg(\cdot)$ is the stop-gradient operator, and $[a]_{\epsilon}=\max(a,\epsilon)$ is used for numerical stability with a small $\epsilon > 0$.
The log mapping makes the region-wise average scale-agnostic: within a region, a multiplicative rescaling of the albedo becomes an additive offset in log space, which cancels out in the difference $\hat{\mathbf{g}}-\mathrm{JBF}(\hat{\mathbf{g}};\mathbf{g})$.
However, the log function also rescales gradients by a factor proportional to $1/\hat{\gbufferalbedo}$. Multiplying by the detached $\sg(\hat{\gbufferalbedo})$ cancels this gradient scaling and appropriately downweights dark regions.
Ultimately, we replace $\hat{\gbufferalbedo}$ in $\hat{\mathbf{g}}$ with $\psi(\hat{\gbufferalbedo})$ when evaluating $\Lmat$.

Current diffusion models predict per-view G-buffers that may be inconsistent across views. Our regularization is relatively robust to such inconsistencies because they often manifest as per-region shifts or scalings, which are largely addressed by our design.

\section{Experiments}
\label{sec:experiments}

After describing implementation details and datasets, we compare against prior inverse-rendering and relighting methods on Stanford-ORB, Synthetic4Relight, and DTC-Synthetic, explore alternative regularization choices, ablate key components, and provide additional relighting comparisons.
The supplementary material provides additional kernel-design ablations, implementation details, the RGB$\leftrightarrow$X upstream-model experiment, experiments applying our material regularizer to a Gaussian Splatting-based inverse-rendering pipeline~\cite{DBLP:conf/cvpr/GuWZY025}, and extensive intrinsic G-buffer visualizations.
The supplementary video shows dynamic relighting and visualizations difficult to assess from static figures.

\begin{figure}[!t]
  \centering
\newcommand{\eccvpanelcaps}{%
\par\smallskip
\begin{tabular}{@{}c@{}c@{}c@{}c@{}}
\makebox[0.245\linewidth][c]{\footnotesize Relight-GT} &
\makebox[0.245\linewidth][c]{\footnotesize Ours} &
\makebox[0.245\linewidth][c]{\footnotesize N.-PBIR} &
\makebox[0.245\linewidth][c]{\footnotesize M. Fusion}
\end{tabular}}
{
\renewcommand{\normalsize}{\footnotesize}
\renewcommand{\small}{\footnotesize}
\begin{tabular}{@{}c@{\hspace{0.01\linewidth}}c@{}}
  \begin{minipage}[t]{0.49\linewidth}\centering
\begin{tabular}{@{}c@{}c@{}c@{}c@{}}
  \multicolumn{4}{l}{\normalsize\texttt{\detokenize{Stanford-ORB baking_scene003}}} \\
  \includegraphics[width=0.245\linewidth]{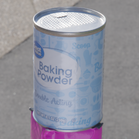} & \includegraphics[width=0.245\linewidth]{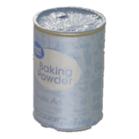} & \includegraphics[width=0.245\linewidth]{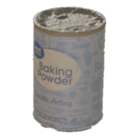} & \includegraphics[width=0.245\linewidth]{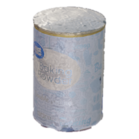}
  
\end{tabular}
\end{minipage} & \begin{minipage}[t]{0.49\linewidth}\centering
\begin{tabular}{@{}c@{}c@{}c@{}c@{}}
  \multicolumn{4}{l}{\normalsize\texttt{\detokenize{Stanford-ORB ball_scene003}}} \\
  \includegraphics[width=0.245\linewidth]{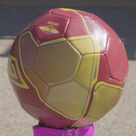} & \includegraphics[width=0.245\linewidth]{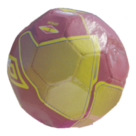} & \includegraphics[width=0.245\linewidth]{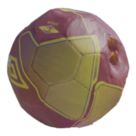} & \includegraphics[width=0.245\linewidth]{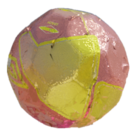}
  
\end{tabular}
\end{minipage}
\end{tabular}
\par\smallskip
\begin{tabular}{@{}c@{\hspace{0.01\linewidth}}c@{}}
  \begin{minipage}[t]{0.49\linewidth}\centering
\begin{tabular}{@{}c@{}c@{}c@{}c@{}}
  \multicolumn{4}{l}{\normalsize\texttt{\detokenize{Stanford-ORB blocks_scene005}}} \\
  \includegraphics[width=0.245\linewidth]{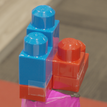} & \includegraphics[width=0.245\linewidth]{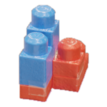} & \includegraphics[width=0.245\linewidth]{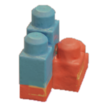} & \includegraphics[width=0.245\linewidth]{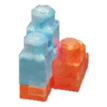}
  
\end{tabular}
\end{minipage} & \begin{minipage}[t]{0.49\linewidth}\centering
\begin{tabular}{@{}c@{}c@{}c@{}c@{}}
  \multicolumn{4}{l}{\normalsize\texttt{\detokenize{Stanford-ORB cactus_scene005}}} \\
  \includegraphics[width=0.245\linewidth]{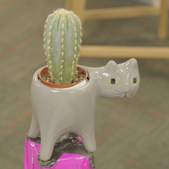} & \includegraphics[width=0.245\linewidth]{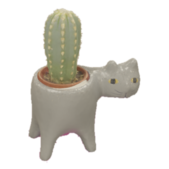} & \includegraphics[width=0.245\linewidth]{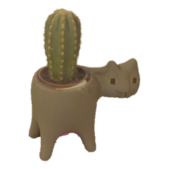} & \includegraphics[width=0.245\linewidth]{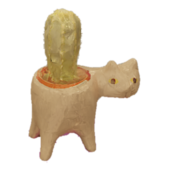}
  
\end{tabular}
\end{minipage}
\end{tabular}
\par\smallskip
\begin{tabular}{@{}c@{\hspace{0.01\linewidth}}c@{}}
  \begin{minipage}[t]{0.49\linewidth}\centering
\begin{tabular}{@{}c@{}c@{}c@{}c@{}}
  \multicolumn{4}{l}{\normalsize\texttt{\detokenize{Stanford-ORB car_scene002}}} \\
  \includegraphics[width=0.245\linewidth]{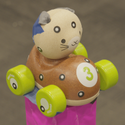} & \includegraphics[width=0.245\linewidth]{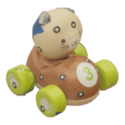} & \includegraphics[width=0.245\linewidth]{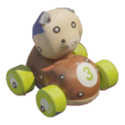} & \includegraphics[width=0.245\linewidth]{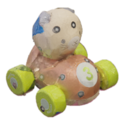}
  
\end{tabular}
\end{minipage} & \begin{minipage}[t]{0.49\linewidth}\centering
\begin{tabular}{@{}c@{}c@{}c@{}c@{}}
  \multicolumn{4}{l}{\normalsize\texttt{\detokenize{Stanford-ORB chips_scene003}}} \\
  \includegraphics[width=0.245\linewidth]{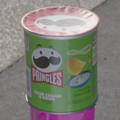} & \includegraphics[width=0.245\linewidth]{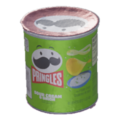} & \includegraphics[width=0.245\linewidth]{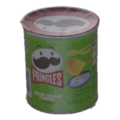} & \includegraphics[width=0.245\linewidth]{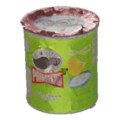}
  
\end{tabular}
\end{minipage}
\end{tabular}
\par\smallskip
\begin{tabular}{@{}c@{\hspace{0.01\linewidth}}c@{}}
  \begin{minipage}[t]{0.49\linewidth}\centering
\begin{tabular}{@{}c@{}c@{}c@{}c@{}}
  \multicolumn{4}{l}{\normalsize\texttt{\detokenize{Stanford-ORB grogu_scene001}}} \\
  \includegraphics[width=0.245\linewidth]{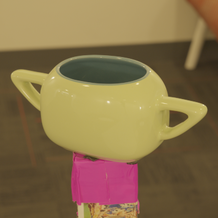} & \includegraphics[width=0.245\linewidth]{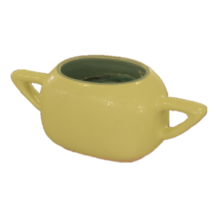} & \includegraphics[width=0.245\linewidth]{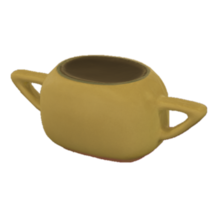} & \includegraphics[width=0.245\linewidth]{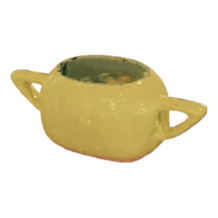}
  
\end{tabular}
\end{minipage} & \begin{minipage}[t]{0.49\linewidth}\centering
\begin{tabular}{@{}c@{}c@{}c@{}c@{}}
  \multicolumn{4}{l}{\normalsize\texttt{\detokenize{Stanford-ORB grogu_scene003}}} \\
  \includegraphics[width=0.245\linewidth]{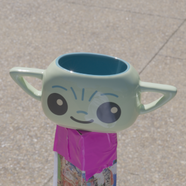} & \includegraphics[width=0.245\linewidth]{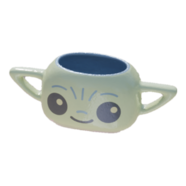} & \includegraphics[width=0.245\linewidth]{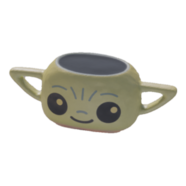} & \includegraphics[width=0.245\linewidth]{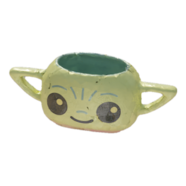}
  
\end{tabular}
\end{minipage}
\end{tabular}
\par\smallskip
\begin{tabular}{@{}c@{\hspace{0.01\linewidth}}c@{}}
  \begin{minipage}[t]{0.49\linewidth}\centering
\begin{tabular}{@{}c@{}c@{}c@{}c@{}}
  \multicolumn{4}{l}{\normalsize\texttt{\detokenize{Stanford-ORB pitcher_scene001}}} \\
  \includegraphics[width=0.245\linewidth]{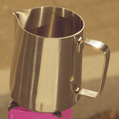} & \includegraphics[width=0.245\linewidth]{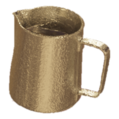} & \includegraphics[width=0.245\linewidth]{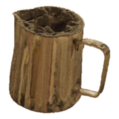} & \includegraphics[width=0.245\linewidth]{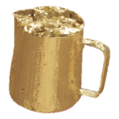}
  
\end{tabular}
\end{minipage} & \begin{minipage}[t]{0.49\linewidth}\centering
\begin{tabular}{@{}c@{}c@{}c@{}c@{}}
  \multicolumn{4}{l}{\normalsize\texttt{\detokenize{Block_RedBlue}}} \\
  \includegraphics[width=0.245\linewidth]{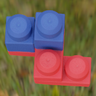} & \includegraphics[width=0.245\linewidth]{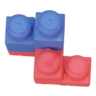} & \includegraphics[width=0.245\linewidth]{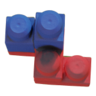} & \includegraphics[width=0.245\linewidth]{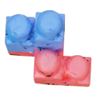}
  
\end{tabular}
\end{minipage}
\end{tabular}
\par\smallskip
\begin{tabular}{@{}c@{\hspace{0.01\linewidth}}c@{}}
  \begin{minipage}[t]{0.49\linewidth}\centering
\begin{tabular}{@{}c@{}c@{}c@{}c@{}}
  \multicolumn{4}{l}{\normalsize\texttt{\detokenize{TeaPot_YellowSunflowers}}} \\
  \includegraphics[width=0.245\linewidth]{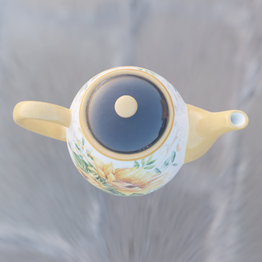} & \includegraphics[width=0.245\linewidth]{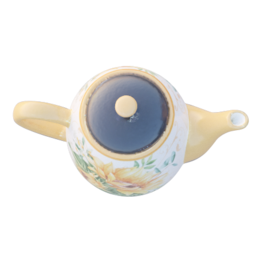} & \includegraphics[width=0.245\linewidth]{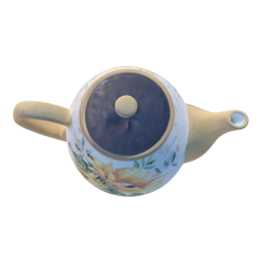} & \includegraphics[width=0.245\linewidth]{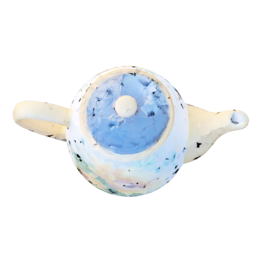}
  
\end{tabular}
\eccvpanelcaps\end{minipage} & \begin{minipage}[t]{0.49\linewidth}\centering
\begin{tabular}{@{}c@{}c@{}c@{}c@{}}
  \multicolumn{4}{l}{\normalsize\texttt{\detokenize{TeaPot_EmeraldGoldTop}}} \\
  \includegraphics[width=0.245\linewidth]{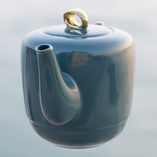} & \includegraphics[width=0.245\linewidth]{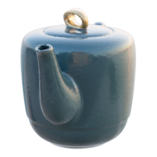} & \includegraphics[width=0.245\linewidth]{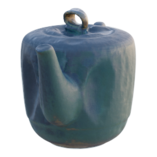} & \includegraphics[width=0.245\linewidth]{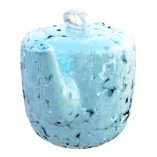}
  
\end{tabular}
\eccvpanelcaps\end{minipage}
\end{tabular}
}

  \caption{Qualitative comparison of relighting on the Stanford-ORB dataset~\cite{DBLP:conf/nips/KuangZYAWW23} and DTC-Synthetic dataset~\cite{DBLP:conf/cvpr/DongCLYZZZTLMCF25} between our method, Neural-PBIR~\cite{DBLP:conf/iccv/0004CLYZM0Z023}, and MaterialFusion~\cite{DBLP:conf/3dim/LitmanPDAZTT25}.}
  \label{fig:relight-curated-comparison}
\end{figure}

\subsection{Implementation Details}
For the SDF stage, we largely follow \citet{DBLP:conf/iccv/0004CLYZM0Z023} and add the normal regularization term with weight $10^{-5}$.
For the PBIR stage, we represent the shape as a mesh, the materials with the Disney principled BRDF~\cite{Burley2012PhysicallyBasedSA} (optimizing base color, metallic, and roughness), and the lighting as an environment map. We parameterize the spatially varying BRDF parameters and the environment map using Dictionary Fields~\cite{DBLP:journals/corr/abs-2302-01226,DBLP:journals/tog/ChenXWTSG23}, a neural field that shares a set of basis functions across locations (we use their 2D configuration). The environment map uses an exponential activation. BRDF outputs use no activation; we clamp them to $[0,1]$ and apply an $L_1$ penalty to out-of-range values with a weight of $10^{-2}$.

We optimize using an initial learning rate of $3\times10^{-2}$ with cosine annealing to $10^{-3}$. We set $\lambda_{\text{mat}}=0.1$, $\sigma_g=0.02$ and $\epsilon=10^{-2}$, and run 900 iterations. We initialize the BRDF and environment-map parameters to uniform values near 0.5. Our pipeline is implemented in Mitsuba 3~\cite{DBLP:journals/tog/JakobSRV22} and PyTorch. We use 256 samples per pixel for primal rendering and 64 samples per pixel for backward-mode automatic differentiation; each iteration renders 6 randomly sampled views. On a single RTX 5090 GPU, the SDF stage takes around 10 minutes, DiffusionRenderer preprocessing around 15 minutes, and the PBIR stage around 7 minutes per case.

\subsection{Datasets and Metrics}
We evaluate our method and baselines on a real-world benchmark, Stanford-ORB~\cite{DBLP:conf/nips/KuangZYAWW23}, and two synthetic datasets, Synthetic4Relight~\cite{DBLP:conf/cvpr/ZhangSHFJZ22} and DTC-Synthetic.
DTC-Synthetic contains seven scenes that we render using objects from DigitalTwinCatalog~\cite{DBLP:conf/cvpr/DongCLYZZZTLMCF25}, designed to more thoroughly test reconstruction of glossy objects and strong cast shadows.

Stanford-ORB contains 42 scenes with captured relighting ground truth; we use the official open-source benchmark code for evaluation.
For Synthetic4Relight, we follow the evaluation protocol of \citet{DBLP:conf/cvpr/ZhangSHFJZ22}.
For DTC-Synthetic, we render 60 relit images under three novel environment maps from \citet{DBLP:conf/cvpr/ZhangSHFJZ22} (20 images per lighting condition) and report relighting metrics in the low dynamic range (LDR) domain.
We report PSNR, SSIM, and LPIPS~\cite{DBLP:journals/corr/abs-1801-03924}.

\begin{figure}[!t]
  \centering
\newcommand{\eccvpanelcaps}{%
\par\smallskip
\begin{tabular}{@{}c@{}c@{}c@{}c@{}}
\makebox[0.245\linewidth][c]{\footnotesize Relight-GT} &
\makebox[0.245\linewidth][c]{\footnotesize Ours} &
\makebox[0.245\linewidth][c]{\footnotesize N.-PBIR} &
\makebox[0.245\linewidth][c]{\footnotesize M. Fusion}
\end{tabular}}
{
\renewcommand{\normalsize}{\footnotesize}
\renewcommand{\small}{\footnotesize}
\begin{tabular}{@{}c@{\hspace{0.01\linewidth}}c@{}}
  \begin{minipage}[t]{0.49\linewidth}\centering
\begin{tabular}{@{}c@{}c@{}c@{}c@{}}
  \multicolumn{4}{l}{\normalsize\texttt{\detokenize{chair}}} \\
  \includegraphics[width=0.245\linewidth]{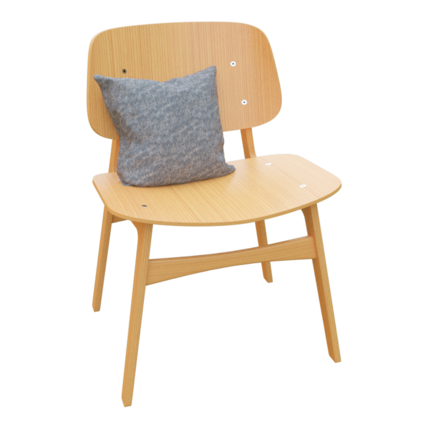} & \includegraphics[width=0.245\linewidth]{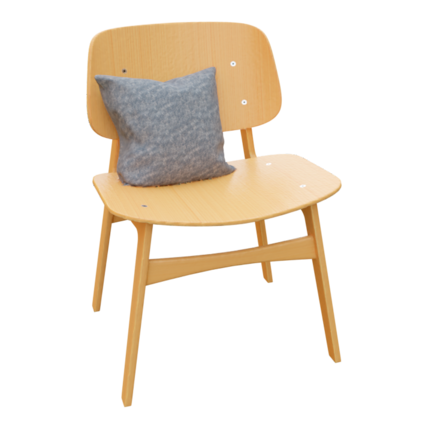} & \includegraphics[width=0.245\linewidth]{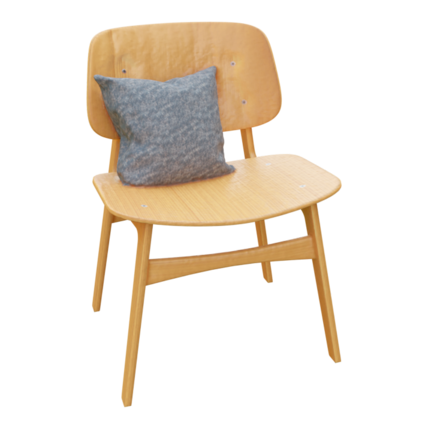} & \includegraphics[width=0.245\linewidth]{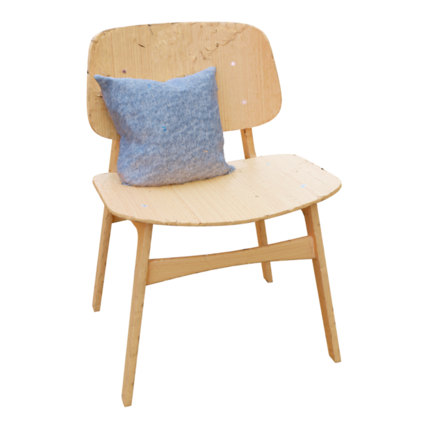}
  
\end{tabular}
\end{minipage} & \begin{minipage}[t]{0.49\linewidth}\centering
\begin{tabular}{@{}c@{}c@{}c@{}c@{}}
  \multicolumn{4}{l}{\normalsize\texttt{\detokenize{air_baloons}}} \\
  \includegraphics[width=0.245\linewidth]{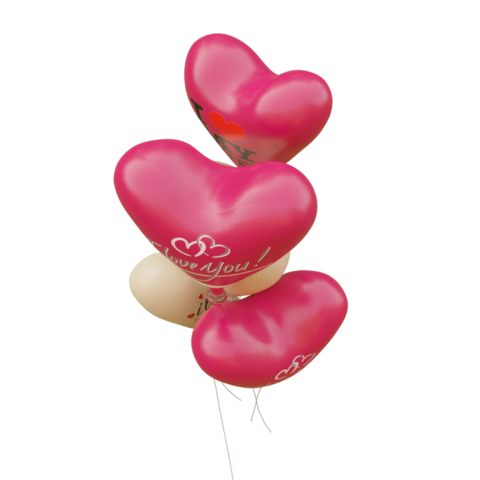} & \includegraphics[width=0.245\linewidth]{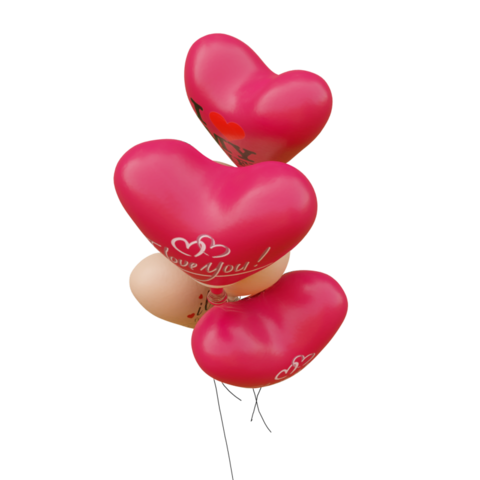} & \includegraphics[width=0.245\linewidth]{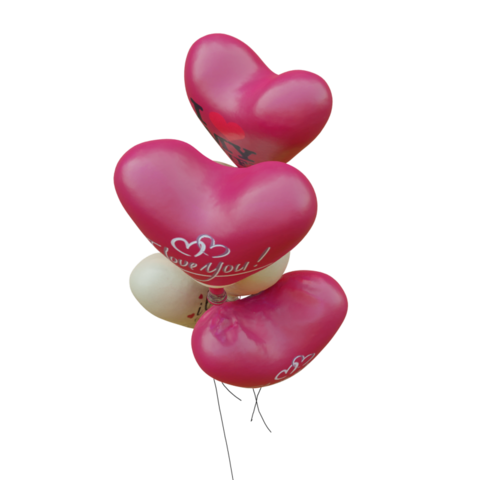} & \includegraphics[width=0.245\linewidth]{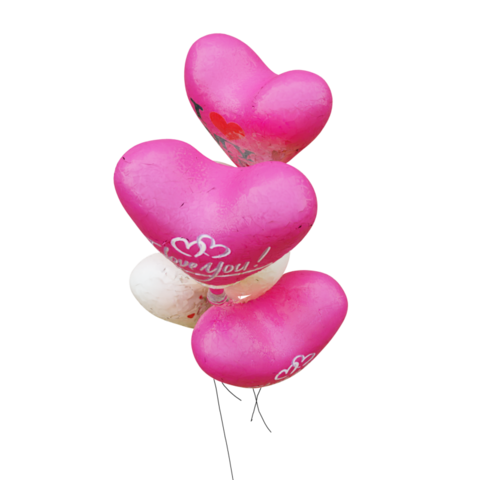}
  
\end{tabular}
\end{minipage}
\end{tabular}
\par\smallskip
\begin{tabular}{@{}c@{\hspace{0.01\linewidth}}c@{}}
  \begin{minipage}[t]{0.49\linewidth}\centering
\begin{tabular}{@{}c@{}c@{}c@{}c@{}}
  \multicolumn{4}{l}{\normalsize\texttt{\detokenize{hotdog}}} \\
  \includegraphics[width=0.245\linewidth]{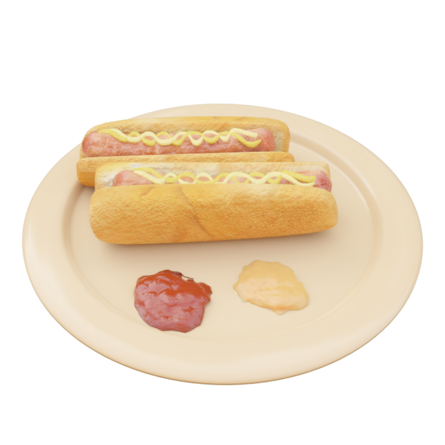} & \includegraphics[width=0.245\linewidth]{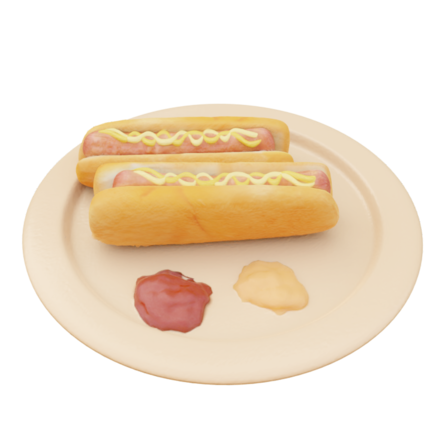} & \includegraphics[width=0.245\linewidth]{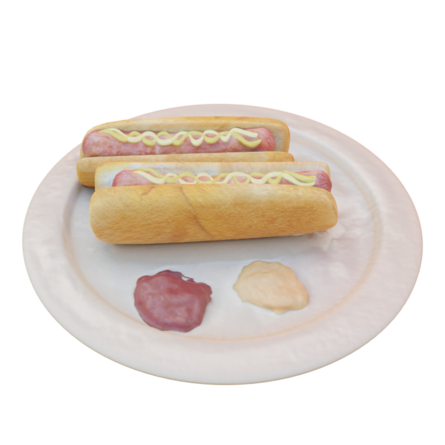} & \includegraphics[width=0.245\linewidth]{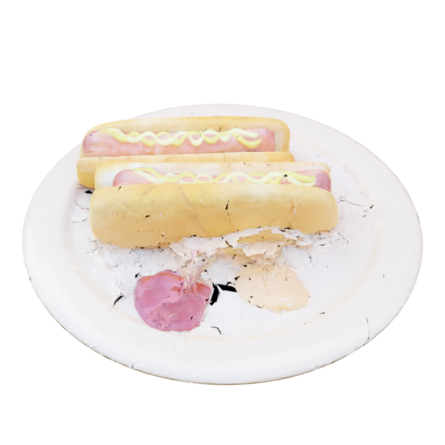}
  
\end{tabular}
\eccvpanelcaps\end{minipage} & \begin{minipage}[t]{0.49\linewidth}\centering
\begin{tabular}{@{}c@{}c@{}c@{}c@{}}
  \multicolumn{4}{l}{\normalsize\texttt{\detokenize{jugs}}} \\
  \includegraphics[width=0.245\linewidth]{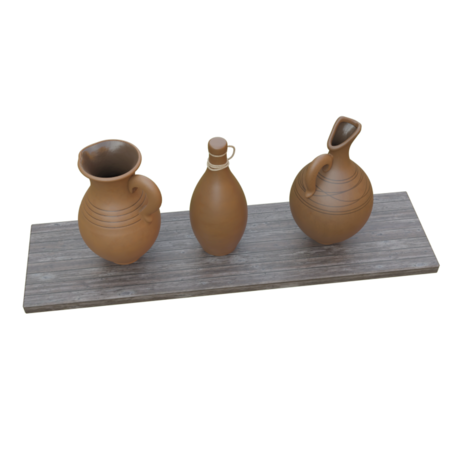} & \includegraphics[width=0.245\linewidth]{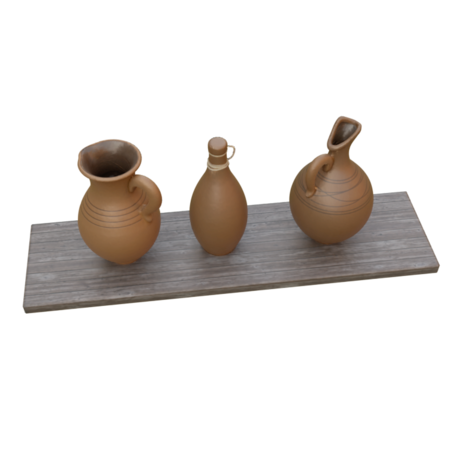} & \includegraphics[width=0.245\linewidth]{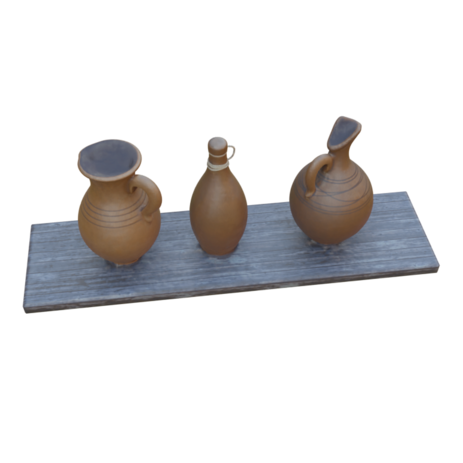} & \includegraphics[width=0.245\linewidth]{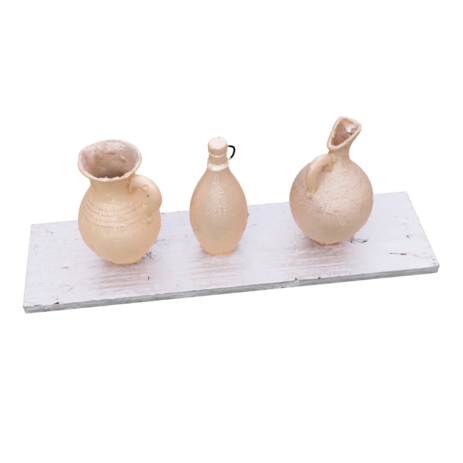}
  
\end{tabular}
\eccvpanelcaps\end{minipage}
\end{tabular}
}

  \caption{Qualitative comparison of relighting on the Synthetic4Relight dataset~\cite{DBLP:conf/cvpr/ZhangSHFJZ22} between our method, Neural-PBIR~\cite{DBLP:conf/iccv/0004CLYZM0Z023}, and MaterialFusion~\cite{DBLP:conf/3dim/LitmanPDAZTT25}.}
  \label{fig:relight-summary-mii}
\end{figure}

\subsection{Comparison}
\label{sec:comparison}
We compare our method against the following approaches:
\emph{Neural-PBIR}~\cite{DBLP:conf/iccv/0004CLYZM0Z023}, as a representative of purely analysis-by-synthesis inverse rendering that optimizes shape, spatially varying materials, and illumination without learned priors, and \emph{MaterialFusion}~\cite{DBLP:conf/3dim/LitmanPDAZTT25}, as a representative of diffusion-prior-aided inverse rendering via score distillation~\cite{DBLP:conf/iclr/ZhuZK24}.
We use the official implementations and follow the recommended hyperparameters.
We also test several previous regularization techniques on our pipeline; please see \cref{sec:alternative_methods}.

We show a qualitative comparison of novel-scene relighting on the Stanford-ORB and DTC-Synthetic datasets in \cref{fig:relight-curated-comparison-teaser,fig:relight-curated-comparison}.

\begin{table}[t]
\centering
\caption{Quantitative results of our method and baselines on the Stanford-ORB dataset.}
\label{tab:stanford-orb}
{\scriptsize\setlength{\tabcolsep}{1.8pt}\resizebox{\textwidth}{!}{\begin{tabular}{@{}l*{11}{c}@{}}
\toprule
  & \multicolumn{3}{c}{Geometry} & \multicolumn{4}{c}{Novel Scene Relighting} & \multicolumn{4}{c}{Novel View Synthesis} \\
\cmidrule(lr){2-4}
\cmidrule(lr){5-8}
\cmidrule(lr){9-12}
  & \shortstack{Depth\\$\downarrow$} & \shortstack{Normal\\$\downarrow$} & \shortstack{Shape\\$\downarrow$} & \shortstack{PSNR-H\\$\uparrow$} & \shortstack{PSNR-L\\$\uparrow$} & \shortstack{SSIM\\$\uparrow$} & \shortstack{LPIPS\\$\downarrow$} & \shortstack{PSNR-H\\$\uparrow$} & \shortstack{PSNR-L\\$\uparrow$} & \shortstack{SSIM\\$\uparrow$} & \shortstack{LPIPS\\$\downarrow$} \\
\midrule
  Neural-PBIR & 0.32 & 0.061 & \underline{0.42} & \underline{26.07} & \underline{33.39} & \underline{0.980} & \underline{0.023} & \underline{28.90} & \underline{36.92} & \underline{0.987} & \underline{0.019} \\
  MaterialFusion & \underline{0.30} & \underline{0.044} & 1.08 & 23.52 & 31.18 & 0.968 & 0.037 & 26.17 & 34.67 & 0.975 & 0.032 \\
  Ours & \textbf{0.25} & \textbf{0.014} & \textbf{0.30} & \textbf{27.22} & \textbf{34.98} & \textbf{0.981} & \textbf{0.021} & \textbf{29.58} & \textbf{38.46} & \textbf{0.987} & \textbf{0.017} \\
\bottomrule
\end{tabular}
}}
\end{table}

\begin{table}[t]
    \centering
    \caption{Quantitative results of our method and baselines on the Synthetic4Relight and DTC-Synthetic datasets.}
    \label{tab:synthetic4relight}
    {\scriptsize\setlength{\tabcolsep}{1.3pt}\resizebox{\textwidth}{!}{\begin{tabular}{@{}l*{13}{c}@{}}
\toprule
  & \multicolumn{10}{c}{Synthetic4Relight} & \multicolumn{3}{c}{DTC-Synthetic} \\
\cmidrule(lr){2-11}
\cmidrule(lr){12-14}
  & \multicolumn{3}{c}{Relighting} & \multicolumn{3}{c}{Aligned albedo} & \multicolumn{3}{c}{Albedo} & \multicolumn{1}{c}{Rough.} & \multicolumn{3}{c}{Relighting} \\
\cmidrule(lr){2-4}
\cmidrule(lr){5-7}
\cmidrule(lr){8-10}
\cmidrule(lr){11-11}
\cmidrule(lr){12-14}
  Method & \shortstack{PSNR\\$\uparrow$} & \shortstack{SSIM\\$\uparrow$} & \shortstack{LPIPS\\$\downarrow$} & \shortstack{PSNR\\$\uparrow$} & \shortstack{SSIM\\$\uparrow$} & \shortstack{LPIPS\\$\downarrow$} & \shortstack{PSNR\\$\uparrow$} & \shortstack{SSIM\\$\uparrow$} & \shortstack{LPIPS\\$\downarrow$} & \shortstack{MSE\\$\downarrow$} & \shortstack{PSNR\\$\uparrow$} & \shortstack{SSIM\\$\uparrow$} & \shortstack{LPIPS\\$\downarrow$} \\
\midrule
  Neural-PBIR & 27.83 & \underline{0.974} & 0.058 & 25.72 & 0.951 & 0.074 & 22.67 & 0.951 & 0.091 & 0.016 & 39.18 & 0.9928 & 0.0100 \\
  MaterialFusion & 20.20 & 0.944 & 0.091 & 24.28 & 0.952 & 0.070 & 15.65 & 0.931 & 0.094 & \underline{0.015} & 28.63 & 0.9746 & 0.0237 \\
  Ours & \textbf{32.02} & \textbf{0.975} & \textbf{0.049} & \textbf{27.83} & \textbf{0.960} & \textbf{0.045} & \textbf{27.04} & \textbf{0.960} & \textbf{0.048} & \textbf{0.013} & \textbf{43.21} & \textbf{0.9961} & \textbf{0.0063} \\
  Ours (w/o reg.) & \underline{28.94} & 0.973 & \underline{0.055} & \underline{26.63} & \underline{0.956} & \underline{0.065} & \underline{23.79} & \underline{0.954} & \underline{0.084} & 0.021 & \underline{40.23} & \underline{0.9954} & \underline{0.0065} \\
\bottomrule
\end{tabular}
}}
  \end{table}

Our method is robust to strong directional lighting in the input images, where hard cast shadows induce high-contrast appearance variation across an object. In this setting, purely analysis-by-synthesis optimization (e.g., Neural-PBIR) can absorb shadowing into the estimated materials, yielding baked artifacts that persist under relighting. By grouping observations within likely same-material regions, our regularization effectively reduces the degrees of freedom in spatially varying materials and discourages fitting cast shadows as texture; as a result, we remove baked shadows on the side of \casename{car_scene002}, on the red cube in \casename{Block_RedBlue}, and in the cast-shadow pattern around the cup handle in \casename{cup_scene006}, which are typically visible in the Neural-PBIR results.
This same-material grouping also improves material estimation. We recover more accurate roughness, as reflected by specular highlights that better match the reference relighting images in \casename{ball_scene003}, \casename{blocks_scene005}, on the flower pot of \casename{cactus_scene005}, and on the top of \casename{chips_scene003}. We also recover more accurate metallic parameters, reproducing the shiny metallic appearance during relighting on top of \casename{baking_scene003}, in \casename{pitcher_scene001} and \casename{pepsi_scene004}, and on top of \casename{TeaPot_EmeraldGoldTop}. With the help of our regularization on normals, our method further improves surface shape and mitigates concave-shape artifacts.

\begin{figure}[!t]
  \centering
\begin{tabular}{@{}c@{}}
  \begin{minipage}[t]{1.0\linewidth}\centering
\begin{tabular}{@{}c@{}c@{}c@{}c@{}c@{}c@{}}
  \multicolumn{6}{l}{\footnotesize\texttt{\detokenize{Block_RedBlue}}} \\
  \includegraphics[width=0.164\linewidth]{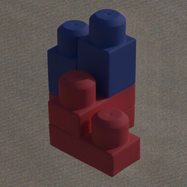} & \includegraphics[width=0.164\linewidth]{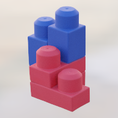} & \includegraphics[width=0.164\linewidth]{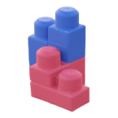} & \includegraphics[width=0.164\linewidth]{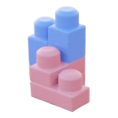} & \includegraphics[width=0.164\linewidth]{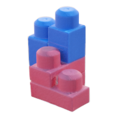} & \includegraphics[width=0.164\linewidth]{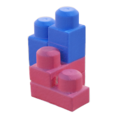}
\end{tabular}
\end{minipage}
\end{tabular}
\par\smallskip
\begin{tabular}{@{}c@{}}
  \begin{minipage}[t]{1.0\linewidth}\centering
\begin{tabular}{@{}c@{}c@{}c@{}c@{}c@{}c@{}}
  \multicolumn{6}{l}{\footnotesize\texttt{\detokenize{grogu_scene003}}} \\
  \includegraphics[width=0.164\linewidth]{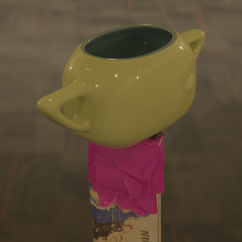} & \includegraphics[width=0.164\linewidth]{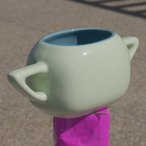} & \includegraphics[width=0.164\linewidth]{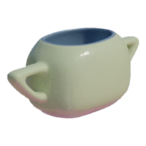} & \includegraphics[width=0.164\linewidth]{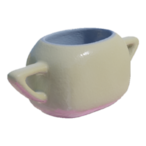} & \includegraphics[width=0.164\linewidth]{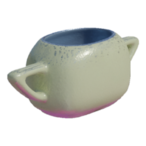} & \includegraphics[width=0.164\linewidth]{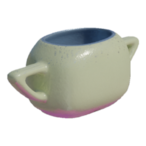} \\
  \footnotesize Input & \footnotesize Relight-GT & \footnotesize Ours & \footnotesize Diffusion-BP & \footnotesize w/o reg. & \footnotesize d-s corr.
  
\end{tabular}
\end{minipage}
\end{tabular}

  \caption{Qualitative comparison of relighting between our method and alternative material regularization strategies, including directly back-projecting DiffusionRenderer predictions (Diffusion-BP), optimizing from this initialization without regularization (w/o reg.), and using a non–data-driven albedo–specular correlation regularizer (d-s corr.).}
  \label{fig:relight-curated-ablation}
\end{figure}

\begin{figure}[t]
  \centering
\begin{tabular}{@{}c@{\hspace{0.01\linewidth}}c@{}}
  \begin{minipage}[t]{0.49\linewidth}\centering
\begin{tabular}{@{}c@{}c@{}c@{}c@{}}
  \multicolumn{4}{l}{\scriptsize\texttt{\detokenize{cup_scene007}}} \\
  \includegraphics[width=0.235\linewidth]{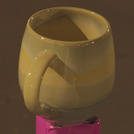} & \includegraphics[width=0.235\linewidth]{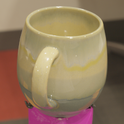} & \includegraphics[width=0.235\linewidth]{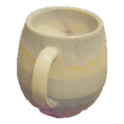} & \includegraphics[width=0.235\linewidth]{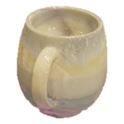} \\
  \scriptsize Input & \scriptsize Relight-GT & \scriptsize Ours & \scriptsize scale inv.
\end{tabular}
\end{minipage} &
  \begin{minipage}[t]{0.49\linewidth}\centering
\begin{tabular}{@{}c@{}c@{}c@{}c@{}}
  \multicolumn{4}{l}{\scriptsize\texttt{\detokenize{curry_scene001}}} \\
  \includegraphics[width=0.235\linewidth]{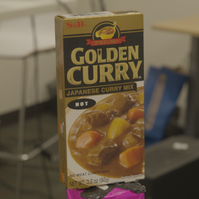} & \includegraphics[width=0.235\linewidth]{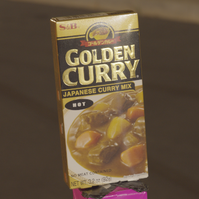} & \includegraphics[width=0.235\linewidth]{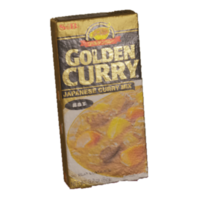} & \includegraphics[width=0.235\linewidth]{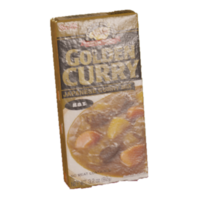} \\
  \scriptsize Input & \scriptsize Relight-GT & \scriptsize Ours & \scriptsize scale inv.
\end{tabular}
\end{minipage}
\end{tabular}

  \caption{Qualitative relighting comparison between our method and the global scale-invariant loss from VideoMat~\cite{DBLP:journals/cgf/MunkbergWLSH25}.}
  \label{fig:relight-ablation-stanford-orb-scale-invariant-curated}
\end{figure}

MaterialFusion is based on a different shape reconstruction pipeline \cite{DBLP:conf/nips/HasselgrenHM22}, so shape quality is not directly comparable. It often eliminates baked-in shadows but tends to underfit the input images, failing to reconstruct spatially varying color patterns in \casename{cup_scene006} and on the side of \casename{baking_scene003}, yielding over-smoothed materials, less accurate relighting, and occasional color drift. We attribute these failures to the score-distillation target not always aligning with the input scene and the lack of an explicit mechanism to correct per-region inaccuracies, which can cause over-regularization.

We show a qualitative comparison of relighting on the Synthetic4Relight dataset in \cref{fig:relight-summary-mii}.
Compared to Neural-PBIR, our method avoids baking on the \casename{chair} back and the \casename{hotdog} plate. Our method also achieves a more accurate overall color tone and produces more accurate reflections on \casename{air_baloons} and the inside wall of \casename{jugs}. MaterialFusion tends to cause color shifts in relighting on \casename{air_baloons} and \casename{jugs}.

We present the quantitative comparison results on Stanford-ORB in \cref{tab:stanford-orb} and on Synthetic4Relight and DTC-Synthetic in \cref{tab:synthetic4relight}.

Our method outperforms the baselines on relighting accuracy as well as albedo/roughness estimation and shape reconstruction. As Neural-PBIR is the current state of the art on Stanford-ORB, surpassing it sets a new state of the art on this dataset by a notable margin.

\begin{table}[t]
\centering
\makebox[0.58\textwidth][c]{%
\begin{minipage}[c]{0.57\textwidth}
  \centering
  \caption{Stanford-ORB relighting comparison: (a) alternative regularization choices and (b) previous-method baselines.}
  \label{tab:stanford-orb-additional}
  \label{tab:stanford-orb-previous-methods}
  {\small\setlength{\tabcolsep}{2.0pt}\resizebox{\linewidth}{!}{\begin{tabular}{@{}lcccc@{}}
\toprule
Method & PSNR-H$\uparrow$ & PSNR-L$\uparrow$ & SSIM$\uparrow$ & LPIPS$\downarrow$ \\
\midrule
\multicolumn{5}{@{}l}{\textit{(a) Alternative regularization choices}} \\
Diffusion-BP & 26.06 & 33.09 & 0.976 & 0.0333 \\
w/o reg. & 26.11 & 33.78 & 0.978 & 0.0223 \\
d-s corr. & 26.37 & 34.10 & 0.979 & 0.0215 \\
scale inv. & 26.38 & 34.21 & 0.980 & 0.0225 \\
Ours & \textbf{27.22} & \textbf{34.98} & \textbf{0.981} & \textbf{0.0213} \\
\midrule
\multicolumn{5}{@{}l}{\textit{(b) Comparison with previous methods}} \\
NVDiffRecMC~\cite{DBLP:conf/nips/HasselgrenHM22} & 24.43 & 31.60 & 0.972 & 0.0360 \\
R3DG~\cite{DBLP:conf/eccv/GaoGLLZCZY24} & 23.40 & 30.96 & 0.968 & 0.0348 \\
Material Anything~\cite{DBLP:conf/cvpr/HuangWLW25} & 24.46 & 31.40 & 0.968 & 0.0387 \\
Neural Gaffer~\cite{DBLP:conf/nips/JinLLXBZX0S24} & 24.32 & 31.46 & 0.963 & 0.0447 \\
IllumiNeRF~\cite{DBLP:conf/nips/ZhaoSVPMH24} & 25.56 & 32.74 & 0.976 & 0.0270 \\
Vanilla Mitsuba~\cite{DBLP:journals/tog/JakobSRV22} & 25.62 & 33.16 & 0.977 & 0.0232 \\
\bottomrule
\end{tabular}
}}
\end{minipage}
}%
\makebox[0.42\textwidth][c]{%
\begin{minipage}[c]{0.41\textwidth}
  \centering
  \resizebox{\linewidth}{!}{\begin{tabular}{@{}c@{\hspace{0.01\linewidth}}c@{\hspace{0.01\linewidth}}c@{}}
  \includegraphics[width=0.327\linewidth]{gt_normal.png} & \includegraphics[width=0.327\linewidth]{ours_normal.png} & \includegraphics[width=0.327\linewidth]{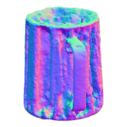} \\
  \normalsize GT & \normalsize Ours & \normalsize w/o normal reg.
  
\end{tabular}
}
  \refstepcounter{figure}
  \label{fig:normal-pitcher-scene001}
  {\small\textbf{Fig.~\thefigure:} Ablation study on the normal-supervision loss (\cref{eq:Lshape}).\par}
  \vspace{0.4em}
  \resizebox{\linewidth}{!}{\begin{tabular}{@{}c@{\hspace{0.01\linewidth}}c@{\hspace{0.01\linewidth}}c@{}}
  \includegraphics[width=0.327\linewidth]{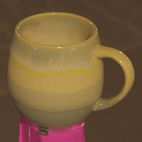} & \includegraphics[width=0.327\linewidth]{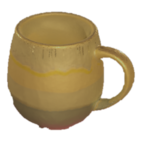} & \includegraphics[width=0.327\linewidth]{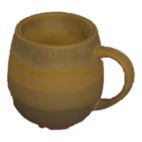} \\
  \\
  \includegraphics[width=0.327\linewidth]{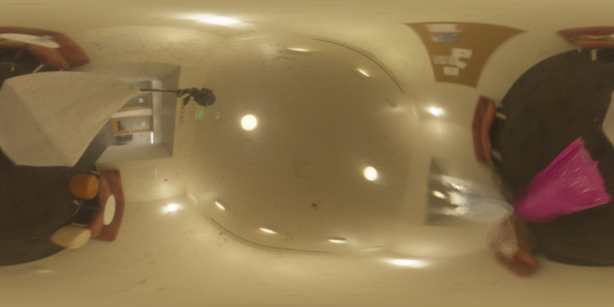} & \includegraphics[width=0.327\linewidth]{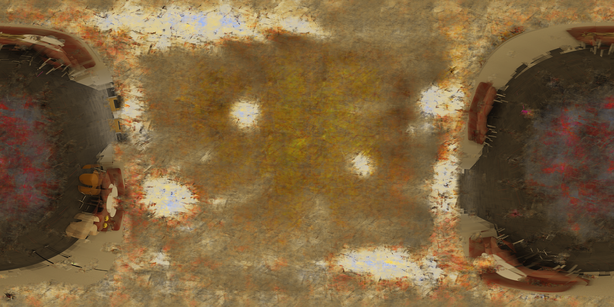} & \includegraphics[width=0.327\linewidth]{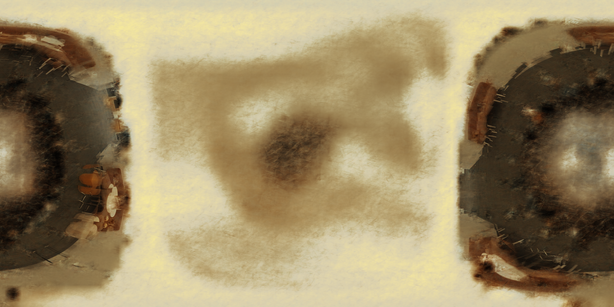} \\
  \normalsize Relight-GT & \normalsize Ours & \normalsize No albedo trans.
  
\end{tabular}
}
  \refstepcounter{figure}
  \label{fig:ablation_albedo_transformation}
  {\small\textbf{Fig.~\thefigure:} Ablating scale-agnostic albedo transformation. Rows show relighting and environment maps.\par}
\end{minipage}
}%
\end{table}

\subsection{Exploring Alternative Choices}
\label{sec:alternative_methods}

We evaluate alternative material-regularization designs against our implicit material clustering regularizer. As a first baseline, we directly back-project our predicted material maps onto our reconstructed shape. The resulting materials are visually plausible but do not match the relighting ground truth (\cref{fig:relight-curated-ablation}), motivating an analysis-by-synthesis stage that enforces consistency between rendered and observed images. However, simply initializing the optimization with these maps and optimizing without our regularization quickly reintroduces baking artifacts, such as the cast-shadow pattern in \casename{Block_RedBlue} and dotted specular highlights in \casename{grogu_scene003} (the w/o reg case in \cref{fig:relight-curated-ablation}), indicating that the initialization remains far from convergence and the optimization can fall into local minima that trade off image fit against material coherence.

We also compare against a global scale-invariant loss, in the spirit of VideoMat~\cite{DBLP:journals/cgf/MunkbergWLSH25} and IntrinsicAnything~\cite{DBLP:conf/eccv/ChenPYLPLZ24}, implemented following VideoMat with our predicted material maps as guidance (\cref{fig:relight-ablation-stanford-orb-scale-invariant-curated}). A global adjustment is less effective at suppressing baked-in shadows (e.g., \casename{cup_scene007}) and can underfit localized material details (e.g., the metallic foil lettering in \casename{curry_scene001}), since it cannot accommodate region-specific corrections.

Finally, we evaluate a non-data-driven diffuse-specular self-correlation regularizer~\cite{DBLP:journals/cgf/LuanZBD21}. We construct the correlation signal from the current optimized base color and use it to regularize the metallic and roughness buffers. However, when the base color already contains baked artifacts, the regularizer can propagate these errors and become less effective (\cref{fig:relight-curated-ablation}, denoted as d-s corr.), as evidenced by cast shadows in \casename{Block_RedBlue} (top row) and dotted artifacts in \casename{grogu_scene003} (bottom row).

The same trends hold quantitatively on Stanford-ORB: all alternative choices in \cref{tab:stanford-orb-additional}(a) underperform our implicit material clustering regularizer in relighting accuracy.

\subsection{Ablation Studies}

We evaluate several components of our method in this section.
In \cref{fig:normal-pitcher-scene001}, we compare reconstructions with and without our regularization on normals. Without it, the reconstructed surface can exhibit concave artifacts, especially on glossy surfaces, which in turn introduce severe highlight-related artifacts. Adding the normal regularization yields a more faithful shape that better matches the reference.
We ablate our scale-agnostic albedo transformation in \cref{fig:ablation_albedo_transformation}.
In the first row, we show the relighting reference and results, and in the second row, we show the reference and reconstructed environment map of the original lighting condition.
Without the transformation, the optimization tends to scale down base color to reduce the regularization, which is compensated by a brighter and more spatially spread-out lighting estimate (second row), deviating from the reference lighting (left) and degrading reconstruction quality.

Finally, \cref{tab:synthetic4relight} reports a quantitative ablation on Synthetic4Relight and DTC-Synthetic. Overall, the improvement of our full system comes from three parts: our improved inverse-rendering pipeline, our regularization on normals, and our material regularization. The ablation isolates the last part: removing our material regularization consistently reduces relighting accuracy across both datasets.

\begin{figure}[t]
\centering
\begingroup
\graphicspath{{figures/additional/stanford_orb_previous_methods_cases/}}
\begin{tabular}{@{}c@{\hspace{0.003\linewidth}}c@{\hspace{0.003\linewidth}}c@{\hspace{0.003\linewidth}}c@{\hspace{0.003\linewidth}}c@{\hspace{0.003\linewidth}}c@{\hspace{0.003\linewidth}}c@{}}
  \multicolumn{7}{l}{\normalsize\texttt{\detokenize{Stanford-ORB grogu_scene001}}} \\
  \includegraphics[width=0.136\linewidth]{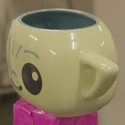} & \includegraphics[width=0.136\linewidth]{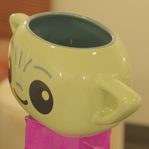} & \includegraphics[width=0.136\linewidth]{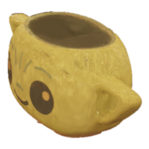} & \includegraphics[width=0.136\linewidth]{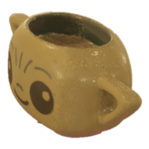} & \includegraphics[width=0.136\linewidth]{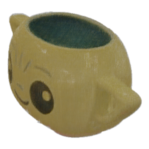} & \includegraphics[width=0.136\linewidth]{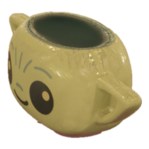} & \includegraphics[width=0.136\linewidth]{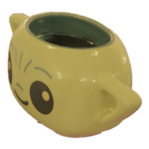} \\[-0.2em]
  \multicolumn{7}{l}{\normalsize\texttt{\detokenize{Stanford-ORB pepsi_scene002}}} \\
  \includegraphics[width=0.136\linewidth]{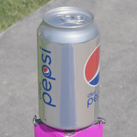} & \includegraphics[width=0.136\linewidth]{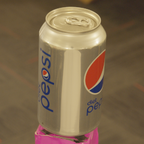} & \includegraphics[width=0.136\linewidth]{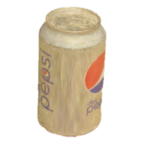} & \includegraphics[width=0.136\linewidth]{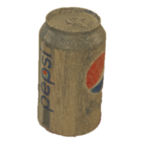} & \includegraphics[width=0.136\linewidth]{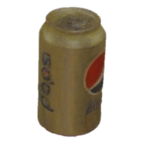} & \includegraphics[width=0.136\linewidth]{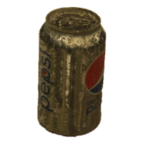} & \includegraphics[width=0.136\linewidth]{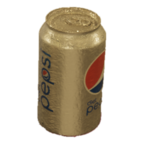} \\
  \footnotesize Input & \footnotesize Relight-GT & \footnotesize R3DG & \begin{tabular}{@{}c@{}}\footnotesize Material\\[-0.25em]\footnotesize Anything\end{tabular} & \begin{tabular}{@{}c@{}}\footnotesize Neural\\[-0.25em]\footnotesize Gaffer\end{tabular} & \begin{tabular}{@{}c@{}}\footnotesize Vanilla\\[-0.25em]\footnotesize Mitsuba\end{tabular} & \footnotesize Ours
\end{tabular}

\endgroup
\caption{Additional qualitative Stanford-ORB relighting results grouped by case. Each row shows the shared input and relighting target once, followed by previous-method baselines and our result.}
\label{fig:stanford-orb-additional-qualitative}
\label{fig:stanford-orb-previous-methods}
\end{figure}


\subsection{Additional Relighting Comparisons}
\label{sec:more_previous_methods}

We further compare with additional relighting and inverse-rendering baselines on Stanford-ORB: NVDiffRecMC~\cite{DBLP:conf/nips/HasselgrenHM22}, R3DG~\cite{DBLP:conf/eccv/GaoGLLZCZY24}, Material Anything~\cite{DBLP:conf/cvpr/HuangWLW25}, Neural Gaffer~\cite{DBLP:conf/nips/JinLLXBZX0S24}, IllumiNeRF~\cite{DBLP:conf/nips/ZhaoSVPMH24}, and a vanilla Mitsuba inverse-rendering baseline~\cite{DBLP:journals/tog/JakobSRV22}.
Unlike the main comparison with Neural-PBIR and MaterialFusion, these methods do not all reconstruct comparable shape and PBR assets; some directly perform relighting or use representations outside the mesh-based setting, so we report only held-out relighting quality.

As shown in \cref{tab:stanford-orb-previous-methods}(b) and \cref{fig:stanford-orb-previous-methods}, our method achieves the best relighting accuracy.
The vanilla Mitsuba baseline shows that a physically based renderer alone is insufficient under sparse-view inverse rendering, where material-lighting ambiguities cause baked-in appearance.
Unlike image-based relighting methods such as Neural Gaffer and IllumiNeRF, our method also produces relightable PBR assets while improving held-out relighting metrics.

\FloatBarrier

\section{Discussion and Conclusion}
\label{sec:conclusion}
\paragraph{Limitations and future work.}
Our approach assumes DiffusionRenderer predicts material maps that are locally consistent within same-material regions. When this breaks down---e.g., for rare appearances outside the training distribution---the prior becomes unreliable and can over-regularize, blurring materials. DiffusionRenderer also operates at limited resolution, so matching its predictions leaves high-frequency details less constrained; higher-resolution diffusion models would address this. Finally, we derive our similarity kernel from a concatenation of material channels; alternative combinations may prove preferable as diffusion models evolve. Residual artifacts also remain near shadow boundaries, which we attribute to imperfect geometry; future work could refine shape using signals from our regularization to better align it with cast-shadow boundaries.

\paragraph{Conclusion.}
We presented an end-to-end pipeline that jointly recovers geometry, spatially varying materials, and illumination from multi-view images. Rather than replacing physics-based optimization with a data-driven predictor, we bridge the two: we treat DiffusionRenderer's predictions not as target material values but as a similarity kernel, penalizing deviations in the optimized material over regions where the predictions are near-constant while still fitting the input images. This suppresses baking artifacts and yields relightable assets, significantly outperforming state-of-the-art baselines on the Synthetic4Relight, Stanford-ORB, and DTC-Synthetic datasets.

\section*{Acknowledgments}
We thank the anonymous reviewers for their feedback.
This work was partially supported by NSF grant 2553564.
This work used NCSA Delta [award OAC 2005572] through ACCESS allocation CIS260003, supported by NSF grants \#2138259, \#2138286, \#2138307, \#2137603, and \#2138296.

\clearpage
\bibliographystyle{splncs04}
\bibliography{sample-base}

\clearpage
\appendix
\begin{center}
  {\Large\bfseries Supplementary Material}
\end{center}

This supplementary material includes an ablation of our similarity-kernel design in \cref{sec:supp-split-channel-guidance-mii}, a discussion of our practical use of DiffusionRenderer in \cref{sec:supp-diffrenderer-video-vs-image}, and additional implementation details in \cref{sec:impl_details_supp}.
\cref{sec:supp-rgbx-upstream} evaluates replacing DiffusionRenderer with RGB$\leftrightarrow$X as the upstream model.
\cref{sec:supp-irgs-jbf-ablation} evaluates the transfer of our material regularization to IRGS~\cite{DBLP:conf/cvpr/GuWZY025}, and \cref{sec:supp-intrinsic-gbuffer-comparison-ablation} provides extensive per-scene intrinsic visualizations for comparisons and ablations.

For dynamic relighting results and additional G-buffer visualizations, please also refer to the supplementary video.

\section{Ablating Concatenated vs.\ Split-Channel Similarity Kernel}
\label{sec:supp-split-channel-guidance-mii}

As described in the main paper (Sec.~3.1), our implicit material clustering regularization builds a single similarity kernel from the \emph{concatenated} predicted G-buffer $\mathbf{g} = [\gbufferalbedo, \gbufferrough, \gbuffermetal]$ and uses it to regularize all material channels jointly.
This design lets the kernel capture differences across all channels, reducing over-regularization when some predicted channels are overly smooth.

\begin{figure*}[t]
  \centering
  \newcommand{\intrinsicrowlabel}[1]{\rotatebox[origin=c]{90}{\scriptsize\strut #1}}
  {\scriptsize\texttt{\detokenize{Synthetic4Relight air_baloons (Split Channel Guidance Ablation)}}}\\[-0.2em]
  \parbox[c]{0.05\linewidth}{\centering \intrinsicrowlabel{Input and Prediction}}%
  \hspace{2pt}%
  \parbox[c]{0.172\linewidth}{\centering \includegraphics[width=\linewidth]{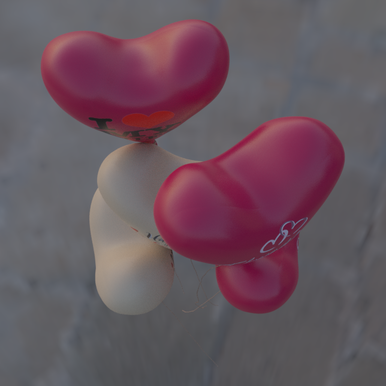}}%
  \hspace{2pt}%
  \parbox[c]{0.172\linewidth}{\centering \includegraphics[width=\linewidth]{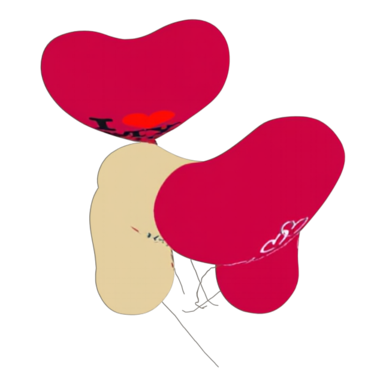}}%
  \hspace{2pt}%
  \parbox[c]{0.172\linewidth}{\centering \includegraphics[width=\linewidth]{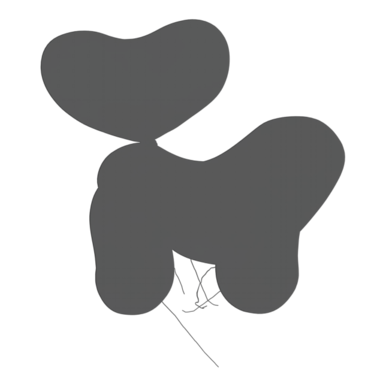}}%
  \hspace{2pt}%
  \parbox[c]{0.172\linewidth}{\centering \includegraphics[width=\linewidth]{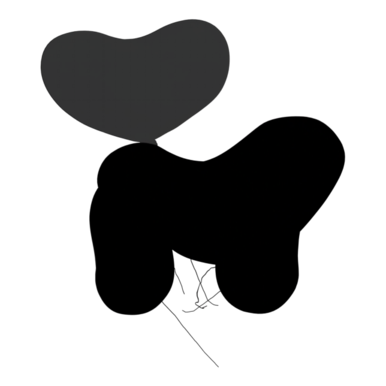}}%
  \hspace{2pt}%
  \parbox[c]{0.172\linewidth}{\centering \includegraphics[width=\linewidth]{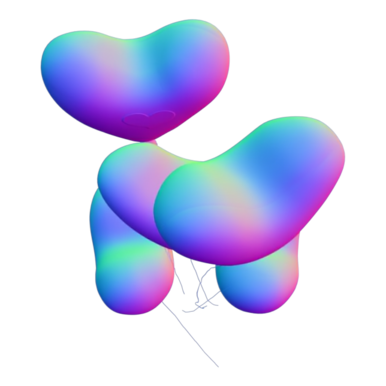}}%
  \\[1pt]
  \makebox[0.05\linewidth][c]{}%
  \hspace{2pt}%
  \makebox[0.172\linewidth][c]{\scriptsize Input}%
  \hspace{2pt}%
  \makebox[0.172\linewidth][c]{\scriptsize DR-base color}%
  \hspace{2pt}%
  \makebox[0.172\linewidth][c]{\scriptsize DR-roughness}%
  \hspace{2pt}%
  \makebox[0.172\linewidth][c]{\scriptsize DR-metallic}%
  \hspace{2pt}%
  \makebox[0.172\linewidth][c]{\scriptsize DR-normal}%
  \\[1pt]
  \vspace{0.15em}
  \par\noindent\rule{\linewidth}{0.35pt}
  \vspace{0.05em}
  \vspace{-0.6em}
  \parbox[c]{0.09\linewidth}{\centering \intrinsicrowlabel{Roughness}}%
  \hspace{2pt}%
  \parbox[c]{0.215\linewidth}{\centering \includegraphics[width=\linewidth]{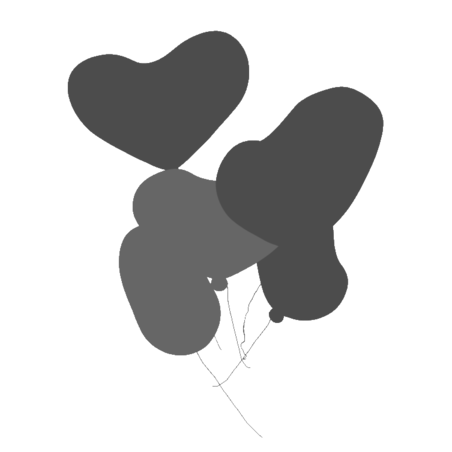}}%
  \hspace{2pt}%
  \parbox[c]{0.215\linewidth}{\centering \includegraphics[width=\linewidth]{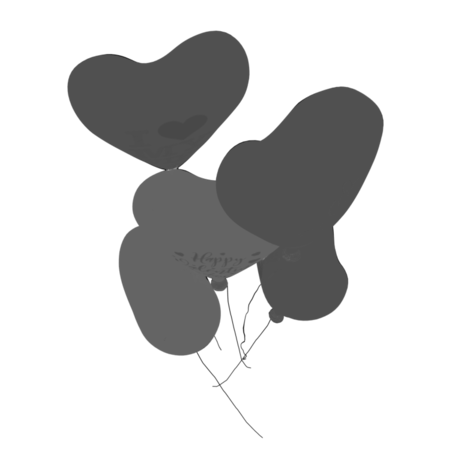}}%
  \hspace{2pt}%
  \parbox[c]{0.215\linewidth}{\centering \includegraphics[width=\linewidth]{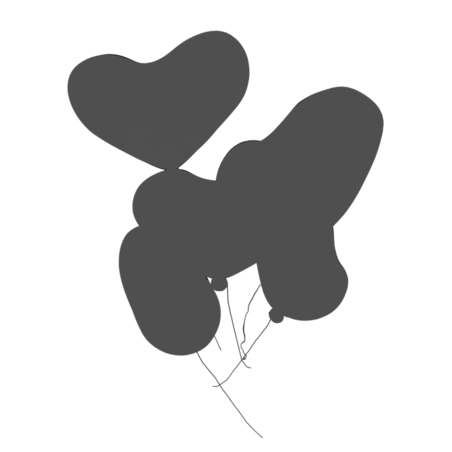}}%
  \\[1pt]
  \makebox[0.09\linewidth][c]{}%
  \hspace{2pt}%
  \makebox[0.215\linewidth][c]{\scriptsize Reference}%
  \hspace{2pt}%
  \makebox[0.215\linewidth][c]{\scriptsize Ours (Concatenation)}%
  \hspace{2pt}%
  \makebox[0.215\linewidth][c]{\scriptsize Split Channel Guidance}%
  \\[1pt]
  \caption{Synthetic4Relight \casename{air_baloons} split-channel-guidance ablation qualitative comparison (input view \texttt{012}, roughness view \texttt{000}).}
  \label{fig:supp-mii-intrinsics-ablation-split-channel-air-baloons}
\end{figure*}

A natural alternative is a \emph{split-channel} design: a separate kernel per channel, each built only from its own predicted values (i.e., base color guides base color, roughness guides roughness, metallic guides metallic).
We compare these two designs; quantitative results on Synthetic4Relight are shown in \cref{tab:supp-mii-split-channel-guidance} and qualitative results in \cref{fig:supp-mii-intrinsics-ablation-split-channel-air-baloons}.

\begin{table}[h]
  \centering
  \caption{Ablation of the similarity kernel design on Synthetic4Relight, averaged over 4 scenes (\texttt{air\_balloons}, \texttt{chair}, \texttt{hotdog}, \texttt{jugs}). \emph{Ours} uses a single kernel built from the concatenated G-buffer $\mathbf{g}=[\gbufferalbedo,\gbufferrough,\gbuffermetal]$; \emph{Split Channel} uses one kernel per material channel. Metric grouping follows Tab.~2 of the main paper.}
  \label{tab:supp-mii-split-channel-guidance}
  {\scriptsize\setlength{\tabcolsep}{1.8pt}
  \begin{tabular}{@{}l*{10}{c}@{}}
    \toprule
      & \multicolumn{10}{c}{Synthetic4Relight} \\
    \cmidrule(lr){2-11}
      & \multicolumn{3}{c}{Relighting} & \multicolumn{3}{c}{Aligned albedo} & \multicolumn{3}{c}{Albedo} & \multicolumn{1}{c}{Rough.} \\
    \cmidrule(lr){2-4}
    \cmidrule(lr){5-7}
    \cmidrule(lr){8-10}
    \cmidrule(lr){11-11}
      Method & \shortstack{PSNR\\$\uparrow$} & \shortstack{SSIM\\$\uparrow$} & \shortstack{LPIPS\\$\downarrow$} & \shortstack{PSNR\\$\uparrow$} & \shortstack{SSIM\\$\uparrow$} & \shortstack{LPIPS\\$\downarrow$} & \shortstack{PSNR\\$\uparrow$} & \shortstack{SSIM\\$\uparrow$} & \shortstack{LPIPS\\$\downarrow$} & \shortstack{MSE\\$\downarrow$} \\
    \midrule
    Ours (Concatenation) & \textbf{32.26} & 0.975 & \textbf{0.049} & \textbf{27.90} & \textbf{0.960} & \textbf{0.045} & \textbf{27.15} & \textbf{0.960} & \textbf{0.048} & \textbf{0.013} \\
    Split Channel & 32.18 & \textbf{0.976} & 0.050 & 27.67 & \textbf{0.960} & 0.046 & 26.46 & 0.959 & 0.049 & 0.024 \\
    \bottomrule
  \end{tabular}}
\end{table}

\noindent The concatenated design outperforms the split-channel design on most metrics, with a notable advantage in albedo and roughness estimation.
This is further evidenced qualitatively in \cref{fig:supp-mii-intrinsics-ablation-split-channel-air-baloons}: the split-channel design yields a near-uniform roughness map that fails to distinguish the differing roughness across individual balloons, whereas our concatenated kernel recovers spatially varying roughness consistent with the reference.

\section{DiffusionRenderer: Image Mode vs.\ Video Mode}
\label{sec:supp-diffrenderer-video-vs-image}

DiffusionRenderer~\cite{DBLP:journals/corr/abs-2501-18590} supports two inference modes.
In \emph{image mode}, each view is processed independently, yielding sharp per-view G-buffer predictions.
In \emph{video mode}, views are processed jointly as a temporally coherent sequence, which improves cross-view consistency at the cost of reduced spatial sharpness.

Multi-view captures consist of sparse, unordered viewpoints rather than a continuous video, so video mode requires preprocessing to form a suitable sequence.
We first compute a Hamiltonian path over the input views using a weighted combination of camera-position and orientation distances, ordering views to minimize total traversal cost.
We then fit a rotation spline to produce smooth, continuous camera motion along this ordering, and fill the gaps between input views with NeRF-based novel-view synthesis for intermediate frames.
To handle sequences of arbitrary length, we apply temporal diffusion synchronization inspired by the spatial synchronization of \citet{DBLP:conf/nips/LeeKKS23}.

\begin{figure*}[t]
  \centering
  \newcommand{\intrinsicrowlabel}[1]{\rotatebox[origin=c]{90}{\scriptsize\strut #1}}
  {\scriptsize\texttt{\detokenize{Stanford-ORB car_scene002 (DiffusionRenderer mode)}}}\\[-0.2em]
  \parbox[c]{0.05\linewidth}{\centering \intrinsicrowlabel{View 1 Video}}%
  \hspace{2pt}%
  \parbox[c]{0.172\linewidth}{\centering \includegraphics[width=\linewidth]{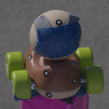}}%
  \hspace{2pt}%
  \parbox[c]{0.172\linewidth}{\centering \includegraphics[width=\linewidth]{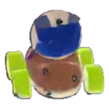}}%
  \hspace{2pt}%
  \parbox[c]{0.172\linewidth}{\centering \includegraphics[width=\linewidth]{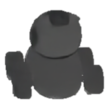}}%
  \hspace{2pt}%
  \parbox[c]{0.172\linewidth}{\centering \includegraphics[width=\linewidth]{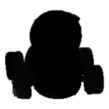}}%
  \hspace{2pt}%
  \parbox[c]{0.172\linewidth}{\centering \includegraphics[width=\linewidth]{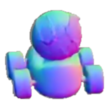}}%
  \\[1pt]
  \parbox[c]{0.05\linewidth}{\centering \intrinsicrowlabel{View 1 Image}}%
  \hspace{2pt}%
  \parbox[c]{0.172\linewidth}{\centering \includegraphics[width=\linewidth]{figures/summary/stanford_orb_car_scene002_diffrenderer_mode/assets/input_0027.png}}%
  \hspace{2pt}%
  \parbox[c]{0.172\linewidth}{\centering \includegraphics[width=\linewidth]{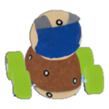}}%
  \hspace{2pt}%
  \parbox[c]{0.172\linewidth}{\centering \includegraphics[width=\linewidth]{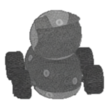}}%
  \hspace{2pt}%
  \parbox[c]{0.172\linewidth}{\centering \includegraphics[width=\linewidth]{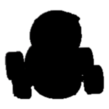}}%
  \hspace{2pt}%
  \parbox[c]{0.172\linewidth}{\centering \includegraphics[width=\linewidth]{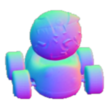}}%
  \\[1pt]
  \parbox[c]{0.05\linewidth}{\centering \intrinsicrowlabel{View 2 Video}}%
  \hspace{2pt}%
  \parbox[c]{0.172\linewidth}{\centering \includegraphics[width=\linewidth]{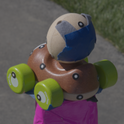}}%
  \hspace{2pt}%
  \parbox[c]{0.172\linewidth}{\centering \includegraphics[width=\linewidth]{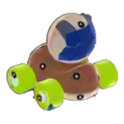}}%
  \hspace{2pt}%
  \parbox[c]{0.172\linewidth}{\centering \includegraphics[width=\linewidth]{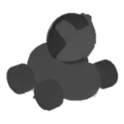}}%
  \hspace{2pt}%
  \parbox[c]{0.172\linewidth}{\centering \includegraphics[width=\linewidth]{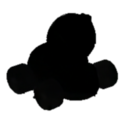}}%
  \hspace{2pt}%
  \parbox[c]{0.172\linewidth}{\centering \includegraphics[width=\linewidth]{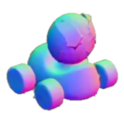}}%
  \\[1pt]
  \parbox[c]{0.05\linewidth}{\centering \intrinsicrowlabel{View 2 Image}}%
  \hspace{2pt}%
  \parbox[c]{0.172\linewidth}{\centering \includegraphics[width=\linewidth]{figures/summary/stanford_orb_car_scene002_diffrenderer_mode/assets/input_0029.png}}%
  \hspace{2pt}%
  \parbox[c]{0.172\linewidth}{\centering \includegraphics[width=\linewidth]{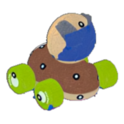}}%
  \hspace{2pt}%
  \parbox[c]{0.172\linewidth}{\centering \includegraphics[width=\linewidth]{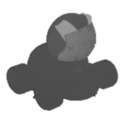}}%
  \hspace{2pt}%
  \parbox[c]{0.172\linewidth}{\centering \includegraphics[width=\linewidth]{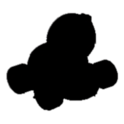}}%
  \hspace{2pt}%
  \parbox[c]{0.172\linewidth}{\centering \includegraphics[width=\linewidth]{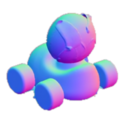}}%
  \\[1pt]
  \makebox[0.05\linewidth][c]{}%
  \hspace{2pt}%
  \makebox[0.172\linewidth][c]{\scriptsize Input}%
  \hspace{2pt}%
  \makebox[0.172\linewidth][c]{\scriptsize DR-base color}%
  \hspace{2pt}%
  \makebox[0.172\linewidth][c]{\scriptsize DR-roughness}%
  \hspace{2pt}%
  \makebox[0.172\linewidth][c]{\scriptsize DR-metallic}%
  \hspace{2pt}%
  \makebox[0.172\linewidth][c]{\scriptsize DR-normal}%
  \\[1pt]
  \caption{Comparison of DiffusionRenderer predictions in video mode and image mode on \casename{car_scene002} (Stanford-ORB), shown for two training views. Video mode (rows~1 and~3) produces more cross-view consistent G-buffers but at the cost of spatial sharpness; image mode (rows~2 and~4) yields sharper per-view predictions.}
  \label{fig:supp-stanford-car-scene002-diffrenderer-mode}
\end{figure*}

As shown in \cref{fig:supp-stanford-car-scene002-diffrenderer-mode}, video mode achieves stronger cross-view consistency but at the cost of spatial sharpness, whereas image mode produces crisper per-view details.
We therefore use image mode for our method and all baselines except Diffusion-BP, which uses video mode.
For our method, this choice is justified by the robustness of our regularization: as discussed in the main paper (Sec.~3.2), cross-view inconsistencies in the predicted G-buffers typically manifest as per-region shifts or scalings, which our regularization handles, making the sharper per-view details from image mode preferable.
Diffusion-BP directly back-projects diffusion predictions onto the reconstructed surface without further optimization and therefore has no mechanism to correct view inconsistencies, making video mode's stronger cross-view consistency beneficial; on Stanford-ORB, video mode also improves Diffusion-BP relighting PSNR-L over image mode (33.09 vs.\ 32.67).

\section{Additional Implementation Details}
\label{sec:impl_details_supp}

\paragraph{Renderer configuration.}
Our physics-based inverse rendering stage uses Mitsuba~3~\cite{DBLP:journals/tog/JakobSRV22} with the \texttt{prb} (path replay backpropagation) integrator~\cite{Vicini2022EfficientDifferentiableRendering} and a maximum path depth of 3, enabling one-bounce indirect illumination and inter-reflections.
Our code will be made publicly available.

\paragraph{Evaluation protocol.}
For Stanford-ORB~\cite{DBLP:conf/nips/KuangZYAWW23}, the background is first masked out using the ground-truth mask, and metrics are then computed on the full masked image; this protocol is applied consistently to all compared methods.

\paragraph{Baselines.}
For R3DG~\cite{DBLP:conf/eccv/GaoGLLZCZY24} and Neural Gaffer~\cite{DBLP:conf/nips/JinLLXBZX0S24} we run the authors' released code per scene, adapting only their data loaders to the Stanford-ORB intrinsics, poses, and environment-map convention.
For Material Anything~\cite{DBLP:conf/cvpr/HuangWLW25} we use its textured-input variant on \emph{our} reconstructed mesh and relight the predicted materials with the same renderer and evaluation pipeline as our method, so the comparison isolates its predicted materials.
The Vanilla Mitsuba~\cite{DBLP:journals/tog/JakobSRV22} baseline is our own pipeline optimized from scratch without the learned G-buffer initialization or any of our regularization terms.
For NVDiffRecMC~\cite{DBLP:conf/nips/HasselgrenHM22} and IllumiNeRF~\cite{DBLP:conf/nips/ZhaoSVPMH24} we report the relighting numbers directly from the official Stanford-ORB leaderboard.

\paragraph{Albedo--lighting ambiguity in Diffusion-BP.}
The aligned-albedo metric in Tab.~2 of the main paper applies a global per-channel scale factor to the reconstructed albedo after optimization, compensating for the overall scale ambiguity between albedo and illumination intensity.
For Diffusion-BP, however, errors are not limited to a global scale offset: diffusion predictions contain per-region albedo discrepancies and roughness inaccuracies that a single global factor cannot correct.
This is precisely the failure mode our regularization addresses by enforcing per-region material consistency rather than a global adjustment.

\section{Changing the Upstream Model}
\label{sec:supp-rgbx-upstream}

Our regularizer is not tied to DiffusionRenderer.
To test whether the same design transfers to another upstream model, we replace DiffusionRenderer~\cite{DBLP:journals/corr/abs-2501-18590} with RGB$\leftrightarrow$X~\cite{DBLP:conf/siggraph/0005DGHHLYH24}, which differs in both architecture and training data.
We compare three ways of using the RGB$\leftrightarrow$X predictions: direct back-projection onto the reconstructed surface, a global scale-invariant guidance loss, and our implicit material clustering regularizer with its similarity kernel built from these predictions.

\begin{table}[t]
  \centering
  \caption{Stanford-ORB relighting results when replacing DiffusionRenderer with RGB$\leftrightarrow$X as the upstream model.}
  \label{tab:supp-stanford-orb-rgbx}
  {\scriptsize\setlength{\tabcolsep}{2.0pt}\resizebox{0.58\textwidth}{!}{\begin{tabular}{@{}lcccc@{}}
\toprule
Method & PSNR-H$\uparrow$ & PSNR-L$\uparrow$ & SSIM$\uparrow$ & LPIPS$\downarrow$ \\
\midrule
RGB$\leftrightarrow$X-BP & 25.78 & 32.51 & 0.976 & 0.0333 \\
RGB$\leftrightarrow$X + scale inv. & 26.02 & 33.87 & 0.978 & 0.0240 \\
RGB$\leftrightarrow$X + Ours & \textbf{26.58} & \textbf{34.25} & \textbf{0.979} & \textbf{0.0238} \\
\bottomrule
\end{tabular}
}}
\end{table}

\begin{figure}[t]
\centering
\begingroup
\graphicspath{{figures/additional/stanford_orb_upstream_rgbx_cases/}}
\begin{tabular}{@{}c@{\hspace{0.003\linewidth}}c@{\hspace{0.003\linewidth}}c@{\hspace{0.003\linewidth}}c@{\hspace{0.003\linewidth}}c@{}}
  \multicolumn{5}{l}{\normalsize\texttt{\detokenize{Stanford-ORB grogu_scene001}}} \\
  \includegraphics[width=0.190\linewidth]{grogu_scene001_-_grogu_scene003_0067/Input.png} & \includegraphics[width=0.190\linewidth]{grogu_scene001_-_grogu_scene003_0067/Relight-GT.png} & \includegraphics[width=0.190\linewidth]{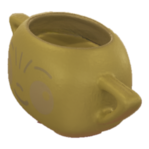} & \includegraphics[width=0.190\linewidth]{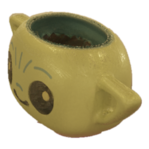} & \includegraphics[width=0.190\linewidth]{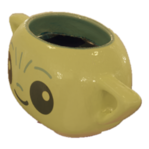} \\[-0.2em]
  \multicolumn{5}{l}{\normalsize\texttt{\detokenize{Stanford-ORB pepsi_scene002}}} \\
  \includegraphics[width=0.190\linewidth]{pepsi_scene002_-_pepsi_scene003_0061/Input.png} & \includegraphics[width=0.190\linewidth]{pepsi_scene002_-_pepsi_scene003_0061/Relight-GT.png} & \includegraphics[width=0.190\linewidth]{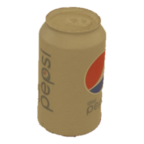} & \includegraphics[width=0.190\linewidth]{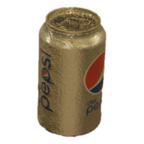} & \includegraphics[width=0.190\linewidth]{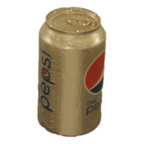} \\
  \footnotesize Input & \footnotesize Relight-GT & \footnotesize RGB$\leftrightarrow$X-BP & \begin{tabular}{@{}c@{}}\footnotesize RGB$\leftrightarrow$X\\[-0.25em]\footnotesize scale inv.\end{tabular} & \begin{tabular}{@{}c@{}}\footnotesize RGB$\leftrightarrow$X\\[-0.25em]\footnotesize Ours\end{tabular}
\end{tabular}

\endgroup
\caption{Qualitative Stanford-ORB relighting results when replacing DiffusionRenderer with RGB$\leftrightarrow$X as the upstream model.}
\label{fig:supp-stanford-orb-rgbx}
\end{figure}

\Cref{tab:supp-stanford-orb-rgbx,fig:supp-stanford-orb-rgbx} show the same trend as our DiffusionRenderer-based experiments: direct back-projection is insufficient, while our regularizer improves relighting quality over the scale-invariant alternative.
This indicates that the main gain comes from the regularizer design rather than from a property specific to one upstream diffusion model.

\section{Transfer of Our Material Regularization to IRGS}
\label{sec:supp-irgs-jbf-ablation}

To assess whether our regularization transfers beyond our PBIR pipeline, we apply it to IRGS~\cite{DBLP:conf/cvpr/GuWZY025}, a strong inverse-rendering baseline based on 2D Gaussian splatting with inter-reflective ray tracing.
IRGS includes non-data-driven smoothness terms on base color and roughness.
We replace these smoothness terms with our diffusion-guided material regularization, keeping the rest of the IRGS optimization unchanged.
We compare against the original IRGS baseline on all four Synthetic4Relight~\cite{DBLP:conf/cvpr/ZhangSHFJZ22} scenes.

\paragraph{Setup.}
In Stage~1, RefGS reconstruction is shared with the unmodified IRGS baseline.
In Stage~2, we replace IRGS's original base-color and roughness smoothness terms with our material regularization term and set its weight to $\lambda_{\text{mat}}=100$ for all scenes.
Relighting is evaluated under two environment maps (envmap6, envmap12) and averaged.

\paragraph{Quantitative results.}
\Cref{tab:supp-irgs-jbf-ablation} reports per-scene relighting PSNR and SSIM, together with the average over the four scenes.
Replacing IRGS's original smoothness terms with our material regularization consistently improves relighting on all four scenes, with gains of $+0.45$ to $+1.59$~dB in PSNR.
The average improvement is $+0.80$~dB PSNR.
These gains indicate that our regularization is not tied to our own PBIR formulation and can also benefit a different inverse-rendering pipeline.

\begin{table}[h]
  \centering
  \caption{Transfer of our material regularization to IRGS on Synthetic4Relight.}
  \label{tab:supp-irgs-jbf-ablation}
  {\scriptsize\setlength{\tabcolsep}{1.8pt}
  \begin{tabular}{@{}l*{10}{c}@{}}
    \toprule
      & \multicolumn{10}{c}{Synthetic4Relight} \\
    \cmidrule(lr){2-11}
      & \multicolumn{10}{c}{Relighting} \\
    \cmidrule(lr){2-11}
      & \multicolumn{2}{c}{\texttt{air\_balloons}} & \multicolumn{2}{c}{\texttt{chair}} & \multicolumn{2}{c}{\texttt{hotdog}} & \multicolumn{2}{c}{\texttt{jugs}} & \multicolumn{2}{c}{Average} \\
    \cmidrule(lr){2-3}
    \cmidrule(lr){4-5}
    \cmidrule(lr){6-7}
    \cmidrule(lr){8-9}
    \cmidrule(lr){10-11}
      Method & \shortstack{PSNR\\$\uparrow$} & \shortstack{SSIM\\$\uparrow$} & \shortstack{PSNR\\$\uparrow$} & \shortstack{SSIM\\$\uparrow$} & \shortstack{PSNR\\$\uparrow$} & \shortstack{SSIM\\$\uparrow$} & \shortstack{PSNR\\$\uparrow$} & \shortstack{SSIM\\$\uparrow$} & \shortstack{PSNR\\$\uparrow$} & \shortstack{SSIM\\$\uparrow$} \\
    \midrule
    IRGS            & 32.85 & 0.956 & 36.32 & 0.970 & 32.48 & 0.949 & 37.49 & 0.978 & 34.78 & 0.963 \\
    IRGS + Our Reg. & \textbf{33.46} & \textbf{0.959} & \textbf{37.91} & \textbf{0.973} & \textbf{33.02} & \textbf{0.950} & \textbf{37.94} & \textbf{0.979} & \textbf{35.58} & \textbf{0.965} \\
    \bottomrule
  \end{tabular}}
\end{table}

\begin{figure*}[t]
  \centering
  \newcommand{\intrinsicrowlabel}[1]{\rotatebox[origin=c]{90}{\scriptsize\strut #1}}
  {\scriptsize\texttt{\detokenize{Synthetic4Relight air_baloons}}}\\[-0.2em]
  \parbox[c]{0.05\linewidth}{\centering \intrinsicrowlabel{Input and Prediction}}%
  \hspace{2pt}%
  \parbox[c]{0.172\linewidth}{\centering \includegraphics[width=\linewidth]{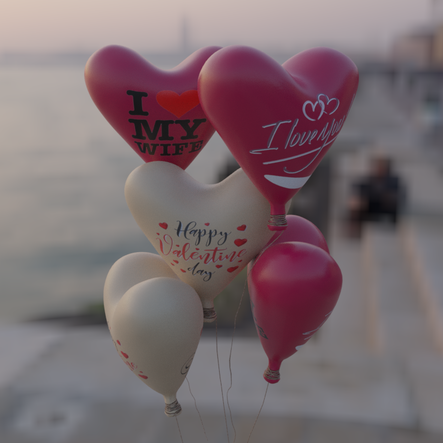}}%
  \hspace{2pt}%
  \parbox[c]{0.172\linewidth}{\centering \includegraphics[width=\linewidth]{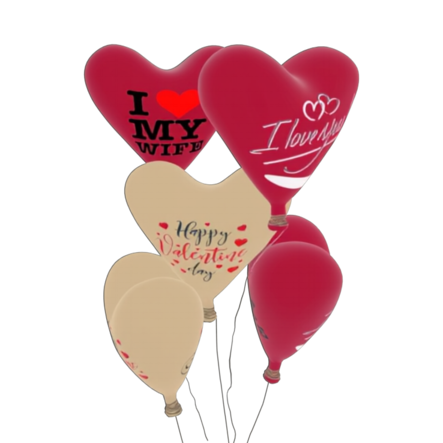}}%
  \hspace{2pt}%
  \parbox[c]{0.172\linewidth}{\centering \includegraphics[width=\linewidth]{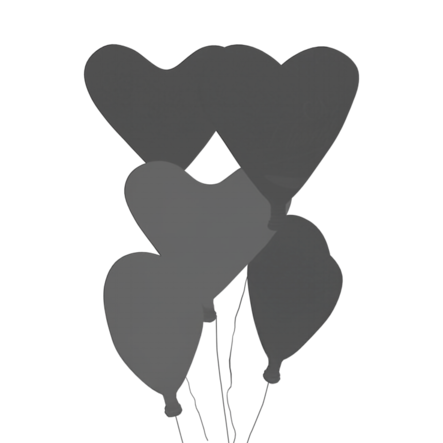}}%
  \hspace{2pt}%
  \parbox[c]{0.172\linewidth}{\centering \includegraphics[width=\linewidth]{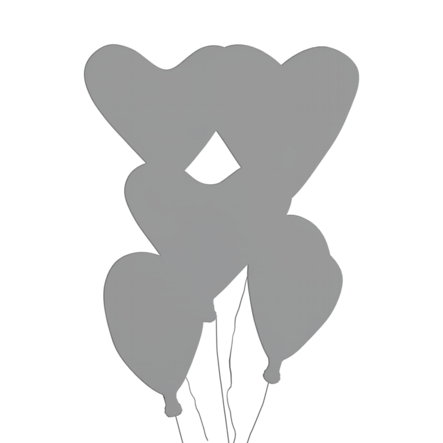}}%
  \hspace{2pt}%
  \parbox[c]{0.172\linewidth}{\centering \includegraphics[width=\linewidth]{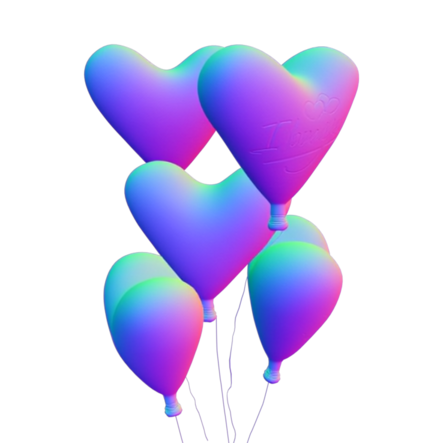}}%
  \\[1pt]
  \makebox[0.05\linewidth][c]{}%
  \hspace{2pt}%
  \makebox[0.172\linewidth][c]{\scriptsize Input}%
  \hspace{2pt}%
  \makebox[0.172\linewidth][c]{\scriptsize DR-base color}%
  \hspace{2pt}%
  \makebox[0.172\linewidth][c]{\scriptsize DR-roughness}%
  \hspace{2pt}%
  \makebox[0.172\linewidth][c]{\scriptsize DR-metallic}%
  \hspace{2pt}%
  \makebox[0.172\linewidth][c]{\scriptsize DR-normal}%
  \\[1pt]
  \vspace{0.15em}
  \par\noindent\rule{\linewidth}{0.35pt}
  \vspace{0.05em}
  \vspace{-0.6em}
  \parbox[c]{0.09\linewidth}{\centering \intrinsicrowlabel{Relight}}%
  \hspace{2pt}%
  \parbox[c]{0.215\linewidth}{\centering \begin{overpic}[width=\linewidth]{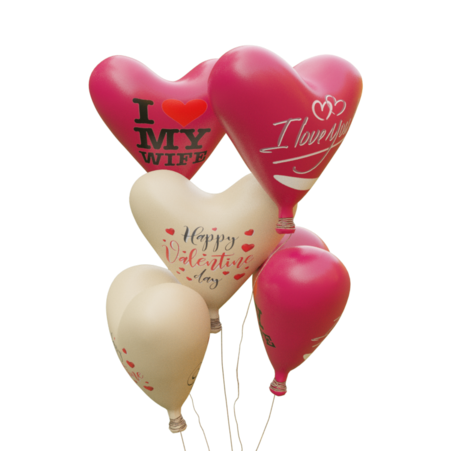}\put(41.678,34.354){\color{red}\linethickness{0.8pt}\framebox(15.180,15.180){}}\end{overpic}}%
  \hspace{2pt}%
  \parbox[c]{0.215\linewidth}{\centering \begin{overpic}[width=\linewidth]{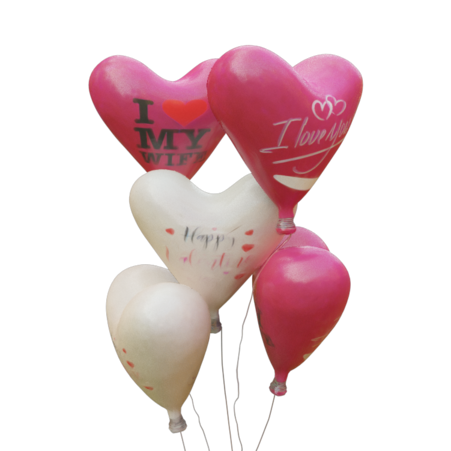}\put(41.678,34.354){\color{red}\linethickness{0.8pt}\framebox(15.180,15.180){}}\end{overpic}}%
  \hspace{2pt}%
  \parbox[c]{0.215\linewidth}{\centering \begin{overpic}[width=\linewidth]{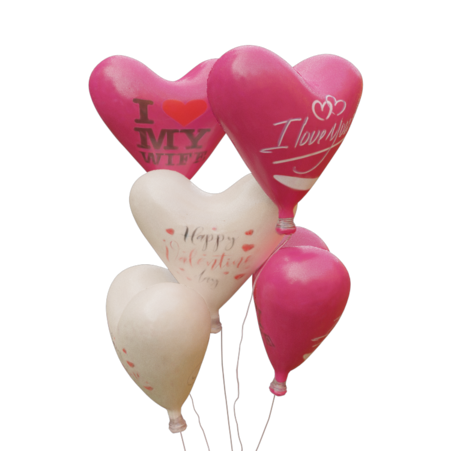}\put(41.678,34.354){\color{red}\linethickness{0.8pt}\framebox(15.180,15.180){}}\end{overpic}}%
  \\[0.1em]
  \parbox[c]{0.09\linewidth}{\centering \intrinsicrowlabel{Relight Zoom}}%
  \hspace{2pt}%
  \parbox[c]{0.215\linewidth}{\centering \includegraphics[width=\linewidth]{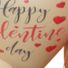}}%
  \hspace{2pt}%
  \parbox[c]{0.215\linewidth}{\centering \includegraphics[width=\linewidth]{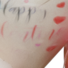}}%
  \hspace{2pt}%
  \parbox[c]{0.215\linewidth}{\centering \includegraphics[width=\linewidth]{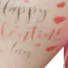}}%
  \\[0.1em]
  \parbox[c]{0.09\linewidth}{\centering \intrinsicrowlabel{Base Color}}%
  \hspace{2pt}%
  \parbox[c]{0.215\linewidth}{\centering \includegraphics[width=\linewidth]{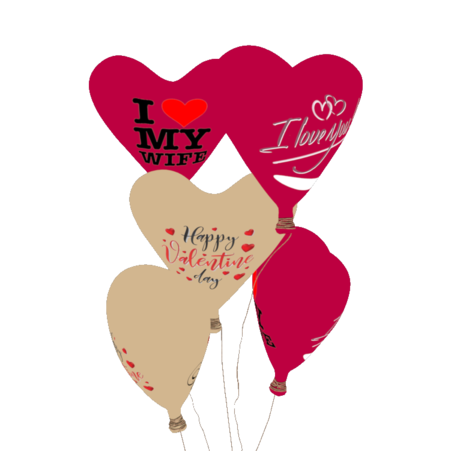}}%
  \hspace{2pt}%
  \parbox[c]{0.215\linewidth}{\centering \includegraphics[width=\linewidth]{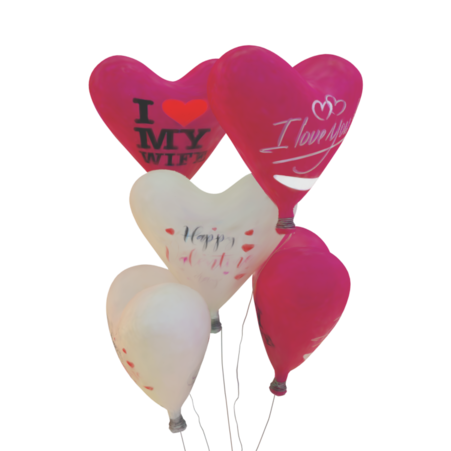}}%
  \hspace{2pt}%
  \parbox[c]{0.215\linewidth}{\centering \includegraphics[width=\linewidth]{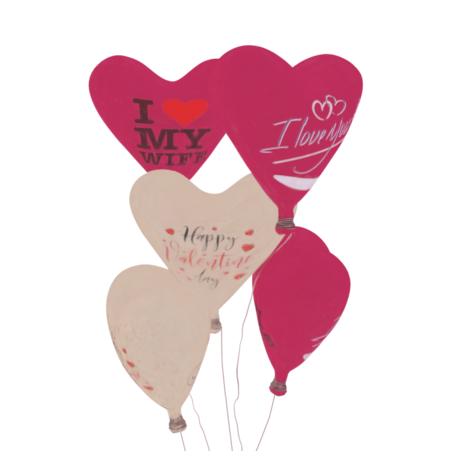}}%
  \\[0.1em]
  \parbox[c]{0.09\linewidth}{\centering \intrinsicrowlabel{Roughness}}%
  \hspace{2pt}%
  \parbox[c]{0.215\linewidth}{\centering \includegraphics[width=\linewidth]{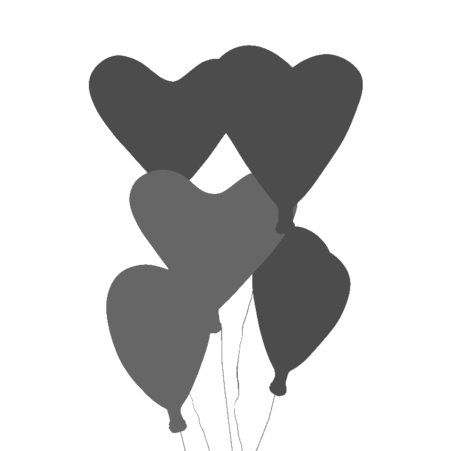}}%
  \hspace{2pt}%
  \parbox[c]{0.215\linewidth}{\centering \includegraphics[width=\linewidth]{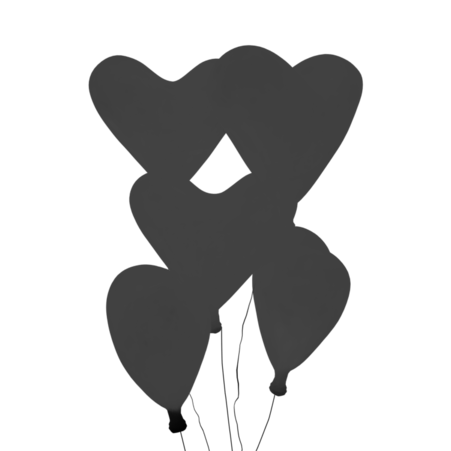}}%
  \hspace{2pt}%
  \parbox[c]{0.215\linewidth}{\centering \includegraphics[width=\linewidth]{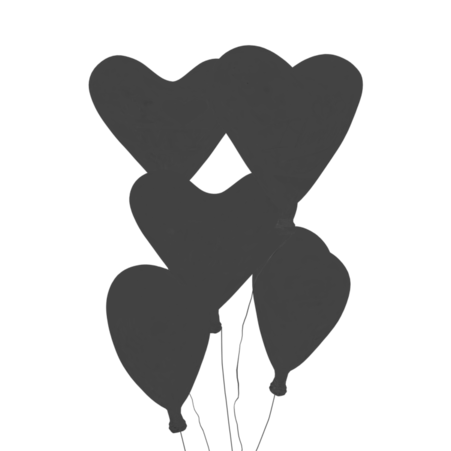}}%
  \\[0.1em]
  \parbox[c]{0.09\linewidth}{\centering \intrinsicrowlabel{Normal}}%
  \hspace{2pt}%
  \parbox[c]{0.215\linewidth}{\centering \begin{overpic}[width=\linewidth]{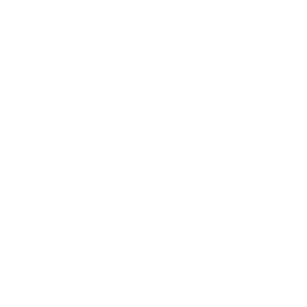}\put(50,50){\makebox(0,0){\scriptsize N/A}}\end{overpic}}%
  \hspace{2pt}%
  \parbox[c]{0.215\linewidth}{\centering \includegraphics[width=\linewidth]{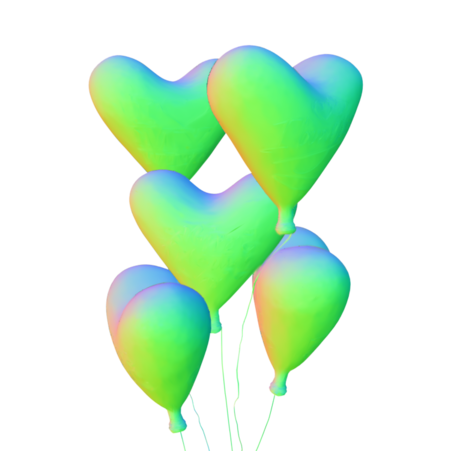}}%
  \hspace{2pt}%
  \parbox[c]{0.215\linewidth}{\centering \includegraphics[width=\linewidth]{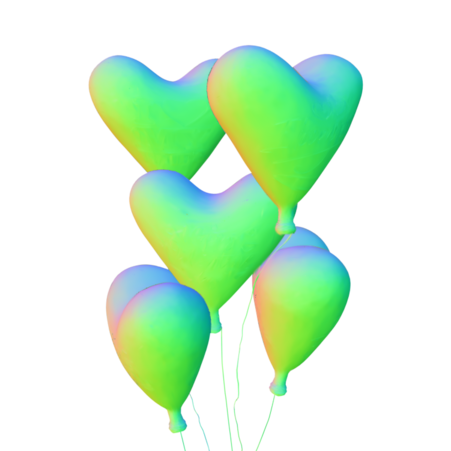}}%
  \\[1pt]
  \makebox[0.09\linewidth][c]{}%
  \hspace{2pt}%
  \makebox[0.215\linewidth][c]{\scriptsize Reference}%
  \hspace{2pt}%
  \makebox[0.215\linewidth][c]{\scriptsize IRGS}%
  \hspace{2pt}%
  \makebox[0.215\linewidth][c]{\scriptsize IRGS + JBF (Ours)}%
  \\[1pt]
  \caption{Synthetic4Relight \casename{air_baloons} (view \texttt{101}). JBF material regularization smooths base color and roughness G-buffers, yielding improved relighting quality.}
  \label{fig:supp-irgs-jbf-air-baloons}
\end{figure*}

\begin{figure*}[t]
  \centering
  \newcommand{\intrinsicrowlabel}[1]{\rotatebox[origin=c]{90}{\scriptsize\strut #1}}
  {\scriptsize\texttt{\detokenize{Synthetic4Relight chair}}}\\[-0.2em]
  \parbox[c]{0.05\linewidth}{\centering \intrinsicrowlabel{Input and Prediction}}%
  \hspace{2pt}%
  \parbox[c]{0.172\linewidth}{\centering \includegraphics[width=\linewidth]{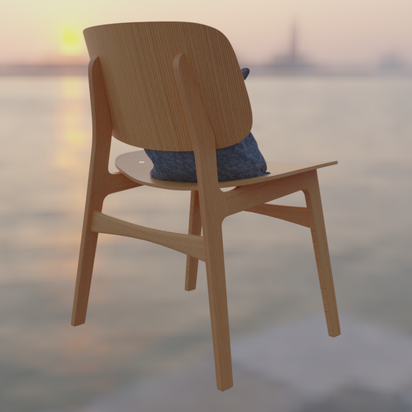}}%
  \hspace{2pt}%
  \parbox[c]{0.172\linewidth}{\centering \includegraphics[width=\linewidth]{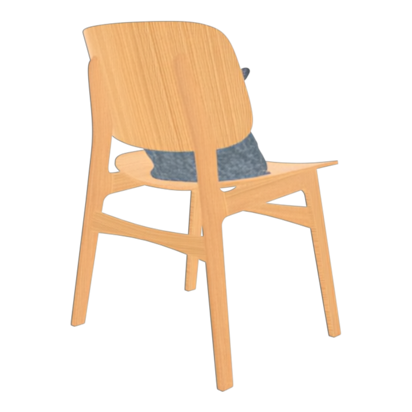}}%
  \hspace{2pt}%
  \parbox[c]{0.172\linewidth}{\centering \includegraphics[width=\linewidth]{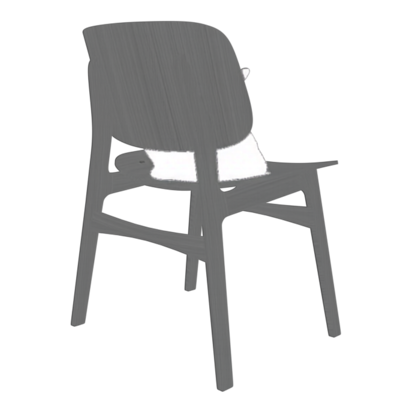}}%
  \hspace{2pt}%
  \parbox[c]{0.172\linewidth}{\centering \includegraphics[width=\linewidth]{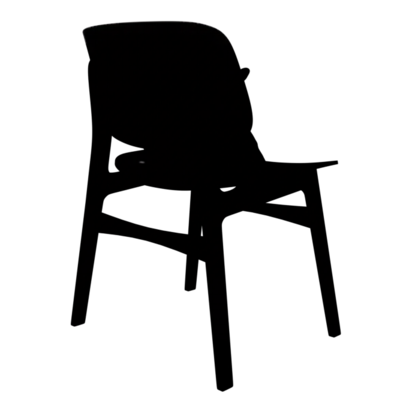}}%
  \hspace{2pt}%
  \parbox[c]{0.172\linewidth}{\centering \includegraphics[width=\linewidth]{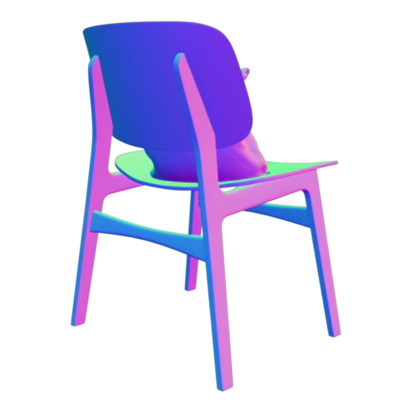}}%
  \\[1pt]
  \makebox[0.05\linewidth][c]{}%
  \hspace{2pt}%
  \makebox[0.172\linewidth][c]{\scriptsize Input}%
  \hspace{2pt}%
  \makebox[0.172\linewidth][c]{\scriptsize DR-base color}%
  \hspace{2pt}%
  \makebox[0.172\linewidth][c]{\scriptsize DR-roughness}%
  \hspace{2pt}%
  \makebox[0.172\linewidth][c]{\scriptsize DR-metallic}%
  \hspace{2pt}%
  \makebox[0.172\linewidth][c]{\scriptsize DR-normal}%
  \\[1pt]
  \vspace{0.15em}
  \par\noindent\rule{\linewidth}{0.35pt}
  \vspace{0.05em}
  \vspace{-0.6em}
  \parbox[c]{0.09\linewidth}{\centering \intrinsicrowlabel{Relight}}%
  \hspace{2pt}%
  \parbox[c]{0.215\linewidth}{\centering \includegraphics[width=\linewidth]{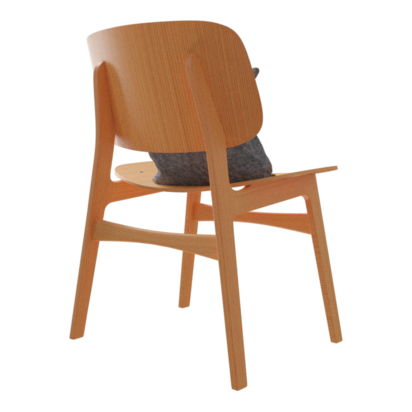}}%
  \hspace{2pt}%
  \parbox[c]{0.215\linewidth}{\centering \includegraphics[width=\linewidth]{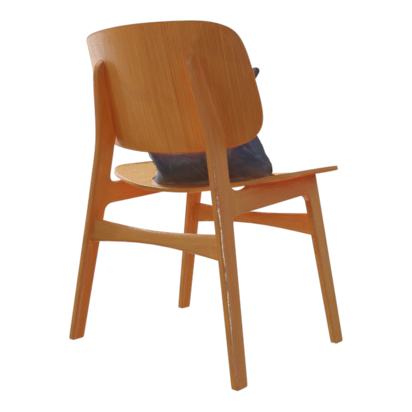}}%
  \hspace{2pt}%
  \parbox[c]{0.215\linewidth}{\centering \includegraphics[width=\linewidth]{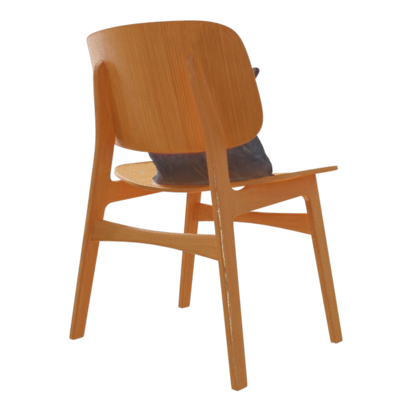}}%
  \\[0.1em]
  \parbox[c]{0.09\linewidth}{\centering \intrinsicrowlabel{Base Color}}%
  \hspace{2pt}%
  \parbox[c]{0.215\linewidth}{\centering \includegraphics[width=\linewidth]{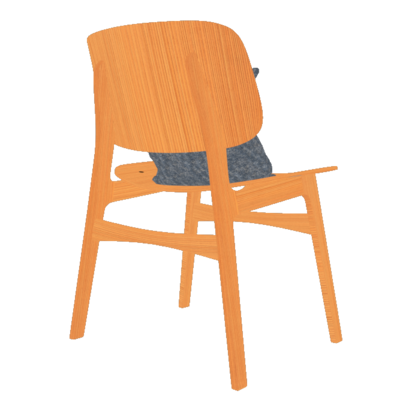}}%
  \hspace{2pt}%
  \parbox[c]{0.215\linewidth}{\centering \includegraphics[width=\linewidth]{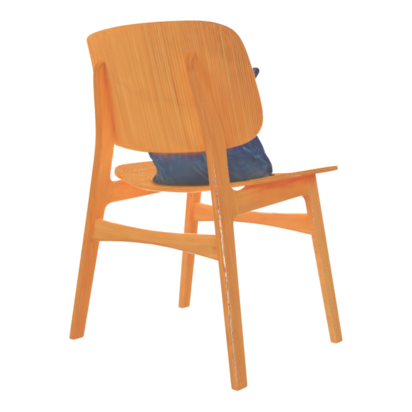}}%
  \hspace{2pt}%
  \parbox[c]{0.215\linewidth}{\centering \includegraphics[width=\linewidth]{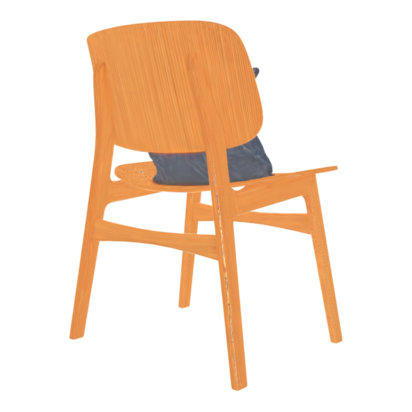}}%
  \\[0.1em]
  \parbox[c]{0.09\linewidth}{\centering \intrinsicrowlabel{Roughness}}%
  \hspace{2pt}%
  \parbox[c]{0.215\linewidth}{\centering \begin{overpic}[width=\linewidth]{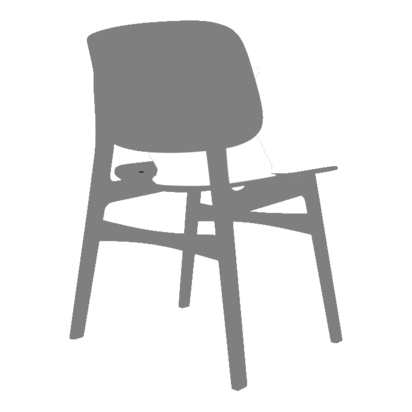}\put(36.099,44.978){\color{red}\linethickness{0.8pt}\framebox(30.568,30.568){}}\end{overpic}}%
  \hspace{2pt}%
  \parbox[c]{0.215\linewidth}{\centering \begin{overpic}[width=\linewidth]{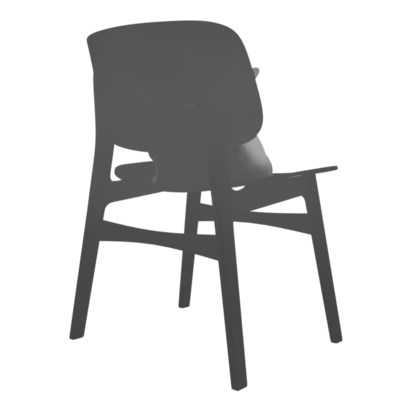}\put(36.099,44.978){\color{red}\linethickness{0.8pt}\framebox(30.568,30.568){}}\end{overpic}}%
  \hspace{2pt}%
  \parbox[c]{0.215\linewidth}{\centering \begin{overpic}[width=\linewidth]{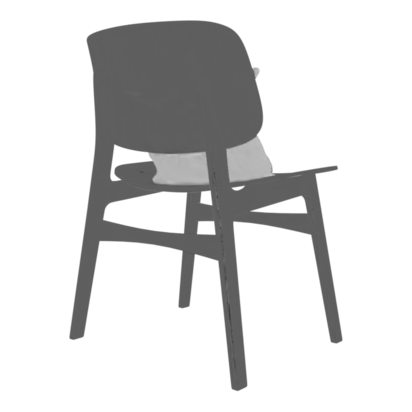}\put(36.099,44.978){\color{red}\linethickness{0.8pt}\framebox(30.568,30.568){}}\end{overpic}}%
  \\[0.1em]
  \parbox[c]{0.09\linewidth}{\centering \intrinsicrowlabel{Roughness Zoom}}%
  \hspace{2pt}%
  \parbox[c]{0.215\linewidth}{\centering \includegraphics[width=\linewidth]{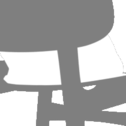}}%
  \hspace{2pt}%
  \parbox[c]{0.215\linewidth}{\centering \includegraphics[width=\linewidth]{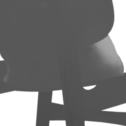}}%
  \hspace{2pt}%
  \parbox[c]{0.215\linewidth}{\centering \includegraphics[width=\linewidth]{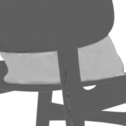}}%
  \\[0.1em]
  \parbox[c]{0.09\linewidth}{\centering \intrinsicrowlabel{Normal}}%
  \hspace{2pt}%
  \parbox[c]{0.215\linewidth}{\centering \begin{overpic}[width=\linewidth]{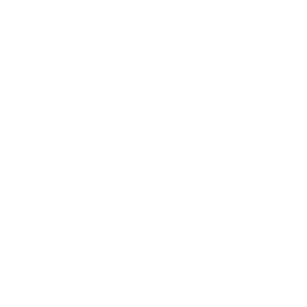}\put(50,50){\makebox(0,0){\scriptsize N/A}}\end{overpic}}%
  \hspace{2pt}%
  \parbox[c]{0.215\linewidth}{\centering \includegraphics[width=\linewidth]{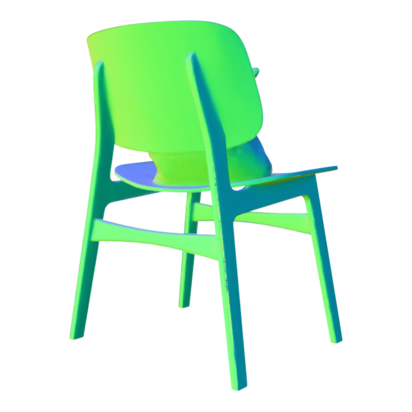}}%
  \hspace{2pt}%
  \parbox[c]{0.215\linewidth}{\centering \includegraphics[width=\linewidth]{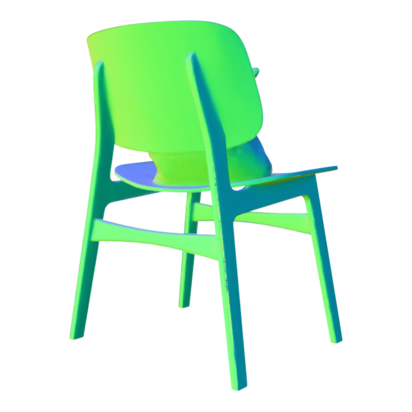}}%
  \\[1pt]
  \makebox[0.09\linewidth][c]{}%
  \hspace{2pt}%
  \makebox[0.215\linewidth][c]{\scriptsize Reference}%
  \hspace{2pt}%
  \makebox[0.215\linewidth][c]{\scriptsize IRGS}%
  \hspace{2pt}%
  \makebox[0.215\linewidth][c]{\scriptsize IRGS + JBF (Ours)}%
  \\[1pt]
  \caption{Synthetic4Relight \casename{chair} (view \texttt{111}). JBF material regularization smooths base color and roughness G-buffers, yielding improved relighting quality.}
  \label{fig:supp-irgs-jbf-chair}
\end{figure*}

\begin{figure*}[t]
  \centering
  \newcommand{\intrinsicrowlabel}[1]{\rotatebox[origin=c]{90}{\scriptsize\strut #1}}
  {\scriptsize\texttt{\detokenize{Synthetic4Relight hotdog}}}\\[-0.2em]
  \parbox[c]{0.05\linewidth}{\centering \intrinsicrowlabel{Input and Prediction}}%
  \hspace{2pt}%
  \parbox[c]{0.172\linewidth}{\centering \includegraphics[width=\linewidth]{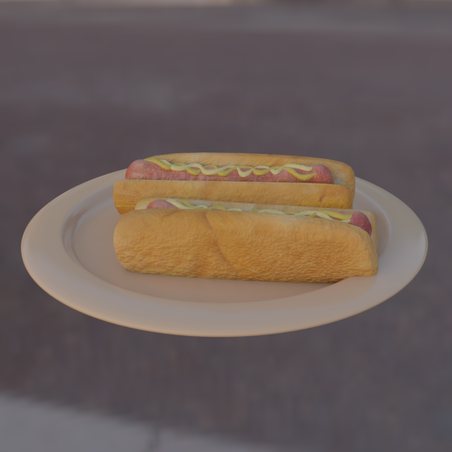}}%
  \hspace{2pt}%
  \parbox[c]{0.172\linewidth}{\centering \includegraphics[width=\linewidth]{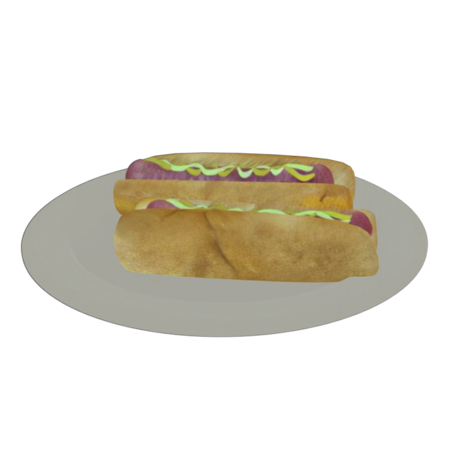}}%
  \hspace{2pt}%
  \parbox[c]{0.172\linewidth}{\centering \includegraphics[width=\linewidth]{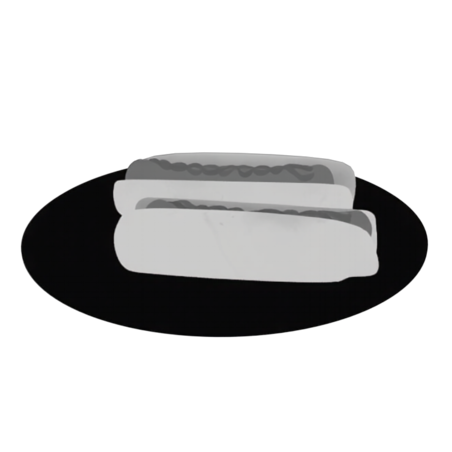}}%
  \hspace{2pt}%
  \parbox[c]{0.172\linewidth}{\centering \includegraphics[width=\linewidth]{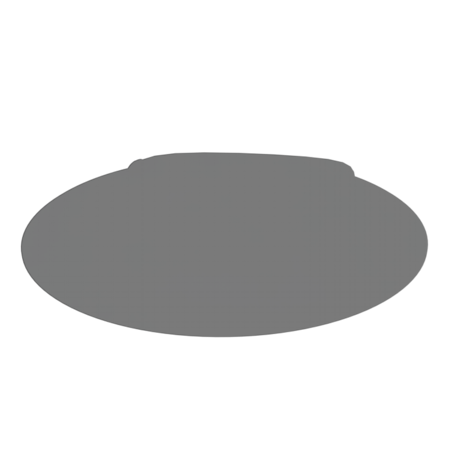}}%
  \hspace{2pt}%
  \parbox[c]{0.172\linewidth}{\centering \includegraphics[width=\linewidth]{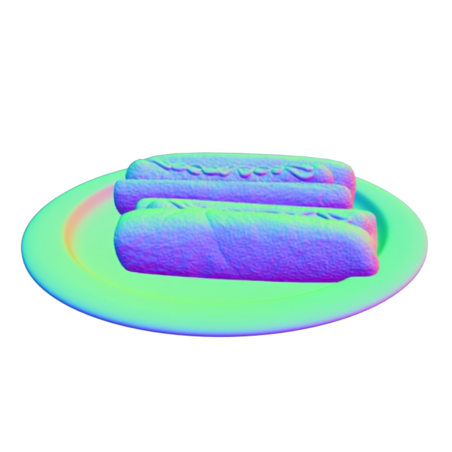}}%
  \\[1pt]
  \makebox[0.05\linewidth][c]{}%
  \hspace{2pt}%
  \makebox[0.172\linewidth][c]{\scriptsize Input}%
  \hspace{2pt}%
  \makebox[0.172\linewidth][c]{\scriptsize DR-base color}%
  \hspace{2pt}%
  \makebox[0.172\linewidth][c]{\scriptsize DR-roughness}%
  \hspace{2pt}%
  \makebox[0.172\linewidth][c]{\scriptsize DR-metallic}%
  \hspace{2pt}%
  \makebox[0.172\linewidth][c]{\scriptsize DR-normal}%
  \\[1pt]
  \vspace{0.15em}
  \par\noindent\rule{\linewidth}{0.35pt}
  \vspace{0.05em}
  \vspace{-0.6em}
  \parbox[c]{0.09\linewidth}{\centering \intrinsicrowlabel{Relight}}%
  \hspace{2pt}%
  \parbox[c]{0.215\linewidth}{\centering \includegraphics[width=\linewidth]{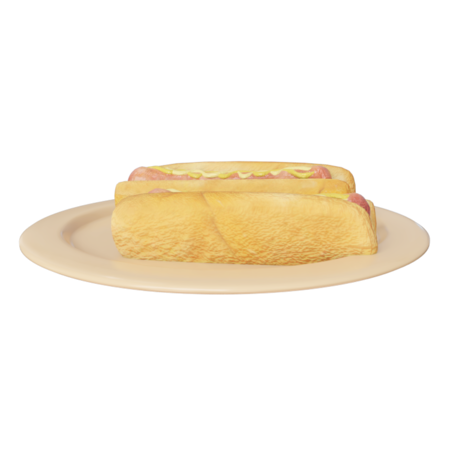}}%
  \hspace{2pt}%
  \parbox[c]{0.215\linewidth}{\centering \includegraphics[width=\linewidth]{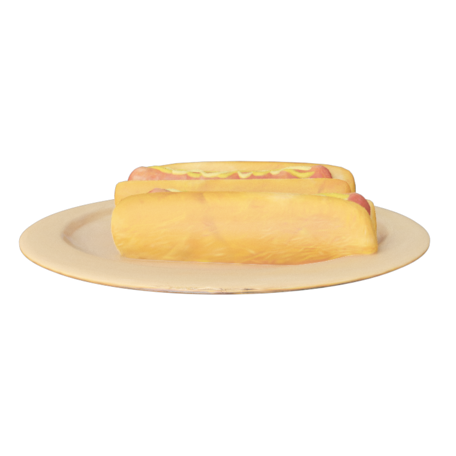}}%
  \hspace{2pt}%
  \parbox[c]{0.215\linewidth}{\centering \includegraphics[width=\linewidth]{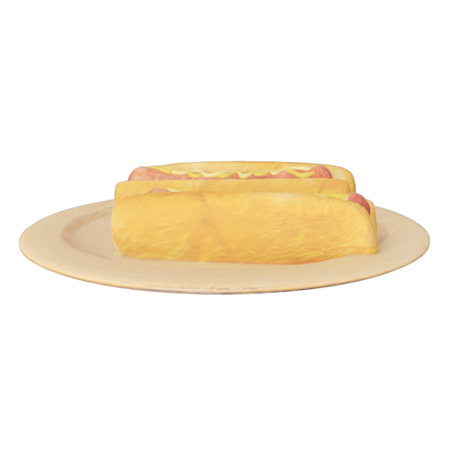}}%
  \\[0.1em]
  \parbox[c]{0.09\linewidth}{\centering \intrinsicrowlabel{Base Color}}%
  \hspace{2pt}%
  \parbox[c]{0.215\linewidth}{\centering \includegraphics[width=\linewidth]{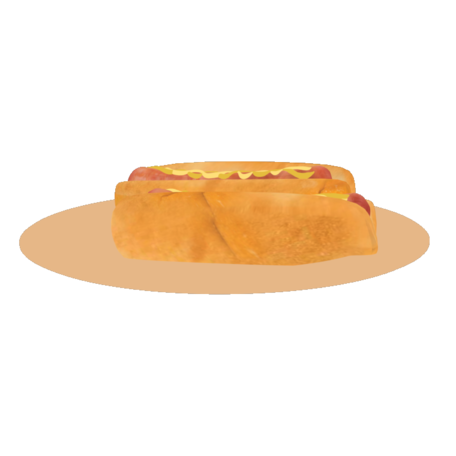}}%
  \hspace{2pt}%
  \parbox[c]{0.215\linewidth}{\centering \includegraphics[width=\linewidth]{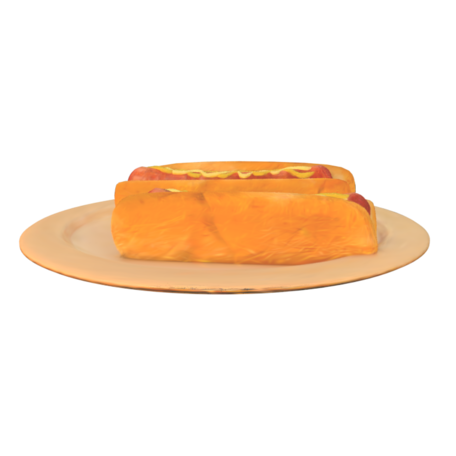}}%
  \hspace{2pt}%
  \parbox[c]{0.215\linewidth}{\centering \includegraphics[width=\linewidth]{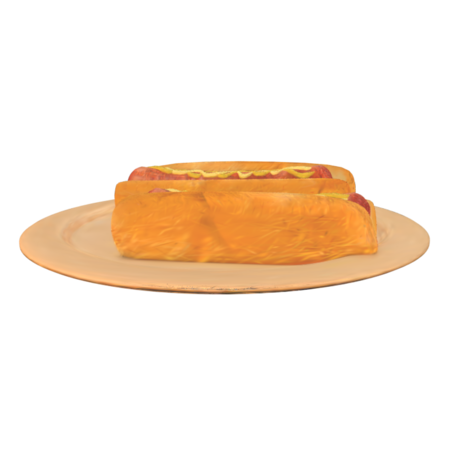}}%
  \\[0.1em]
  \parbox[c]{0.09\linewidth}{\centering \intrinsicrowlabel{Roughness}}%
  \hspace{2pt}%
  \parbox[c]{0.215\linewidth}{\centering \begin{overpic}[width=\linewidth]{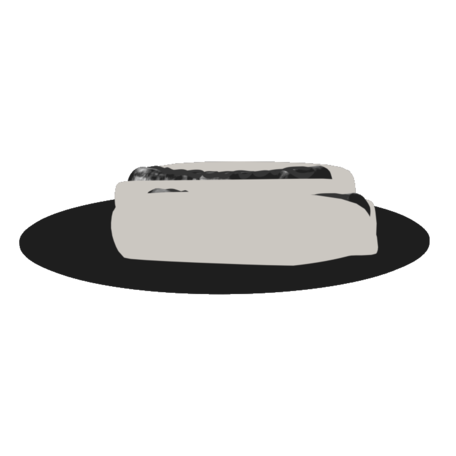}\put(23.816,20.921){\color{red}\linethickness{0.8pt}\framebox(59.211,59.211){}}\end{overpic}}%
  \hspace{2pt}%
  \parbox[c]{0.215\linewidth}{\centering \begin{overpic}[width=\linewidth]{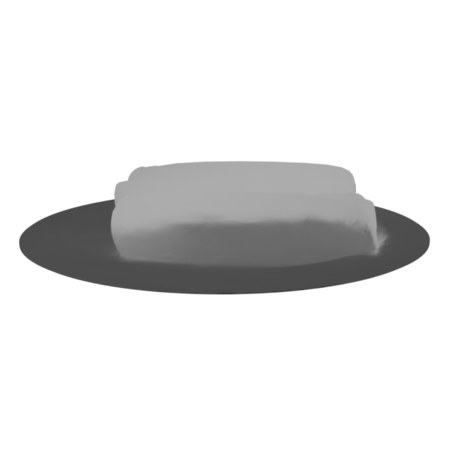}\put(23.816,20.921){\color{red}\linethickness{0.8pt}\framebox(59.211,59.211){}}\end{overpic}}%
  \hspace{2pt}%
  \parbox[c]{0.215\linewidth}{\centering \begin{overpic}[width=\linewidth]{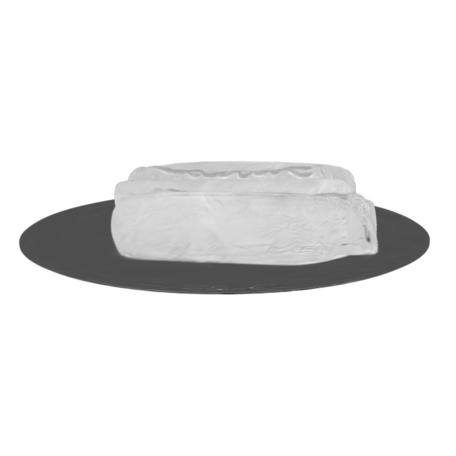}\put(23.816,20.921){\color{red}\linethickness{0.8pt}\framebox(59.211,59.211){}}\end{overpic}}%
  \\[0.1em]
  \parbox[c]{0.09\linewidth}{\centering \intrinsicrowlabel{Roughness Zoom}}%
  \hspace{2pt}%
  \parbox[c]{0.215\linewidth}{\centering \includegraphics[width=\linewidth]{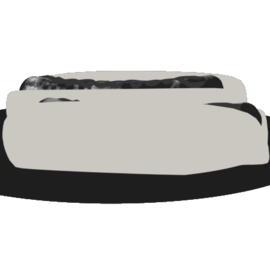}}%
  \hspace{2pt}%
  \parbox[c]{0.215\linewidth}{\centering \includegraphics[width=\linewidth]{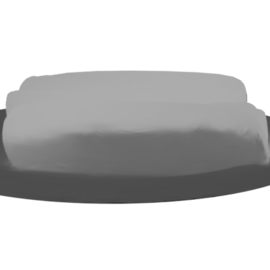}}%
  \hspace{2pt}%
  \parbox[c]{0.215\linewidth}{\centering \includegraphics[width=\linewidth]{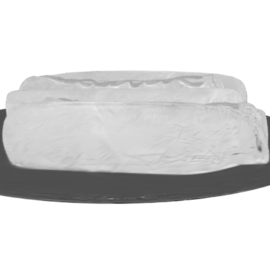}}%
  \\[0.1em]
  \parbox[c]{0.09\linewidth}{\centering \intrinsicrowlabel{Normal}}%
  \hspace{2pt}%
  \parbox[c]{0.215\linewidth}{\centering \begin{overpic}[width=\linewidth]{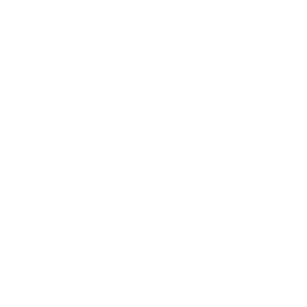}\put(50,50){\makebox(0,0){\scriptsize N/A}}\end{overpic}}%
  \hspace{2pt}%
  \parbox[c]{0.215\linewidth}{\centering \includegraphics[width=\linewidth]{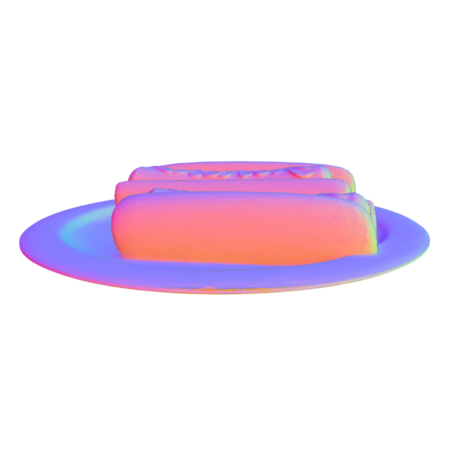}}%
  \hspace{2pt}%
  \parbox[c]{0.215\linewidth}{\centering \includegraphics[width=\linewidth]{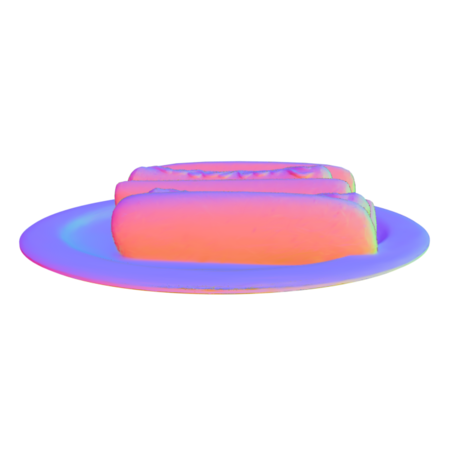}}%
  \\[1pt]
  \makebox[0.09\linewidth][c]{}%
  \hspace{2pt}%
  \makebox[0.215\linewidth][c]{\scriptsize Reference}%
  \hspace{2pt}%
  \makebox[0.215\linewidth][c]{\scriptsize IRGS}%
  \hspace{2pt}%
  \makebox[0.215\linewidth][c]{\scriptsize IRGS + JBF (Ours)}%
  \\[1pt]
  \caption{Synthetic4Relight \casename{hotdog} (view \texttt{078}). JBF material regularization smooths base color and roughness G-buffers, yielding improved relighting quality.}
  \label{fig:supp-irgs-jbf-hotdog}
\end{figure*}

\begin{figure*}[t]
  \centering
  \newcommand{\intrinsicrowlabel}[1]{\rotatebox[origin=c]{90}{\scriptsize\strut #1}}
  {\scriptsize\texttt{\detokenize{Synthetic4Relight jugs}}}\\[-0.2em]
  \parbox[c]{0.05\linewidth}{\centering \intrinsicrowlabel{Input and Prediction}}%
  \hspace{2pt}%
  \parbox[c]{0.172\linewidth}{\centering \includegraphics[width=\linewidth]{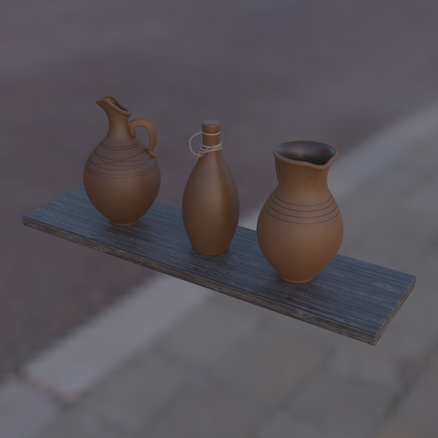}}%
  \hspace{2pt}%
  \parbox[c]{0.172\linewidth}{\centering \includegraphics[width=\linewidth]{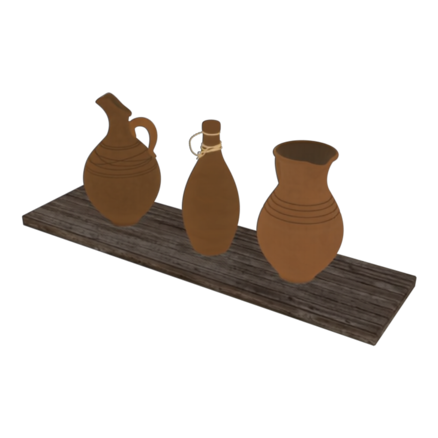}}%
  \hspace{2pt}%
  \parbox[c]{0.172\linewidth}{\centering \includegraphics[width=\linewidth]{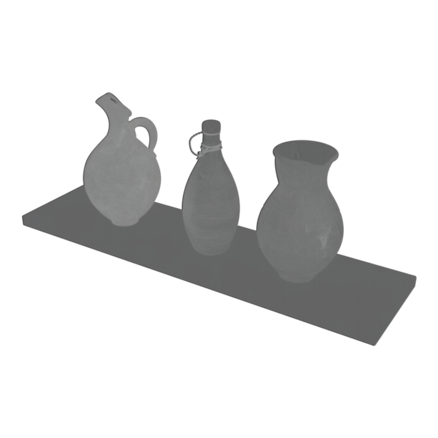}}%
  \hspace{2pt}%
  \parbox[c]{0.172\linewidth}{\centering \includegraphics[width=\linewidth]{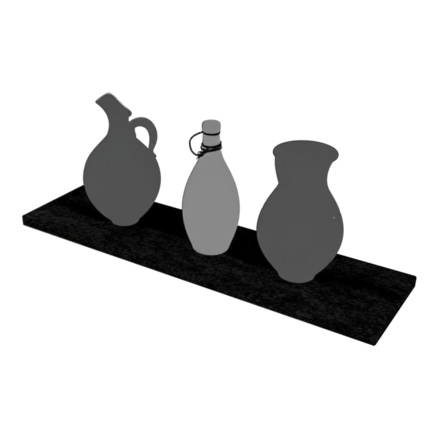}}%
  \hspace{2pt}%
  \parbox[c]{0.172\linewidth}{\centering \includegraphics[width=\linewidth]{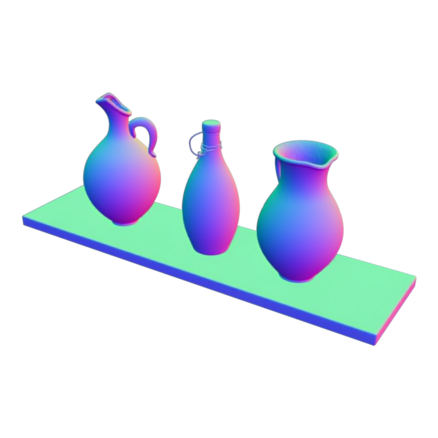}}%
  \\[1pt]
  \makebox[0.05\linewidth][c]{}%
  \hspace{2pt}%
  \makebox[0.172\linewidth][c]{\scriptsize Input}%
  \hspace{2pt}%
  \makebox[0.172\linewidth][c]{\scriptsize DR-base color}%
  \hspace{2pt}%
  \makebox[0.172\linewidth][c]{\scriptsize DR-roughness}%
  \hspace{2pt}%
  \makebox[0.172\linewidth][c]{\scriptsize DR-metallic}%
  \hspace{2pt}%
  \makebox[0.172\linewidth][c]{\scriptsize DR-normal}%
  \\[1pt]
  \vspace{0.15em}
  \par\noindent\rule{\linewidth}{0.35pt}
  \vspace{0.05em}
  \vspace{-0.6em}
  \parbox[c]{0.09\linewidth}{\centering \intrinsicrowlabel{Relight}}%
  \hspace{2pt}%
  \parbox[c]{0.215\linewidth}{\centering \includegraphics[width=\linewidth]{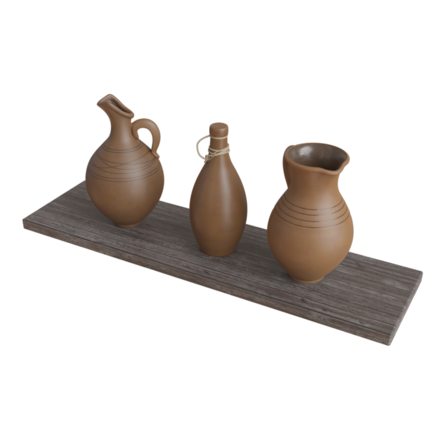}}%
  \hspace{2pt}%
  \parbox[c]{0.215\linewidth}{\centering \includegraphics[width=\linewidth]{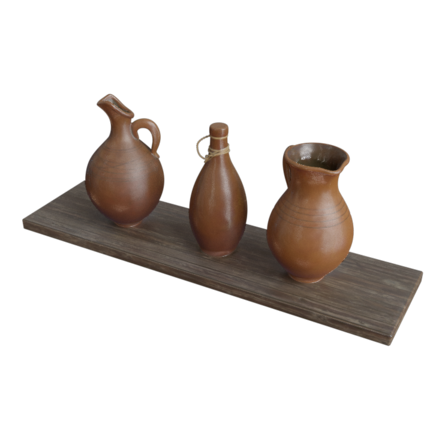}}%
  \hspace{2pt}%
  \parbox[c]{0.215\linewidth}{\centering \includegraphics[width=\linewidth]{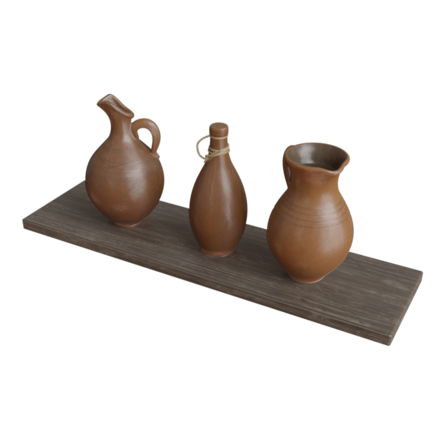}}%
  \\[0.1em]
  \parbox[c]{0.09\linewidth}{\centering \intrinsicrowlabel{Base Color}}%
  \hspace{2pt}%
  \parbox[c]{0.215\linewidth}{\centering \includegraphics[width=\linewidth]{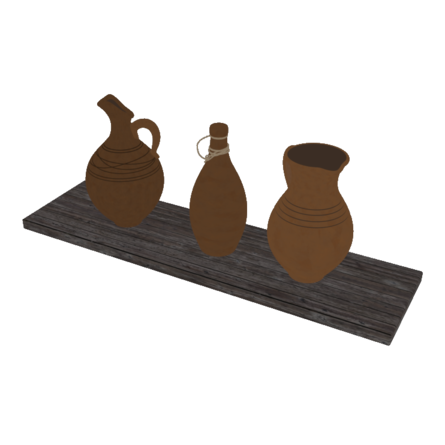}}%
  \hspace{2pt}%
  \parbox[c]{0.215\linewidth}{\centering \includegraphics[width=\linewidth]{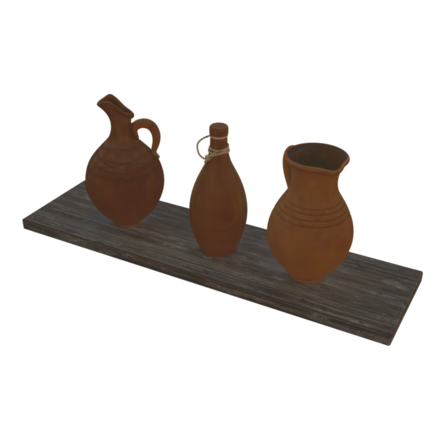}}%
  \hspace{2pt}%
  \parbox[c]{0.215\linewidth}{\centering \includegraphics[width=\linewidth]{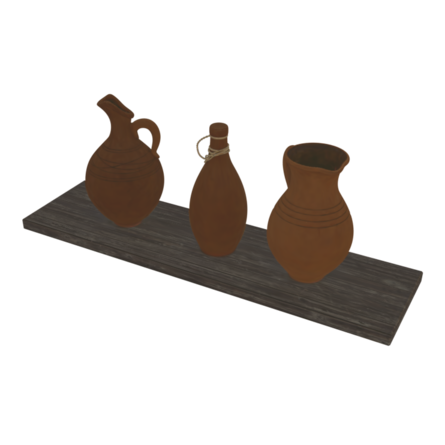}}%
  \\[0.1em]
  \parbox[c]{0.09\linewidth}{\centering \intrinsicrowlabel{Roughness}}%
  \hspace{2pt}%
  \parbox[c]{0.215\linewidth}{\centering \begin{overpic}[width=\linewidth]{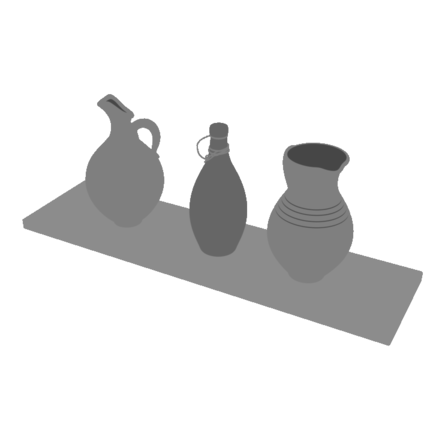}\put(18.598,53.639){\color{red}\linethickness{0.8pt}\framebox(20.216,20.216){}}\end{overpic}}%
  \hspace{2pt}%
  \parbox[c]{0.215\linewidth}{\centering \begin{overpic}[width=\linewidth]{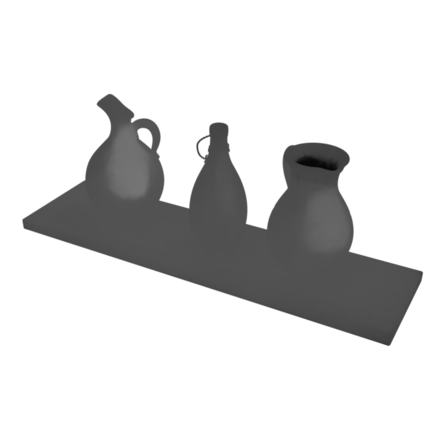}\put(18.598,53.639){\color{red}\linethickness{0.8pt}\framebox(20.216,20.216){}}\end{overpic}}%
  \hspace{2pt}%
  \parbox[c]{0.215\linewidth}{\centering \begin{overpic}[width=\linewidth]{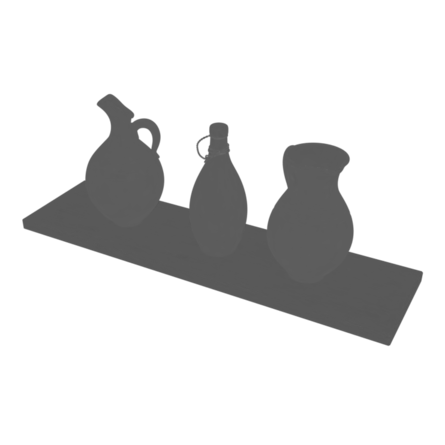}\put(18.598,53.639){\color{red}\linethickness{0.8pt}\framebox(20.216,20.216){}}\end{overpic}}%
  \\[0.1em]
  \parbox[c]{0.09\linewidth}{\centering \intrinsicrowlabel{Roughness Zoom}}%
  \hspace{2pt}%
  \parbox[c]{0.215\linewidth}{\centering \includegraphics[width=\linewidth]{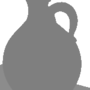}}%
  \hspace{2pt}%
  \parbox[c]{0.215\linewidth}{\centering \includegraphics[width=\linewidth]{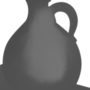}}%
  \hspace{2pt}%
  \parbox[c]{0.215\linewidth}{\centering \includegraphics[width=\linewidth]{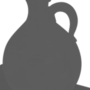}}%
  \\[0.1em]
  \parbox[c]{0.09\linewidth}{\centering \intrinsicrowlabel{Normal}}%
  \hspace{2pt}%
  \parbox[c]{0.215\linewidth}{\centering \begin{overpic}[width=\linewidth]{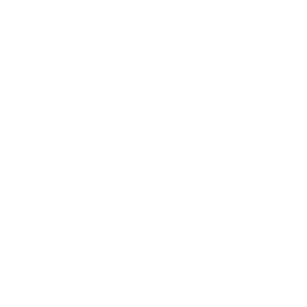}\put(50,50){\makebox(0,0){\scriptsize N/A}}\end{overpic}}%
  \hspace{2pt}%
  \parbox[c]{0.215\linewidth}{\centering \includegraphics[width=\linewidth]{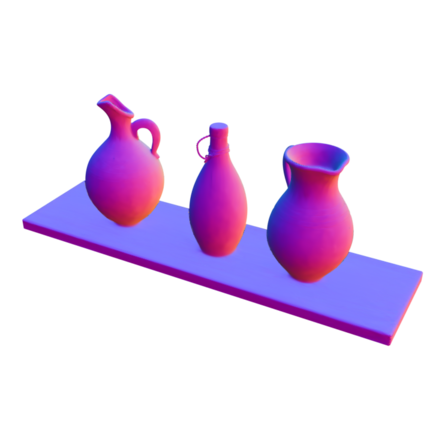}}%
  \hspace{2pt}%
  \parbox[c]{0.215\linewidth}{\centering \includegraphics[width=\linewidth]{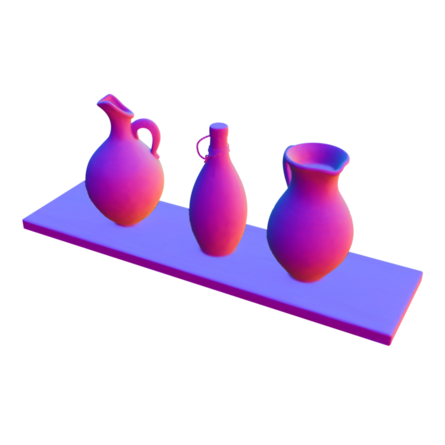}}%
  \\[1pt]
  \makebox[0.09\linewidth][c]{}%
  \hspace{2pt}%
  \makebox[0.215\linewidth][c]{\scriptsize Reference}%
  \hspace{2pt}%
  \makebox[0.215\linewidth][c]{\scriptsize IRGS}%
  \hspace{2pt}%
  \makebox[0.215\linewidth][c]{\scriptsize IRGS + JBF (Ours)}%
  \\[1pt]
  \caption{Synthetic4Relight \casename{jugs} (view \texttt{158}). JBF material regularization smooths base color and roughness G-buffers, yielding improved relighting quality.}
  \label{fig:supp-irgs-jbf-jugs}
\end{figure*}

\paragraph{Qualitative results.}
\Cref{fig:supp-irgs-jbf-air-baloons,fig:supp-irgs-jbf-chair,fig:supp-irgs-jbf-hotdog,fig:supp-irgs-jbf-jugs} show representative relighting and intrinsic comparisons for all four scenes.
The top row shows the input image together with DiffusionRenderer-predicted G-buffers, which provide the guidance signal for our regularization.
Compared with the original IRGS smoothness terms, our regularization reconstructs richer base-color details while suppressing baked-in illumination, yielding cleaner relighting results.
This behavior is especially visible in regions where the original IRGS reconstruction either over-smooths texture or absorbs shading into the optimized roughness.

\section{Intrinsic G-Buffer Visualization for Comparison and Ablation}
\label{sec:supp-intrinsic-gbuffer-comparison-ablation}

This section provides per-scene intrinsic G-buffer visualizations supplementing the qualitative figures in the main paper.
Each figure shows, for all compared methods, the diffusion-predicted G-buffers alongside the optimized base color, roughness, reconstructed surface normal, and estimated lighting. N/A indicates that the corresponding reference result is not available.
For dynamic relighting and video-based visualizations, please refer to the supplementary video.

\begin{figure*}[t]
  \centering
  \newcommand{\intrinsicrowlabel}[1]{\rotatebox[origin=c]{90}{\scriptsize\strut #1}}
  {\scriptsize\texttt{\detokenize{Stanford-ORB car_scene002}}}\\[-0.2em]
  \parbox[c]{0.05\linewidth}{\centering \intrinsicrowlabel{Input and Prediction}}%
  \hspace{2pt}%
  \parbox[c]{0.172\linewidth}{\centering \includegraphics[width=\linewidth]{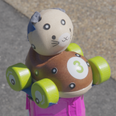}}%
  \hspace{2pt}%
  \parbox[c]{0.172\linewidth}{\centering \includegraphics[width=\linewidth]{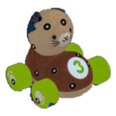}}%
  \hspace{2pt}%
  \parbox[c]{0.172\linewidth}{\centering \includegraphics[width=\linewidth]{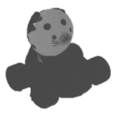}}%
  \hspace{2pt}%
  \parbox[c]{0.172\linewidth}{\centering \includegraphics[width=\linewidth]{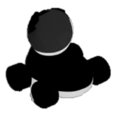}}%
  \hspace{2pt}%
  \parbox[c]{0.172\linewidth}{\centering \includegraphics[width=\linewidth]{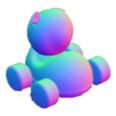}}%
  \\[1pt]
  \makebox[0.05\linewidth][c]{}%
  \hspace{2pt}%
  \makebox[0.172\linewidth][c]{\scriptsize Input}%
  \hspace{2pt}%
  \makebox[0.172\linewidth][c]{\scriptsize DR-base color}%
  \hspace{2pt}%
  \makebox[0.172\linewidth][c]{\scriptsize DR-roughness}%
  \hspace{2pt}%
  \makebox[0.172\linewidth][c]{\scriptsize DR-metallic}%
  \hspace{2pt}%
  \makebox[0.172\linewidth][c]{\scriptsize DR-normal}%
  \\[1pt]
  \vspace{0.15em}
  \par\noindent\rule{\linewidth}{0.35pt}
  \vspace{0.05em}
  \vspace{-0.6em}
  \parbox[c]{0.05\linewidth}{\centering \intrinsicrowlabel{Relight}}%
  \hspace{2pt}%
  \parbox[c]{0.215\linewidth}{\centering \begin{overpic}[width=\linewidth]{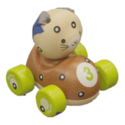}\put(0,0){\includegraphics[width=0.666\linewidth]{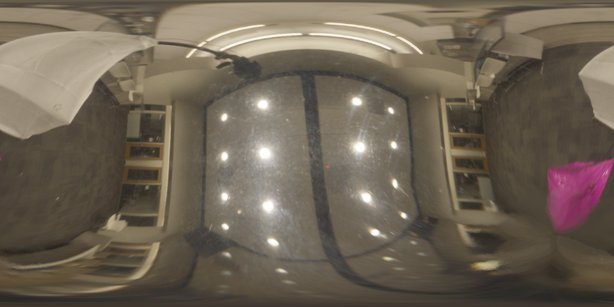}}\end{overpic}}%
  \hspace{2pt}%
  \parbox[c]{0.215\linewidth}{\centering \includegraphics[width=\linewidth]{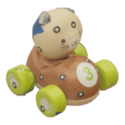}}%
  \hspace{2pt}%
  \parbox[c]{0.215\linewidth}{\centering \includegraphics[width=\linewidth]{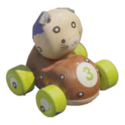}}%
  \hspace{2pt}%
  \parbox[c]{0.215\linewidth}{\centering \includegraphics[width=\linewidth]{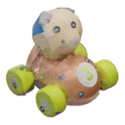}}%
  \\[0.1em]
  \parbox[c]{0.05\linewidth}{\centering \intrinsicrowlabel{Base Color}}%
  \hspace{2pt}%
  \parbox[c]{0.215\linewidth}{\centering \includegraphics[width=\linewidth]{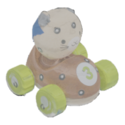}}%
  \hspace{2pt}%
  \parbox[c]{0.215\linewidth}{\centering \includegraphics[width=\linewidth]{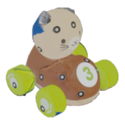}}%
  \hspace{2pt}%
  \parbox[c]{0.215\linewidth}{\centering \includegraphics[width=\linewidth]{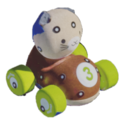}}%
  \hspace{2pt}%
  \parbox[c]{0.215\linewidth}{\centering \includegraphics[width=\linewidth]{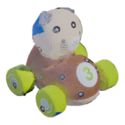}}%
  \\[0.1em]
  \parbox[c]{0.05\linewidth}{\centering \intrinsicrowlabel{Roughness}}%
  \hspace{2pt}%
  \parbox[c]{0.215\linewidth}{\centering \begin{overpic}[width=\linewidth]{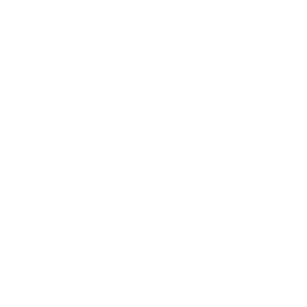}\put(50,50){\makebox(0,0){\scriptsize N/A}}\end{overpic}}%
  \hspace{2pt}%
  \parbox[c]{0.215\linewidth}{\centering \includegraphics[width=\linewidth]{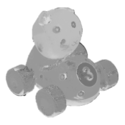}}%
  \hspace{2pt}%
  \parbox[c]{0.215\linewidth}{\centering \includegraphics[width=\linewidth]{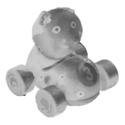}}%
  \hspace{2pt}%
  \parbox[c]{0.215\linewidth}{\centering \includegraphics[width=\linewidth]{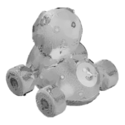}}%
  \\[0.1em]
  \parbox[c]{0.05\linewidth}{\centering \intrinsicrowlabel{Normal}}%
  \hspace{2pt}%
  \parbox[c]{0.215\linewidth}{\centering \includegraphics[width=\linewidth]{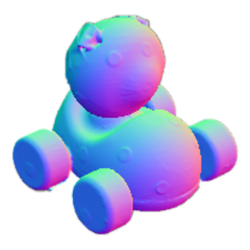}}%
  \hspace{2pt}%
  \parbox[c]{0.215\linewidth}{\centering \includegraphics[width=\linewidth]{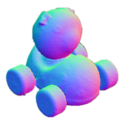}}%
  \hspace{2pt}%
  \parbox[c]{0.215\linewidth}{\centering \includegraphics[width=\linewidth]{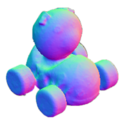}}%
  \hspace{2pt}%
  \parbox[c]{0.215\linewidth}{\centering \includegraphics[width=\linewidth]{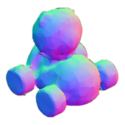}}%
  \\[0.1em]
  \parbox[c]{0.05\linewidth}{\centering \intrinsicrowlabel{Lighting}}%
  \hspace{2pt}%
  \parbox[c]{0.215\linewidth}{\centering \includegraphics[width=\linewidth]{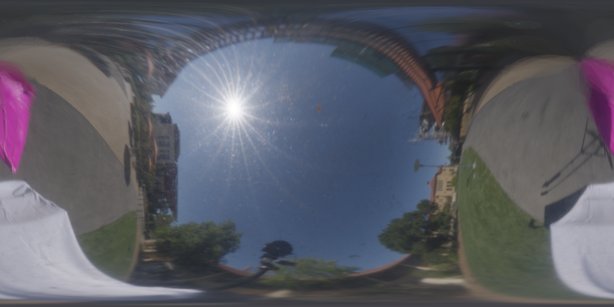}}%
  \hspace{2pt}%
  \parbox[c]{0.215\linewidth}{\centering \includegraphics[width=\linewidth]{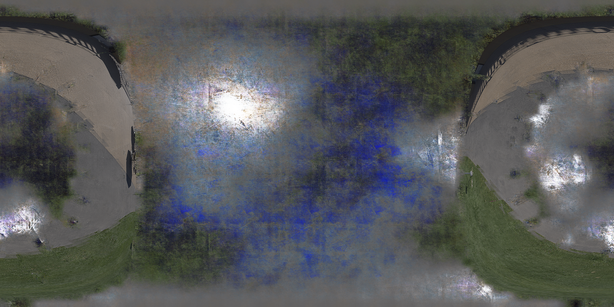}}%
  \hspace{2pt}%
  \parbox[c]{0.215\linewidth}{\centering \includegraphics[width=\linewidth]{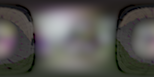}}%
  \hspace{2pt}%
  \parbox[c]{0.215\linewidth}{\centering \includegraphics[width=\linewidth]{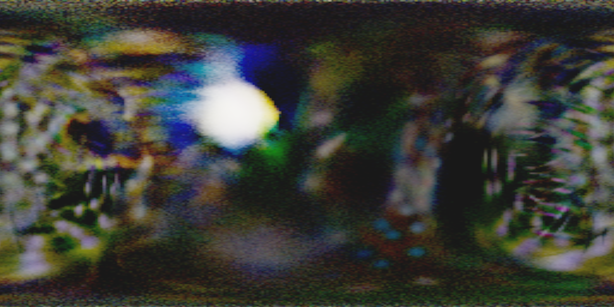}}%
  \\[1pt]
  \makebox[0.05\linewidth][c]{}%
  \hspace{2pt}%
  \makebox[0.215\linewidth][c]{\scriptsize Reference}%
  \hspace{2pt}%
  \makebox[0.215\linewidth][c]{\scriptsize Ours}%
  \hspace{2pt}%
  \makebox[0.215\linewidth][c]{\scriptsize Neural-PBIR}%
  \hspace{2pt}%
  \makebox[0.215\linewidth][c]{\scriptsize MaterialFusion}%
  \\[1pt]
  \caption{Stanford-ORB \casename{car_scene002}.}
  \label{fig:supp-stanford-intrinsics-car-scene002}
\end{figure*}

\begin{figure*}[t]
  \centering
  \newcommand{\intrinsicrowlabel}[1]{\rotatebox[origin=c]{90}{\scriptsize\strut #1}}
  {\scriptsize\texttt{\detokenize{Stanford-ORB cup_scene006}}}\\[-0.2em]
  \parbox[c]{0.05\linewidth}{\centering \intrinsicrowlabel{Input and Prediction}}%
  \hspace{2pt}%
  \parbox[c]{0.172\linewidth}{\centering \includegraphics[width=\linewidth]{figures/summary/stanford_orb_cup_scene006_intrinsic_assets/input_dr/input.png}}%
  \hspace{2pt}%
  \parbox[c]{0.172\linewidth}{\centering \includegraphics[width=\linewidth]{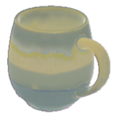}}%
  \hspace{2pt}%
  \parbox[c]{0.172\linewidth}{\centering \includegraphics[width=\linewidth]{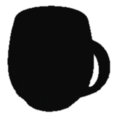}}%
  \hspace{2pt}%
  \parbox[c]{0.172\linewidth}{\centering \includegraphics[width=\linewidth]{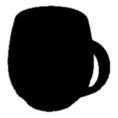}}%
  \hspace{2pt}%
  \parbox[c]{0.172\linewidth}{\centering \includegraphics[width=\linewidth]{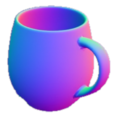}}%
  \\[1pt]
  \makebox[0.05\linewidth][c]{}%
  \hspace{2pt}%
  \makebox[0.172\linewidth][c]{\scriptsize Input}%
  \hspace{2pt}%
  \makebox[0.172\linewidth][c]{\scriptsize DR-base color}%
  \hspace{2pt}%
  \makebox[0.172\linewidth][c]{\scriptsize DR-roughness}%
  \hspace{2pt}%
  \makebox[0.172\linewidth][c]{\scriptsize DR-metallic}%
  \hspace{2pt}%
  \makebox[0.172\linewidth][c]{\scriptsize DR-normal}%
  \\[1pt]
  \vspace{0.15em}
  \par\noindent\rule{\linewidth}{0.35pt}
  \vspace{0.05em}
  \vspace{-0.6em}
  \parbox[c]{0.05\linewidth}{\centering \intrinsicrowlabel{Relight}}%
  \hspace{2pt}%
  \parbox[c]{0.215\linewidth}{\centering \begin{overpic}[width=\linewidth]{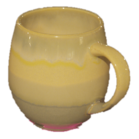}\put(0,0){\includegraphics[width=0.666\linewidth]{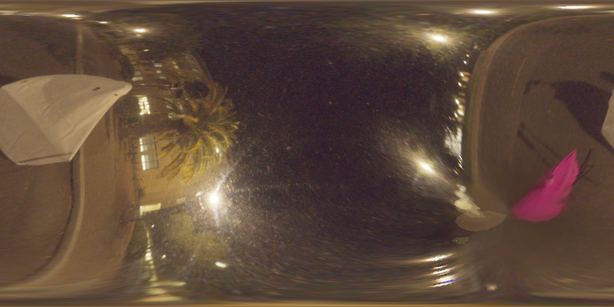}}\end{overpic}}%
  \hspace{2pt}%
  \parbox[c]{0.215\linewidth}{\centering \includegraphics[width=\linewidth]{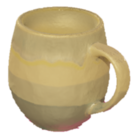}}%
  \hspace{2pt}%
  \parbox[c]{0.215\linewidth}{\centering \includegraphics[width=\linewidth]{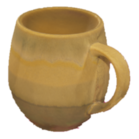}}%
  \hspace{2pt}%
  \parbox[c]{0.215\linewidth}{\centering \includegraphics[width=\linewidth]{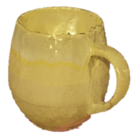}}%
  \\[0.1em]
  \parbox[c]{0.05\linewidth}{\centering \intrinsicrowlabel{Base Color}}%
  \hspace{2pt}%
  \parbox[c]{0.215\linewidth}{\centering \includegraphics[width=\linewidth]{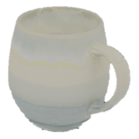}}%
  \hspace{2pt}%
  \parbox[c]{0.215\linewidth}{\centering \includegraphics[width=\linewidth]{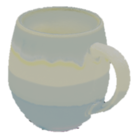}}%
  \hspace{2pt}%
  \parbox[c]{0.215\linewidth}{\centering \includegraphics[width=\linewidth]{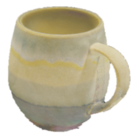}}%
  \hspace{2pt}%
  \parbox[c]{0.215\linewidth}{\centering \includegraphics[width=\linewidth]{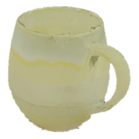}}%
  \\[0.1em]
  \parbox[c]{0.05\linewidth}{\centering \intrinsicrowlabel{Roughness}}%
  \hspace{2pt}%
  \parbox[c]{0.215\linewidth}{\centering \begin{overpic}[width=\linewidth]{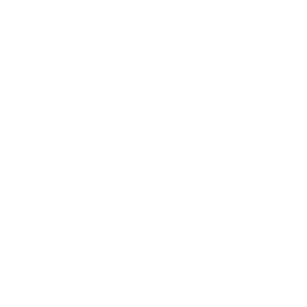}\put(50,50){\makebox(0,0){\scriptsize N/A}}\end{overpic}}%
  \hspace{2pt}%
  \parbox[c]{0.215\linewidth}{\centering \includegraphics[width=\linewidth]{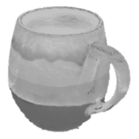}}%
  \hspace{2pt}%
  \parbox[c]{0.215\linewidth}{\centering \includegraphics[width=\linewidth]{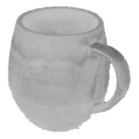}}%
  \hspace{2pt}%
  \parbox[c]{0.215\linewidth}{\centering \includegraphics[width=\linewidth]{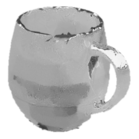}}%
  \\[0.1em]
  \parbox[c]{0.05\linewidth}{\centering \intrinsicrowlabel{Normal}}%
  \hspace{2pt}%
  \parbox[c]{0.215\linewidth}{\centering \includegraphics[width=\linewidth]{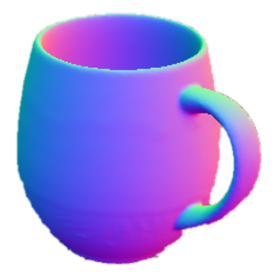}}%
  \hspace{2pt}%
  \parbox[c]{0.215\linewidth}{\centering \includegraphics[width=\linewidth]{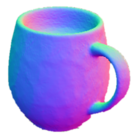}}%
  \hspace{2pt}%
  \parbox[c]{0.215\linewidth}{\centering \includegraphics[width=\linewidth]{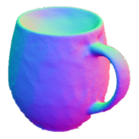}}%
  \hspace{2pt}%
  \parbox[c]{0.215\linewidth}{\centering \includegraphics[width=\linewidth]{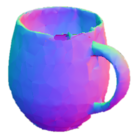}}%
  \\[0.1em]
  \parbox[c]{0.05\linewidth}{\centering \intrinsicrowlabel{Lighting}}%
  \hspace{2pt}%
  \parbox[c]{0.215\linewidth}{\centering \includegraphics[width=\linewidth]{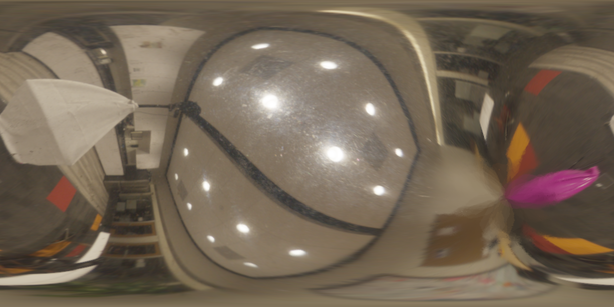}}%
  \hspace{2pt}%
  \parbox[c]{0.215\linewidth}{\centering \includegraphics[width=\linewidth]{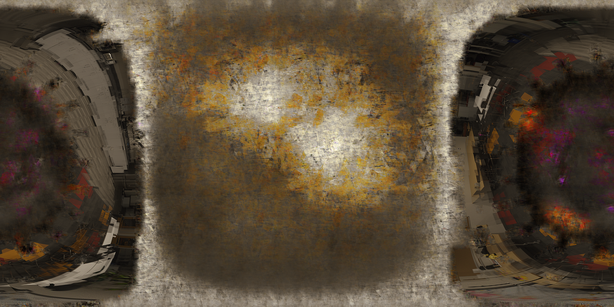}}%
  \hspace{2pt}%
  \parbox[c]{0.215\linewidth}{\centering \includegraphics[width=\linewidth]{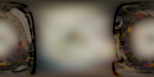}}%
  \hspace{2pt}%
  \parbox[c]{0.215\linewidth}{\centering \includegraphics[width=\linewidth]{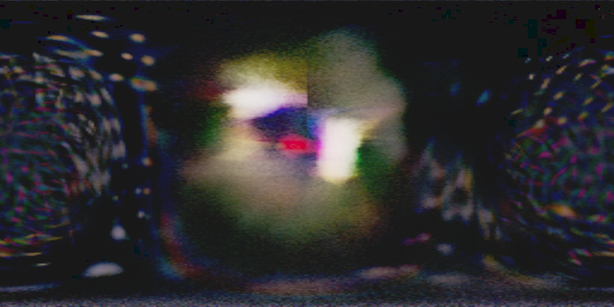}}%
  \\[1pt]
  \makebox[0.05\linewidth][c]{}%
  \hspace{2pt}%
  \makebox[0.215\linewidth][c]{\scriptsize Reference}%
  \hspace{2pt}%
  \makebox[0.215\linewidth][c]{\scriptsize Ours}%
  \hspace{2pt}%
  \makebox[0.215\linewidth][c]{\scriptsize Neural-PBIR}%
  \hspace{2pt}%
  \makebox[0.215\linewidth][c]{\scriptsize MaterialFusion}%
  \\[1pt]
  \caption{Stanford-ORB \casename{cup_scene006}.}
  \label{fig:supp-stanford-intrinsics-cup-scene006}
\end{figure*}

\begin{figure*}[t]
  \centering
  \newcommand{\intrinsicrowlabel}[1]{\rotatebox[origin=c]{90}{\scriptsize\strut #1}}
  {\scriptsize\texttt{\detokenize{DTC-Synthetic TeaPot_B084G3K8TD_YellowBlackSunflowers_TU_scene002}}}\\[-0.2em]
  \parbox[c]{0.05\linewidth}{\centering \intrinsicrowlabel{Input and Prediction}}%
  \hspace{2pt}%
  \parbox[c]{0.172\linewidth}{\centering \includegraphics[width=\linewidth]{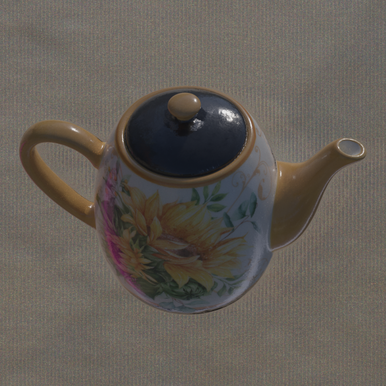}}%
  \hspace{2pt}%
  \parbox[c]{0.172\linewidth}{\centering \includegraphics[width=\linewidth]{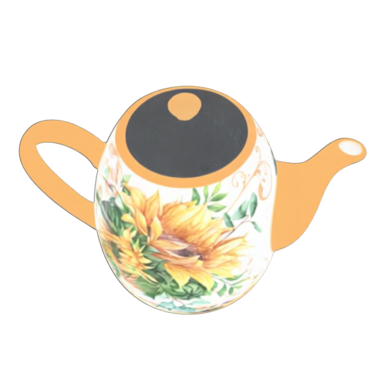}}%
  \hspace{2pt}%
  \parbox[c]{0.172\linewidth}{\centering \includegraphics[width=\linewidth]{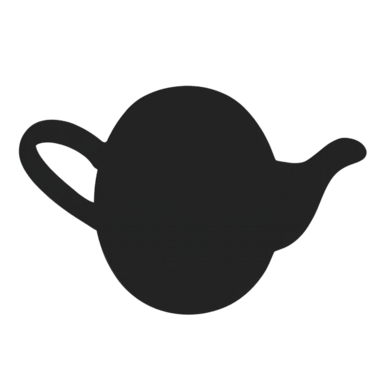}}%
  \hspace{2pt}%
  \parbox[c]{0.172\linewidth}{\centering \includegraphics[width=\linewidth]{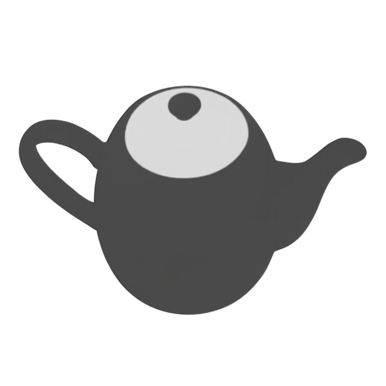}}%
  \hspace{2pt}%
  \parbox[c]{0.172\linewidth}{\centering \includegraphics[width=\linewidth]{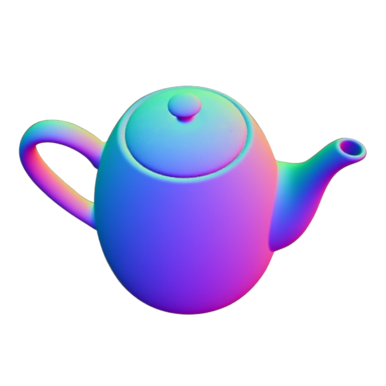}}%
  \\[1pt]
  \makebox[0.05\linewidth][c]{}%
  \hspace{2pt}%
  \makebox[0.172\linewidth][c]{\scriptsize Input}%
  \hspace{2pt}%
  \makebox[0.172\linewidth][c]{\scriptsize DR-base color}%
  \hspace{2pt}%
  \makebox[0.172\linewidth][c]{\scriptsize DR-roughness}%
  \hspace{2pt}%
  \makebox[0.172\linewidth][c]{\scriptsize DR-metallic}%
  \hspace{2pt}%
  \makebox[0.172\linewidth][c]{\scriptsize DR-normal}%
  \\[1pt]
  \vspace{0.15em}
  \par\noindent\rule{\linewidth}{0.35pt}
  \vspace{0.05em}
  \vspace{-0.6em}
  \parbox[c]{0.05\linewidth}{\centering \intrinsicrowlabel{Relight}}%
  \hspace{2pt}%
  \parbox[c]{0.215\linewidth}{\centering \begin{overpic}[width=\linewidth]{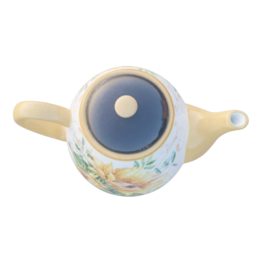}\put(0,0){\includegraphics[width=0.666\linewidth]{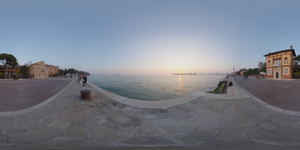}}\end{overpic}}%
  \hspace{2pt}%
  \parbox[c]{0.215\linewidth}{\centering \includegraphics[width=\linewidth]{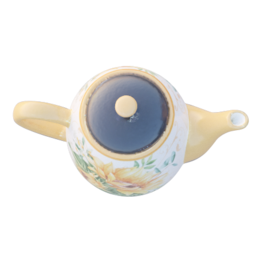}}%
  \hspace{2pt}%
  \parbox[c]{0.215\linewidth}{\centering \includegraphics[width=\linewidth]{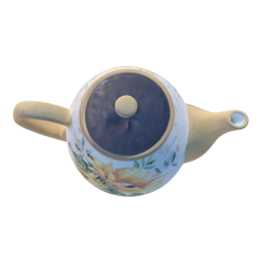}}%
  \hspace{2pt}%
  \parbox[c]{0.215\linewidth}{\centering \includegraphics[width=\linewidth]{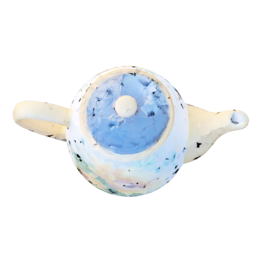}}%
  \\[0.1em]
  \parbox[c]{0.05\linewidth}{\centering \intrinsicrowlabel{Base Color}}%
  \hspace{2pt}%
  \parbox[c]{0.215\linewidth}{\centering \includegraphics[width=\linewidth]{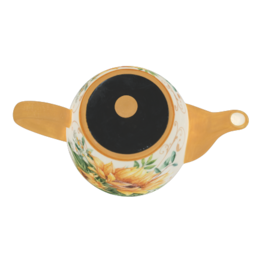}}%
  \hspace{2pt}%
  \parbox[c]{0.215\linewidth}{\centering \includegraphics[width=\linewidth]{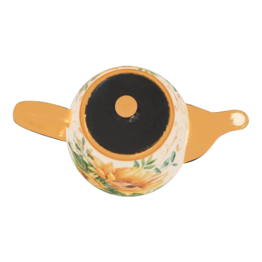}}%
  \hspace{2pt}%
  \parbox[c]{0.215\linewidth}{\centering \includegraphics[width=\linewidth]{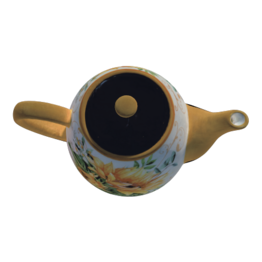}}%
  \hspace{2pt}%
  \parbox[c]{0.215\linewidth}{\centering \includegraphics[width=\linewidth]{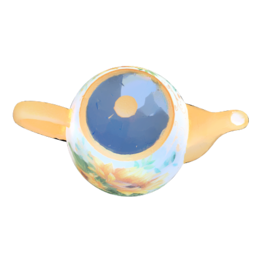}}%
  \\[0.1em]
  \parbox[c]{0.05\linewidth}{\centering \intrinsicrowlabel{Roughness}}%
  \hspace{2pt}%
  \parbox[c]{0.215\linewidth}{\centering \includegraphics[width=\linewidth]{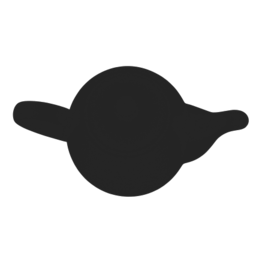}}%
  \hspace{2pt}%
  \parbox[c]{0.215\linewidth}{\centering \includegraphics[width=\linewidth]{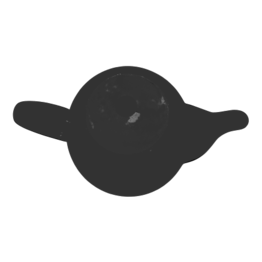}}%
  \hspace{2pt}%
  \parbox[c]{0.215\linewidth}{\centering \includegraphics[width=\linewidth]{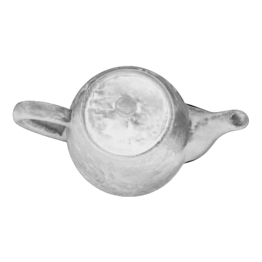}}%
  \hspace{2pt}%
  \parbox[c]{0.215\linewidth}{\centering \includegraphics[width=\linewidth]{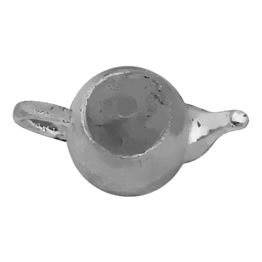}}%
  \\[0.1em]
  \parbox[c]{0.05\linewidth}{\centering \intrinsicrowlabel{Metallic}}%
  \hspace{2pt}%
  \parbox[c]{0.215\linewidth}{\centering \includegraphics[width=\linewidth]{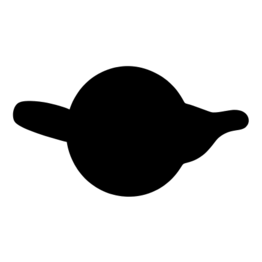}}%
  \hspace{2pt}%
  \parbox[c]{0.215\linewidth}{\centering \includegraphics[width=\linewidth]{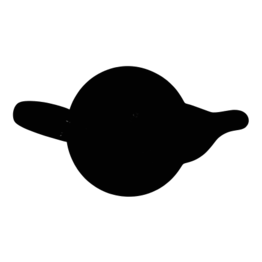}}%
  \hspace{2pt}%
  \parbox[c]{0.215\linewidth}{\centering \includegraphics[width=\linewidth]{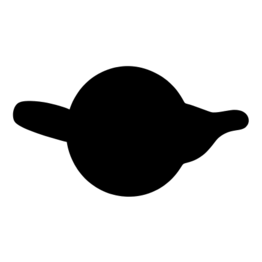}}%
  \hspace{2pt}%
  \parbox[c]{0.215\linewidth}{\centering \includegraphics[width=\linewidth]{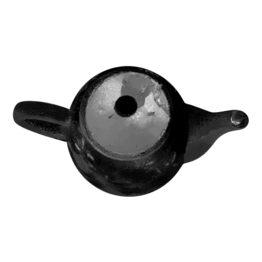}}%
  \\[0.1em]
  \parbox[c]{0.05\linewidth}{\centering \intrinsicrowlabel{Normal}}%
  \hspace{2pt}%
  \parbox[c]{0.215\linewidth}{\centering \includegraphics[width=\linewidth]{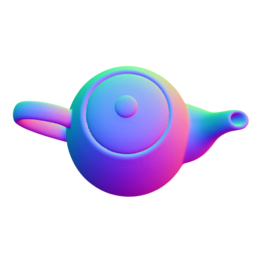}}%
  \hspace{2pt}%
  \parbox[c]{0.215\linewidth}{\centering \includegraphics[width=\linewidth]{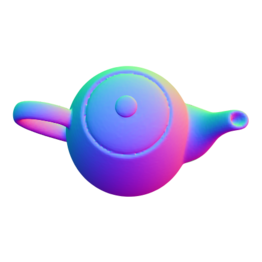}}%
  \hspace{2pt}%
  \parbox[c]{0.215\linewidth}{\centering \includegraphics[width=\linewidth]{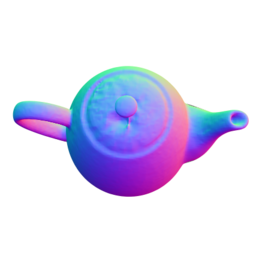}}%
  \hspace{2pt}%
  \parbox[c]{0.215\linewidth}{\centering \includegraphics[width=\linewidth]{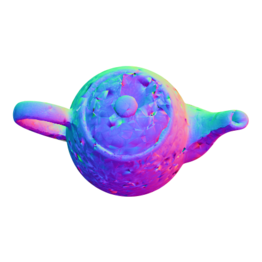}}%
  \\[0.1em]
  \parbox[c]{0.05\linewidth}{\centering \intrinsicrowlabel{Lighting}}%
  \hspace{2pt}%
  \parbox[c]{0.215\linewidth}{\centering \includegraphics[width=\linewidth]{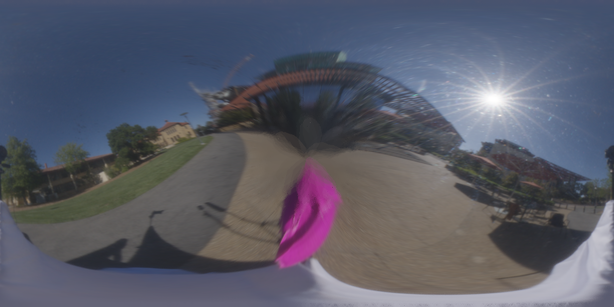}}%
  \hspace{2pt}%
  \parbox[c]{0.215\linewidth}{\centering \includegraphics[width=\linewidth]{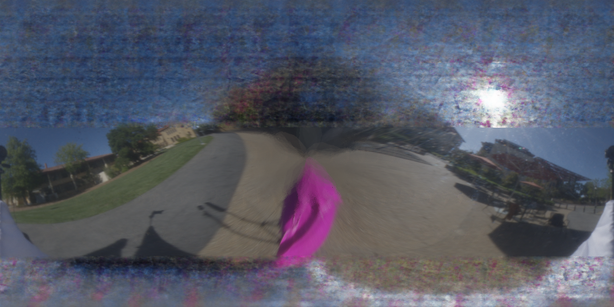}}%
  \hspace{2pt}%
  \parbox[c]{0.215\linewidth}{\centering \includegraphics[width=\linewidth]{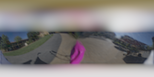}}%
  \hspace{2pt}%
  \parbox[c]{0.215\linewidth}{\centering \includegraphics[width=\linewidth]{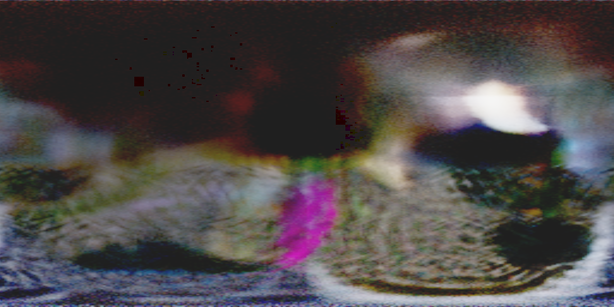}}%
  \\[1pt]
  \makebox[0.05\linewidth][c]{}%
  \hspace{2pt}%
  \makebox[0.215\linewidth][c]{\scriptsize Reference}%
  \hspace{2pt}%
  \makebox[0.215\linewidth][c]{\scriptsize Ours}%
  \hspace{2pt}%
  \makebox[0.215\linewidth][c]{\scriptsize Neural-PBIR}%
  \hspace{2pt}%
  \makebox[0.215\linewidth][c]{\scriptsize MaterialFusion}%
  \\[1pt]
  \caption{DTC-Synthetic \casename{TeaPot_B084G3K8TD_YellowBlackSunflowers_TU_scene002}.}
  \label{fig:supp-dtc-intrinsics-teapot_b084g3k8td_yellowblacksunflowers_tu_scene002}
\end{figure*}

The input images are included to make the capture conditions visible; for example, \cref{fig:supp-stanford-intrinsics-car-scene002,fig:supp-stanford-intrinsics-cup-scene006} show hard cast shadows under strong directional lighting, while our reconstruction removes the shadow baking that persists in Neural-PBIR.
The estimated environment maps should be interpreted together with the object's reflectance: weakly glossy scenes, such as \cref{fig:supp-stanford-intrinsics-cup-scene006}, provide limited constraints on high-frequency illumination, whereas glossier scenes, such as \cref{fig:supp-dtc-intrinsics-teapot_b084g3k8td_yellowblacksunflowers_tu_scene002}, allow our reconstruction to recover sharper lighting than the baselines.
In the video, ground truth is omitted only when the dataset does not release it, such as Stanford-ORB relighting videos or roughness maps and Synthetic4Relight normal maps.

\begin{figure*}[t]
  \centering
  \newcommand{\intrinsicrowlabel}[1]{\rotatebox[origin=c]{90}{\scriptsize\strut #1}}
  {\scriptsize\texttt{\detokenize{Stanford-ORB pitcher_scene001}}}\\[-0.2em]
  \parbox[c]{0.05\linewidth}{\centering \intrinsicrowlabel{Input and Prediction}}%
  \hspace{2pt}%
  \parbox[c]{0.172\linewidth}{\centering \includegraphics[width=\linewidth]{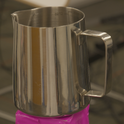}}%
  \hspace{2pt}%
  \parbox[c]{0.172\linewidth}{\centering \includegraphics[width=\linewidth]{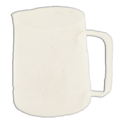}}%
  \hspace{2pt}%
  \parbox[c]{0.172\linewidth}{\centering \includegraphics[width=\linewidth]{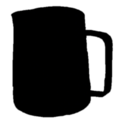}}%
  \hspace{2pt}%
  \parbox[c]{0.172\linewidth}{\centering \includegraphics[width=\linewidth]{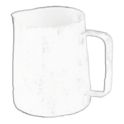}}%
  \hspace{2pt}%
  \parbox[c]{0.172\linewidth}{\centering \includegraphics[width=\linewidth]{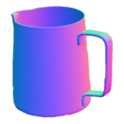}}%
  \\[1pt]
  \makebox[0.05\linewidth][c]{}%
  \hspace{2pt}%
  \makebox[0.172\linewidth][c]{\scriptsize Input}%
  \hspace{2pt}%
  \makebox[0.172\linewidth][c]{\scriptsize DR-base color}%
  \hspace{2pt}%
  \makebox[0.172\linewidth][c]{\scriptsize DR-roughness}%
  \hspace{2pt}%
  \makebox[0.172\linewidth][c]{\scriptsize DR-metallic}%
  \hspace{2pt}%
  \makebox[0.172\linewidth][c]{\scriptsize DR-normal}%
  \\[1pt]
  \vspace{0.15em}
  \par\noindent\rule{\linewidth}{0.35pt}
  \vspace{0.05em}
  \vspace{-0.6em}
  \parbox[c]{0.05\linewidth}{\centering \intrinsicrowlabel{Relight}}%
  \hspace{2pt}%
  \parbox[c]{0.215\linewidth}{\centering \begin{overpic}[width=\linewidth]{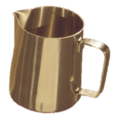}\put(0,0){\includegraphics[width=0.666\linewidth]{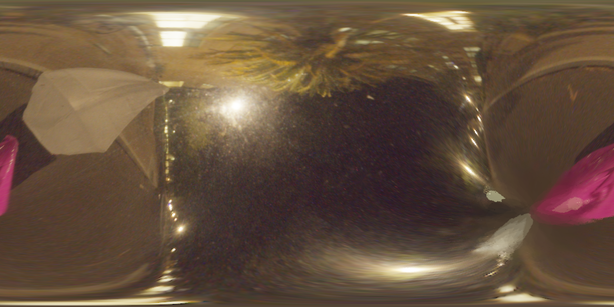}}\end{overpic}}%
  \hspace{2pt}%
  \parbox[c]{0.215\linewidth}{\centering \includegraphics[width=\linewidth]{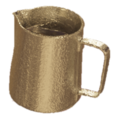}}%
  \hspace{2pt}%
  \parbox[c]{0.215\linewidth}{\centering \includegraphics[width=\linewidth]{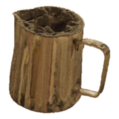}}%
  \hspace{2pt}%
  \parbox[c]{0.215\linewidth}{\centering \includegraphics[width=\linewidth]{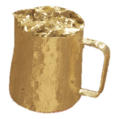}}%
  \\[0.1em]
  \parbox[c]{0.05\linewidth}{\centering \intrinsicrowlabel{Base Color}}%
  \hspace{2pt}%
  \parbox[c]{0.215\linewidth}{\centering \includegraphics[width=\linewidth]{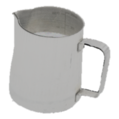}}%
  \hspace{2pt}%
  \parbox[c]{0.215\linewidth}{\centering \includegraphics[width=\linewidth]{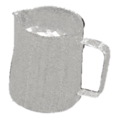}}%
  \hspace{2pt}%
  \parbox[c]{0.215\linewidth}{\centering \includegraphics[width=\linewidth]{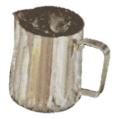}}%
  \hspace{2pt}%
  \parbox[c]{0.215\linewidth}{\centering \includegraphics[width=\linewidth]{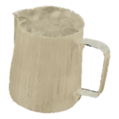}}%
  \\[0.1em]
  \parbox[c]{0.05\linewidth}{\centering \intrinsicrowlabel{Roughness}}%
  \hspace{2pt}%
  \parbox[c]{0.215\linewidth}{\centering \begin{overpic}[width=\linewidth]{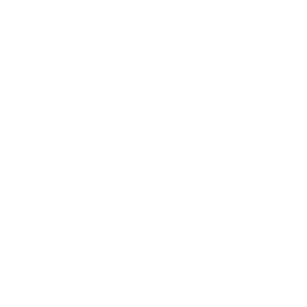}\put(50,50){\makebox(0,0){\scriptsize N/A}}\end{overpic}}%
  \hspace{2pt}%
  \parbox[c]{0.215\linewidth}{\centering \includegraphics[width=\linewidth]{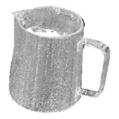}}%
  \hspace{2pt}%
  \parbox[c]{0.215\linewidth}{\centering \includegraphics[width=\linewidth]{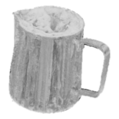}}%
  \hspace{2pt}%
  \parbox[c]{0.215\linewidth}{\centering \includegraphics[width=\linewidth]{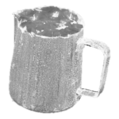}}%
  \\[0.1em]
  \parbox[c]{0.05\linewidth}{\centering \intrinsicrowlabel{Normal}}%
  \hspace{2pt}%
  \parbox[c]{0.215\linewidth}{\centering \includegraphics[width=\linewidth]{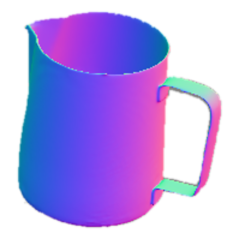}}%
  \hspace{2pt}%
  \parbox[c]{0.215\linewidth}{\centering \includegraphics[width=\linewidth]{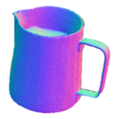}}%
  \hspace{2pt}%
  \parbox[c]{0.215\linewidth}{\centering \includegraphics[width=\linewidth]{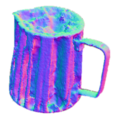}}%
  \hspace{2pt}%
  \parbox[c]{0.215\linewidth}{\centering \includegraphics[width=\linewidth]{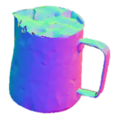}}%
  \\[0.1em]
  \parbox[c]{0.05\linewidth}{\centering \intrinsicrowlabel{Lighting}}%
  \hspace{2pt}%
  \parbox[c]{0.215\linewidth}{\centering \includegraphics[width=\linewidth]{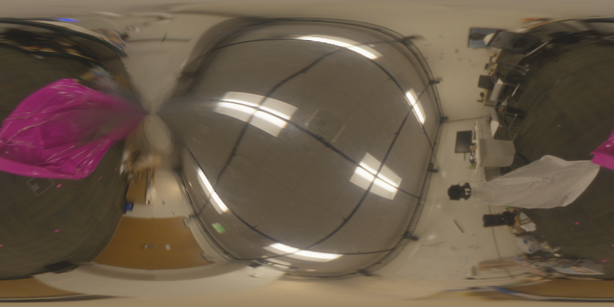}}%
  \hspace{2pt}%
  \parbox[c]{0.215\linewidth}{\centering \includegraphics[width=\linewidth]{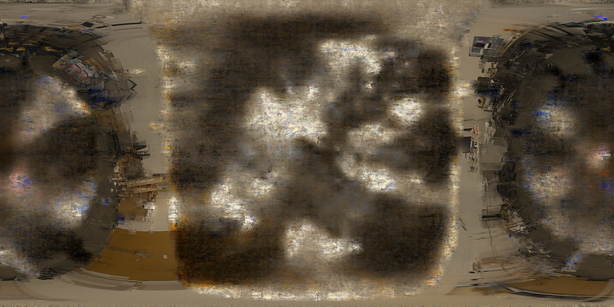}}%
  \hspace{2pt}%
  \parbox[c]{0.215\linewidth}{\centering \includegraphics[width=\linewidth]{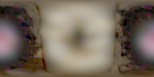}}%
  \hspace{2pt}%
  \parbox[c]{0.215\linewidth}{\centering \includegraphics[width=\linewidth]{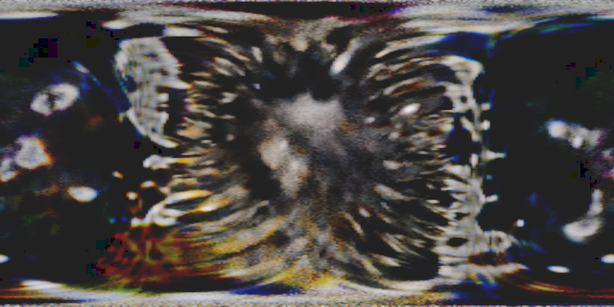}}%
  \\[1pt]
  \makebox[0.05\linewidth][c]{}%
  \hspace{2pt}%
  \makebox[0.215\linewidth][c]{\scriptsize Reference}%
  \hspace{2pt}%
  \makebox[0.215\linewidth][c]{\scriptsize Ours}%
  \hspace{2pt}%
  \makebox[0.215\linewidth][c]{\scriptsize Neural-PBIR}%
  \hspace{2pt}%
  \makebox[0.215\linewidth][c]{\scriptsize MaterialFusion}%
  \\[1pt]
  \caption{Stanford-ORB \casename{pitcher_scene001}.}
  \label{fig:supp-stanford-intrinsics-pitcher-scene001}
\end{figure*}

\begin{figure*}[t]
  \centering
  \newcommand{\intrinsicrowlabel}[1]{\rotatebox[origin=c]{90}{\scriptsize\strut #1}}
  {\scriptsize\texttt{\detokenize{DTC-Synthetic TeaPot_B094FQW6Q4_EmeraldGoldTop_scene002}}}\\[-0.2em]
  \parbox[c]{0.05\linewidth}{\centering \intrinsicrowlabel{Input and Prediction}}%
  \hspace{2pt}%
  \parbox[c]{0.172\linewidth}{\centering \includegraphics[width=\linewidth]{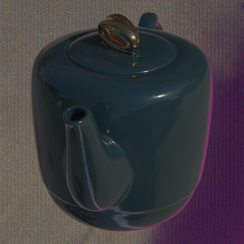}}%
  \hspace{2pt}%
  \parbox[c]{0.172\linewidth}{\centering \includegraphics[width=\linewidth]{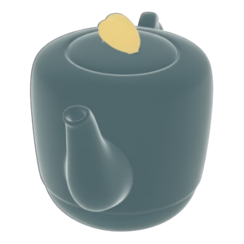}}%
  \hspace{2pt}%
  \parbox[c]{0.172\linewidth}{\centering \includegraphics[width=\linewidth]{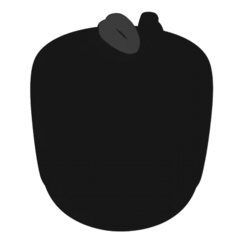}}%
  \hspace{2pt}%
  \parbox[c]{0.172\linewidth}{\centering \includegraphics[width=\linewidth]{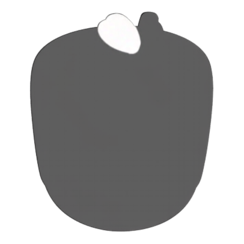}}%
  \hspace{2pt}%
  \parbox[c]{0.172\linewidth}{\centering \includegraphics[width=\linewidth]{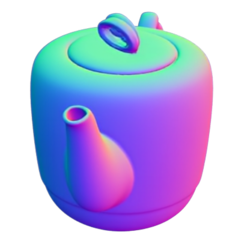}}%
  \\[1pt]
  \makebox[0.05\linewidth][c]{}%
  \hspace{2pt}%
  \makebox[0.172\linewidth][c]{\scriptsize Input}%
  \hspace{2pt}%
  \makebox[0.172\linewidth][c]{\scriptsize DR-base color}%
  \hspace{2pt}%
  \makebox[0.172\linewidth][c]{\scriptsize DR-roughness}%
  \hspace{2pt}%
  \makebox[0.172\linewidth][c]{\scriptsize DR-metallic}%
  \hspace{2pt}%
  \makebox[0.172\linewidth][c]{\scriptsize DR-normal}%
  \\[1pt]
  \vspace{0.15em}
  \par\noindent\rule{\linewidth}{0.35pt}
  \vspace{0.05em}
  \vspace{-0.6em}
  \parbox[c]{0.05\linewidth}{\centering \intrinsicrowlabel{Relight}}%
  \hspace{2pt}%
  \parbox[c]{0.215\linewidth}{\centering \begin{overpic}[width=\linewidth]{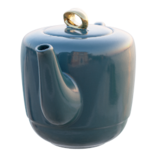}\put(0,0){\includegraphics[width=0.666\linewidth]{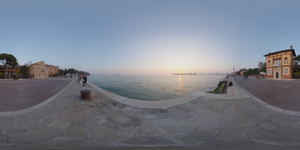}}\end{overpic}}%
  \hspace{2pt}%
  \parbox[c]{0.215\linewidth}{\centering \includegraphics[width=\linewidth]{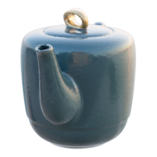}}%
  \hspace{2pt}%
  \parbox[c]{0.215\linewidth}{\centering \includegraphics[width=\linewidth]{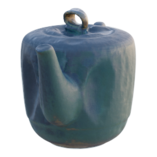}}%
  \hspace{2pt}%
  \parbox[c]{0.215\linewidth}{\centering \includegraphics[width=\linewidth]{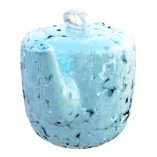}}%
  \\[0.1em]
  \parbox[c]{0.05\linewidth}{\centering \intrinsicrowlabel{Base Color}}%
  \hspace{2pt}%
  \parbox[c]{0.215\linewidth}{\centering \includegraphics[width=\linewidth]{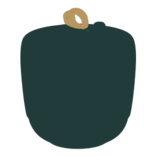}}%
  \hspace{2pt}%
  \parbox[c]{0.215\linewidth}{\centering \includegraphics[width=\linewidth]{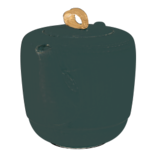}}%
  \hspace{2pt}%
  \parbox[c]{0.215\linewidth}{\centering \includegraphics[width=\linewidth]{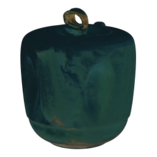}}%
  \hspace{2pt}%
  \parbox[c]{0.215\linewidth}{\centering \includegraphics[width=\linewidth]{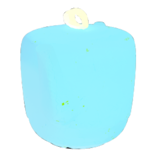}}%
  \\[0.1em]
  \parbox[c]{0.05\linewidth}{\centering \intrinsicrowlabel{Roughness}}%
  \hspace{2pt}%
  \parbox[c]{0.215\linewidth}{\centering \includegraphics[width=\linewidth]{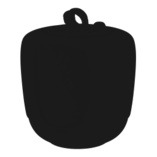}}%
  \hspace{2pt}%
  \parbox[c]{0.215\linewidth}{\centering \includegraphics[width=\linewidth]{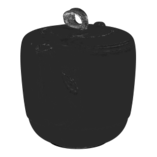}}%
  \hspace{2pt}%
  \parbox[c]{0.215\linewidth}{\centering \includegraphics[width=\linewidth]{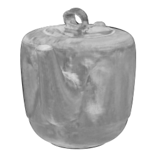}}%
  \hspace{2pt}%
  \parbox[c]{0.215\linewidth}{\centering \includegraphics[width=\linewidth]{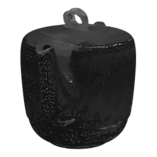}}%
  \\[0.1em]
  \parbox[c]{0.05\linewidth}{\centering \intrinsicrowlabel{Metallic}}%
  \hspace{2pt}%
  \parbox[c]{0.215\linewidth}{\centering \includegraphics[width=\linewidth]{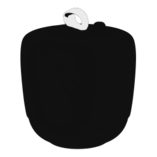}}%
  \hspace{2pt}%
  \parbox[c]{0.215\linewidth}{\centering \includegraphics[width=\linewidth]{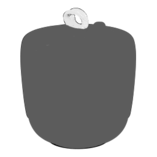}}%
  \hspace{2pt}%
  \parbox[c]{0.215\linewidth}{\centering \includegraphics[width=\linewidth]{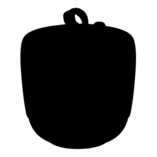}}%
  \hspace{2pt}%
  \parbox[c]{0.215\linewidth}{\centering \includegraphics[width=\linewidth]{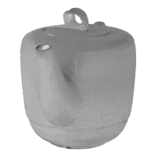}}%
  \\[0.1em]
  \parbox[c]{0.05\linewidth}{\centering \intrinsicrowlabel{Normal}}%
  \hspace{2pt}%
  \parbox[c]{0.215\linewidth}{\centering \includegraphics[width=\linewidth]{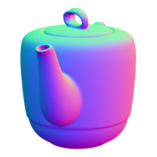}}%
  \hspace{2pt}%
  \parbox[c]{0.215\linewidth}{\centering \includegraphics[width=\linewidth]{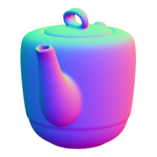}}%
  \hspace{2pt}%
  \parbox[c]{0.215\linewidth}{\centering \includegraphics[width=\linewidth]{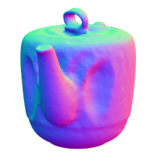}}%
  \hspace{2pt}%
  \parbox[c]{0.215\linewidth}{\centering \includegraphics[width=\linewidth]{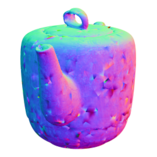}}%
  \\[0.1em]
  \parbox[c]{0.05\linewidth}{\centering \intrinsicrowlabel{Lighting}}%
  \hspace{2pt}%
  \parbox[c]{0.215\linewidth}{\centering \includegraphics[width=\linewidth]{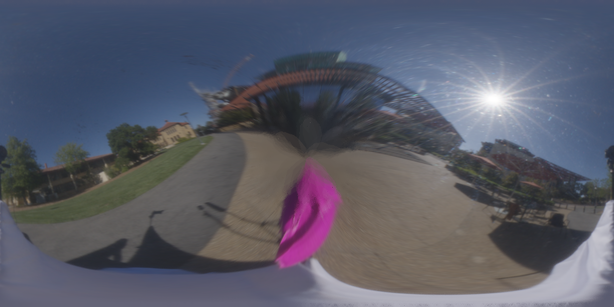}}%
  \hspace{2pt}%
  \parbox[c]{0.215\linewidth}{\centering \includegraphics[width=\linewidth]{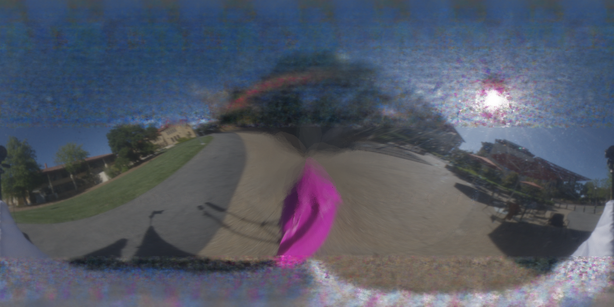}}%
  \hspace{2pt}%
  \parbox[c]{0.215\linewidth}{\centering \includegraphics[width=\linewidth]{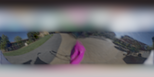}}%
  \hspace{2pt}%
  \parbox[c]{0.215\linewidth}{\centering \includegraphics[width=\linewidth]{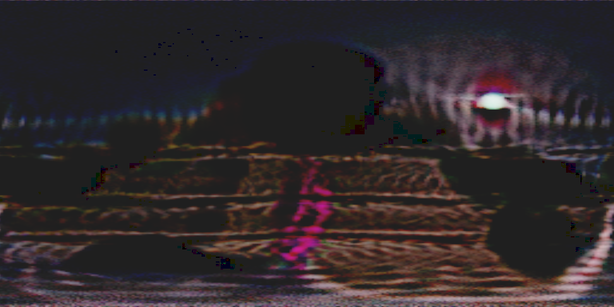}}%
  \\[1pt]
  \makebox[0.05\linewidth][c]{}%
  \hspace{2pt}%
  \makebox[0.215\linewidth][c]{\scriptsize Reference}%
  \hspace{2pt}%
  \makebox[0.215\linewidth][c]{\scriptsize Ours}%
  \hspace{2pt}%
  \makebox[0.215\linewidth][c]{\scriptsize Neural-PBIR}%
  \hspace{2pt}%
  \makebox[0.215\linewidth][c]{\scriptsize MaterialFusion}%
  \\[1pt]
  \caption{DTC-Synthetic \casename{TeaPot_B094FQW6Q4_EmeraldGoldTop_scene002}.}
  \label{fig:supp-dtc-intrinsics-teapot_b094fqw6q4_emeraldgoldtop_scene002}
\end{figure*}

\begin{figure*}[t]
  \centering
  \newcommand{\intrinsicrowlabel}[1]{\rotatebox[origin=c]{90}{\scriptsize\strut #1}}
  {\scriptsize\texttt{\detokenize{Synthetic4Relight air_baloons}}}\\[-0.2em]
  \parbox[c]{0.05\linewidth}{\centering \intrinsicrowlabel{Input and Prediction}}%
  \hspace{2pt}%
  \parbox[c]{0.172\linewidth}{\centering \includegraphics[width=\linewidth]{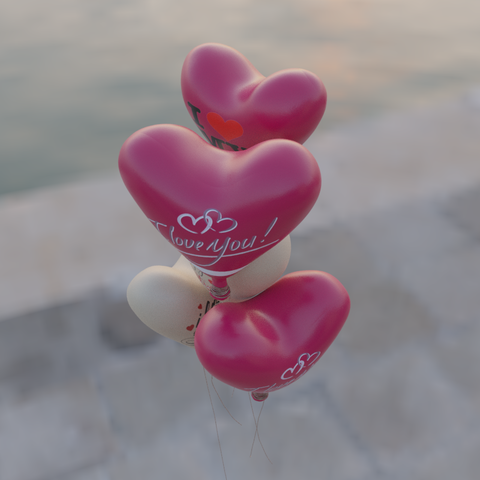}}%
  \hspace{2pt}%
  \parbox[c]{0.172\linewidth}{\centering \includegraphics[width=\linewidth]{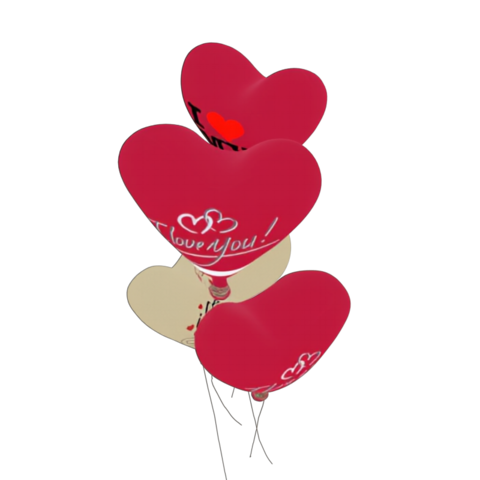}}%
  \hspace{2pt}%
  \parbox[c]{0.172\linewidth}{\centering \includegraphics[width=\linewidth]{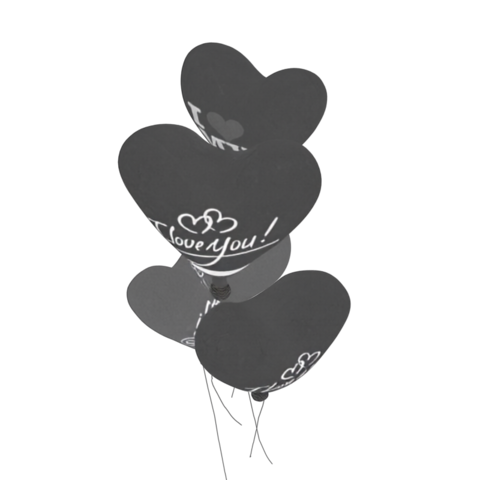}}%
  \hspace{2pt}%
  \parbox[c]{0.172\linewidth}{\centering \includegraphics[width=\linewidth]{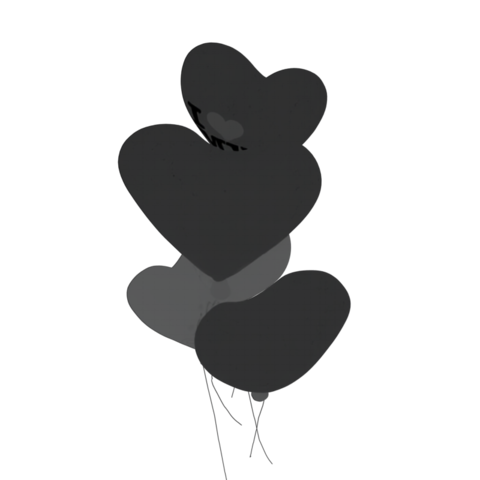}}%
  \hspace{2pt}%
  \parbox[c]{0.172\linewidth}{\centering \includegraphics[width=\linewidth]{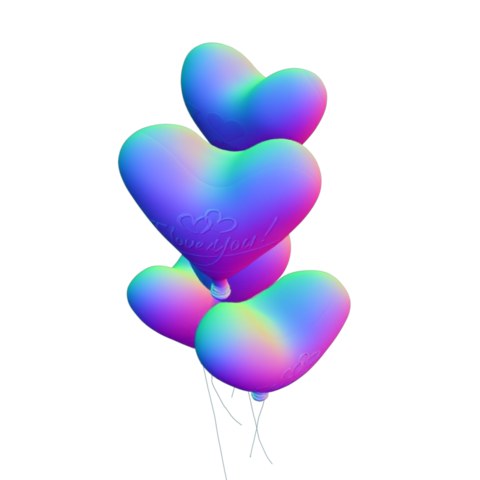}}%
  \\[1pt]
  \makebox[0.05\linewidth][c]{}%
  \hspace{2pt}%
  \makebox[0.172\linewidth][c]{\scriptsize Input}%
  \hspace{2pt}%
  \makebox[0.172\linewidth][c]{\scriptsize DR-base color}%
  \hspace{2pt}%
  \makebox[0.172\linewidth][c]{\scriptsize DR-roughness}%
  \hspace{2pt}%
  \makebox[0.172\linewidth][c]{\scriptsize DR-metallic}%
  \hspace{2pt}%
  \makebox[0.172\linewidth][c]{\scriptsize DR-normal}%
  \\[1pt]
  \vspace{0.15em}
  \par\noindent\rule{\linewidth}{0.35pt}
  \vspace{0.05em}
  \vspace{-0.6em}
  \parbox[c]{0.05\linewidth}{\centering \intrinsicrowlabel{Relight}}%
  \hspace{2pt}%
  \parbox[c]{0.215\linewidth}{\centering \begin{overpic}[width=\linewidth]{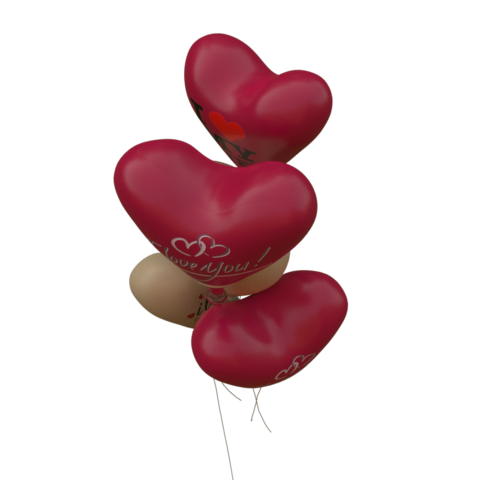}\put(0,0){\includegraphics[width=0.666\linewidth]{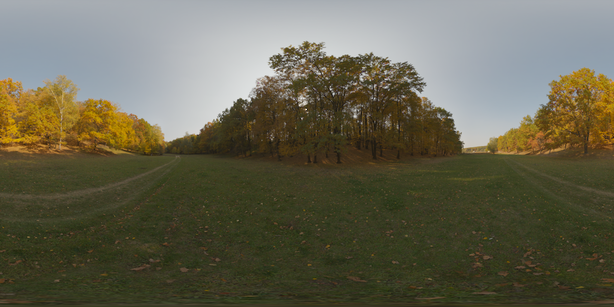}}\end{overpic}}%
  \hspace{2pt}%
  \parbox[c]{0.215\linewidth}{\centering \includegraphics[width=\linewidth]{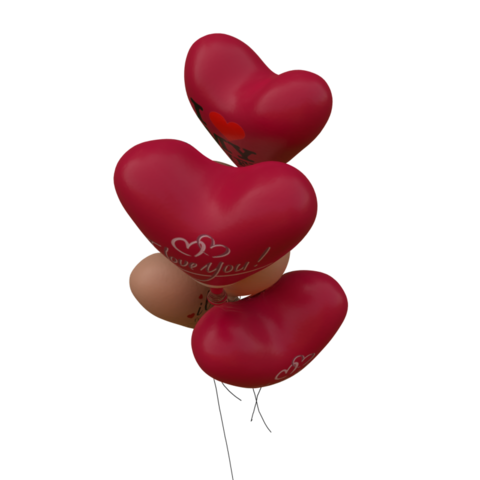}}%
  \hspace{2pt}%
  \parbox[c]{0.215\linewidth}{\centering \includegraphics[width=\linewidth]{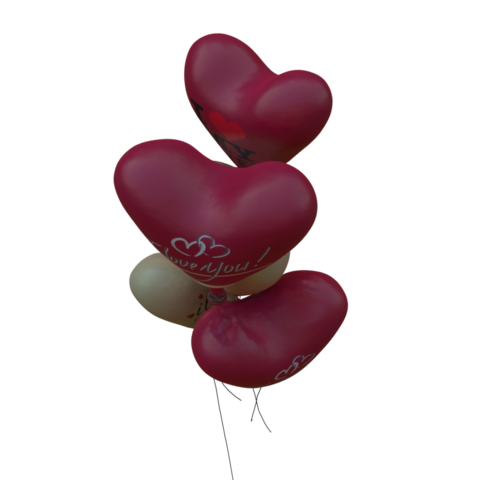}}%
  \hspace{2pt}%
  \parbox[c]{0.215\linewidth}{\centering \includegraphics[width=\linewidth]{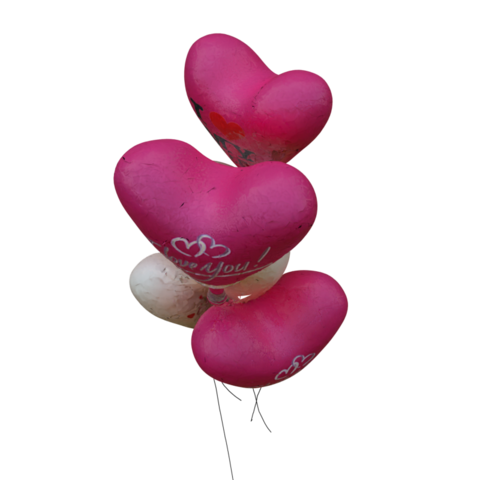}}%
  \\[0.1em]
  \parbox[c]{0.05\linewidth}{\centering \intrinsicrowlabel{Base Color}}%
  \hspace{2pt}%
  \parbox[c]{0.215\linewidth}{\centering \includegraphics[width=\linewidth]{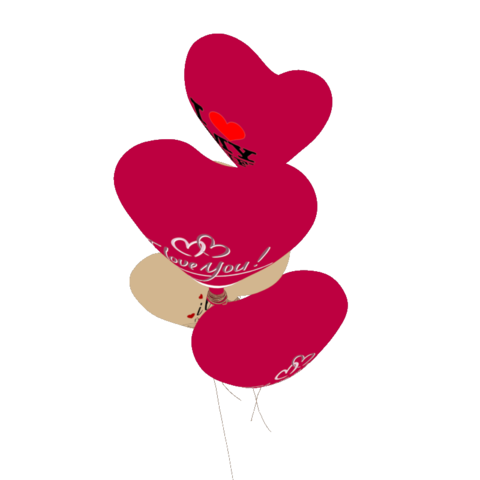}}%
  \hspace{2pt}%
  \parbox[c]{0.215\linewidth}{\centering \includegraphics[width=\linewidth]{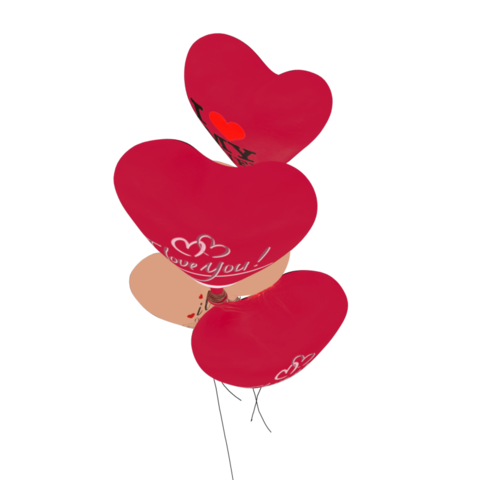}}%
  \hspace{2pt}%
  \parbox[c]{0.215\linewidth}{\centering \includegraphics[width=\linewidth]{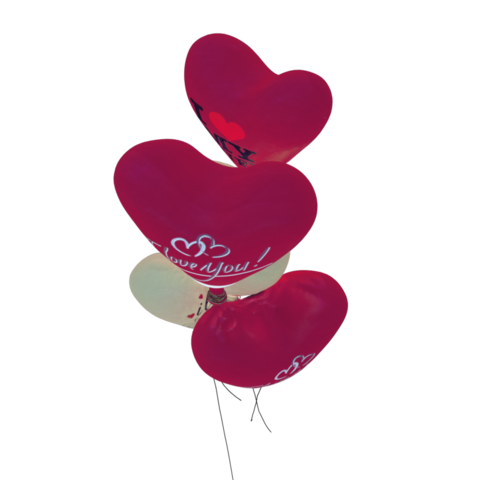}}%
  \hspace{2pt}%
  \parbox[c]{0.215\linewidth}{\centering \includegraphics[width=\linewidth]{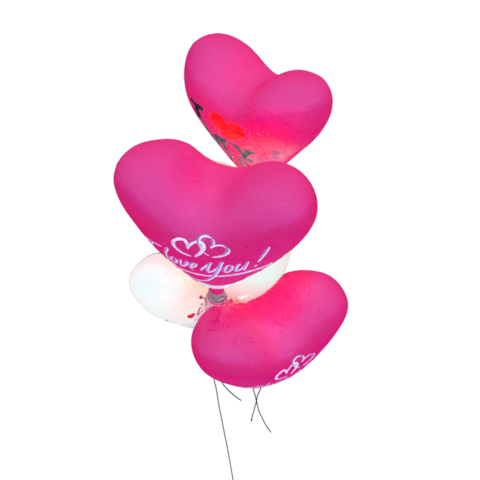}}%
  \\[0.1em]
  \parbox[c]{0.05\linewidth}{\centering \intrinsicrowlabel{Roughness}}%
  \hspace{2pt}%
  \parbox[c]{0.215\linewidth}{\centering \includegraphics[width=\linewidth]{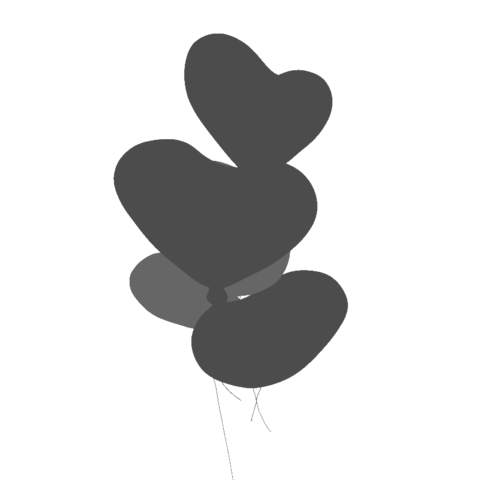}}%
  \hspace{2pt}%
  \parbox[c]{0.215\linewidth}{\centering \includegraphics[width=\linewidth]{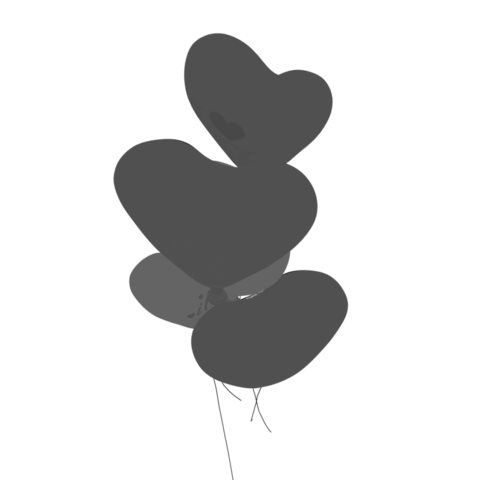}}%
  \hspace{2pt}%
  \parbox[c]{0.215\linewidth}{\centering \includegraphics[width=\linewidth]{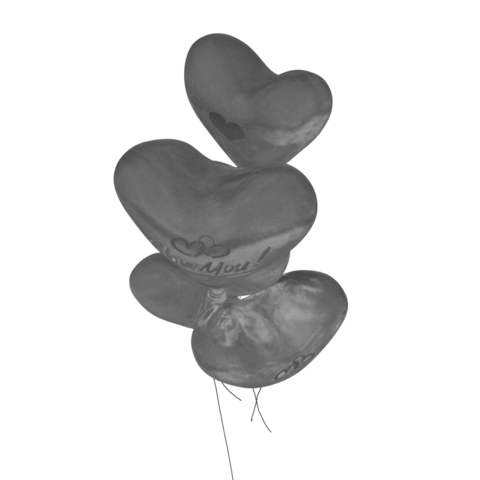}}%
  \hspace{2pt}%
  \parbox[c]{0.215\linewidth}{\centering \includegraphics[width=\linewidth]{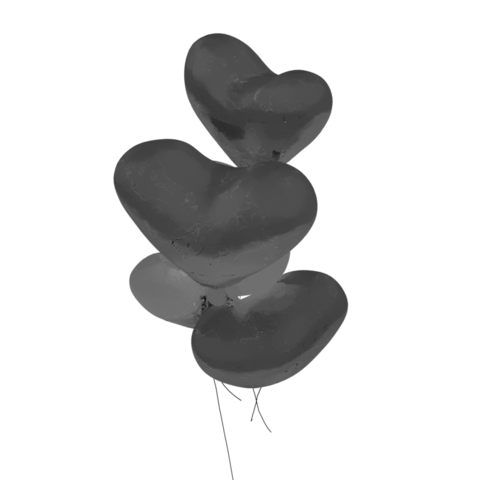}}%
  \\[0.1em]
  \parbox[c]{0.05\linewidth}{\centering \intrinsicrowlabel{Normal}}%
  \hspace{2pt}%
  \parbox[c]{0.215\linewidth}{\centering \begin{overpic}[width=\linewidth]{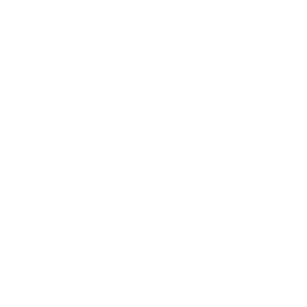}\put(50,50){\makebox(0,0){\scriptsize N/A}}\end{overpic}}%
  \hspace{2pt}%
  \parbox[c]{0.215\linewidth}{\centering \includegraphics[width=\linewidth]{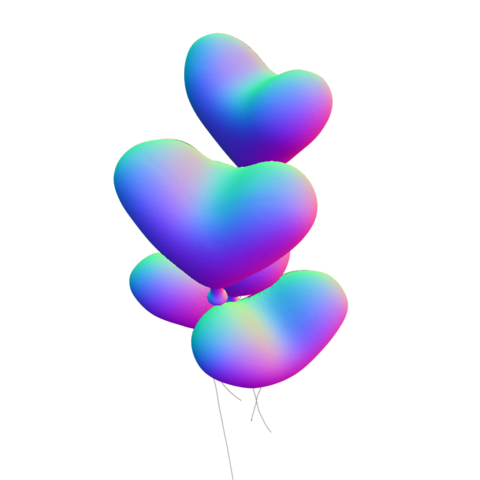}}%
  \hspace{2pt}%
  \parbox[c]{0.215\linewidth}{\centering \includegraphics[width=\linewidth]{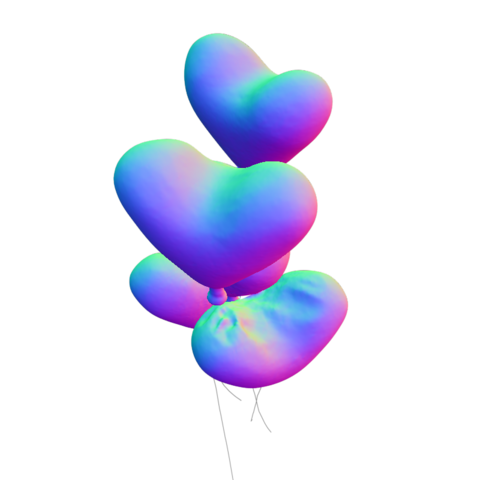}}%
  \hspace{2pt}%
  \parbox[c]{0.215\linewidth}{\centering \includegraphics[width=\linewidth]{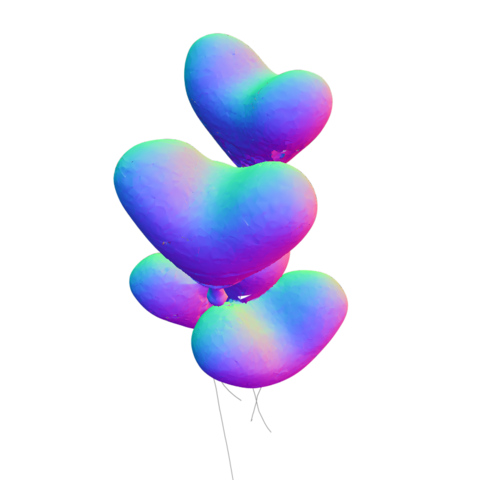}}%
  \\[0.1em]
  \parbox[c]{0.05\linewidth}{\centering \intrinsicrowlabel{Lighting}}%
  \hspace{2pt}%
  \parbox[c]{0.215\linewidth}{\centering \includegraphics[width=\linewidth]{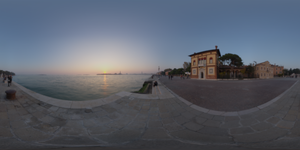}}%
  \hspace{2pt}%
  \parbox[c]{0.215\linewidth}{\centering \includegraphics[width=\linewidth]{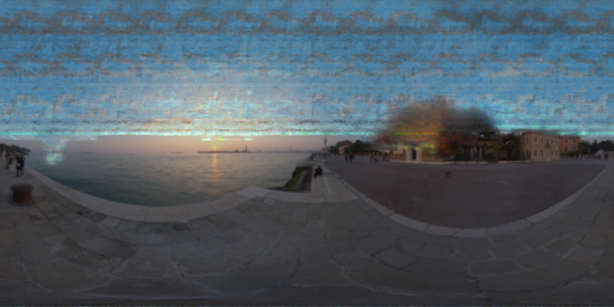}}%
  \hspace{2pt}%
  \parbox[c]{0.215\linewidth}{\centering \includegraphics[width=\linewidth]{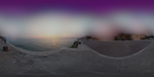}}%
  \hspace{2pt}%
  \parbox[c]{0.215\linewidth}{\centering \includegraphics[width=\linewidth]{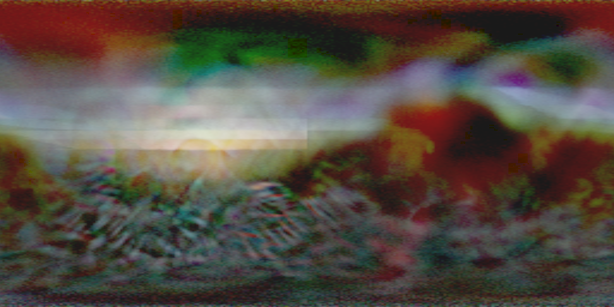}}%
  \\[1pt]
  \makebox[0.05\linewidth][c]{}%
  \hspace{2pt}%
  \makebox[0.215\linewidth][c]{\scriptsize Reference}%
  \hspace{2pt}%
  \makebox[0.215\linewidth][c]{\scriptsize Ours}%
  \hspace{2pt}%
  \makebox[0.215\linewidth][c]{\scriptsize Neural-PBIR}%
  \hspace{2pt}%
  \makebox[0.215\linewidth][c]{\scriptsize MaterialFusion}%
  \\[1pt]
  \caption{Synthetic4Relight \casename{air_baloons}.}
  \label{fig:supp-mii-intrinsics-air-baloons}
\end{figure*}

\begin{figure*}[t]
  \centering
  \newcommand{\intrinsicrowlabel}[1]{\rotatebox[origin=c]{90}{\scriptsize\strut #1}}
  {\scriptsize\texttt{\detokenize{Synthetic4Relight hotdog}}}\\[-0.2em]
  \parbox[c]{0.05\linewidth}{\centering \intrinsicrowlabel{Input and Prediction}}%
  \hspace{2pt}%
  \parbox[c]{0.172\linewidth}{\centering \includegraphics[width=\linewidth]{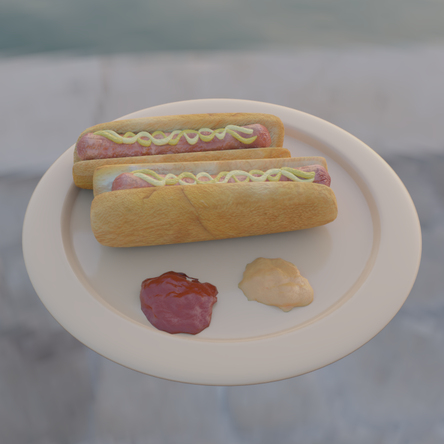}}%
  \hspace{2pt}%
  \parbox[c]{0.172\linewidth}{\centering \includegraphics[width=\linewidth]{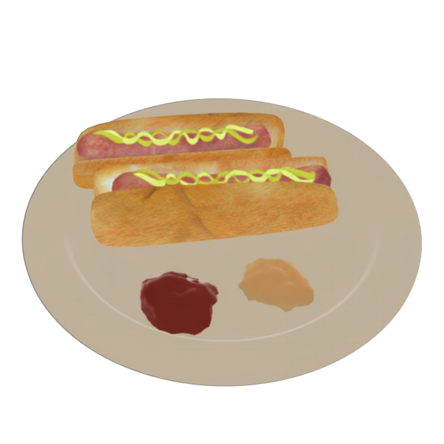}}%
  \hspace{2pt}%
  \parbox[c]{0.172\linewidth}{\centering \includegraphics[width=\linewidth]{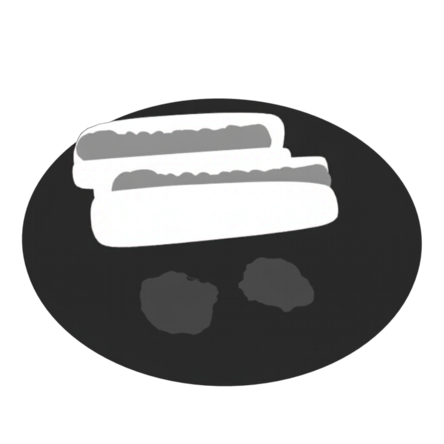}}%
  \hspace{2pt}%
  \parbox[c]{0.172\linewidth}{\centering \includegraphics[width=\linewidth]{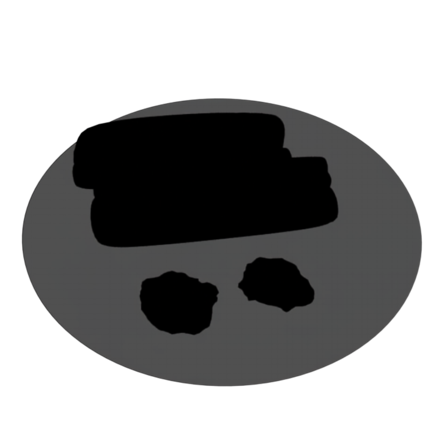}}%
  \hspace{2pt}%
  \parbox[c]{0.172\linewidth}{\centering \includegraphics[width=\linewidth]{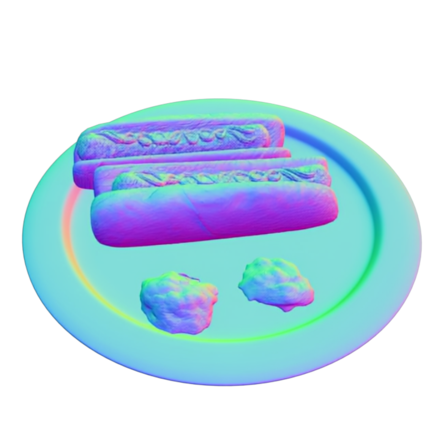}}%
  \\[1pt]
  \makebox[0.05\linewidth][c]{}%
  \hspace{2pt}%
  \makebox[0.172\linewidth][c]{\scriptsize Input}%
  \hspace{2pt}%
  \makebox[0.172\linewidth][c]{\scriptsize DR-base color}%
  \hspace{2pt}%
  \makebox[0.172\linewidth][c]{\scriptsize DR-roughness}%
  \hspace{2pt}%
  \makebox[0.172\linewidth][c]{\scriptsize DR-metallic}%
  \hspace{2pt}%
  \makebox[0.172\linewidth][c]{\scriptsize DR-normal}%
  \\[1pt]
  \vspace{0.15em}
  \par\noindent\rule{\linewidth}{0.35pt}
  \vspace{0.05em}
  \vspace{-0.6em}
  \parbox[c]{0.05\linewidth}{\centering \intrinsicrowlabel{Relight}}%
  \hspace{2pt}%
  \parbox[c]{0.215\linewidth}{\centering \begin{overpic}[width=\linewidth]{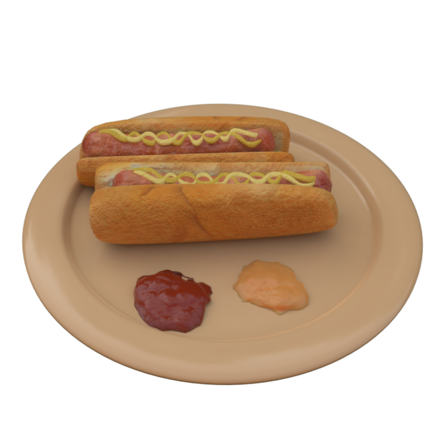}\put(0,0){\includegraphics[width=0.666\linewidth]{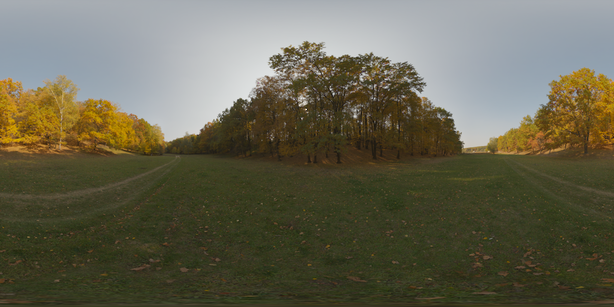}}\end{overpic}}%
  \hspace{2pt}%
  \parbox[c]{0.215\linewidth}{\centering \includegraphics[width=\linewidth]{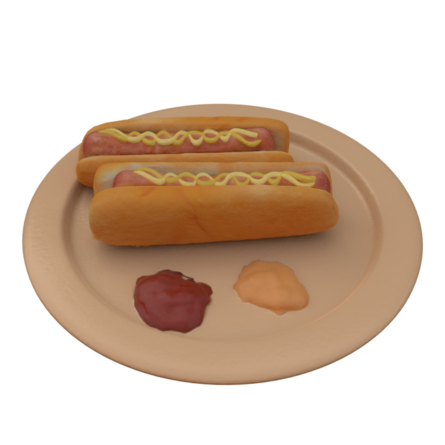}}%
  \hspace{2pt}%
  \parbox[c]{0.215\linewidth}{\centering \includegraphics[width=\linewidth]{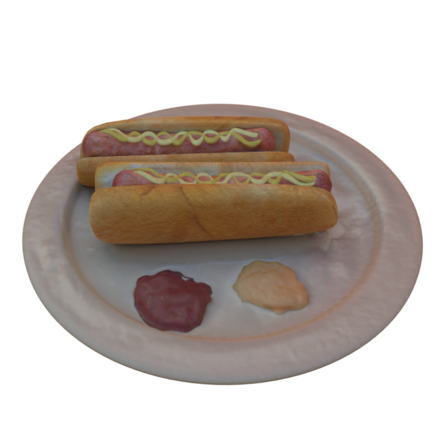}}%
  \hspace{2pt}%
  \parbox[c]{0.215\linewidth}{\centering \includegraphics[width=\linewidth]{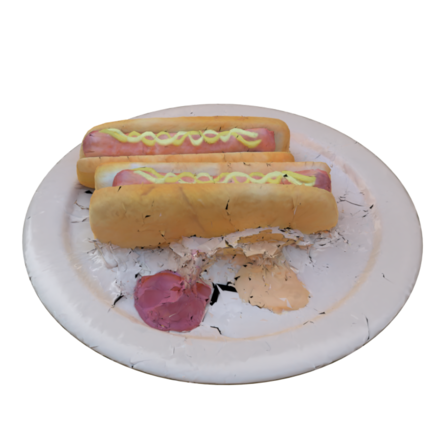}}%
  \\[0.1em]
  \parbox[c]{0.05\linewidth}{\centering \intrinsicrowlabel{Base Color}}%
  \hspace{2pt}%
  \parbox[c]{0.215\linewidth}{\centering \includegraphics[width=\linewidth]{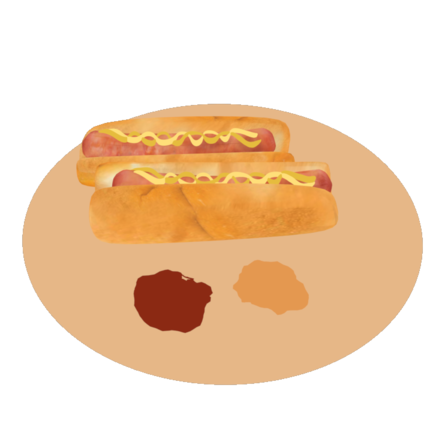}}%
  \hspace{2pt}%
  \parbox[c]{0.215\linewidth}{\centering \includegraphics[width=\linewidth]{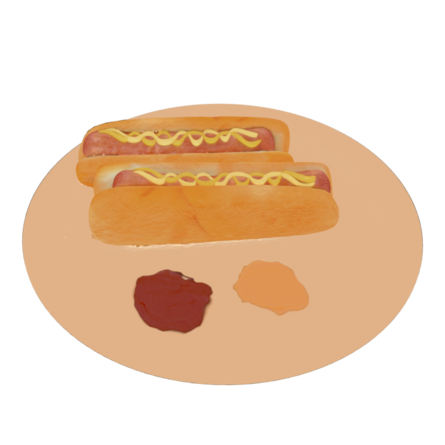}}%
  \hspace{2pt}%
  \parbox[c]{0.215\linewidth}{\centering \includegraphics[width=\linewidth]{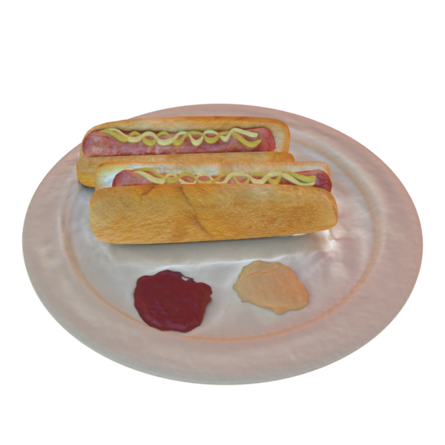}}%
  \hspace{2pt}%
  \parbox[c]{0.215\linewidth}{\centering \includegraphics[width=\linewidth]{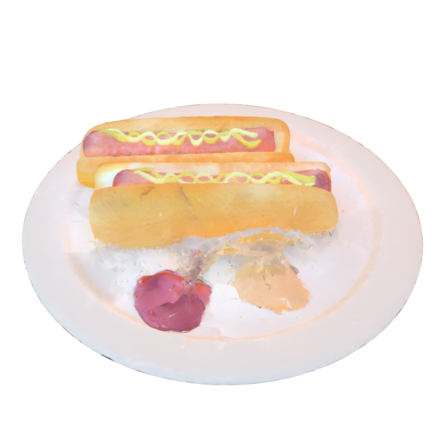}}%
  \\[0.1em]
  \parbox[c]{0.05\linewidth}{\centering \intrinsicrowlabel{Roughness}}%
  \hspace{2pt}%
  \parbox[c]{0.215\linewidth}{\centering \includegraphics[width=\linewidth]{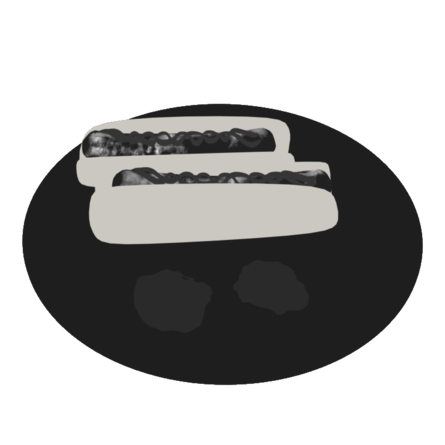}}%
  \hspace{2pt}%
  \parbox[c]{0.215\linewidth}{\centering \includegraphics[width=\linewidth]{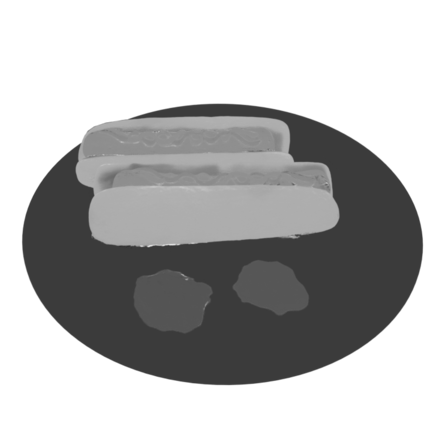}}%
  \hspace{2pt}%
  \parbox[c]{0.215\linewidth}{\centering \includegraphics[width=\linewidth]{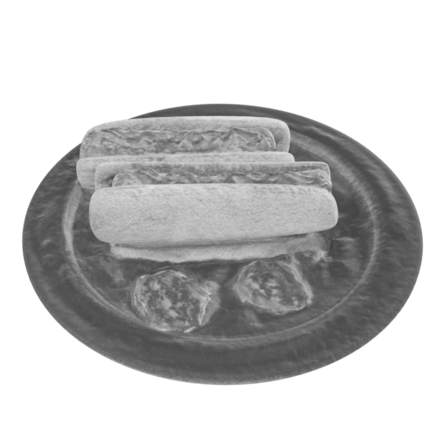}}%
  \hspace{2pt}%
  \parbox[c]{0.215\linewidth}{\centering \includegraphics[width=\linewidth]{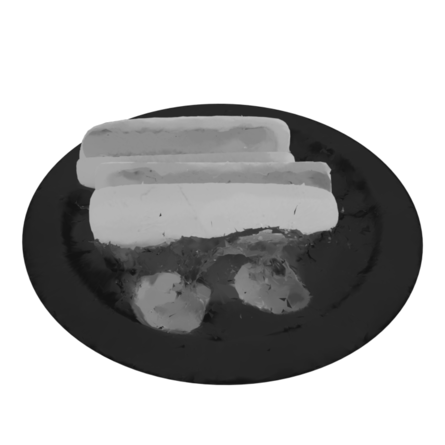}}%
  \\[0.1em]
  \parbox[c]{0.05\linewidth}{\centering \intrinsicrowlabel{Normal}}%
  \hspace{2pt}%
  \parbox[c]{0.215\linewidth}{\centering \begin{overpic}[width=\linewidth]{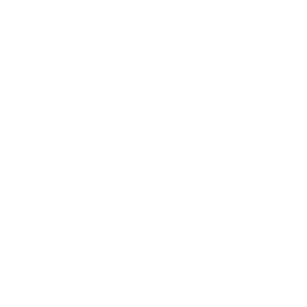}\put(50,50){\makebox(0,0){\scriptsize N/A}}\end{overpic}}%
  \hspace{2pt}%
  \parbox[c]{0.215\linewidth}{\centering \includegraphics[width=\linewidth]{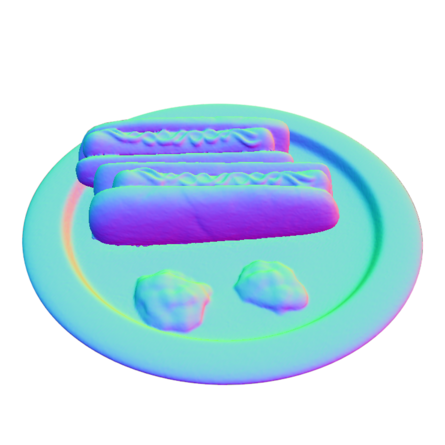}}%
  \hspace{2pt}%
  \parbox[c]{0.215\linewidth}{\centering \includegraphics[width=\linewidth]{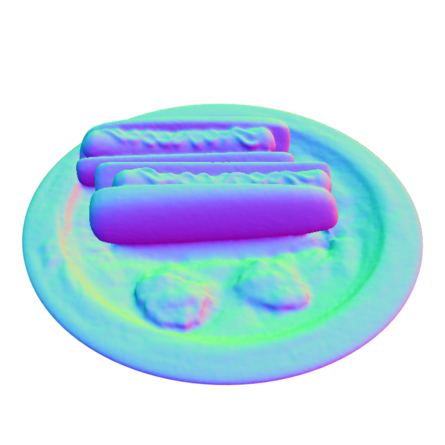}}%
  \hspace{2pt}%
  \parbox[c]{0.215\linewidth}{\centering \includegraphics[width=\linewidth]{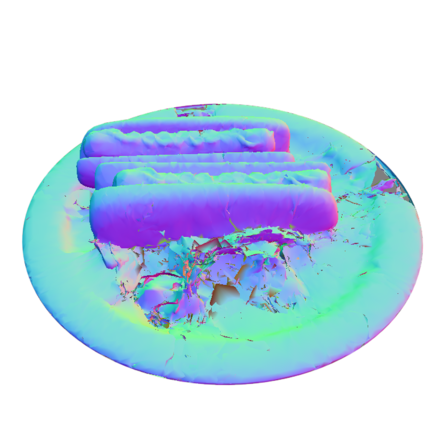}}%
  \\[0.1em]
  \parbox[c]{0.05\linewidth}{\centering \intrinsicrowlabel{Lighting}}%
  \hspace{2pt}%
  \parbox[c]{0.215\linewidth}{\centering \includegraphics[width=\linewidth]{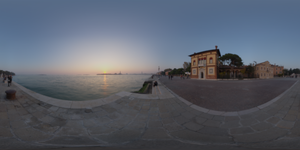}}%
  \hspace{2pt}%
  \parbox[c]{0.215\linewidth}{\centering \includegraphics[width=\linewidth]{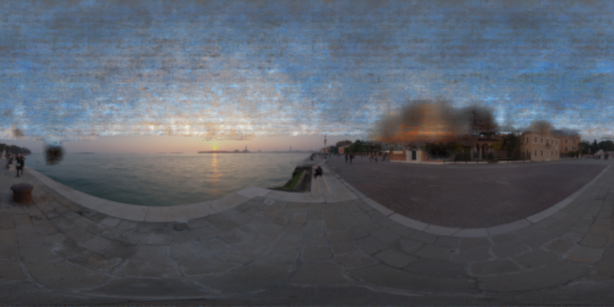}}%
  \hspace{2pt}%
  \parbox[c]{0.215\linewidth}{\centering \includegraphics[width=\linewidth]{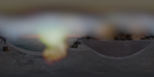}}%
  \hspace{2pt}%
  \parbox[c]{0.215\linewidth}{\centering \includegraphics[width=\linewidth]{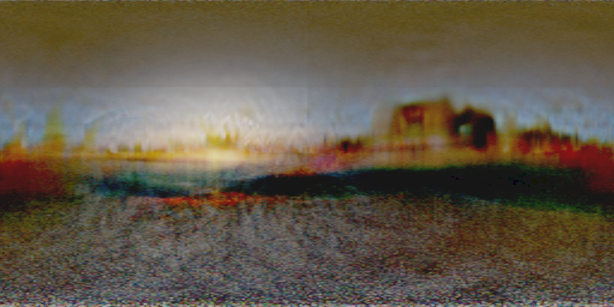}}%
  \\[1pt]
  \makebox[0.05\linewidth][c]{}%
  \hspace{2pt}%
  \makebox[0.215\linewidth][c]{\scriptsize Reference}%
  \hspace{2pt}%
  \makebox[0.215\linewidth][c]{\scriptsize Ours}%
  \hspace{2pt}%
  \makebox[0.215\linewidth][c]{\scriptsize Neural-PBIR}%
  \hspace{2pt}%
  \makebox[0.215\linewidth][c]{\scriptsize MaterialFusion}%
  \\[1pt]
  \caption{Synthetic4Relight \casename{hotdog}.}
  \label{fig:supp-mii-intrinsics-hotdog}
\end{figure*}

\begin{figure*}[t]
  \centering
  \newcommand{\intrinsicrowlabel}[1]{\rotatebox[origin=c]{90}{\scriptsize\strut #1}}
  {\scriptsize\texttt{\detokenize{Synthetic4Relight chair}}}\\[-0.2em]
  \parbox[c]{0.05\linewidth}{\centering \intrinsicrowlabel{Input and Prediction}}%
  \hspace{2pt}%
  \parbox[c]{0.172\linewidth}{\centering \includegraphics[width=\linewidth]{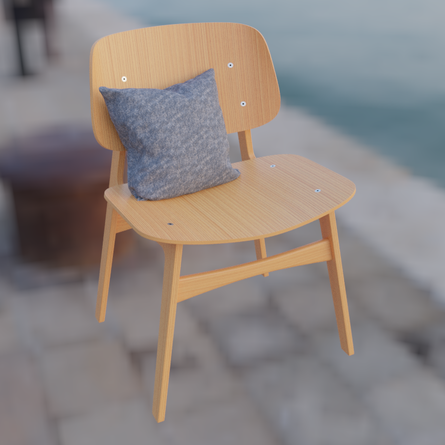}}%
  \hspace{2pt}%
  \parbox[c]{0.172\linewidth}{\centering \includegraphics[width=\linewidth]{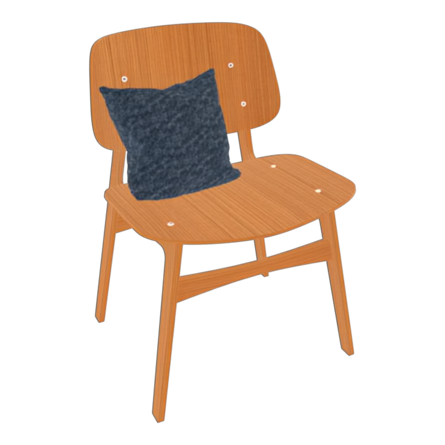}}%
  \hspace{2pt}%
  \parbox[c]{0.172\linewidth}{\centering \includegraphics[width=\linewidth]{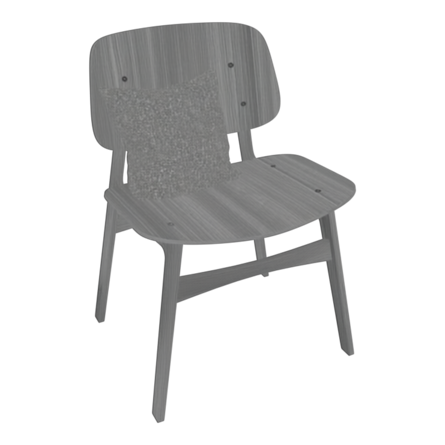}}%
  \hspace{2pt}%
  \parbox[c]{0.172\linewidth}{\centering \includegraphics[width=\linewidth]{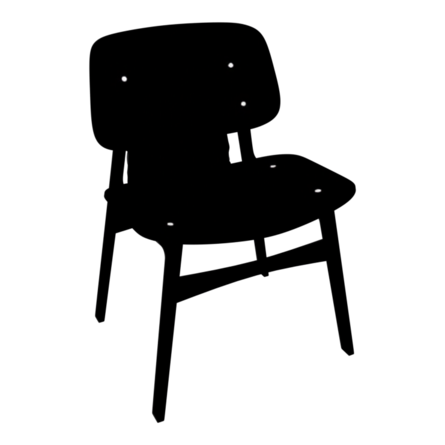}}%
  \hspace{2pt}%
  \parbox[c]{0.172\linewidth}{\centering \includegraphics[width=\linewidth]{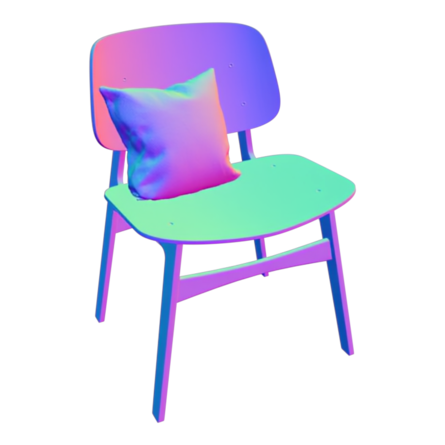}}%
  \\[1pt]
  \makebox[0.05\linewidth][c]{}%
  \hspace{2pt}%
  \makebox[0.172\linewidth][c]{\scriptsize Input}%
  \hspace{2pt}%
  \makebox[0.172\linewidth][c]{\scriptsize DR-base color}%
  \hspace{2pt}%
  \makebox[0.172\linewidth][c]{\scriptsize DR-roughness}%
  \hspace{2pt}%
  \makebox[0.172\linewidth][c]{\scriptsize DR-metallic}%
  \hspace{2pt}%
  \makebox[0.172\linewidth][c]{\scriptsize DR-normal}%
  \\[1pt]
  \vspace{0.15em}
  \par\noindent\rule{\linewidth}{0.35pt}
  \vspace{0.05em}
  \vspace{-0.6em}
  \parbox[c]{0.05\linewidth}{\centering \intrinsicrowlabel{Relight}}%
  \hspace{2pt}%
  \parbox[c]{0.215\linewidth}{\centering \begin{overpic}[width=\linewidth]{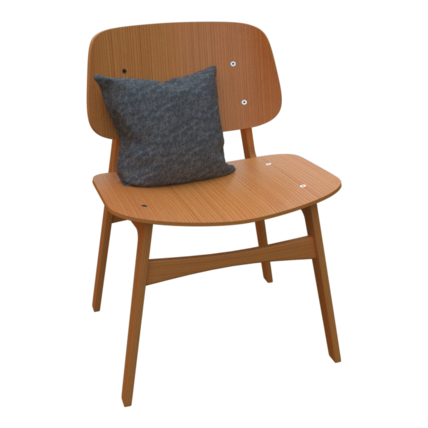}\put(0,0){\includegraphics[width=0.666\linewidth]{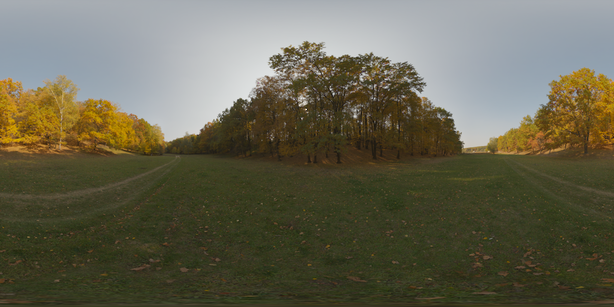}}\end{overpic}}%
  \hspace{2pt}%
  \parbox[c]{0.215\linewidth}{\centering \includegraphics[width=\linewidth]{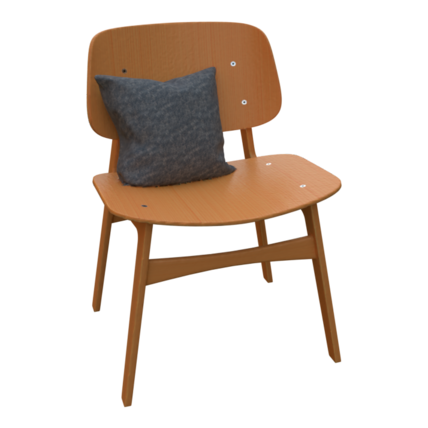}}%
  \hspace{2pt}%
  \parbox[c]{0.215\linewidth}{\centering \includegraphics[width=\linewidth]{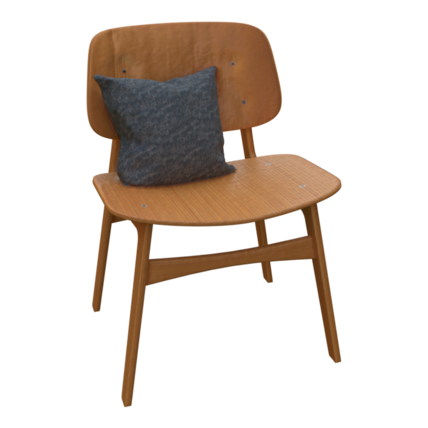}}%
  \hspace{2pt}%
  \parbox[c]{0.215\linewidth}{\centering \includegraphics[width=\linewidth]{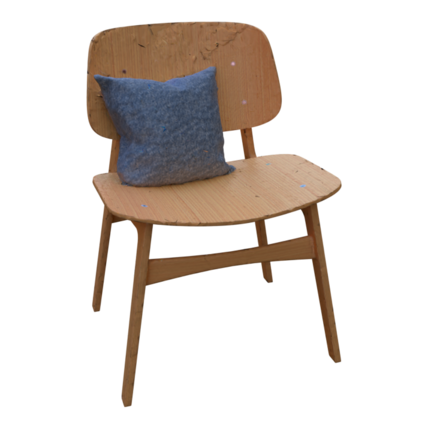}}%
  \\[0.1em]
  \parbox[c]{0.05\linewidth}{\centering \intrinsicrowlabel{Base Color}}%
  \hspace{2pt}%
  \parbox[c]{0.215\linewidth}{\centering \includegraphics[width=\linewidth]{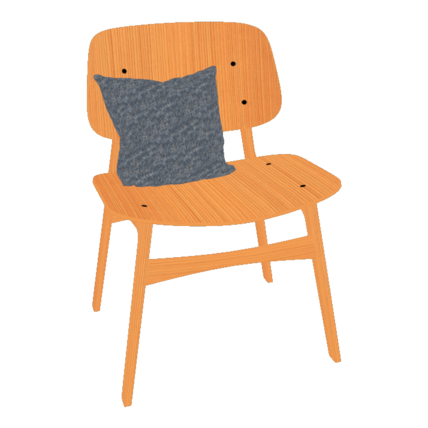}}%
  \hspace{2pt}%
  \parbox[c]{0.215\linewidth}{\centering \includegraphics[width=\linewidth]{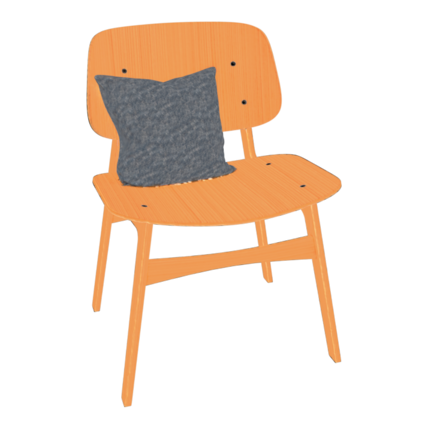}}%
  \hspace{2pt}%
  \parbox[c]{0.215\linewidth}{\centering \includegraphics[width=\linewidth]{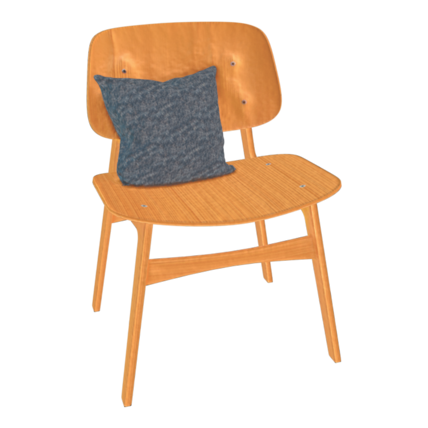}}%
  \hspace{2pt}%
  \parbox[c]{0.215\linewidth}{\centering \includegraphics[width=\linewidth]{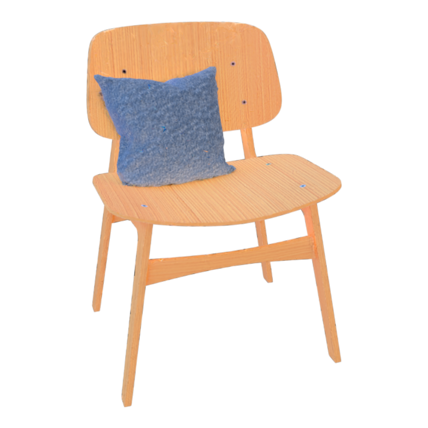}}%
  \\[0.1em]
  \parbox[c]{0.05\linewidth}{\centering \intrinsicrowlabel{Roughness}}%
  \hspace{2pt}%
  \parbox[c]{0.215\linewidth}{\centering \includegraphics[width=\linewidth]{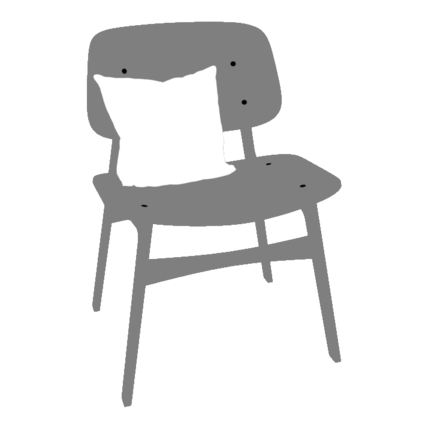}}%
  \hspace{2pt}%
  \parbox[c]{0.215\linewidth}{\centering \includegraphics[width=\linewidth]{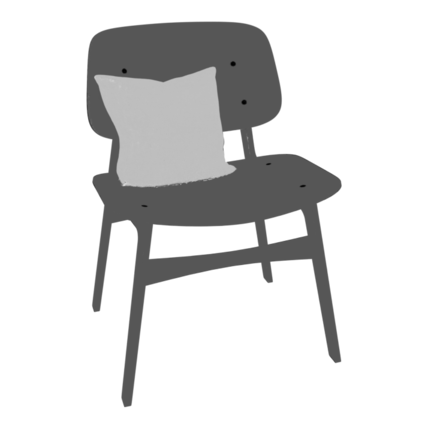}}%
  \hspace{2pt}%
  \parbox[c]{0.215\linewidth}{\centering \includegraphics[width=\linewidth]{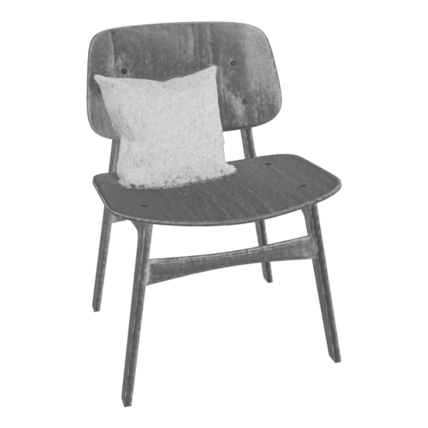}}%
  \hspace{2pt}%
  \parbox[c]{0.215\linewidth}{\centering \includegraphics[width=\linewidth]{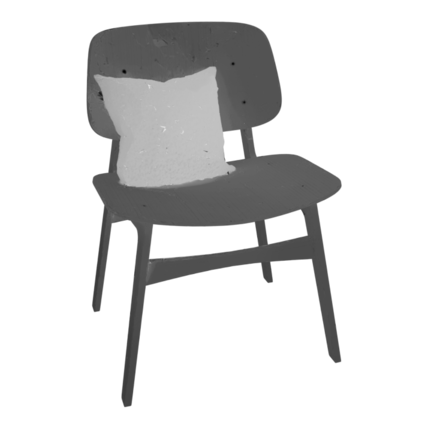}}%
  \\[0.1em]
  \parbox[c]{0.05\linewidth}{\centering \intrinsicrowlabel{Normal}}%
  \hspace{2pt}%
  \parbox[c]{0.215\linewidth}{\centering \begin{overpic}[width=\linewidth]{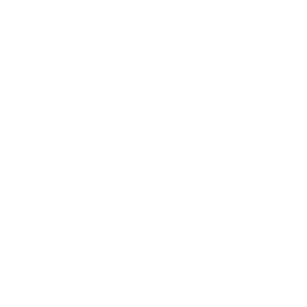}\put(50,50){\makebox(0,0){\scriptsize N/A}}\end{overpic}}%
  \hspace{2pt}%
  \parbox[c]{0.215\linewidth}{\centering \includegraphics[width=\linewidth]{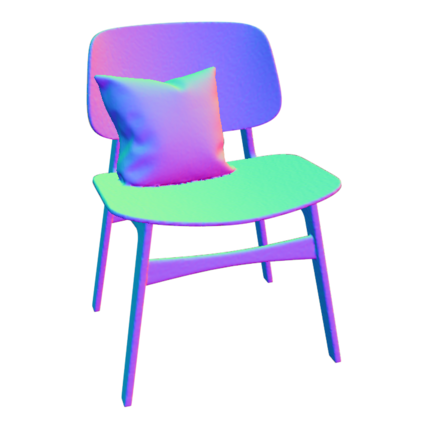}}%
  \hspace{2pt}%
  \parbox[c]{0.215\linewidth}{\centering \includegraphics[width=\linewidth]{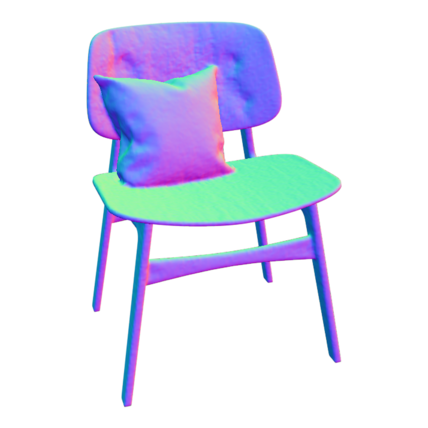}}%
  \hspace{2pt}%
  \parbox[c]{0.215\linewidth}{\centering \includegraphics[width=\linewidth]{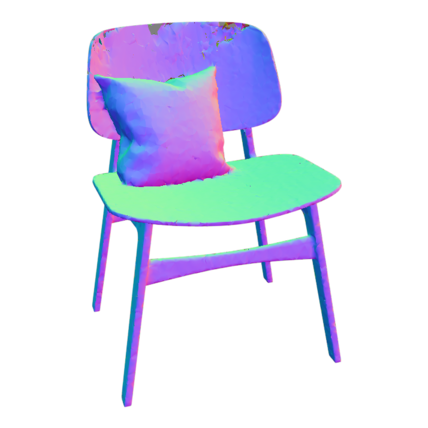}}%
  \\[0.1em]
  \parbox[c]{0.05\linewidth}{\centering \intrinsicrowlabel{Lighting}}%
  \hspace{2pt}%
  \parbox[c]{0.215\linewidth}{\centering \includegraphics[width=\linewidth]{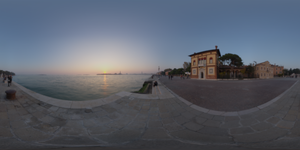}}%
  \hspace{2pt}%
  \parbox[c]{0.215\linewidth}{\centering \includegraphics[width=\linewidth]{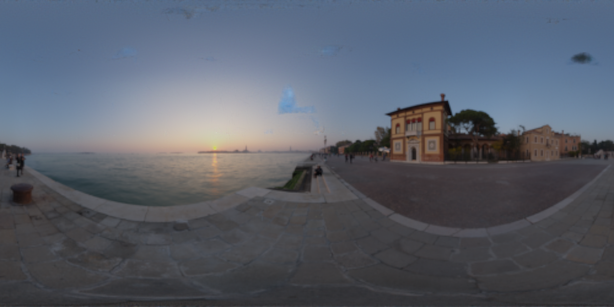}}%
  \hspace{2pt}%
  \parbox[c]{0.215\linewidth}{\centering \includegraphics[width=\linewidth]{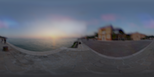}}%
  \hspace{2pt}%
  \parbox[c]{0.215\linewidth}{\centering \includegraphics[width=\linewidth]{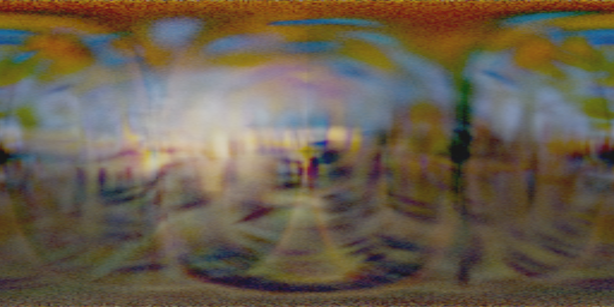}}%
  \\[1pt]
  \makebox[0.05\linewidth][c]{}%
  \hspace{2pt}%
  \makebox[0.215\linewidth][c]{\scriptsize Reference}%
  \hspace{2pt}%
  \makebox[0.215\linewidth][c]{\scriptsize Ours}%
  \hspace{2pt}%
  \makebox[0.215\linewidth][c]{\scriptsize Neural-PBIR}%
  \hspace{2pt}%
  \makebox[0.215\linewidth][c]{\scriptsize MaterialFusion}%
  \\[1pt]
  \caption{Synthetic4Relight \casename{chair}.}
  \label{fig:supp-mii-intrinsics-chair}
\end{figure*}

\begin{figure*}[t]
  \centering
  \newcommand{\intrinsicrowlabel}[1]{\rotatebox[origin=c]{90}{\scriptsize\strut #1}}
  {\scriptsize\texttt{\detokenize{Synthetic4Relight jugs}}}\\[-0.2em]
  \parbox[c]{0.05\linewidth}{\centering \intrinsicrowlabel{Input and Prediction}}%
  \hspace{2pt}%
  \parbox[c]{0.172\linewidth}{\centering \includegraphics[width=\linewidth]{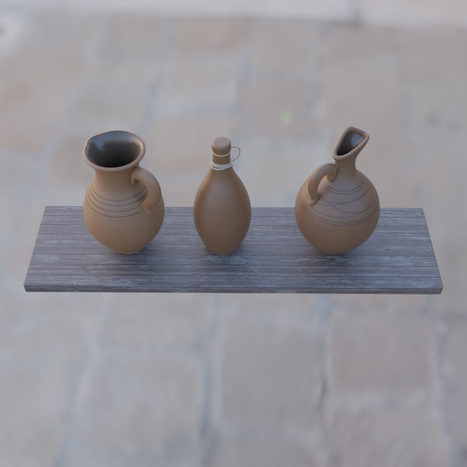}}%
  \hspace{2pt}%
  \parbox[c]{0.172\linewidth}{\centering \includegraphics[width=\linewidth]{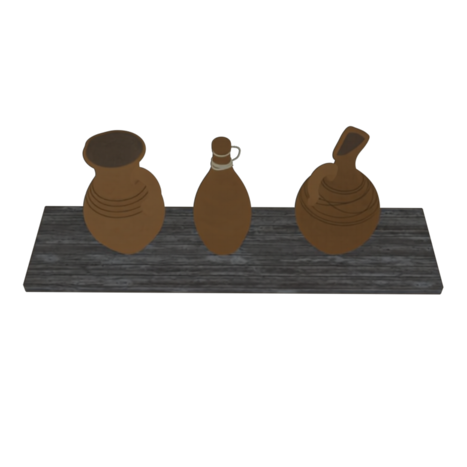}}%
  \hspace{2pt}%
  \parbox[c]{0.172\linewidth}{\centering \includegraphics[width=\linewidth]{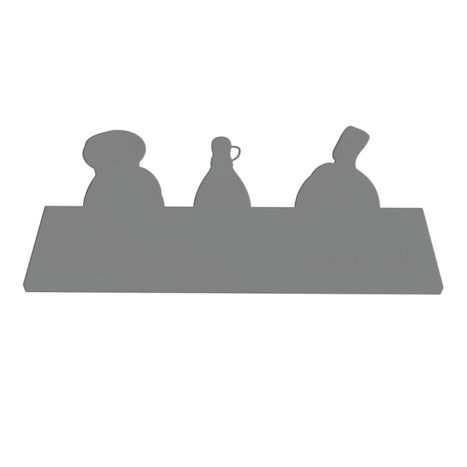}}%
  \hspace{2pt}%
  \parbox[c]{0.172\linewidth}{\centering \includegraphics[width=\linewidth]{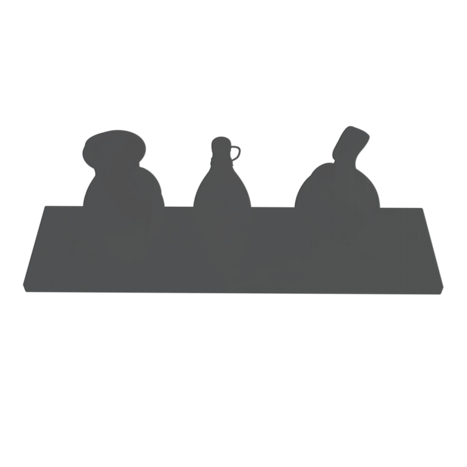}}%
  \hspace{2pt}%
  \parbox[c]{0.172\linewidth}{\centering \includegraphics[width=\linewidth]{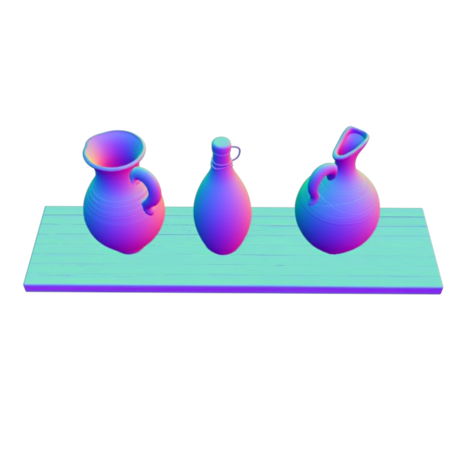}}%
  \\[1pt]
  \makebox[0.05\linewidth][c]{}%
  \hspace{2pt}%
  \makebox[0.172\linewidth][c]{\scriptsize Input}%
  \hspace{2pt}%
  \makebox[0.172\linewidth][c]{\scriptsize DR-base color}%
  \hspace{2pt}%
  \makebox[0.172\linewidth][c]{\scriptsize DR-roughness}%
  \hspace{2pt}%
  \makebox[0.172\linewidth][c]{\scriptsize DR-metallic}%
  \hspace{2pt}%
  \makebox[0.172\linewidth][c]{\scriptsize DR-normal}%
  \\[1pt]
  \vspace{0.15em}
  \par\noindent\rule{\linewidth}{0.35pt}
  \vspace{0.05em}
  \vspace{-0.6em}
  \parbox[c]{0.05\linewidth}{\centering \intrinsicrowlabel{Relight}}%
  \hspace{2pt}%
  \parbox[c]{0.215\linewidth}{\centering \begin{overpic}[width=\linewidth]{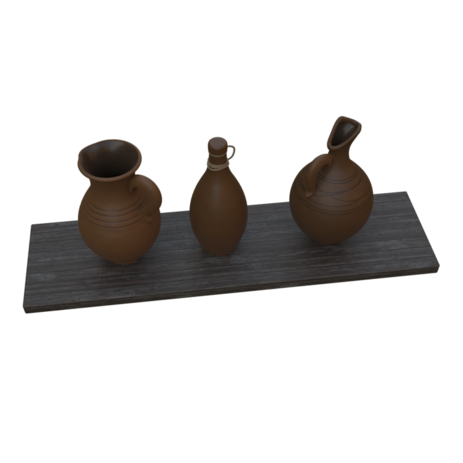}\put(0,0){\includegraphics[width=0.666\linewidth]{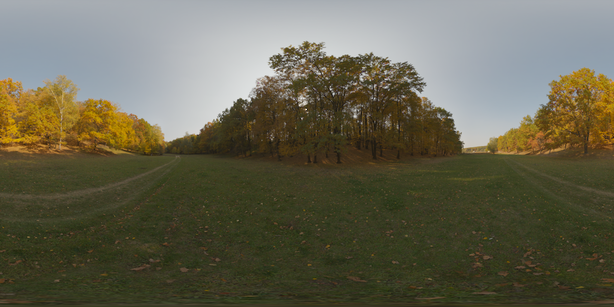}}\end{overpic}}%
  \hspace{2pt}%
  \parbox[c]{0.215\linewidth}{\centering \includegraphics[width=\linewidth]{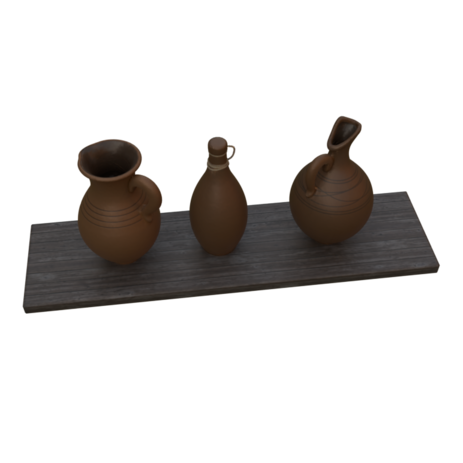}}%
  \hspace{2pt}%
  \parbox[c]{0.215\linewidth}{\centering \includegraphics[width=\linewidth]{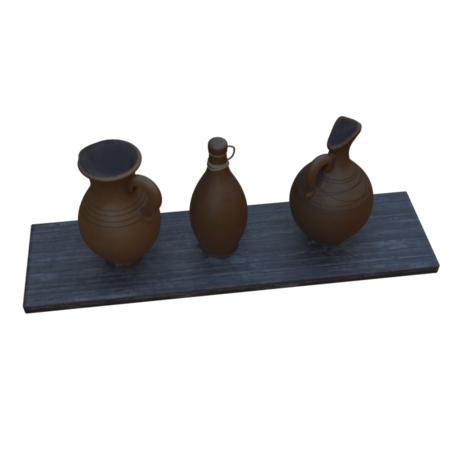}}%
  \hspace{2pt}%
  \parbox[c]{0.215\linewidth}{\centering \includegraphics[width=\linewidth]{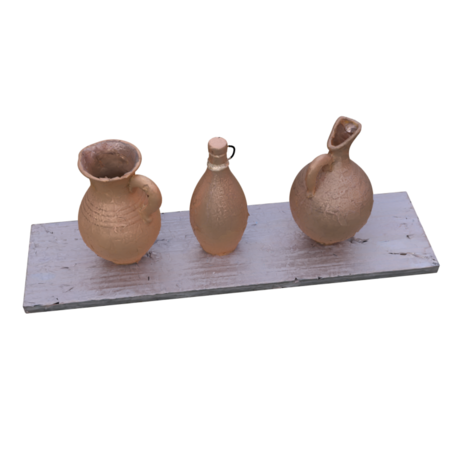}}%
  \\[0.1em]
  \parbox[c]{0.05\linewidth}{\centering \intrinsicrowlabel{Base Color}}%
  \hspace{2pt}%
  \parbox[c]{0.215\linewidth}{\centering \includegraphics[width=\linewidth]{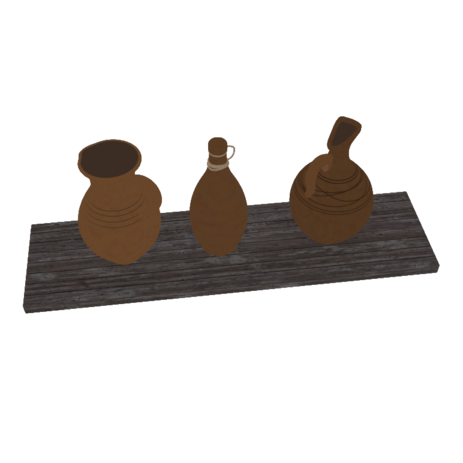}}%
  \hspace{2pt}%
  \parbox[c]{0.215\linewidth}{\centering \includegraphics[width=\linewidth]{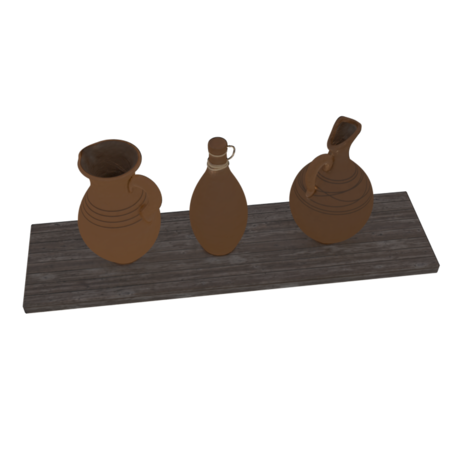}}%
  \hspace{2pt}%
  \parbox[c]{0.215\linewidth}{\centering \includegraphics[width=\linewidth]{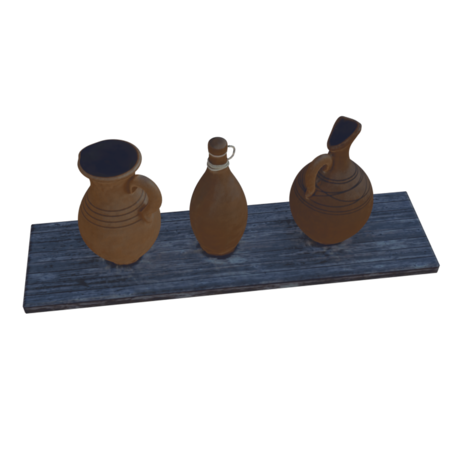}}%
  \hspace{2pt}%
  \parbox[c]{0.215\linewidth}{\centering \includegraphics[width=\linewidth]{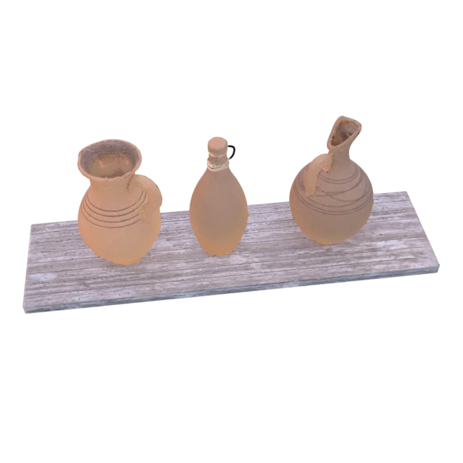}}%
  \\[0.1em]
  \parbox[c]{0.05\linewidth}{\centering \intrinsicrowlabel{Roughness}}%
  \hspace{2pt}%
  \parbox[c]{0.215\linewidth}{\centering \includegraphics[width=\linewidth]{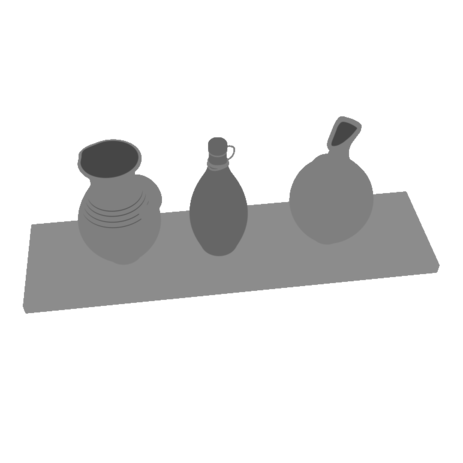}}%
  \hspace{2pt}%
  \parbox[c]{0.215\linewidth}{\centering \includegraphics[width=\linewidth]{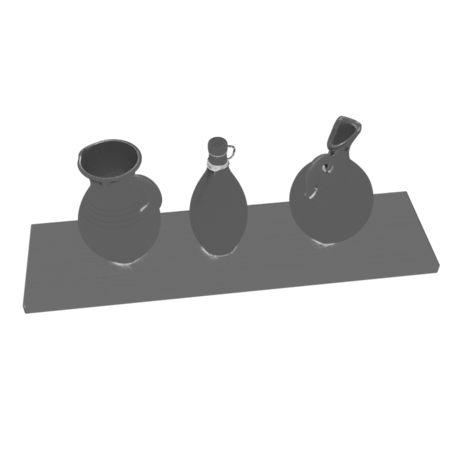}}%
  \hspace{2pt}%
  \parbox[c]{0.215\linewidth}{\centering \includegraphics[width=\linewidth]{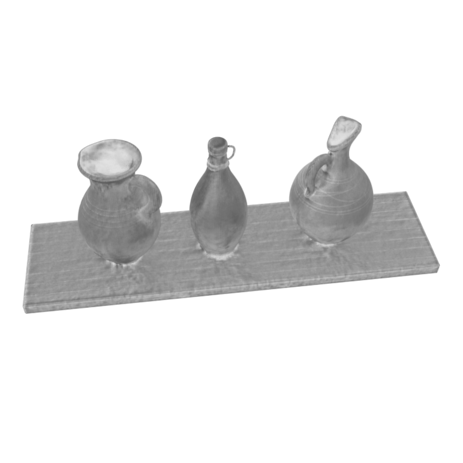}}%
  \hspace{2pt}%
  \parbox[c]{0.215\linewidth}{\centering \includegraphics[width=\linewidth]{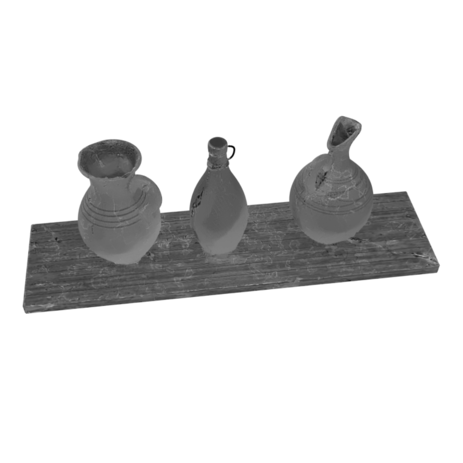}}%
  \\[0.1em]
  \parbox[c]{0.05\linewidth}{\centering \intrinsicrowlabel{Normal}}%
  \hspace{2pt}%
  \parbox[c]{0.215\linewidth}{\centering \begin{overpic}[width=\linewidth]{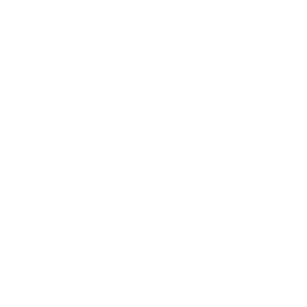}\put(50,50){\makebox(0,0){\scriptsize N/A}}\end{overpic}}%
  \hspace{2pt}%
  \parbox[c]{0.215\linewidth}{\centering \includegraphics[width=\linewidth]{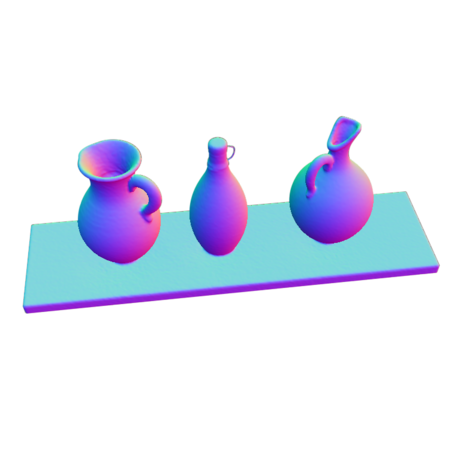}}%
  \hspace{2pt}%
  \parbox[c]{0.215\linewidth}{\centering \includegraphics[width=\linewidth]{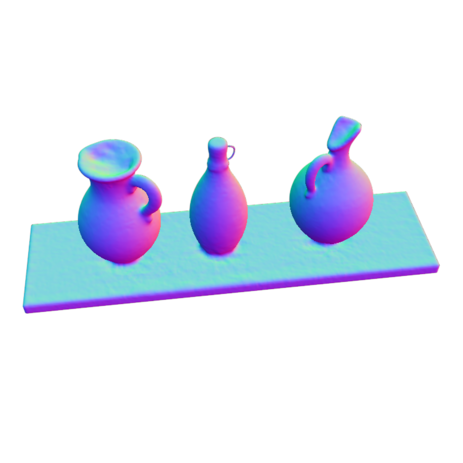}}%
  \hspace{2pt}%
  \parbox[c]{0.215\linewidth}{\centering \includegraphics[width=\linewidth]{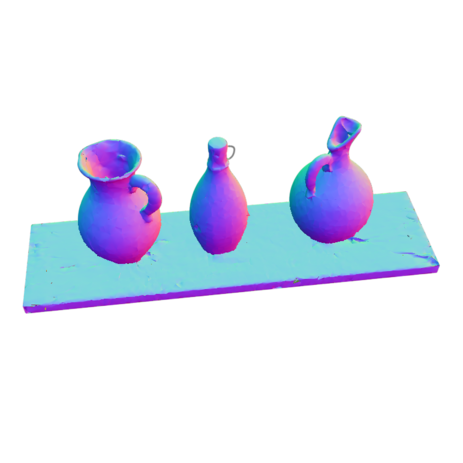}}%
  \\[0.1em]
  \parbox[c]{0.05\linewidth}{\centering \intrinsicrowlabel{Lighting}}%
  \hspace{2pt}%
  \parbox[c]{0.215\linewidth}{\centering \includegraphics[width=\linewidth]{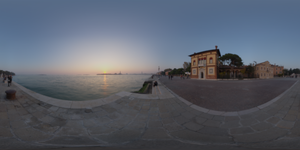}}%
  \hspace{2pt}%
  \parbox[c]{0.215\linewidth}{\centering \includegraphics[width=\linewidth]{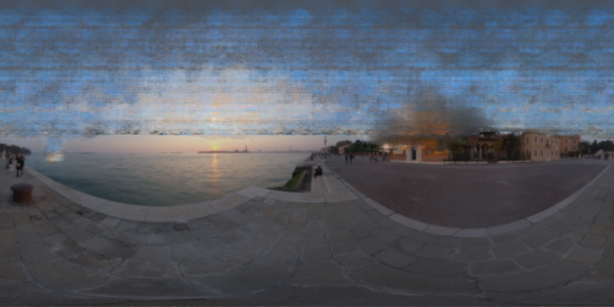}}%
  \hspace{2pt}%
  \parbox[c]{0.215\linewidth}{\centering \includegraphics[width=\linewidth]{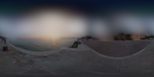}}%
  \hspace{2pt}%
  \parbox[c]{0.215\linewidth}{\centering \includegraphics[width=\linewidth]{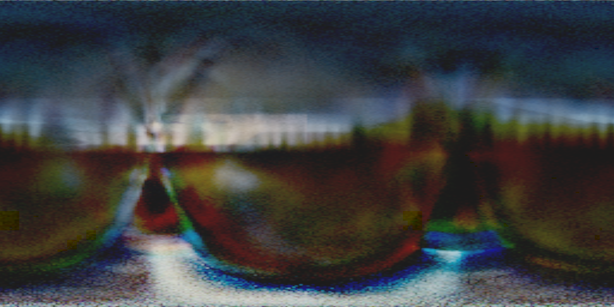}}%
  \\[1pt]
  \makebox[0.05\linewidth][c]{}%
  \hspace{2pt}%
  \makebox[0.215\linewidth][c]{\scriptsize Reference}%
  \hspace{2pt}%
  \makebox[0.215\linewidth][c]{\scriptsize Ours}%
  \hspace{2pt}%
  \makebox[0.215\linewidth][c]{\scriptsize Neural-PBIR}%
  \hspace{2pt}%
  \makebox[0.215\linewidth][c]{\scriptsize MaterialFusion}%
  \\[1pt]
  \caption{Synthetic4Relight \casename{jugs}.}
  \label{fig:supp-mii-intrinsics-jugs}
\end{figure*}

\begin{figure*}[t]
  \centering
  \newcommand{\intrinsicrowlabel}[1]{\rotatebox[origin=c]{90}{\scriptsize\strut #1}}
  {\scriptsize\texttt{\detokenize{Stanford-ORB grogu_scene003 (Ablation)}}}\\[-0.2em]
  \parbox[c]{0.05\linewidth}{\centering \intrinsicrowlabel{Input and Prediction}}%
  \hspace{2pt}%
  \parbox[c]{0.172\linewidth}{\centering \includegraphics[width=\linewidth]{figures/summary/stanford_orb_grogu_scene003_intrinsic_assets_ablation/input_dr/input.png}}%
  \hspace{2pt}%
  \parbox[c]{0.172\linewidth}{\centering \includegraphics[width=\linewidth]{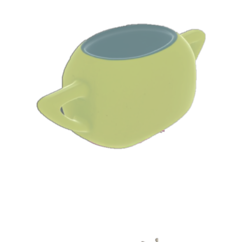}}%
  \hspace{2pt}%
  \parbox[c]{0.172\linewidth}{\centering \includegraphics[width=\linewidth]{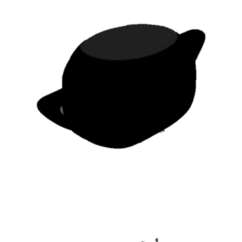}}%
  \hspace{2pt}%
  \parbox[c]{0.172\linewidth}{\centering \includegraphics[width=\linewidth]{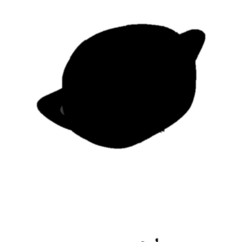}}%
  \hspace{2pt}%
  \parbox[c]{0.172\linewidth}{\centering \includegraphics[width=\linewidth]{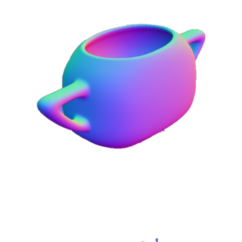}}%
  \\[1pt]
  \makebox[0.05\linewidth][c]{}%
  \hspace{2pt}%
  \makebox[0.172\linewidth][c]{\scriptsize Input}%
  \hspace{2pt}%
  \makebox[0.172\linewidth][c]{\scriptsize DR-base color}%
  \hspace{2pt}%
  \makebox[0.172\linewidth][c]{\scriptsize DR-roughness}%
  \hspace{2pt}%
  \makebox[0.172\linewidth][c]{\scriptsize DR-metallic}%
  \hspace{2pt}%
  \makebox[0.172\linewidth][c]{\scriptsize DR-normal}%
  \\[1pt]
  \vspace{0.15em}
  \par\noindent\rule{\linewidth}{0.35pt}
  \vspace{0.05em}
  \vspace{-0.6em}
  \parbox[c]{0.05\linewidth}{\centering \intrinsicrowlabel{Relight}}%
  \hspace{2pt}%
  \parbox[c]{0.17\linewidth}{\centering \begin{overpic}[width=\linewidth]{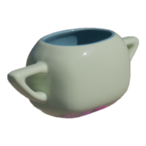}\put(0,0){\includegraphics[width=0.666\linewidth]{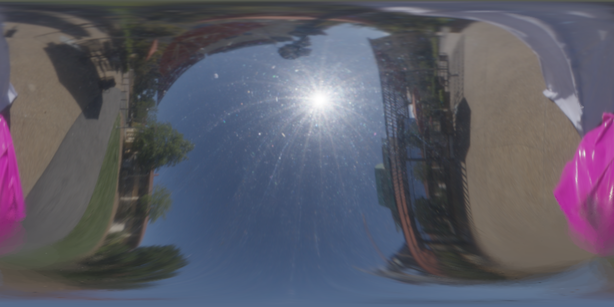}}\end{overpic}}%
  \hspace{2pt}%
  \parbox[c]{0.17\linewidth}{\centering \includegraphics[width=\linewidth]{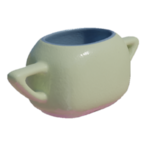}}%
  \hspace{2pt}%
  \parbox[c]{0.17\linewidth}{\centering \includegraphics[width=\linewidth]{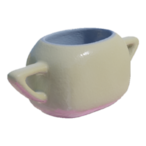}}%
  \hspace{2pt}%
  \parbox[c]{0.17\linewidth}{\centering \includegraphics[width=\linewidth]{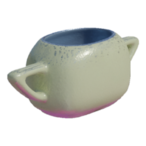}}%
  \hspace{2pt}%
  \parbox[c]{0.17\linewidth}{\centering \includegraphics[width=\linewidth]{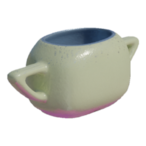}}%
  \\[0.1em]
  \parbox[c]{0.05\linewidth}{\centering \intrinsicrowlabel{Base Color}}%
  \hspace{2pt}%
  \parbox[c]{0.17\linewidth}{\centering \includegraphics[width=\linewidth]{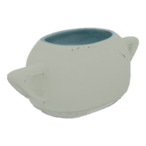}}%
  \hspace{2pt}%
  \parbox[c]{0.17\linewidth}{\centering \includegraphics[width=\linewidth]{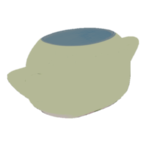}}%
  \hspace{2pt}%
  \parbox[c]{0.17\linewidth}{\centering \includegraphics[width=\linewidth]{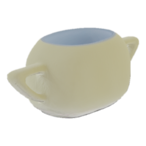}}%
  \hspace{2pt}%
  \parbox[c]{0.17\linewidth}{\centering \includegraphics[width=\linewidth]{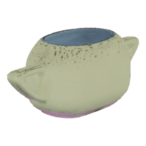}}%
  \hspace{2pt}%
  \parbox[c]{0.17\linewidth}{\centering \includegraphics[width=\linewidth]{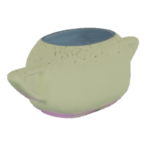}}%
  \\[0.1em]
  \parbox[c]{0.05\linewidth}{\centering \intrinsicrowlabel{Roughness}}%
  \hspace{2pt}%
  \parbox[c]{0.17\linewidth}{\centering \begin{overpic}[width=\linewidth]{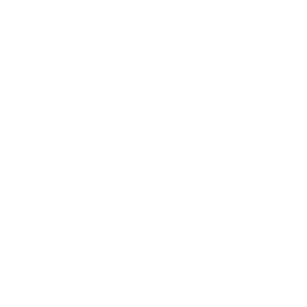}\put(50,50){\makebox(0,0){\scriptsize N/A}}\end{overpic}}%
  \hspace{2pt}%
  \parbox[c]{0.17\linewidth}{\centering \includegraphics[width=\linewidth]{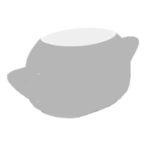}}%
  \hspace{2pt}%
  \parbox[c]{0.17\linewidth}{\centering \includegraphics[width=\linewidth]{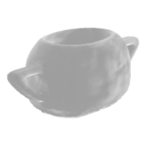}}%
  \hspace{2pt}%
  \parbox[c]{0.17\linewidth}{\centering \includegraphics[width=\linewidth]{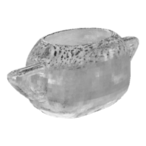}}%
  \hspace{2pt}%
  \parbox[c]{0.17\linewidth}{\centering \includegraphics[width=\linewidth]{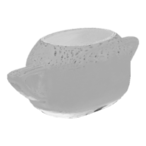}}%
  \\[0.1em]
  \parbox[c]{0.05\linewidth}{\centering \intrinsicrowlabel{Normal}}%
  \hspace{2pt}%
  \parbox[c]{0.17\linewidth}{\centering \includegraphics[width=\linewidth]{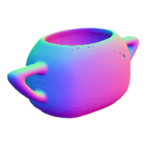}}%
  \hspace{2pt}%
  \parbox[c]{0.17\linewidth}{\centering \includegraphics[width=\linewidth]{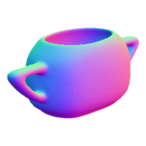}}%
  \hspace{2pt}%
  \parbox[c]{0.17\linewidth}{\centering \includegraphics[width=\linewidth]{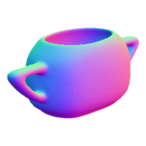}}%
  \hspace{2pt}%
  \parbox[c]{0.17\linewidth}{\centering \includegraphics[width=\linewidth]{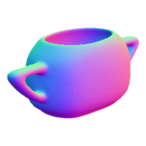}}%
  \hspace{2pt}%
  \parbox[c]{0.17\linewidth}{\centering \includegraphics[width=\linewidth]{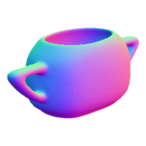}}%
  \\[0.1em]
  \parbox[c]{0.05\linewidth}{\centering \intrinsicrowlabel{Lighting}}%
  \hspace{2pt}%
  \parbox[c]{0.17\linewidth}{\centering \includegraphics[width=\linewidth]{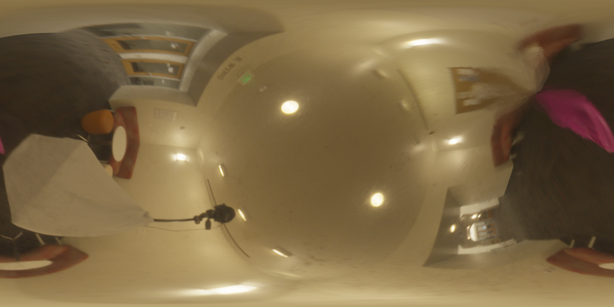}}%
  \hspace{2pt}%
  \parbox[c]{0.17\linewidth}{\centering \includegraphics[width=\linewidth]{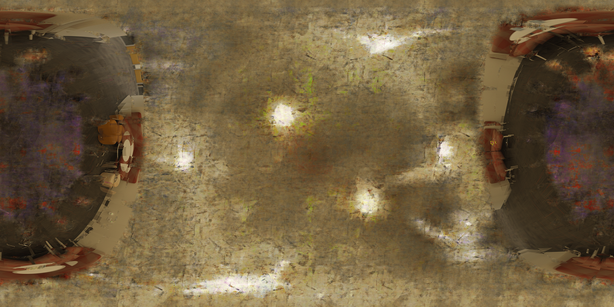}}%
  \hspace{2pt}%
  \parbox[c]{0.17\linewidth}{\centering \includegraphics[width=\linewidth]{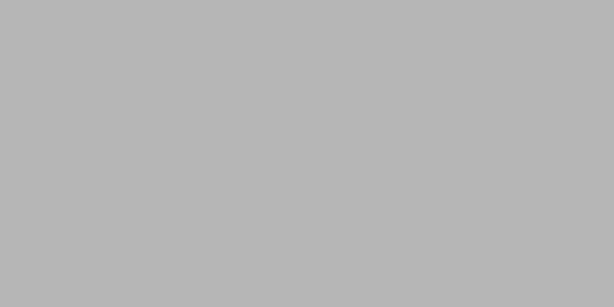}}%
  \hspace{2pt}%
  \parbox[c]{0.17\linewidth}{\centering \includegraphics[width=\linewidth]{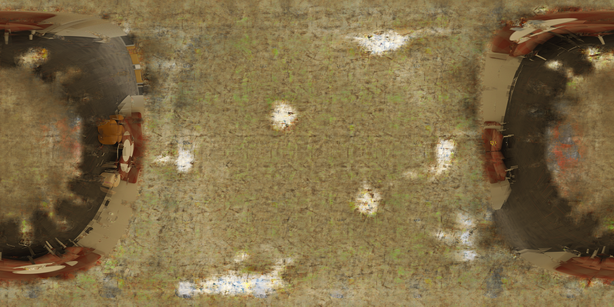}}%
  \hspace{2pt}%
  \parbox[c]{0.17\linewidth}{\centering \includegraphics[width=\linewidth]{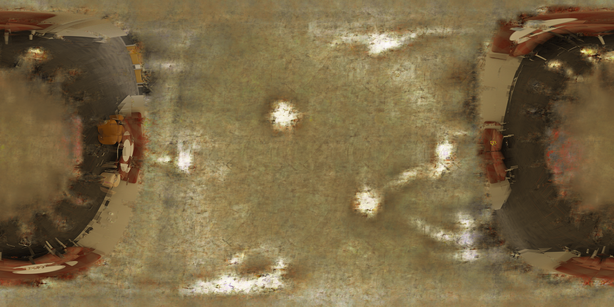}}%
  \\[1pt]
  \makebox[0.05\linewidth][c]{}%
  \hspace{2pt}%
  \makebox[0.17\linewidth][c]{\scriptsize Reference}%
  \hspace{2pt}%
  \makebox[0.17\linewidth][c]{\scriptsize Ours}%
  \hspace{2pt}%
  \makebox[0.17\linewidth][c]{\scriptsize Diffusion-BP}%
  \hspace{2pt}%
  \makebox[0.17\linewidth][c]{\scriptsize w/o reg.}%
  \hspace{2pt}%
  \makebox[0.17\linewidth][c]{\scriptsize d-s corr.}%
  \\[1pt]
  \caption{Stanford-ORB \casename{grogu_scene003} ablation intrinsic comparison.}
  \label{fig:supp-stanford-intrinsics-ablation-grogu-scene003}
\end{figure*}

\begin{figure*}[t]
  \centering
  \newcommand{\intrinsicrowlabel}[1]{\rotatebox[origin=c]{90}{\scriptsize\strut #1}}
  {\scriptsize\texttt{\detokenize{DTC-Synthetic Block_B007GE75HY_RedBlue_scene002 (Ablation)}}}\\[-0.2em]
  \parbox[c]{0.05\linewidth}{\centering \intrinsicrowlabel{Input and Prediction}}%
  \hspace{2pt}%
  \parbox[c]{0.172\linewidth}{\centering \includegraphics[width=\linewidth]{figures/summary/dtc_Block_B007GE75HY_RedBlue_scene002_intrinsic_assets_ablation/input_dr/input.png}}%
  \hspace{2pt}%
  \parbox[c]{0.172\linewidth}{\centering \includegraphics[width=\linewidth]{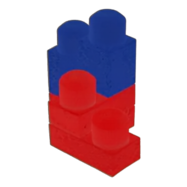}}%
  \hspace{2pt}%
  \parbox[c]{0.172\linewidth}{\centering \includegraphics[width=\linewidth]{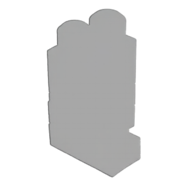}}%
  \hspace{2pt}%
  \parbox[c]{0.172\linewidth}{\centering \includegraphics[width=\linewidth]{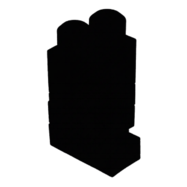}}%
  \hspace{2pt}%
  \parbox[c]{0.172\linewidth}{\centering \includegraphics[width=\linewidth]{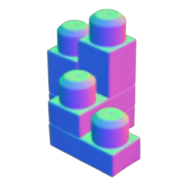}}%
  \\[1pt]
  \makebox[0.05\linewidth][c]{}%
  \hspace{2pt}%
  \makebox[0.172\linewidth][c]{\scriptsize Input}%
  \hspace{2pt}%
  \makebox[0.172\linewidth][c]{\scriptsize DR-base color}%
  \hspace{2pt}%
  \makebox[0.172\linewidth][c]{\scriptsize DR-roughness}%
  \hspace{2pt}%
  \makebox[0.172\linewidth][c]{\scriptsize DR-metallic}%
  \hspace{2pt}%
  \makebox[0.172\linewidth][c]{\scriptsize DR-normal}%
  \\[1pt]
  \vspace{0.15em}
  \par\noindent\rule{\linewidth}{0.35pt}
  \vspace{0.05em}
  \vspace{-0.6em}
  \parbox[c]{0.05\linewidth}{\centering \intrinsicrowlabel{Relight}}%
  \hspace{2pt}%
  \parbox[c]{0.17\linewidth}{\centering \begin{overpic}[width=\linewidth]{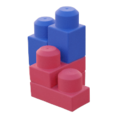}\put(0,0){\includegraphics[width=0.666\linewidth]{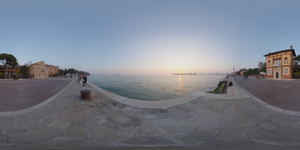}}\end{overpic}}%
  \hspace{2pt}%
  \parbox[c]{0.17\linewidth}{\centering \includegraphics[width=\linewidth]{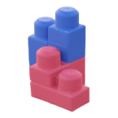}}%
  \hspace{2pt}%
  \parbox[c]{0.17\linewidth}{\centering \includegraphics[width=\linewidth]{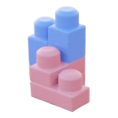}}%
  \hspace{2pt}%
  \parbox[c]{0.17\linewidth}{\centering \includegraphics[width=\linewidth]{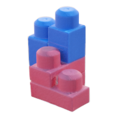}}%
  \hspace{2pt}%
  \parbox[c]{0.17\linewidth}{\centering \includegraphics[width=\linewidth]{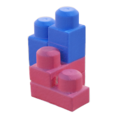}}%
  \\[0.1em]
  \parbox[c]{0.05\linewidth}{\centering \intrinsicrowlabel{Base Color}}%
  \hspace{2pt}%
  \parbox[c]{0.17\linewidth}{\centering \includegraphics[width=\linewidth]{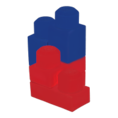}}%
  \hspace{2pt}%
  \parbox[c]{0.17\linewidth}{\centering \includegraphics[width=\linewidth]{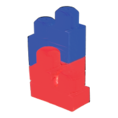}}%
  \hspace{2pt}%
  \parbox[c]{0.17\linewidth}{\centering \includegraphics[width=\linewidth]{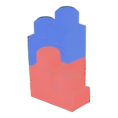}}%
  \hspace{2pt}%
  \parbox[c]{0.17\linewidth}{\centering \includegraphics[width=\linewidth]{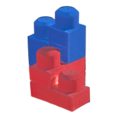}}%
  \hspace{2pt}%
  \parbox[c]{0.17\linewidth}{\centering \includegraphics[width=\linewidth]{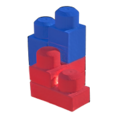}}%
  \\[0.1em]
  \parbox[c]{0.05\linewidth}{\centering \intrinsicrowlabel{Roughness}}%
  \hspace{2pt}%
  \parbox[c]{0.17\linewidth}{\centering \includegraphics[width=\linewidth]{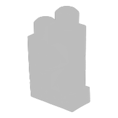}}%
  \hspace{2pt}%
  \parbox[c]{0.17\linewidth}{\centering \includegraphics[width=\linewidth]{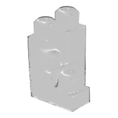}}%
  \hspace{2pt}%
  \parbox[c]{0.17\linewidth}{\centering \includegraphics[width=\linewidth]{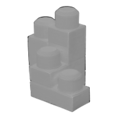}}%
  \hspace{2pt}%
  \parbox[c]{0.17\linewidth}{\centering \includegraphics[width=\linewidth]{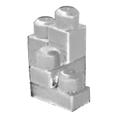}}%
  \hspace{2pt}%
  \parbox[c]{0.17\linewidth}{\centering \includegraphics[width=\linewidth]{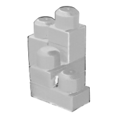}}%
  \\[0.1em]
  \parbox[c]{0.05\linewidth}{\centering \intrinsicrowlabel{Metallic}}%
  \hspace{2pt}%
  \parbox[c]{0.17\linewidth}{\centering \includegraphics[width=\linewidth]{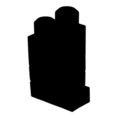}}%
  \hspace{2pt}%
  \parbox[c]{0.17\linewidth}{\centering \includegraphics[width=\linewidth]{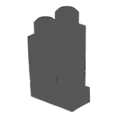}}%
  \hspace{2pt}%
  \parbox[c]{0.17\linewidth}{\centering \includegraphics[width=\linewidth]{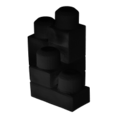}}%
  \hspace{2pt}%
  \parbox[c]{0.17\linewidth}{\centering \includegraphics[width=\linewidth]{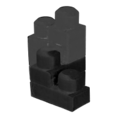}}%
  \hspace{2pt}%
  \parbox[c]{0.17\linewidth}{\centering \includegraphics[width=\linewidth]{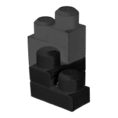}}%
  \\[0.1em]
  \parbox[c]{0.05\linewidth}{\centering \intrinsicrowlabel{Normal}}%
  \hspace{2pt}%
  \parbox[c]{0.17\linewidth}{\centering \includegraphics[width=\linewidth]{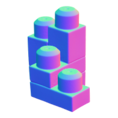}}%
  \hspace{2pt}%
  \parbox[c]{0.17\linewidth}{\centering \includegraphics[width=\linewidth]{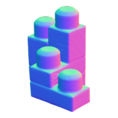}}%
  \hspace{2pt}%
  \parbox[c]{0.17\linewidth}{\centering \includegraphics[width=\linewidth]{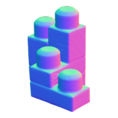}}%
  \hspace{2pt}%
  \parbox[c]{0.17\linewidth}{\centering \includegraphics[width=\linewidth]{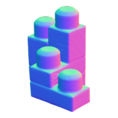}}%
  \hspace{2pt}%
  \parbox[c]{0.17\linewidth}{\centering \includegraphics[width=\linewidth]{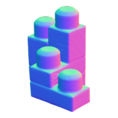}}%
  \\[0.1em]
  \parbox[c]{0.05\linewidth}{\centering \intrinsicrowlabel{Lighting}}%
  \hspace{2pt}%
  \parbox[c]{0.17\linewidth}{\centering \includegraphics[width=\linewidth]{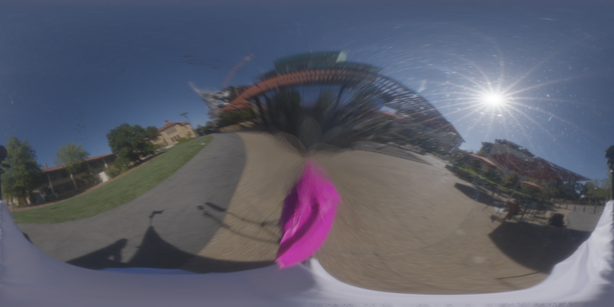}}%
  \hspace{2pt}%
  \parbox[c]{0.17\linewidth}{\centering \includegraphics[width=\linewidth]{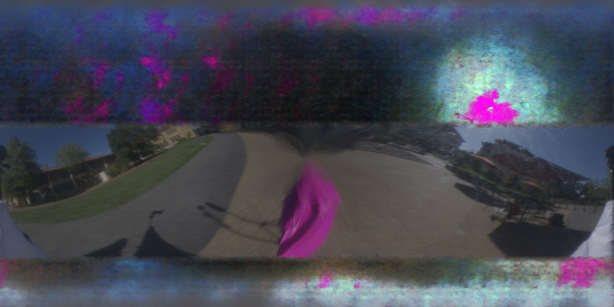}}%
  \hspace{2pt}%
  \parbox[c]{0.17\linewidth}{\centering \includegraphics[width=\linewidth]{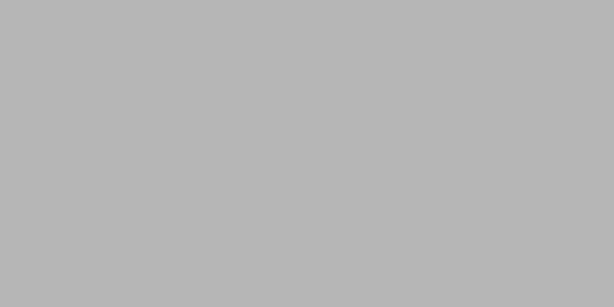}}%
  \hspace{2pt}%
  \parbox[c]{0.17\linewidth}{\centering \includegraphics[width=\linewidth]{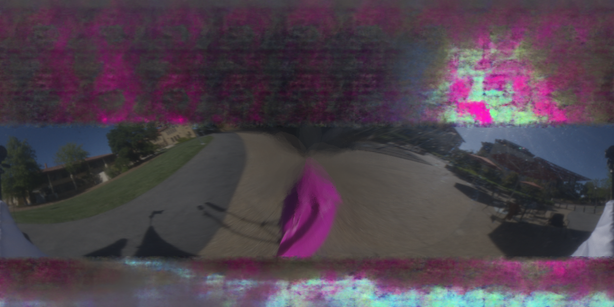}}%
  \hspace{2pt}%
  \parbox[c]{0.17\linewidth}{\centering \includegraphics[width=\linewidth]{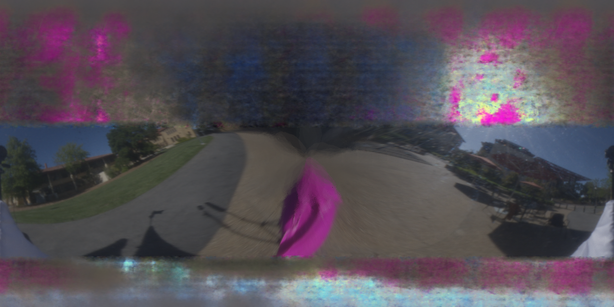}}%
  \\[1pt]
  \makebox[0.05\linewidth][c]{}%
  \hspace{2pt}%
  \makebox[0.17\linewidth][c]{\scriptsize Reference}%
  \hspace{2pt}%
  \makebox[0.17\linewidth][c]{\scriptsize Ours}%
  \hspace{2pt}%
  \makebox[0.17\linewidth][c]{\scriptsize Diffusion-BP}%
  \hspace{2pt}%
  \makebox[0.17\linewidth][c]{\scriptsize w/o reg.}%
  \hspace{2pt}%
  \makebox[0.17\linewidth][c]{\scriptsize d-s corr.}%
  \\[1pt]
  \caption{DTC-Synthetic \casename{Block_B007GE75HY_RedBlue_scene002} ablation intrinsic comparison.}
  \label{fig:supp-dtc-intrinsics-ablation-block_b007ge75hy_redblue_scene002}
\end{figure*}

\begin{figure*}[t]
  \centering
  \newcommand{\intrinsicrowlabel}[1]{\rotatebox[origin=c]{90}{\scriptsize\strut #1}}
  {\scriptsize\texttt{\detokenize{Stanford-ORB cup_scene007 (Scale-Invariant Ablation)}}}\\[-0.2em]
  \parbox[c]{0.05\linewidth}{\centering \intrinsicrowlabel{Input and Prediction}}%
  \hspace{2pt}%
  \parbox[c]{0.172\linewidth}{\centering \includegraphics[width=\linewidth]{figures/summary/stanford_orb_cup_scene007_intrinsic_assets_scale_invariant_ablation/input_dr/input.png}}%
  \hspace{2pt}%
  \parbox[c]{0.172\linewidth}{\centering \includegraphics[width=\linewidth]{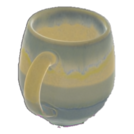}}%
  \hspace{2pt}%
  \parbox[c]{0.172\linewidth}{\centering \includegraphics[width=\linewidth]{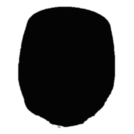}}%
  \hspace{2pt}%
  \parbox[c]{0.172\linewidth}{\centering \includegraphics[width=\linewidth]{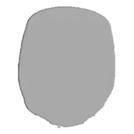}}%
  \hspace{2pt}%
  \parbox[c]{0.172\linewidth}{\centering \includegraphics[width=\linewidth]{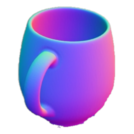}}%
  \\[1pt]
  \makebox[0.05\linewidth][c]{}%
  \hspace{2pt}%
  \makebox[0.172\linewidth][c]{\scriptsize Input}%
  \hspace{2pt}%
  \makebox[0.172\linewidth][c]{\scriptsize DR-base color}%
  \hspace{2pt}%
  \makebox[0.172\linewidth][c]{\scriptsize DR-roughness}%
  \hspace{2pt}%
  \makebox[0.172\linewidth][c]{\scriptsize DR-metallic}%
  \hspace{2pt}%
  \makebox[0.172\linewidth][c]{\scriptsize DR-normal}%
  \\[1pt]
  \vspace{0.15em}
  \par\noindent\rule{\linewidth}{0.35pt}
  \vspace{0.05em}
  \vspace{-0.6em}
  \parbox[c]{0.09\linewidth}{\centering \intrinsicrowlabel{Relight}}%
  \hspace{2pt}%
  \parbox[c]{0.215\linewidth}{\centering \begin{overpic}[width=\linewidth]{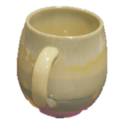}\put(0,0){\includegraphics[width=0.666\linewidth]{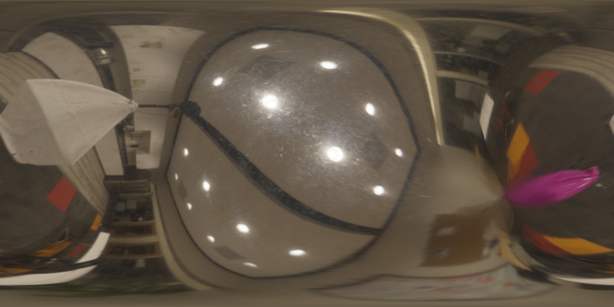}}\end{overpic}}%
  \hspace{2pt}%
  \parbox[c]{0.215\linewidth}{\centering \includegraphics[width=\linewidth]{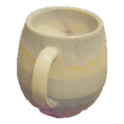}}%
  \hspace{2pt}%
  \parbox[c]{0.215\linewidth}{\centering \includegraphics[width=\linewidth]{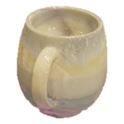}}%
  \\[0.1em]
  \parbox[c]{0.09\linewidth}{\centering \intrinsicrowlabel{Base Color}}%
  \hspace{2pt}%
  \parbox[c]{0.215\linewidth}{\centering \includegraphics[width=\linewidth]{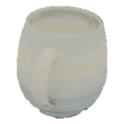}}%
  \hspace{2pt}%
  \parbox[c]{0.215\linewidth}{\centering \includegraphics[width=\linewidth]{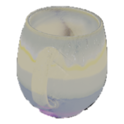}}%
  \hspace{2pt}%
  \parbox[c]{0.215\linewidth}{\centering \includegraphics[width=\linewidth]{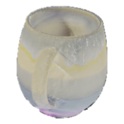}}%
  \\[0.1em]
  \parbox[c]{0.09\linewidth}{\centering \intrinsicrowlabel{Roughness}}%
  \hspace{2pt}%
  \parbox[c]{0.215\linewidth}{\centering \begin{overpic}[width=\linewidth]{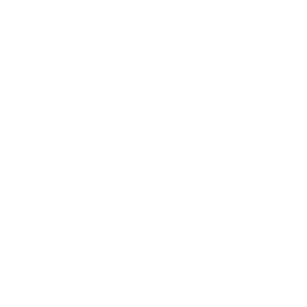}\put(50,50){\makebox(0,0){\scriptsize N/A}}\end{overpic}}%
  \hspace{2pt}%
  \parbox[c]{0.215\linewidth}{\centering \includegraphics[width=\linewidth]{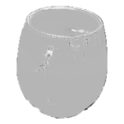}}%
  \hspace{2pt}%
  \parbox[c]{0.215\linewidth}{\centering \includegraphics[width=\linewidth]{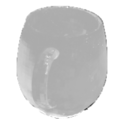}}%
  \\[0.1em]
  \parbox[c]{0.09\linewidth}{\centering \intrinsicrowlabel{Normal}}%
  \hspace{2pt}%
  \parbox[c]{0.215\linewidth}{\centering \includegraphics[width=\linewidth]{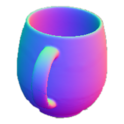}}%
  \hspace{2pt}%
  \parbox[c]{0.215\linewidth}{\centering \includegraphics[width=\linewidth]{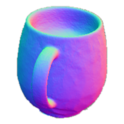}}%
  \hspace{2pt}%
  \parbox[c]{0.215\linewidth}{\centering \includegraphics[width=\linewidth]{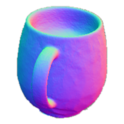}}%
  \\[0.1em]
  \parbox[c]{0.09\linewidth}{\centering \intrinsicrowlabel{Lighting}}%
  \hspace{2pt}%
  \parbox[c]{0.215\linewidth}{\centering \includegraphics[width=\linewidth]{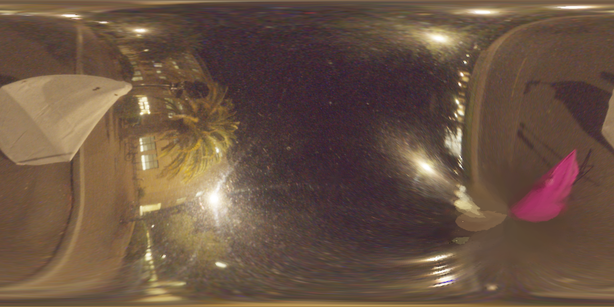}}%
  \hspace{2pt}%
  \parbox[c]{0.215\linewidth}{\centering \includegraphics[width=\linewidth]{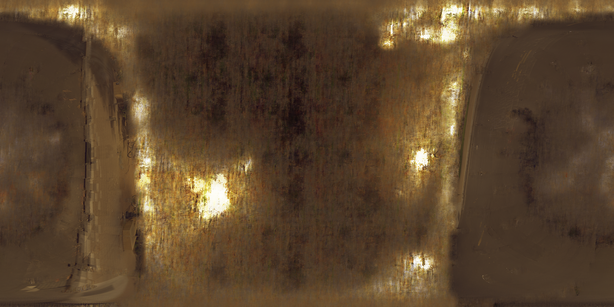}}%
  \hspace{2pt}%
  \parbox[c]{0.215\linewidth}{\centering \includegraphics[width=\linewidth]{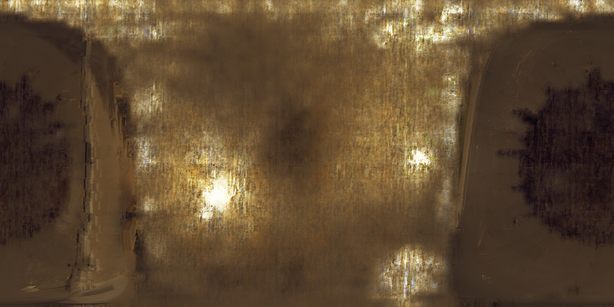}}%
  \\[1pt]
  \makebox[0.09\linewidth][c]{}%
  \hspace{2pt}%
  \makebox[0.215\linewidth][c]{\scriptsize Reference}%
  \hspace{2pt}%
  \makebox[0.215\linewidth][c]{\scriptsize Ours}%
  \hspace{2pt}%
  \makebox[0.215\linewidth][c]{\scriptsize scale inv.}%
  \\[1pt]
  \caption{Stanford-ORB \casename{cup_scene007} scale-invariant ablation intrinsic comparison.}
  \label{fig:supp-stanford-intrinsics-ablation-scale-invariant-cup-scene007}
\end{figure*}

\begin{figure*}[t]
  \centering
  \newcommand{\intrinsicrowlabel}[1]{\rotatebox[origin=c]{90}{\scriptsize\strut #1}}
  {\scriptsize\texttt{\detokenize{Stanford-ORB curry_scene001 (Scale-Invariant Ablation)}}}\\[-0.2em]
  \parbox[c]{0.05\linewidth}{\centering \intrinsicrowlabel{Input and Prediction}}%
  \hspace{2pt}%
  \parbox[c]{0.172\linewidth}{\centering \includegraphics[width=\linewidth]{figures/summary/stanford_orb_curry_scene001_intrinsic_assets_scale_invariant_ablation/input_dr/input.png}}%
  \hspace{2pt}%
  \parbox[c]{0.172\linewidth}{\centering \includegraphics[width=\linewidth]{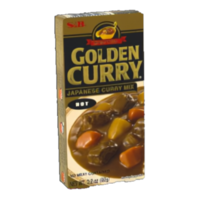}}%
  \hspace{2pt}%
  \parbox[c]{0.172\linewidth}{\centering \includegraphics[width=\linewidth]{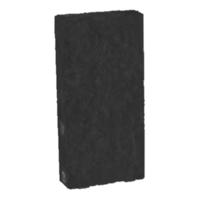}}%
  \hspace{2pt}%
  \parbox[c]{0.172\linewidth}{\centering \includegraphics[width=\linewidth]{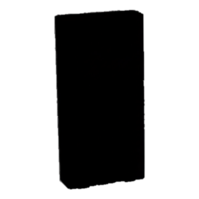}}%
  \hspace{2pt}%
  \parbox[c]{0.172\linewidth}{\centering \includegraphics[width=\linewidth]{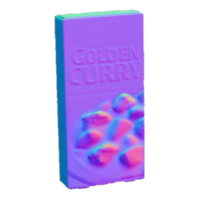}}%
  \\[1pt]
  \makebox[0.05\linewidth][c]{}%
  \hspace{2pt}%
  \makebox[0.172\linewidth][c]{\scriptsize Input}%
  \hspace{2pt}%
  \makebox[0.172\linewidth][c]{\scriptsize DR-base color}%
  \hspace{2pt}%
  \makebox[0.172\linewidth][c]{\scriptsize DR-roughness}%
  \hspace{2pt}%
  \makebox[0.172\linewidth][c]{\scriptsize DR-metallic}%
  \hspace{2pt}%
  \makebox[0.172\linewidth][c]{\scriptsize DR-normal}%
  \\[1pt]
  \vspace{0.15em}
  \par\noindent\rule{\linewidth}{0.35pt}
  \vspace{0.05em}
  \vspace{-0.6em}
  \parbox[c]{0.09\linewidth}{\centering \intrinsicrowlabel{Relight}}%
  \hspace{2pt}%
  \parbox[c]{0.215\linewidth}{\centering \begin{overpic}[width=\linewidth]{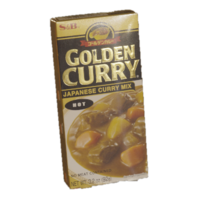}\put(0,0){\includegraphics[width=0.666\linewidth]{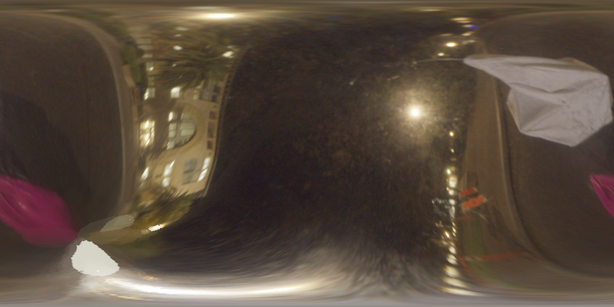}}\end{overpic}}%
  \hspace{2pt}%
  \parbox[c]{0.215\linewidth}{\centering \includegraphics[width=\linewidth]{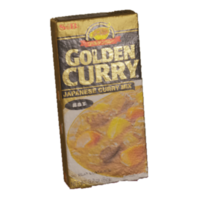}}%
  \hspace{2pt}%
  \parbox[c]{0.215\linewidth}{\centering \includegraphics[width=\linewidth]{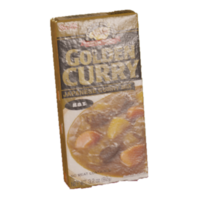}}%
  \\[0.1em]
  \parbox[c]{0.09\linewidth}{\centering \intrinsicrowlabel{Base Color}}%
  \hspace{2pt}%
  \parbox[c]{0.215\linewidth}{\centering \includegraphics[width=\linewidth]{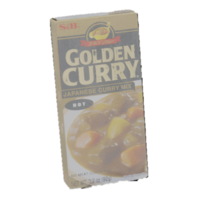}}%
  \hspace{2pt}%
  \parbox[c]{0.215\linewidth}{\centering \includegraphics[width=\linewidth]{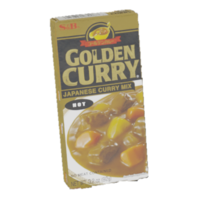}}%
  \hspace{2pt}%
  \parbox[c]{0.215\linewidth}{\centering \includegraphics[width=\linewidth]{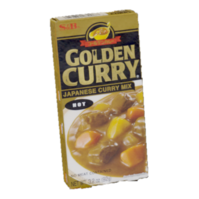}}%
  \\[0.1em]
  \parbox[c]{0.09\linewidth}{\centering \intrinsicrowlabel{Roughness}}%
  \hspace{2pt}%
  \parbox[c]{0.215\linewidth}{\centering \begin{overpic}[width=\linewidth]{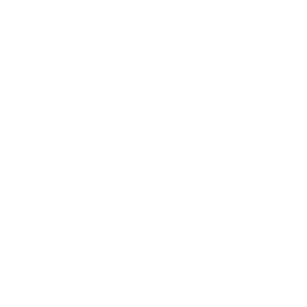}\put(50,50){\makebox(0,0){\scriptsize N/A}}\end{overpic}}%
  \hspace{2pt}%
  \parbox[c]{0.215\linewidth}{\centering \includegraphics[width=\linewidth]{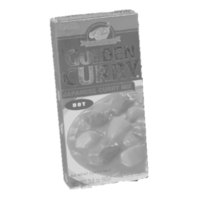}}%
  \hspace{2pt}%
  \parbox[c]{0.215\linewidth}{\centering \includegraphics[width=\linewidth]{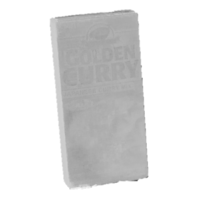}}%
  \\[0.1em]
  \parbox[c]{0.09\linewidth}{\centering \intrinsicrowlabel{Normal}}%
  \hspace{2pt}%
  \parbox[c]{0.215\linewidth}{\centering \includegraphics[width=\linewidth]{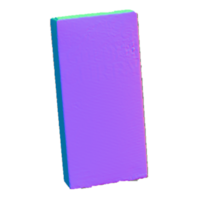}}%
  \hspace{2pt}%
  \parbox[c]{0.215\linewidth}{\centering \includegraphics[width=\linewidth]{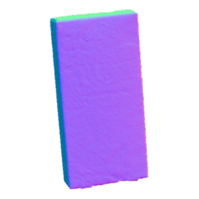}}%
  \hspace{2pt}%
  \parbox[c]{0.215\linewidth}{\centering \includegraphics[width=\linewidth]{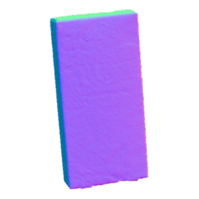}}%
  \\[0.1em]
  \parbox[c]{0.09\linewidth}{\centering \intrinsicrowlabel{Lighting}}%
  \hspace{2pt}%
  \parbox[c]{0.215\linewidth}{\centering \includegraphics[width=\linewidth]{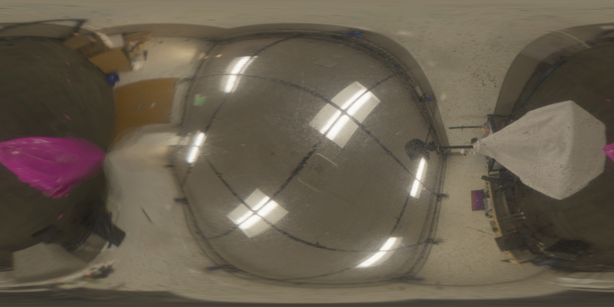}}%
  \hspace{2pt}%
  \parbox[c]{0.215\linewidth}{\centering \includegraphics[width=\linewidth]{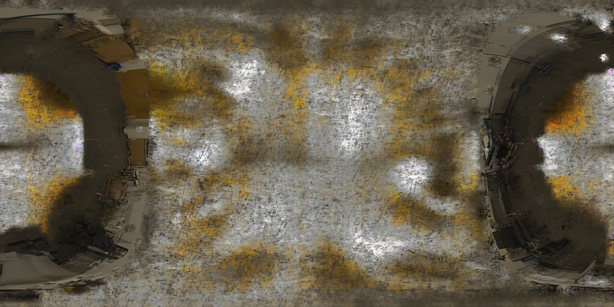}}%
  \hspace{2pt}%
  \parbox[c]{0.215\linewidth}{\centering \includegraphics[width=\linewidth]{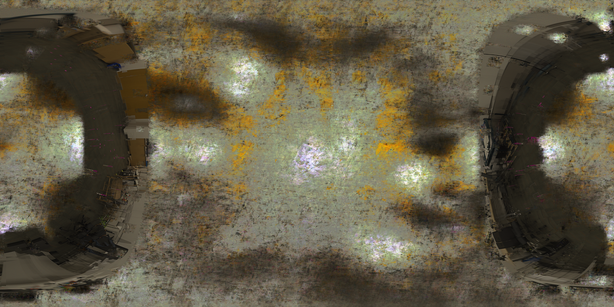}}%
  \\[1pt]
  \makebox[0.09\linewidth][c]{}%
  \hspace{2pt}%
  \makebox[0.215\linewidth][c]{\scriptsize Reference}%
  \hspace{2pt}%
  \makebox[0.215\linewidth][c]{\scriptsize Ours}%
  \hspace{2pt}%
  \makebox[0.215\linewidth][c]{\scriptsize scale inv.}%
  \\[1pt]
  \caption{Stanford-ORB \casename{curry_scene001} scale-invariant ablation intrinsic comparison.}
  \label{fig:supp-stanford-intrinsics-ablation-scale-invariant-curry-scene001}
\end{figure*}

The comparison figures supplement the qualitative relighting comparisons in the main paper, covering Stanford-ORB scenes (\cref{fig:supp-stanford-intrinsics-baking-scene003,fig:supp-stanford-intrinsics-ball-scene003,fig:supp-stanford-intrinsics-blocks-scene005,fig:supp-stanford-intrinsics-cactus-scene005,fig:supp-stanford-intrinsics-car-scene002,fig:supp-stanford-intrinsics-chips-scene003,fig:supp-stanford-intrinsics-cup-scene006,fig:supp-stanford-intrinsics-grogu-scene003,fig:supp-stanford-intrinsics-pitcher-scene001}), DTC-Synthetic scenes (\cref{fig:supp-dtc-intrinsics-teapot_b084g3k8td_yellowblacksunflowers_tu_scene002,fig:supp-dtc-intrinsics-teapot_b094fqw6q4_emeraldgoldtop_scene002}), and four Synthetic4Relight scenes (\cref{fig:supp-mii-intrinsics-air-baloons,fig:supp-mii-intrinsics-chair,fig:supp-mii-intrinsics-hotdog,fig:supp-mii-intrinsics-jugs}).
The ablation figures supplement Sec.~4.4 of the main paper, covering our material regularization ablation (\cref{fig:supp-dtc-intrinsics-ablation-block_b007ge75hy_redblue_scene002,fig:supp-stanford-intrinsics-ablation-grogu-scene003}) and our scale-invariant loss comparison (\cref{fig:supp-stanford-intrinsics-ablation-scale-invariant-cup-scene007,fig:supp-stanford-intrinsics-ablation-scale-invariant-curry-scene001}).
Metallic rows are shown for the DTC-Synthetic dataset since its ground truth is available.
In some scenes with strongly glossy objects (e.g., \cref{fig:supp-stanford-intrinsics-pitcher-scene001}), dotted high-frequency artifacts are visible on the reconstructed surface.
They are most visible around glossy highlights, where small errors in geometry, normals, or material estimates are strongly amplified.
The \emph{Normal} row confirms that these artifacts mainly originate from imperfect surface normals in the SDF reconstruction stage rather than from material estimation; our regularization mitigates but does not eliminate them.

\begin{figure*}[t]
  \centering
  \newcommand{\intrinsicrowlabel}[1]{\rotatebox[origin=c]{90}{\scriptsize\strut #1}}
  {\scriptsize\texttt{\detokenize{Stanford-ORB cactus_scene005}}}\\[-0.2em]
  \parbox[c]{0.05\linewidth}{\centering \intrinsicrowlabel{Input and Prediction}}%
  \hspace{2pt}%
  \parbox[c]{0.172\linewidth}{\centering \includegraphics[width=\linewidth]{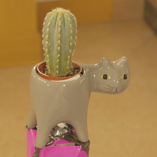}}%
  \hspace{2pt}%
  \parbox[c]{0.172\linewidth}{\centering \includegraphics[width=\linewidth]{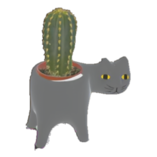}}%
  \hspace{2pt}%
  \parbox[c]{0.172\linewidth}{\centering \includegraphics[width=\linewidth]{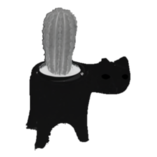}}%
  \hspace{2pt}%
  \parbox[c]{0.172\linewidth}{\centering \includegraphics[width=\linewidth]{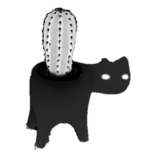}}%
  \hspace{2pt}%
  \parbox[c]{0.172\linewidth}{\centering \includegraphics[width=\linewidth]{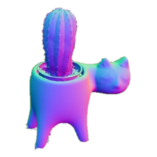}}%
  \\[1pt]
  \makebox[0.05\linewidth][c]{}%
  \hspace{2pt}%
  \makebox[0.172\linewidth][c]{\scriptsize Input}%
  \hspace{2pt}%
  \makebox[0.172\linewidth][c]{\scriptsize DR-base color}%
  \hspace{2pt}%
  \makebox[0.172\linewidth][c]{\scriptsize DR-roughness}%
  \hspace{2pt}%
  \makebox[0.172\linewidth][c]{\scriptsize DR-metallic}%
  \hspace{2pt}%
  \makebox[0.172\linewidth][c]{\scriptsize DR-normal}%
  \\[1pt]
  \vspace{0.15em}
  \par\noindent\rule{\linewidth}{0.35pt}
  \vspace{0.05em}
  \vspace{-0.6em}
  \parbox[c]{0.05\linewidth}{\centering \intrinsicrowlabel{Relight}}%
  \hspace{2pt}%
  \parbox[c]{0.215\linewidth}{\centering \begin{overpic}[width=\linewidth]{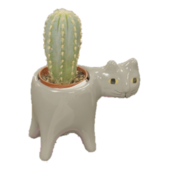}\put(0,0){\includegraphics[width=0.666\linewidth]{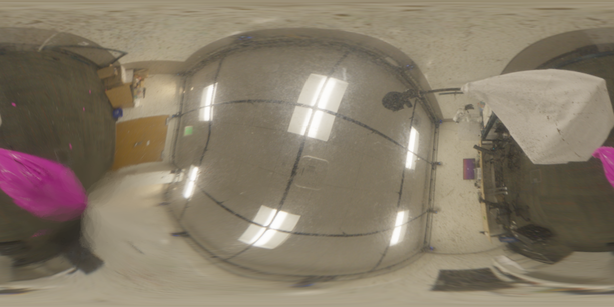}}\end{overpic}}%
  \hspace{2pt}%
  \parbox[c]{0.215\linewidth}{\centering \includegraphics[width=\linewidth]{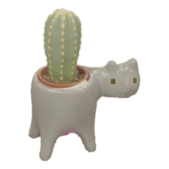}}%
  \hspace{2pt}%
  \parbox[c]{0.215\linewidth}{\centering \includegraphics[width=\linewidth]{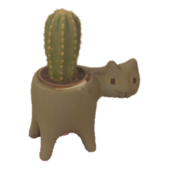}}%
  \hspace{2pt}%
  \parbox[c]{0.215\linewidth}{\centering \includegraphics[width=\linewidth]{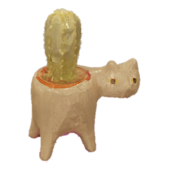}}%
  \\[0.1em]
  \parbox[c]{0.05\linewidth}{\centering \intrinsicrowlabel{Base Color}}%
  \hspace{2pt}%
  \parbox[c]{0.215\linewidth}{\centering \includegraphics[width=\linewidth]{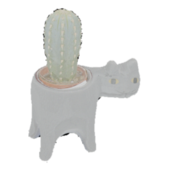}}%
  \hspace{2pt}%
  \parbox[c]{0.215\linewidth}{\centering \includegraphics[width=\linewidth]{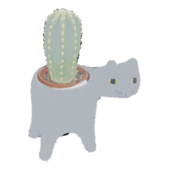}}%
  \hspace{2pt}%
  \parbox[c]{0.215\linewidth}{\centering \includegraphics[width=\linewidth]{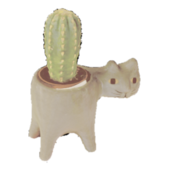}}%
  \hspace{2pt}%
  \parbox[c]{0.215\linewidth}{\centering \includegraphics[width=\linewidth]{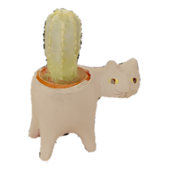}}%
  \\[0.1em]
  \parbox[c]{0.05\linewidth}{\centering \intrinsicrowlabel{Roughness}}%
  \hspace{2pt}%
  \parbox[c]{0.215\linewidth}{\centering \begin{overpic}[width=\linewidth]{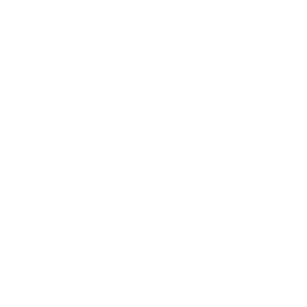}\put(50,50){\makebox(0,0){\scriptsize N/A}}\end{overpic}}%
  \hspace{2pt}%
  \parbox[c]{0.215\linewidth}{\centering \includegraphics[width=\linewidth]{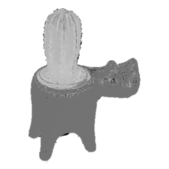}}%
  \hspace{2pt}%
  \parbox[c]{0.215\linewidth}{\centering \includegraphics[width=\linewidth]{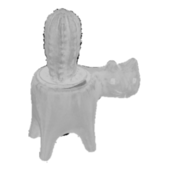}}%
  \hspace{2pt}%
  \parbox[c]{0.215\linewidth}{\centering \includegraphics[width=\linewidth]{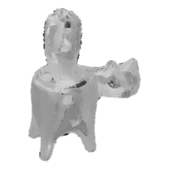}}%
  \\[0.1em]
  \parbox[c]{0.05\linewidth}{\centering \intrinsicrowlabel{Normal}}%
  \hspace{2pt}%
  \parbox[c]{0.215\linewidth}{\centering \includegraphics[width=\linewidth]{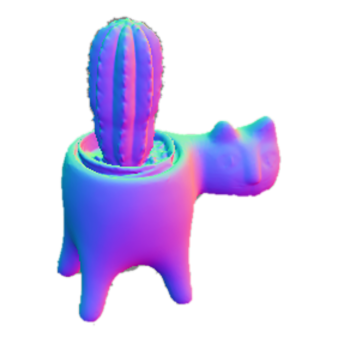}}%
  \hspace{2pt}%
  \parbox[c]{0.215\linewidth}{\centering \includegraphics[width=\linewidth]{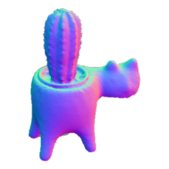}}%
  \hspace{2pt}%
  \parbox[c]{0.215\linewidth}{\centering \includegraphics[width=\linewidth]{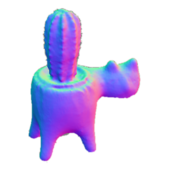}}%
  \hspace{2pt}%
  \parbox[c]{0.215\linewidth}{\centering \includegraphics[width=\linewidth]{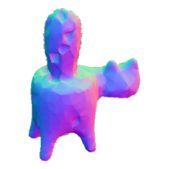}}%
  \\[0.1em]
  \parbox[c]{0.05\linewidth}{\centering \intrinsicrowlabel{Lighting}}%
  \hspace{2pt}%
  \parbox[c]{0.215\linewidth}{\centering \includegraphics[width=\linewidth]{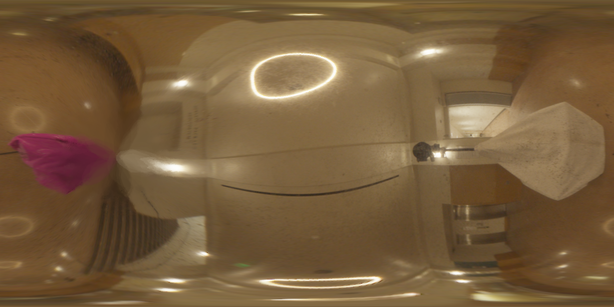}}%
  \hspace{2pt}%
  \parbox[c]{0.215\linewidth}{\centering \includegraphics[width=\linewidth]{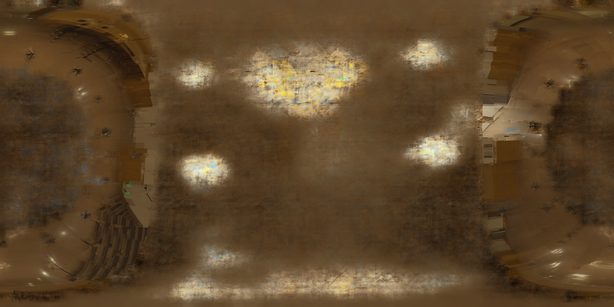}}%
  \hspace{2pt}%
  \parbox[c]{0.215\linewidth}{\centering \includegraphics[width=\linewidth]{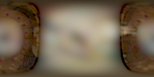}}%
  \hspace{2pt}%
  \parbox[c]{0.215\linewidth}{\centering \includegraphics[width=\linewidth]{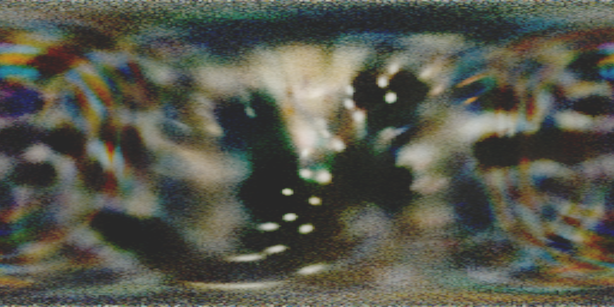}}%
  \\[1pt]
  \makebox[0.05\linewidth][c]{}%
  \hspace{2pt}%
  \makebox[0.215\linewidth][c]{\scriptsize Reference}%
  \hspace{2pt}%
  \makebox[0.215\linewidth][c]{\scriptsize Ours}%
  \hspace{2pt}%
  \makebox[0.215\linewidth][c]{\scriptsize Neural-PBIR}%
  \hspace{2pt}%
  \makebox[0.215\linewidth][c]{\scriptsize MaterialFusion}%
  \\[1pt]
  \caption{Stanford-ORB \casename{cactus_scene005}.}
  \label{fig:supp-stanford-intrinsics-cactus-scene005}
\end{figure*}

\begin{figure*}[t]
  \centering
  \newcommand{\intrinsicrowlabel}[1]{\rotatebox[origin=c]{90}{\scriptsize\strut #1}}
  {\scriptsize\texttt{\detokenize{Stanford-ORB grogu_scene003}}}\\[-0.2em]
  \parbox[c]{0.05\linewidth}{\centering \intrinsicrowlabel{Input and Prediction}}%
  \hspace{2pt}%
  \parbox[c]{0.172\linewidth}{\centering \includegraphics[width=\linewidth]{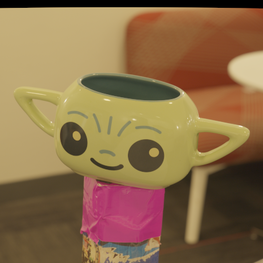}}%
  \hspace{2pt}%
  \parbox[c]{0.172\linewidth}{\centering \includegraphics[width=\linewidth]{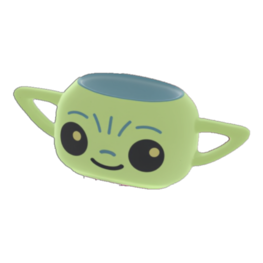}}%
  \hspace{2pt}%
  \parbox[c]{0.172\linewidth}{\centering \includegraphics[width=\linewidth]{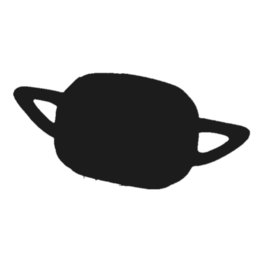}}%
  \hspace{2pt}%
  \parbox[c]{0.172\linewidth}{\centering \includegraphics[width=\linewidth]{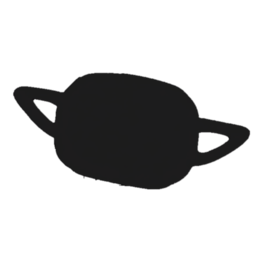}}%
  \hspace{2pt}%
  \parbox[c]{0.172\linewidth}{\centering \includegraphics[width=\linewidth]{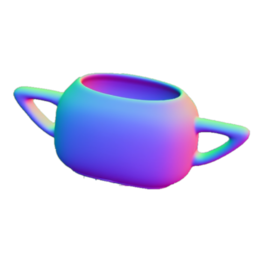}}%
  \\[1pt]
  \makebox[0.05\linewidth][c]{}%
  \hspace{2pt}%
  \makebox[0.172\linewidth][c]{\scriptsize Input}%
  \hspace{2pt}%
  \makebox[0.172\linewidth][c]{\scriptsize DR-base color}%
  \hspace{2pt}%
  \makebox[0.172\linewidth][c]{\scriptsize DR-roughness}%
  \hspace{2pt}%
  \makebox[0.172\linewidth][c]{\scriptsize DR-metallic}%
  \hspace{2pt}%
  \makebox[0.172\linewidth][c]{\scriptsize DR-normal}%
  \\[1pt]
  \vspace{0.15em}
  \par\noindent\rule{\linewidth}{0.35pt}
  \vspace{0.05em}
  \vspace{-0.6em}
  \parbox[c]{0.05\linewidth}{\centering \intrinsicrowlabel{Relight}}%
  \hspace{2pt}%
  \parbox[c]{0.215\linewidth}{\centering \begin{overpic}[width=\linewidth]{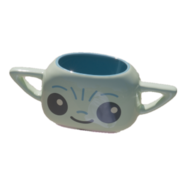}\put(0,0){\includegraphics[width=0.666\linewidth]{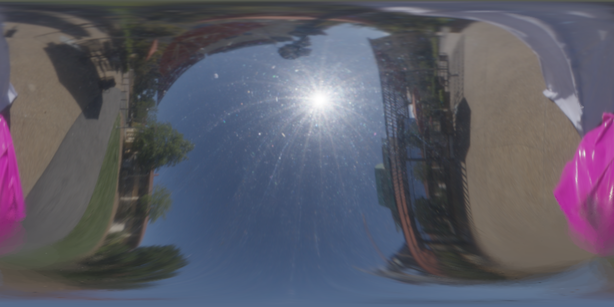}}\end{overpic}}%
  \hspace{2pt}%
  \parbox[c]{0.215\linewidth}{\centering \includegraphics[width=\linewidth]{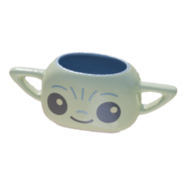}}%
  \hspace{2pt}%
  \parbox[c]{0.215\linewidth}{\centering \includegraphics[width=\linewidth]{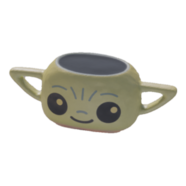}}%
  \hspace{2pt}%
  \parbox[c]{0.215\linewidth}{\centering \includegraphics[width=\linewidth]{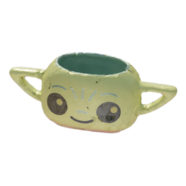}}%
  \\[0.1em]
  \parbox[c]{0.05\linewidth}{\centering \intrinsicrowlabel{Base Color}}%
  \hspace{2pt}%
  \parbox[c]{0.215\linewidth}{\centering \includegraphics[width=\linewidth]{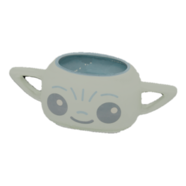}}%
  \hspace{2pt}%
  \parbox[c]{0.215\linewidth}{\centering \includegraphics[width=\linewidth]{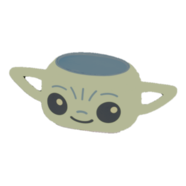}}%
  \hspace{2pt}%
  \parbox[c]{0.215\linewidth}{\centering \includegraphics[width=\linewidth]{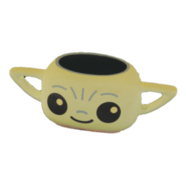}}%
  \hspace{2pt}%
  \parbox[c]{0.215\linewidth}{\centering \includegraphics[width=\linewidth]{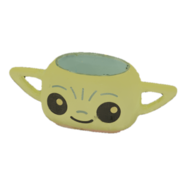}}%
  \\[0.1em]
  \parbox[c]{0.05\linewidth}{\centering \intrinsicrowlabel{Roughness}}%
  \hspace{2pt}%
  \parbox[c]{0.215\linewidth}{\centering \begin{overpic}[width=\linewidth]{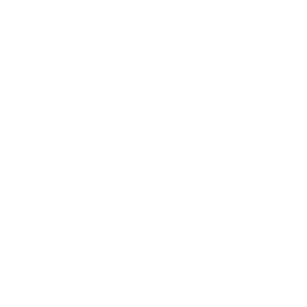}\put(50,50){\makebox(0,0){\scriptsize N/A}}\end{overpic}}%
  \hspace{2pt}%
  \parbox[c]{0.215\linewidth}{\centering \includegraphics[width=\linewidth]{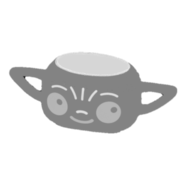}}%
  \hspace{2pt}%
  \parbox[c]{0.215\linewidth}{\centering \includegraphics[width=\linewidth]{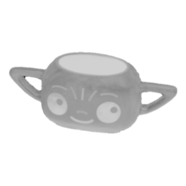}}%
  \hspace{2pt}%
  \parbox[c]{0.215\linewidth}{\centering \includegraphics[width=\linewidth]{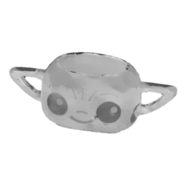}}%
  \\[0.1em]
  \parbox[c]{0.05\linewidth}{\centering \intrinsicrowlabel{Normal}}%
  \hspace{2pt}%
  \parbox[c]{0.215\linewidth}{\centering \includegraphics[width=\linewidth]{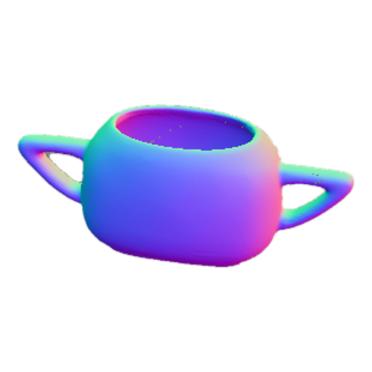}}%
  \hspace{2pt}%
  \parbox[c]{0.215\linewidth}{\centering \includegraphics[width=\linewidth]{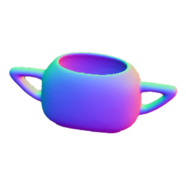}}%
  \hspace{2pt}%
  \parbox[c]{0.215\linewidth}{\centering \includegraphics[width=\linewidth]{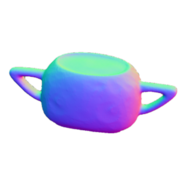}}%
  \hspace{2pt}%
  \parbox[c]{0.215\linewidth}{\centering \includegraphics[width=\linewidth]{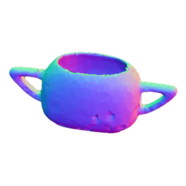}}%
  \\[0.1em]
  \parbox[c]{0.05\linewidth}{\centering \intrinsicrowlabel{Lighting}}%
  \hspace{2pt}%
  \parbox[c]{0.215\linewidth}{\centering \includegraphics[width=\linewidth]{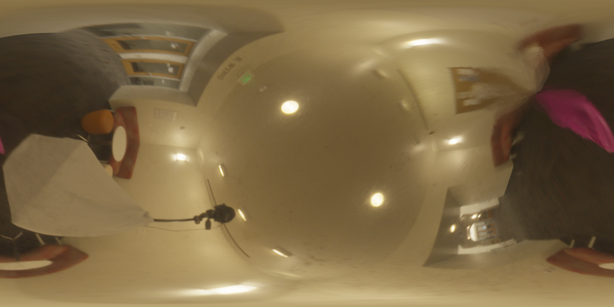}}%
  \hspace{2pt}%
  \parbox[c]{0.215\linewidth}{\centering \includegraphics[width=\linewidth]{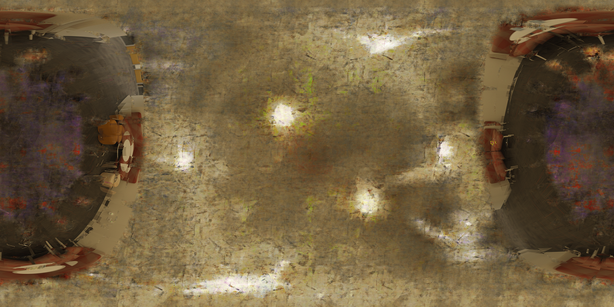}}%
  \hspace{2pt}%
  \parbox[c]{0.215\linewidth}{\centering \includegraphics[width=\linewidth]{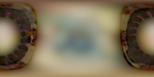}}%
  \hspace{2pt}%
  \parbox[c]{0.215\linewidth}{\centering \includegraphics[width=\linewidth]{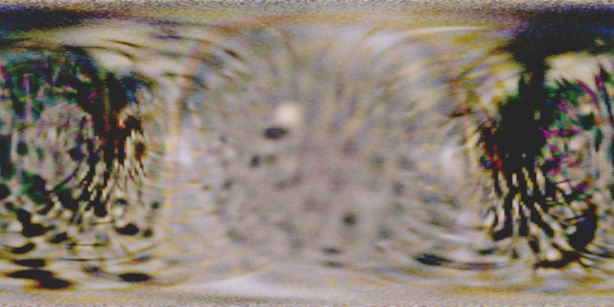}}%
  \\[1pt]
  \makebox[0.05\linewidth][c]{}%
  \hspace{2pt}%
  \makebox[0.215\linewidth][c]{\scriptsize Reference}%
  \hspace{2pt}%
  \makebox[0.215\linewidth][c]{\scriptsize Ours}%
  \hspace{2pt}%
  \makebox[0.215\linewidth][c]{\scriptsize Neural-PBIR}%
  \hspace{2pt}%
  \makebox[0.215\linewidth][c]{\scriptsize MaterialFusion}%
  \\[1pt]
  \caption{Stanford-ORB \casename{grogu_scene003}.}
  \label{fig:supp-stanford-intrinsics-grogu-scene003}
\end{figure*}

\begin{figure*}[t]
  \centering
  \newcommand{\intrinsicrowlabel}[1]{\rotatebox[origin=c]{90}{\scriptsize\strut #1}}
  {\scriptsize\texttt{\detokenize{Stanford-ORB baking_scene003}}}\\[-0.2em]
  \parbox[c]{0.05\linewidth}{\centering \intrinsicrowlabel{Input and Prediction}}%
  \hspace{2pt}%
  \parbox[c]{0.172\linewidth}{\centering \includegraphics[width=\linewidth]{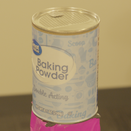}}%
  \hspace{2pt}%
  \parbox[c]{0.172\linewidth}{\centering \includegraphics[width=\linewidth]{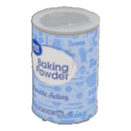}}%
  \hspace{2pt}%
  \parbox[c]{0.172\linewidth}{\centering \includegraphics[width=\linewidth]{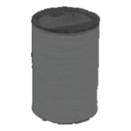}}%
  \hspace{2pt}%
  \parbox[c]{0.172\linewidth}{\centering \includegraphics[width=\linewidth]{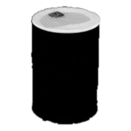}}%
  \hspace{2pt}%
  \parbox[c]{0.172\linewidth}{\centering \includegraphics[width=\linewidth]{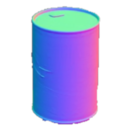}}%
  \\[1pt]
  \makebox[0.05\linewidth][c]{}%
  \hspace{2pt}%
  \makebox[0.172\linewidth][c]{\scriptsize Input}%
  \hspace{2pt}%
  \makebox[0.172\linewidth][c]{\scriptsize DR-base color}%
  \hspace{2pt}%
  \makebox[0.172\linewidth][c]{\scriptsize DR-roughness}%
  \hspace{2pt}%
  \makebox[0.172\linewidth][c]{\scriptsize DR-metallic}%
  \hspace{2pt}%
  \makebox[0.172\linewidth][c]{\scriptsize DR-normal}%
  \\[1pt]
  \vspace{0.15em}
  \par\noindent\rule{\linewidth}{0.35pt}
  \vspace{0.05em}
  \vspace{-0.6em}
  \parbox[c]{0.05\linewidth}{\centering \intrinsicrowlabel{Relight}}%
  \hspace{2pt}%
  \parbox[c]{0.215\linewidth}{\centering \begin{overpic}[width=\linewidth]{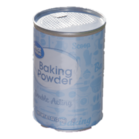}\put(0,0){\includegraphics[width=0.666\linewidth]{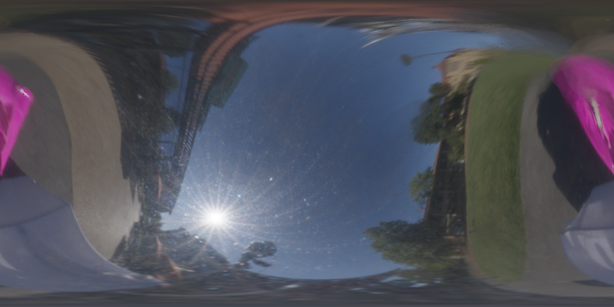}}\end{overpic}}%
  \hspace{2pt}%
  \parbox[c]{0.215\linewidth}{\centering \includegraphics[width=\linewidth]{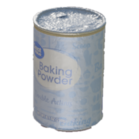}}%
  \hspace{2pt}%
  \parbox[c]{0.215\linewidth}{\centering \includegraphics[width=\linewidth]{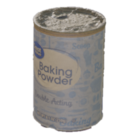}}%
  \hspace{2pt}%
  \parbox[c]{0.215\linewidth}{\centering \includegraphics[width=\linewidth]{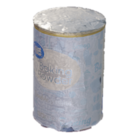}}%
  \\[0.1em]
  \parbox[c]{0.05\linewidth}{\centering \intrinsicrowlabel{Base Color}}%
  \hspace{2pt}%
  \parbox[c]{0.215\linewidth}{\centering \includegraphics[width=\linewidth]{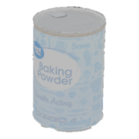}}%
  \hspace{2pt}%
  \parbox[c]{0.215\linewidth}{\centering \includegraphics[width=\linewidth]{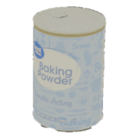}}%
  \hspace{2pt}%
  \parbox[c]{0.215\linewidth}{\centering \includegraphics[width=\linewidth]{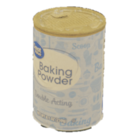}}%
  \hspace{2pt}%
  \parbox[c]{0.215\linewidth}{\centering \includegraphics[width=\linewidth]{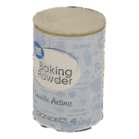}}%
  \\[0.1em]
  \parbox[c]{0.05\linewidth}{\centering \intrinsicrowlabel{Roughness}}%
  \hspace{2pt}%
  \parbox[c]{0.215\linewidth}{\centering \begin{overpic}[width=\linewidth]{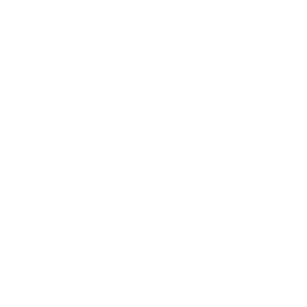}\put(50,50){\makebox(0,0){\scriptsize N/A}}\end{overpic}}%
  \hspace{2pt}%
  \parbox[c]{0.215\linewidth}{\centering \includegraphics[width=\linewidth]{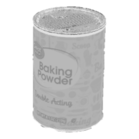}}%
  \hspace{2pt}%
  \parbox[c]{0.215\linewidth}{\centering \includegraphics[width=\linewidth]{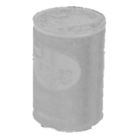}}%
  \hspace{2pt}%
  \parbox[c]{0.215\linewidth}{\centering \includegraphics[width=\linewidth]{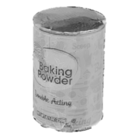}}%
  \\[0.1em]
  \parbox[c]{0.05\linewidth}{\centering \intrinsicrowlabel{Normal}}%
  \hspace{2pt}%
  \parbox[c]{0.215\linewidth}{\centering \includegraphics[width=\linewidth]{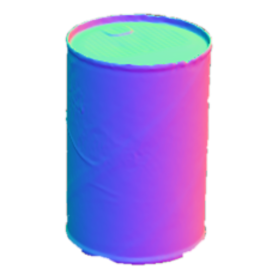}}%
  \hspace{2pt}%
  \parbox[c]{0.215\linewidth}{\centering \includegraphics[width=\linewidth]{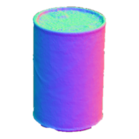}}%
  \hspace{2pt}%
  \parbox[c]{0.215\linewidth}{\centering \includegraphics[width=\linewidth]{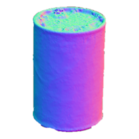}}%
  \hspace{2pt}%
  \parbox[c]{0.215\linewidth}{\centering \includegraphics[width=\linewidth]{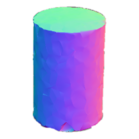}}%
  \\[0.1em]
  \parbox[c]{0.05\linewidth}{\centering \intrinsicrowlabel{Lighting}}%
  \hspace{2pt}%
  \parbox[c]{0.215\linewidth}{\centering \includegraphics[width=\linewidth]{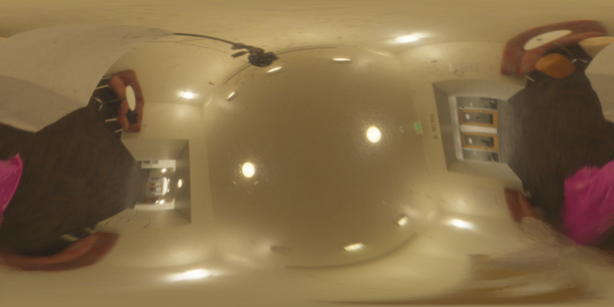}}%
  \hspace{2pt}%
  \parbox[c]{0.215\linewidth}{\centering \includegraphics[width=\linewidth]{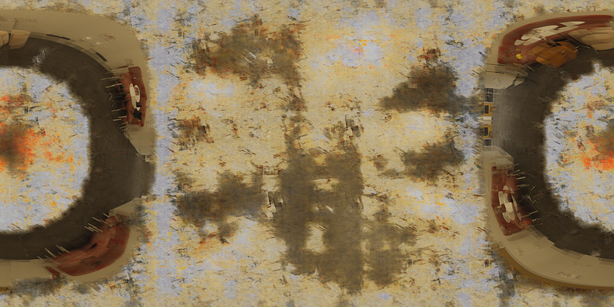}}%
  \hspace{2pt}%
  \parbox[c]{0.215\linewidth}{\centering \includegraphics[width=\linewidth]{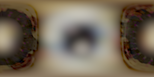}}%
  \hspace{2pt}%
  \parbox[c]{0.215\linewidth}{\centering \includegraphics[width=\linewidth]{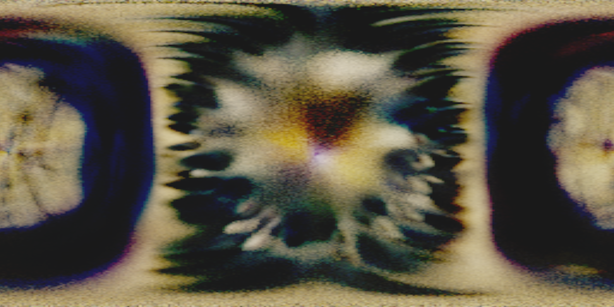}}%
  \\[1pt]
  \makebox[0.05\linewidth][c]{}%
  \hspace{2pt}%
  \makebox[0.215\linewidth][c]{\scriptsize Reference}%
  \hspace{2pt}%
  \makebox[0.215\linewidth][c]{\scriptsize Ours}%
  \hspace{2pt}%
  \makebox[0.215\linewidth][c]{\scriptsize Neural-PBIR}%
  \hspace{2pt}%
  \makebox[0.215\linewidth][c]{\scriptsize MaterialFusion}%
  \\[1pt]
  \caption{Stanford-ORB \casename{baking_scene003}.}
  \label{fig:supp-stanford-intrinsics-baking-scene003}
\end{figure*}

\begin{figure*}[t]
  \centering
  \newcommand{\intrinsicrowlabel}[1]{\rotatebox[origin=c]{90}{\scriptsize\strut #1}}
  {\scriptsize\texttt{\detokenize{Stanford-ORB ball_scene003}}}\\[-0.2em]
  \parbox[c]{0.05\linewidth}{\centering \intrinsicrowlabel{Input and Prediction}}%
  \hspace{2pt}%
  \parbox[c]{0.172\linewidth}{\centering \includegraphics[width=\linewidth]{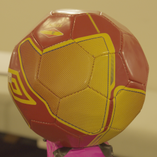}}%
  \hspace{2pt}%
  \parbox[c]{0.172\linewidth}{\centering \includegraphics[width=\linewidth]{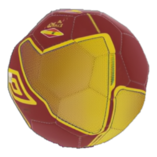}}%
  \hspace{2pt}%
  \parbox[c]{0.172\linewidth}{\centering \includegraphics[width=\linewidth]{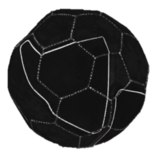}}%
  \hspace{2pt}%
  \parbox[c]{0.172\linewidth}{\centering \includegraphics[width=\linewidth]{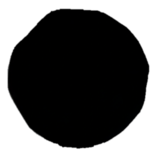}}%
  \hspace{2pt}%
  \parbox[c]{0.172\linewidth}{\centering \includegraphics[width=\linewidth]{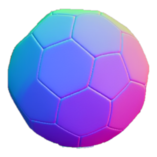}}%
  \\[1pt]
  \makebox[0.05\linewidth][c]{}%
  \hspace{2pt}%
  \makebox[0.172\linewidth][c]{\scriptsize Input}%
  \hspace{2pt}%
  \makebox[0.172\linewidth][c]{\scriptsize DR-base color}%
  \hspace{2pt}%
  \makebox[0.172\linewidth][c]{\scriptsize DR-roughness}%
  \hspace{2pt}%
  \makebox[0.172\linewidth][c]{\scriptsize DR-metallic}%
  \hspace{2pt}%
  \makebox[0.172\linewidth][c]{\scriptsize DR-normal}%
  \\[1pt]
  \vspace{0.15em}
  \par\noindent\rule{\linewidth}{0.35pt}
  \vspace{0.05em}
  \vspace{-0.6em}
  \parbox[c]{0.05\linewidth}{\centering \intrinsicrowlabel{Relight}}%
  \hspace{2pt}%
  \parbox[c]{0.215\linewidth}{\centering \begin{overpic}[width=\linewidth]{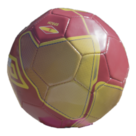}\put(0,0){\includegraphics[width=0.666\linewidth]{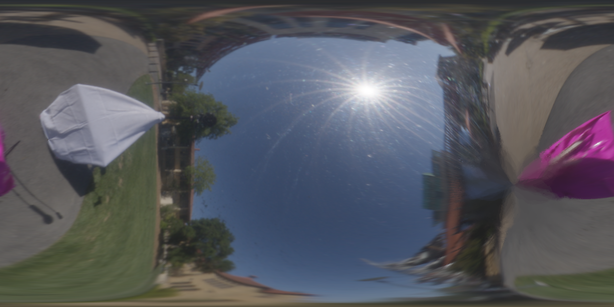}}\end{overpic}}%
  \hspace{2pt}%
  \parbox[c]{0.215\linewidth}{\centering \includegraphics[width=\linewidth]{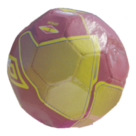}}%
  \hspace{2pt}%
  \parbox[c]{0.215\linewidth}{\centering \includegraphics[width=\linewidth]{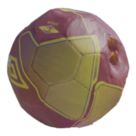}}%
  \hspace{2pt}%
  \parbox[c]{0.215\linewidth}{\centering \includegraphics[width=\linewidth]{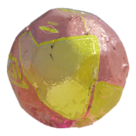}}%
  \\[0.1em]
  \parbox[c]{0.05\linewidth}{\centering \intrinsicrowlabel{Base Color}}%
  \hspace{2pt}%
  \parbox[c]{0.215\linewidth}{\centering \includegraphics[width=\linewidth]{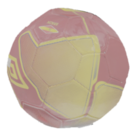}}%
  \hspace{2pt}%
  \parbox[c]{0.215\linewidth}{\centering \includegraphics[width=\linewidth]{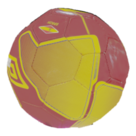}}%
  \hspace{2pt}%
  \parbox[c]{0.215\linewidth}{\centering \includegraphics[width=\linewidth]{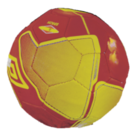}}%
  \hspace{2pt}%
  \parbox[c]{0.215\linewidth}{\centering \includegraphics[width=\linewidth]{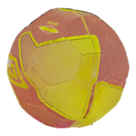}}%
  \\[0.1em]
  \parbox[c]{0.05\linewidth}{\centering \intrinsicrowlabel{Roughness}}%
  \hspace{2pt}%
  \parbox[c]{0.215\linewidth}{\centering \begin{overpic}[width=\linewidth]{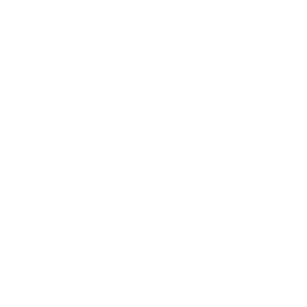}\put(50,50){\makebox(0,0){\scriptsize N/A}}\end{overpic}}%
  \hspace{2pt}%
  \parbox[c]{0.215\linewidth}{\centering \includegraphics[width=\linewidth]{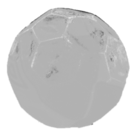}}%
  \hspace{2pt}%
  \parbox[c]{0.215\linewidth}{\centering \includegraphics[width=\linewidth]{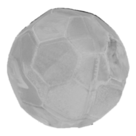}}%
  \hspace{2pt}%
  \parbox[c]{0.215\linewidth}{\centering \includegraphics[width=\linewidth]{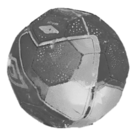}}%
  \\[0.1em]
  \parbox[c]{0.05\linewidth}{\centering \intrinsicrowlabel{Normal}}%
  \hspace{2pt}%
  \parbox[c]{0.215\linewidth}{\centering \includegraphics[width=\linewidth]{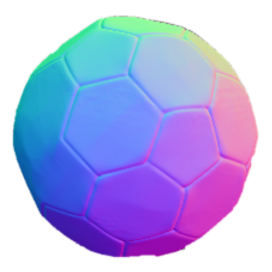}}%
  \hspace{2pt}%
  \parbox[c]{0.215\linewidth}{\centering \includegraphics[width=\linewidth]{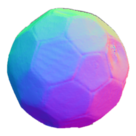}}%
  \hspace{2pt}%
  \parbox[c]{0.215\linewidth}{\centering \includegraphics[width=\linewidth]{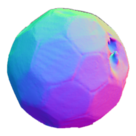}}%
  \hspace{2pt}%
  \parbox[c]{0.215\linewidth}{\centering \includegraphics[width=\linewidth]{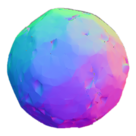}}%
  \\[0.1em]
  \parbox[c]{0.05\linewidth}{\centering \intrinsicrowlabel{Lighting}}%
  \hspace{2pt}%
  \parbox[c]{0.215\linewidth}{\centering \includegraphics[width=\linewidth]{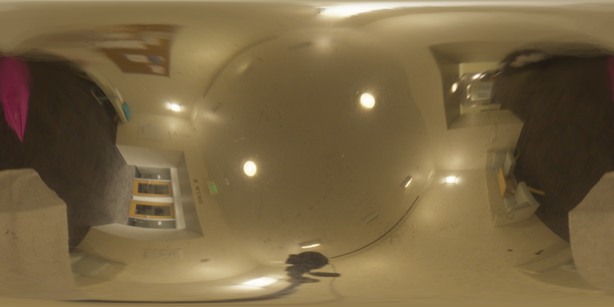}}%
  \hspace{2pt}%
  \parbox[c]{0.215\linewidth}{\centering \includegraphics[width=\linewidth]{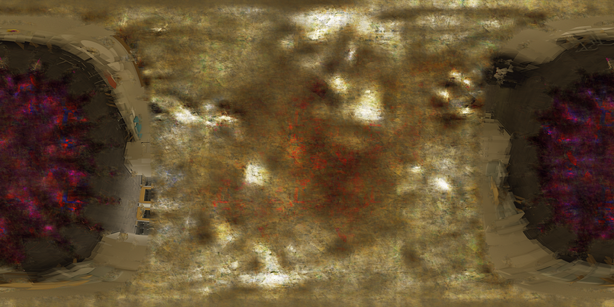}}%
  \hspace{2pt}%
  \parbox[c]{0.215\linewidth}{\centering \includegraphics[width=\linewidth]{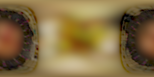}}%
  \hspace{2pt}%
  \parbox[c]{0.215\linewidth}{\centering \includegraphics[width=\linewidth]{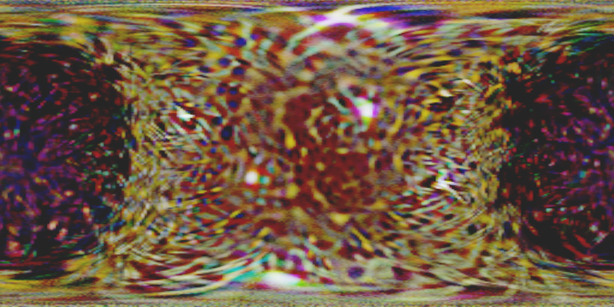}}%
  \\[1pt]
  \makebox[0.05\linewidth][c]{}%
  \hspace{2pt}%
  \makebox[0.215\linewidth][c]{\scriptsize Reference}%
  \hspace{2pt}%
  \makebox[0.215\linewidth][c]{\scriptsize Ours}%
  \hspace{2pt}%
  \makebox[0.215\linewidth][c]{\scriptsize Neural-PBIR}%
  \hspace{2pt}%
  \makebox[0.215\linewidth][c]{\scriptsize MaterialFusion}%
  \\[1pt]
  \caption{Stanford-ORB \casename{ball_scene003}.}
  \label{fig:supp-stanford-intrinsics-ball-scene003}
\end{figure*}

\begin{figure*}[t]
  \centering
  \newcommand{\intrinsicrowlabel}[1]{\rotatebox[origin=c]{90}{\scriptsize\strut #1}}
  {\scriptsize\texttt{\detokenize{Stanford-ORB blocks_scene005}}}\\[-0.2em]
  \parbox[c]{0.05\linewidth}{\centering \intrinsicrowlabel{Input and Prediction}}%
  \hspace{2pt}%
  \parbox[c]{0.172\linewidth}{\centering \includegraphics[width=\linewidth]{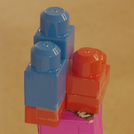}}%
  \hspace{2pt}%
  \parbox[c]{0.172\linewidth}{\centering \includegraphics[width=\linewidth]{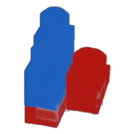}}%
  \hspace{2pt}%
  \parbox[c]{0.172\linewidth}{\centering \includegraphics[width=\linewidth]{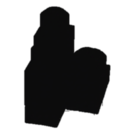}}%
  \hspace{2pt}%
  \parbox[c]{0.172\linewidth}{\centering \includegraphics[width=\linewidth]{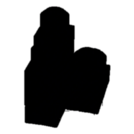}}%
  \hspace{2pt}%
  \parbox[c]{0.172\linewidth}{\centering \includegraphics[width=\linewidth]{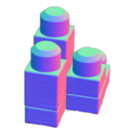}}%
  \\[1pt]
  \makebox[0.05\linewidth][c]{}%
  \hspace{2pt}%
  \makebox[0.172\linewidth][c]{\scriptsize Input}%
  \hspace{2pt}%
  \makebox[0.172\linewidth][c]{\scriptsize DR-base color}%
  \hspace{2pt}%
  \makebox[0.172\linewidth][c]{\scriptsize DR-roughness}%
  \hspace{2pt}%
  \makebox[0.172\linewidth][c]{\scriptsize DR-metallic}%
  \hspace{2pt}%
  \makebox[0.172\linewidth][c]{\scriptsize DR-normal}%
  \\[1pt]
  \vspace{0.15em}
  \par\noindent\rule{\linewidth}{0.35pt}
  \vspace{0.05em}
  \vspace{-0.6em}
  \parbox[c]{0.05\linewidth}{\centering \intrinsicrowlabel{Relight}}%
  \hspace{2pt}%
  \parbox[c]{0.215\linewidth}{\centering \begin{overpic}[width=\linewidth]{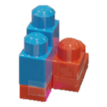}\put(0,0){\includegraphics[width=0.666\linewidth]{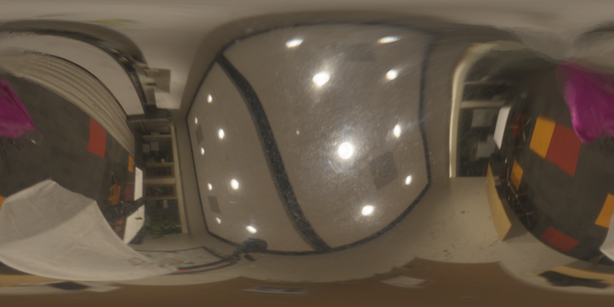}}\end{overpic}}%
  \hspace{2pt}%
  \parbox[c]{0.215\linewidth}{\centering \includegraphics[width=\linewidth]{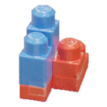}}%
  \hspace{2pt}%
  \parbox[c]{0.215\linewidth}{\centering \includegraphics[width=\linewidth]{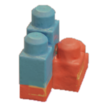}}%
  \hspace{2pt}%
  \parbox[c]{0.215\linewidth}{\centering \includegraphics[width=\linewidth]{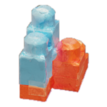}}%
  \\[0.1em]
  \parbox[c]{0.05\linewidth}{\centering \intrinsicrowlabel{Base Color}}%
  \hspace{2pt}%
  \parbox[c]{0.215\linewidth}{\centering \includegraphics[width=\linewidth]{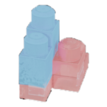}}%
  \hspace{2pt}%
  \parbox[c]{0.215\linewidth}{\centering \includegraphics[width=\linewidth]{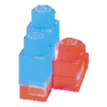}}%
  \hspace{2pt}%
  \parbox[c]{0.215\linewidth}{\centering \includegraphics[width=\linewidth]{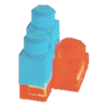}}%
  \hspace{2pt}%
  \parbox[c]{0.215\linewidth}{\centering \includegraphics[width=\linewidth]{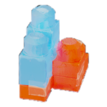}}%
  \\[0.1em]
  \parbox[c]{0.05\linewidth}{\centering \intrinsicrowlabel{Roughness}}%
  \hspace{2pt}%
  \parbox[c]{0.215\linewidth}{\centering \begin{overpic}[width=\linewidth]{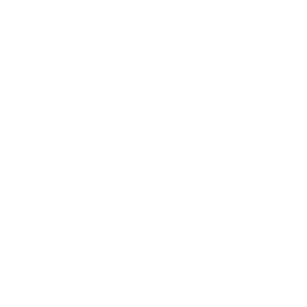}\put(50,50){\makebox(0,0){\scriptsize N/A}}\end{overpic}}%
  \hspace{2pt}%
  \parbox[c]{0.215\linewidth}{\centering \includegraphics[width=\linewidth]{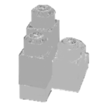}}%
  \hspace{2pt}%
  \parbox[c]{0.215\linewidth}{\centering \includegraphics[width=\linewidth]{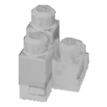}}%
  \hspace{2pt}%
  \parbox[c]{0.215\linewidth}{\centering \includegraphics[width=\linewidth]{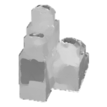}}%
  \\[0.1em]
  \parbox[c]{0.05\linewidth}{\centering \intrinsicrowlabel{Normal}}%
  \hspace{2pt}%
  \parbox[c]{0.215\linewidth}{\centering \includegraphics[width=\linewidth]{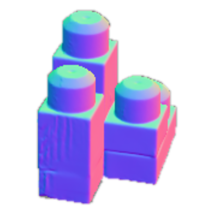}}%
  \hspace{2pt}%
  \parbox[c]{0.215\linewidth}{\centering \includegraphics[width=\linewidth]{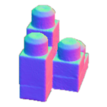}}%
  \hspace{2pt}%
  \parbox[c]{0.215\linewidth}{\centering \includegraphics[width=\linewidth]{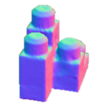}}%
  \hspace{2pt}%
  \parbox[c]{0.215\linewidth}{\centering \includegraphics[width=\linewidth]{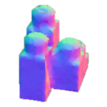}}%
  \\[0.1em]
  \parbox[c]{0.05\linewidth}{\centering \intrinsicrowlabel{Lighting}}%
  \hspace{2pt}%
  \parbox[c]{0.215\linewidth}{\centering \includegraphics[width=\linewidth]{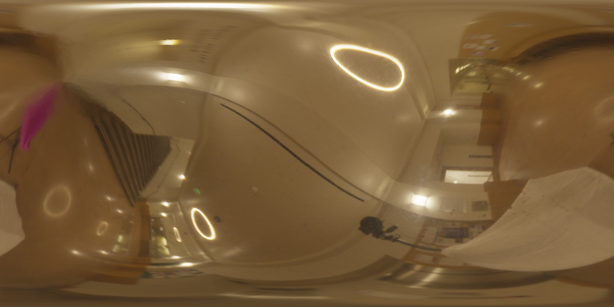}}%
  \hspace{2pt}%
  \parbox[c]{0.215\linewidth}{\centering \includegraphics[width=\linewidth]{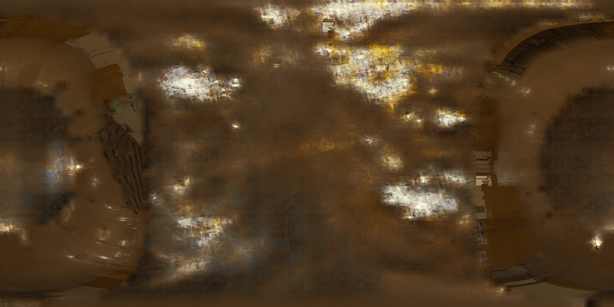}}%
  \hspace{2pt}%
  \parbox[c]{0.215\linewidth}{\centering \includegraphics[width=\linewidth]{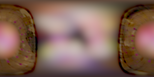}}%
  \hspace{2pt}%
  \parbox[c]{0.215\linewidth}{\centering \includegraphics[width=\linewidth]{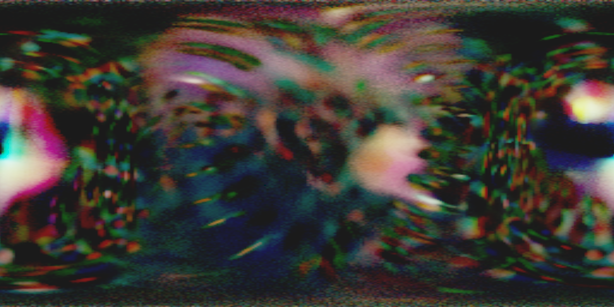}}%
  \\[1pt]
  \makebox[0.05\linewidth][c]{}%
  \hspace{2pt}%
  \makebox[0.215\linewidth][c]{\scriptsize Reference}%
  \hspace{2pt}%
  \makebox[0.215\linewidth][c]{\scriptsize Ours}%
  \hspace{2pt}%
  \makebox[0.215\linewidth][c]{\scriptsize Neural-PBIR}%
  \hspace{2pt}%
  \makebox[0.215\linewidth][c]{\scriptsize MaterialFusion}%
  \\[1pt]
  \caption{Stanford-ORB \casename{blocks_scene005}.}
  \label{fig:supp-stanford-intrinsics-blocks-scene005}
\end{figure*}

\begin{figure*}[t]
  \centering
  \newcommand{\intrinsicrowlabel}[1]{\rotatebox[origin=c]{90}{\scriptsize\strut #1}}
  {\scriptsize\texttt{\detokenize{Stanford-ORB chips_scene003}}}\\[-0.2em]
  \parbox[c]{0.05\linewidth}{\centering \intrinsicrowlabel{Input and Prediction}}%
  \hspace{2pt}%
  \parbox[c]{0.172\linewidth}{\centering \includegraphics[width=\linewidth]{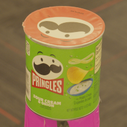}}%
  \hspace{2pt}%
  \parbox[c]{0.172\linewidth}{\centering \includegraphics[width=\linewidth]{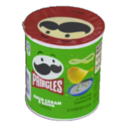}}%
  \hspace{2pt}%
  \parbox[c]{0.172\linewidth}{\centering \includegraphics[width=\linewidth]{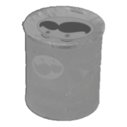}}%
  \hspace{2pt}%
  \parbox[c]{0.172\linewidth}{\centering \includegraphics[width=\linewidth]{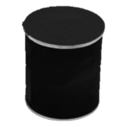}}%
  \hspace{2pt}%
  \parbox[c]{0.172\linewidth}{\centering \includegraphics[width=\linewidth]{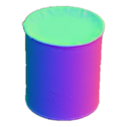}}%
  \\[1pt]
  \makebox[0.05\linewidth][c]{}%
  \hspace{2pt}%
  \makebox[0.172\linewidth][c]{\scriptsize Input}%
  \hspace{2pt}%
  \makebox[0.172\linewidth][c]{\scriptsize DR-base color}%
  \hspace{2pt}%
  \makebox[0.172\linewidth][c]{\scriptsize DR-roughness}%
  \hspace{2pt}%
  \makebox[0.172\linewidth][c]{\scriptsize DR-metallic}%
  \hspace{2pt}%
  \makebox[0.172\linewidth][c]{\scriptsize DR-normal}%
  \\[1pt]
  \vspace{0.15em}
  \par\noindent\rule{\linewidth}{0.35pt}
  \vspace{0.05em}
  \vspace{-0.6em}
  \parbox[c]{0.05\linewidth}{\centering \intrinsicrowlabel{Relight}}%
  \hspace{2pt}%
  \parbox[c]{0.215\linewidth}{\centering \begin{overpic}[width=\linewidth]{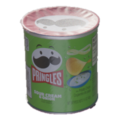}\put(0,0){\includegraphics[width=0.666\linewidth]{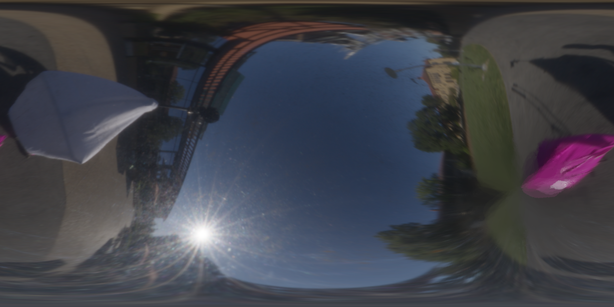}}\end{overpic}}%
  \hspace{2pt}%
  \parbox[c]{0.215\linewidth}{\centering \includegraphics[width=\linewidth]{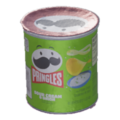}}%
  \hspace{2pt}%
  \parbox[c]{0.215\linewidth}{\centering \includegraphics[width=\linewidth]{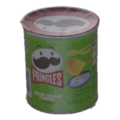}}%
  \hspace{2pt}%
  \parbox[c]{0.215\linewidth}{\centering \includegraphics[width=\linewidth]{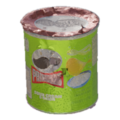}}%
  \\[0.1em]
  \parbox[c]{0.05\linewidth}{\centering \intrinsicrowlabel{Base Color}}%
  \hspace{2pt}%
  \parbox[c]{0.215\linewidth}{\centering \includegraphics[width=\linewidth]{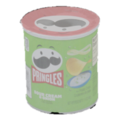}}%
  \hspace{2pt}%
  \parbox[c]{0.215\linewidth}{\centering \includegraphics[width=\linewidth]{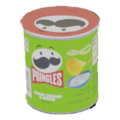}}%
  \hspace{2pt}%
  \parbox[c]{0.215\linewidth}{\centering \includegraphics[width=\linewidth]{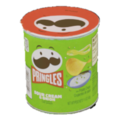}}%
  \hspace{2pt}%
  \parbox[c]{0.215\linewidth}{\centering \includegraphics[width=\linewidth]{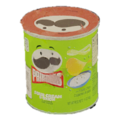}}%
  \\[0.1em]
  \parbox[c]{0.05\linewidth}{\centering \intrinsicrowlabel{Roughness}}%
  \hspace{2pt}%
  \parbox[c]{0.215\linewidth}{\centering \begin{overpic}[width=\linewidth]{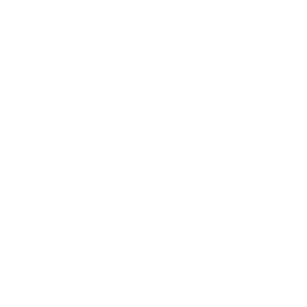}\put(50,50){\makebox(0,0){\scriptsize N/A}}\end{overpic}}%
  \hspace{2pt}%
  \parbox[c]{0.215\linewidth}{\centering \includegraphics[width=\linewidth]{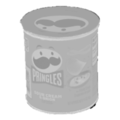}}%
  \hspace{2pt}%
  \parbox[c]{0.215\linewidth}{\centering \includegraphics[width=\linewidth]{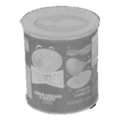}}%
  \hspace{2pt}%
  \parbox[c]{0.215\linewidth}{\centering \includegraphics[width=\linewidth]{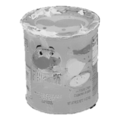}}%
  \\[0.1em]
  \parbox[c]{0.05\linewidth}{\centering \intrinsicrowlabel{Normal}}%
  \hspace{2pt}%
  \parbox[c]{0.215\linewidth}{\centering \includegraphics[width=\linewidth]{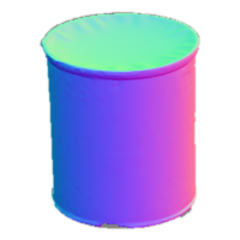}}%
  \hspace{2pt}%
  \parbox[c]{0.215\linewidth}{\centering \includegraphics[width=\linewidth]{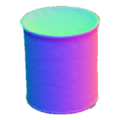}}%
  \hspace{2pt}%
  \parbox[c]{0.215\linewidth}{\centering \includegraphics[width=\linewidth]{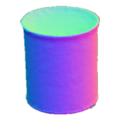}}%
  \hspace{2pt}%
  \parbox[c]{0.215\linewidth}{\centering \includegraphics[width=\linewidth]{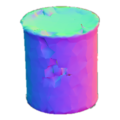}}%
  \\[0.1em]
  \parbox[c]{0.05\linewidth}{\centering \intrinsicrowlabel{Lighting}}%
  \hspace{2pt}%
  \parbox[c]{0.215\linewidth}{\centering \includegraphics[width=\linewidth]{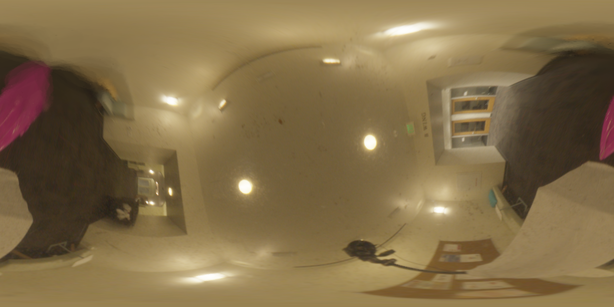}}%
  \hspace{2pt}%
  \parbox[c]{0.215\linewidth}{\centering \includegraphics[width=\linewidth]{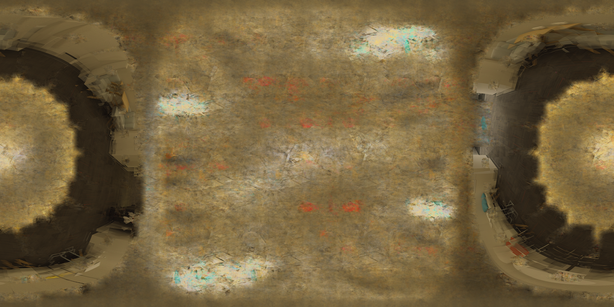}}%
  \hspace{2pt}%
  \parbox[c]{0.215\linewidth}{\centering \includegraphics[width=\linewidth]{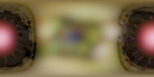}}%
  \hspace{2pt}%
  \parbox[c]{0.215\linewidth}{\centering \includegraphics[width=\linewidth]{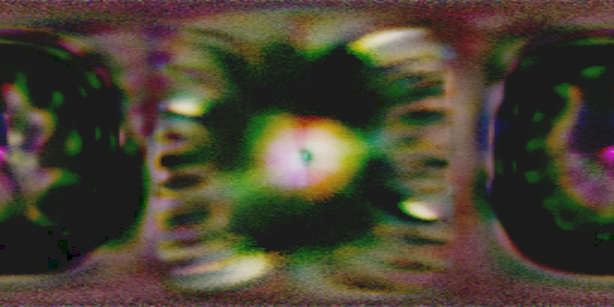}}%
  \\[1pt]
  \makebox[0.05\linewidth][c]{}%
  \hspace{2pt}%
  \makebox[0.215\linewidth][c]{\scriptsize Reference}%
  \hspace{2pt}%
  \makebox[0.215\linewidth][c]{\scriptsize Ours}%
  \hspace{2pt}%
  \makebox[0.215\linewidth][c]{\scriptsize Neural-PBIR}%
  \hspace{2pt}%
  \makebox[0.215\linewidth][c]{\scriptsize MaterialFusion}%
  \\[1pt]
  \caption{Stanford-ORB \casename{chips_scene003}.}
  \label{fig:supp-stanford-intrinsics-chips-scene003}
\end{figure*}

\end{document}